\documentclass[twoside]{article}

\usepackage[accepted]{aistats2023}
\usepackage[round]{natbib}

\usepackage{subcaption}
\usepackage{booktabs}
\usepackage{pifont}
\usepackage{xcolor}
\usepackage{graphicx}
\usepackage{changepage}
\usepackage{bm}

\usepackage{amssymb}
\DeclareMathOperator{\EX}{\mathbb{E}}

\usepackage{amsthm}
\newtheorem{theorem}{Theorem}

\begin{document}

%

%

\twocolumn[

\aistatstitle{NTS-NOTEARS: Learning Nonparametric DBNs With Prior Knowledge}

\aistatsauthor{Xiangyu Sun$^1$ \And Oliver Schulte$^1$ \And  Guiliang Liu$^2$ \And Pascal Poupart$^3$}

\aistatsaddress{$^1$Simon Fraser University \And $^2$The Chinese University of Hong Kong, Shenzhen \And   $^3$University of Waterloo} 

\runningauthor{Xiangyu Sun, Oliver Schulte, Guiliang Liu, Pascal Poupart}

]

\begin{abstract}
We describe NTS-NOTEARS, a score-based structure learning method for time-series data to learn dynamic Bayesian networks (DBNs) that captures nonlinear, lagged (inter-slice) and instantaneous (intra-slice) relations among variables. NTS-NOTEARS utilizes 1D convolutional neural networks (CNNs) to model the dependence of child variables on their parents; 1D CNN is a neural function approximation model well-suited for sequential data. DBN-CNN structure learning is formulated as a continuous optimization problem with an acyclicity constraint, following the NOTEARS DAG learning approach~\citep{zheng2018dags, zheng2020learning}. We show how prior knowledge of dependencies (e.g., forbidden and required edges) can be included as additional optimization constraints.  Empirical evaluation on simulated and benchmark data shows that NTS-NOTEARS achieves state-of-the-art DAG structure quality compared to both parametric and nonparametric baseline methods, with improvement in the range of 10-20\% on the F1-score. We also evaluate NTS-NOTEARS on complex real-world data acquired from professional ice hockey games that contain a mixture of continuous and discrete variables. The code is available online\footnote{https://github.com/xiangyu-sun-789/NTS-NOTEARS}.
\end{abstract}




\section{INTRODUCTION}

\begin{table*}[h]
\centering
\caption{Difference between existing methods and NTS-NOTEARS. Starred methods are evaluation baselines.}
  \begin{tabular}{c|c|c|c|c|c}
    \toprule
    Method & Score-Based & Nonlinear & Temporal & Instantaneous Edges & Acyclic \\
    \midrule
    cMLP & \ding{51} & \ding{51} & \ding{51} & \ding{55} & \ding{51} \\
    \midrule
    Economy-SRU & \ding{51} & \ding{51} & \ding{51} & \ding{55} & \ding{51} \\
    \midrule
    GVAR & \ding{51} & \ding{51} & \ding{51} & \ding{55} & \ding{51} \\
    \midrule
    VAR-LINGAM & \ding{51} & \ding{55} & \ding{51} & \ding{51} & \ding{51} \\
    \midrule
    PCMCI+* & \ding{55} & \ding{51} & \ding{51} & \ding{51} & \ding{51} \\
    \midrule
    TCDF* & \ding{51} & \ding{51} & \ding{51} & \ding{51} & \ding{55} \\
    \midrule
    NOTEARS & \ding{51} & \ding{55} & \ding{55} & \ding{51} & \ding{51} \\
    \midrule
    GraN-DAG & \ding{51} & \ding{51} & \ding{55} & \ding{51} & \ding{51} \\
    \midrule
    NOTEARS-MLP & \ding{51} & \ding{51} & \ding{55} & \ding{51} & \ding{51} \\
    \midrule
    DYNOTEARS* & \ding{51} & \ding{55} & \ding{51} & \ding{51} & \ding{51} \\
    \midrule
    NTS-NOTEARS & \ding{51} & \ding{51} & \ding{51} & \ding{51} & \ding{51} \\
  \bottomrule
\end{tabular}
\label{tab:method_comparison}
\end{table*}



Dynamic Bayesian Networks (DBNs) are graphical models for time-series data. DBNs have many applications in real-world domains such as biology~\citep{sachs2005causal}, finance~\citep{sanford2012bayesian} and economics~\citep{appiah2018investigating}.
The paper addresses the problem of learning DBN structure from time-series data where data samples across time slices are dependent (inter-slice dependencies), and there may also exist instantaneous dependencies among variables at the same time (intra-slice dependencies).


A key issue for directed acyclic graph (DAG) learning is how to model the predictive relationship between a child node and its parents.
Most previous time-series models require the user to select a priori parametric models (e.g., linear). However, when domain knowledge is not available to determine the parametric models, these approaches may lead to model misspecification and incorrect DAG structure. In this paper we develop a new approach to learn nonparametric DBNs where a child value is predicted from its parents using a 1D convolutional neural network (CNN). 
The 1D CNN architecture is designed to model a sequential topology in the input data, and therefore especially suitable for time-series. While a CNN defines a parameter space, it is nonparametric in the sense of being a general function approximator.

\paragraph{Structure Learning} We formulate DBN-CNN model learning as a continuous optimization search for an acyclic weight matrix, by adapting 
 the NOTEARS DAG learning approach for non-temporal data~\citep{zheng2018dags, zheng2020learning}. 
The weight matrix is extracted from the first layer of the trained 1D CNN kernel weights. We show analytically that using the first-layer weights involves no loss of expressive power. We show how the efficient L-BFGS-B optimization algorithm can be leveraged to incorporate useful prior knowledge in the model search, such as forbidden or required edges. 



\paragraph{Evaluation}
Our evaluation focuses on apple-to-apple comparisons within the same model class as NTS-NOTEARS: temporal graphs with both intra-slice and inter-slice dependencies. Our comparison methods include representative methods based on i) nonlinear score optimization using neural networks: TCDF~\citep{nauta2019causal}, 
ii) linear models: DYNOTEARS~\citep{pamfil2020dynotears}, 
and iii) conditional independence (CI) constraints: PCMCI+ with nonlinear CI test ~\citep{runge2020discovering}.


We compare the  learned structures against synthetic and real-world benchmark
ground-truth DBNs (Lorenz 96~\citep{lorenz1996predictability} and fMRI~\citep{smith2011network}), and on a new real-world dataset featuring National Hockey League (NHL) event logs. The hockey data comprise binary, categorical and continuous variables. Compared to the linear DYNOTEARS  model~\citep{pamfil2020dynotears},
NTS-NOTEARS produces structures that are better, 
by as much as 15\% in terms of F1-score on the benchmark datasets. 
We obtain much better improvements over the previous neural-based TCDF method~\citep{nauta2019causal}, due to extracting DAG edge weights from CNNs rather than the attention mechanism. Compared to the constraint-based method PCMCI+~\citep{runge2020discovering} 
with nonlinear CI constraints, NTS-NOTEARS learns substantially better structures and is much more scalable.

\paragraph{Contributions} Our contributions are as follows:

\begin{itemize}
\item We propose 1D CNNs to define a new class of {\em nonparametric DBNs} that capture linear, nonlinear, inter-slice and intra-slice relations among both continuous and discrete variables in time-series data.
    \item We describe NTS-NOTEARS, a continuous optimization approach for learning DBN-CNN models.
    \item We show how prior knowledge of dependencies can be translated into optimization constraints on convolutional weights.
\end{itemize}

The paper is structured as follows. We discuss related works in Section~\ref{section_related_work}, describe NTS-NOTEARS and its training objective in Section~\ref{section_NTS_NOTEARS} and~\ref{section_training_objective}, respectively. Then, we explain how to incorporate prior knowledge in Section~\ref{section_prior_knowledge}. In Section~\ref{section_evaluation}, we evaluate NTS-NOTEARS with simulated data, benchmarks and complex real-world data. 

\section{RELATED WORK}\label{section_related_work}

{\em Non-temporal Nonparametric 
DAG Structure Learning.}
A recent algebraic acyclicity constraint is presented in~\citet{zheng2018dags}, as the basis of the NOTEARS approach to learn instantaneous DAGs in the linear case. Later works such as GraN-DAG~\citep{lachapelle2019gradient} and NOTEARS-MLP~\citep{zheng2020learning} utilize the acylicity constraint for learning nonparametric nonlinear instantaneous DAGs using multilayer perceptrons (MLPs). 

To understand the relationship between  continuous optimization methods and previous DAG structure learning, consider a 2-stage approach: 1) For each variable $X_j$, learn the Markov blanket of $X_j$ using classification/regression methods. Make each member $X_i$ of the Markov blanket a parent of $X_j$. 2) Resolve cycles to produce a DAG with the same Markov blankets. Previous work using the 2-stage approach~\citep{edera2014grow} proposed different discrete algorithms for each stage. Instead of 2 separate stages, the continuous optimization approach introduces 2 different components in the structure learning objective function: 1) A regression component that encourages $X_i$ to be a parent of $X_j$ if $X_i$ improves the prediction of $X_j$. 2) An acyclicity component that discourages cycles in the resulting graph. Both predictive error and cyclicity are jointly minimized using gradient descent. If an MB discovery approach is based on a special differentiable MB predictive loss, it can be incorporated in our method by changing the loss function. 

{\em Temporal DAG Structure Learning.}
Learning DBNs for temporal data is a popular topic. Methods for learning DBN structure can be divided into score-based and constraint-based.

{\em Score-based Methods.} 
Linear autoregressive models include 
DYNOTEARS~\citep{pamfil2020dynotears} and VAR-LINGAM~\citep{hyvarinen2010estimation}. 
DYNOTEARS extends instantaneous linear NOTEARS using autoregression. VAR-LINGAM extends LINGAM~\citep{shimizu2006linear, shimizu2011directlingam}, a linear model class with additive non-Gaussian noise. While linear models generally support fast learning, their capacity is limited compared to our nonlinear model. 

There are several nonlinear neural DBN structure learning methods that estimate inter-slice dependencies only, which is consistent with Granger's approach to causality~\citep{granger1969investigating}. For example, cMLP and cLSTM~\citep{Tank_2021} use MLPs and LSTMs, respectively, to estimate inter-slice DBNs. GVAR~\citep{marcinkevics2021interpretable} estimates 
summary graphs using self-explaining neural networks.
Economy-SRU~\citep{khanna2019economy} is an RNN-based method that learns inter-slice DBNs.
However, ignoring intra-slice dependencies may lead to incorrect estimation of inter-slice relations~\citep{hyvarinen2010estimation}.

To our knowledge, TCDF~\citep{nauta2019causal} is the only other method that also uses CNNs. It constructs a dependency graph structure using attention weights, rather not the NOTEARS method. The attention approach does not guarantee acylicity and is therefore a different model class from DBNs. Previous evaluations~\citep{marcinkevics2021interpretable, khanna2019economy} and our experiments show that the attention weights do not produce accurate graphs. 

{\em Constraint-based Methods} utilize CI tests to estimate graphs. 
PCMCI+~\citep{runge2020discovering} outputs a completed partially directed acyclic graph (CPDAG) with multiple time steps. LPCMCI~\citep{gerhardus2020high} outputs a partial ancestral graph (PAG) that indicates potential latent confounders. Users can choose different CI tests based on linearity assumptions or nonparametric. 
However, nonlinear CI tests are computationally expensive~\citep{zhang2011kernel, runge2018conditional, zheng2020learning, runge2019detecting}.


Table~\ref{tab:method_comparison} summarizes the difference between previous methods and NTS-NOTEARS. Methods excluded from our evaluation are from a different model class; we discuss them further in Appendix~\ref{section_omitted_methods_from_evaluation}.


\section{NTS-NOTEARS MODEL}\label{section_NTS_NOTEARS}


\begin{figure}[ht]
  \centering
  \includegraphics[scale=0.9]{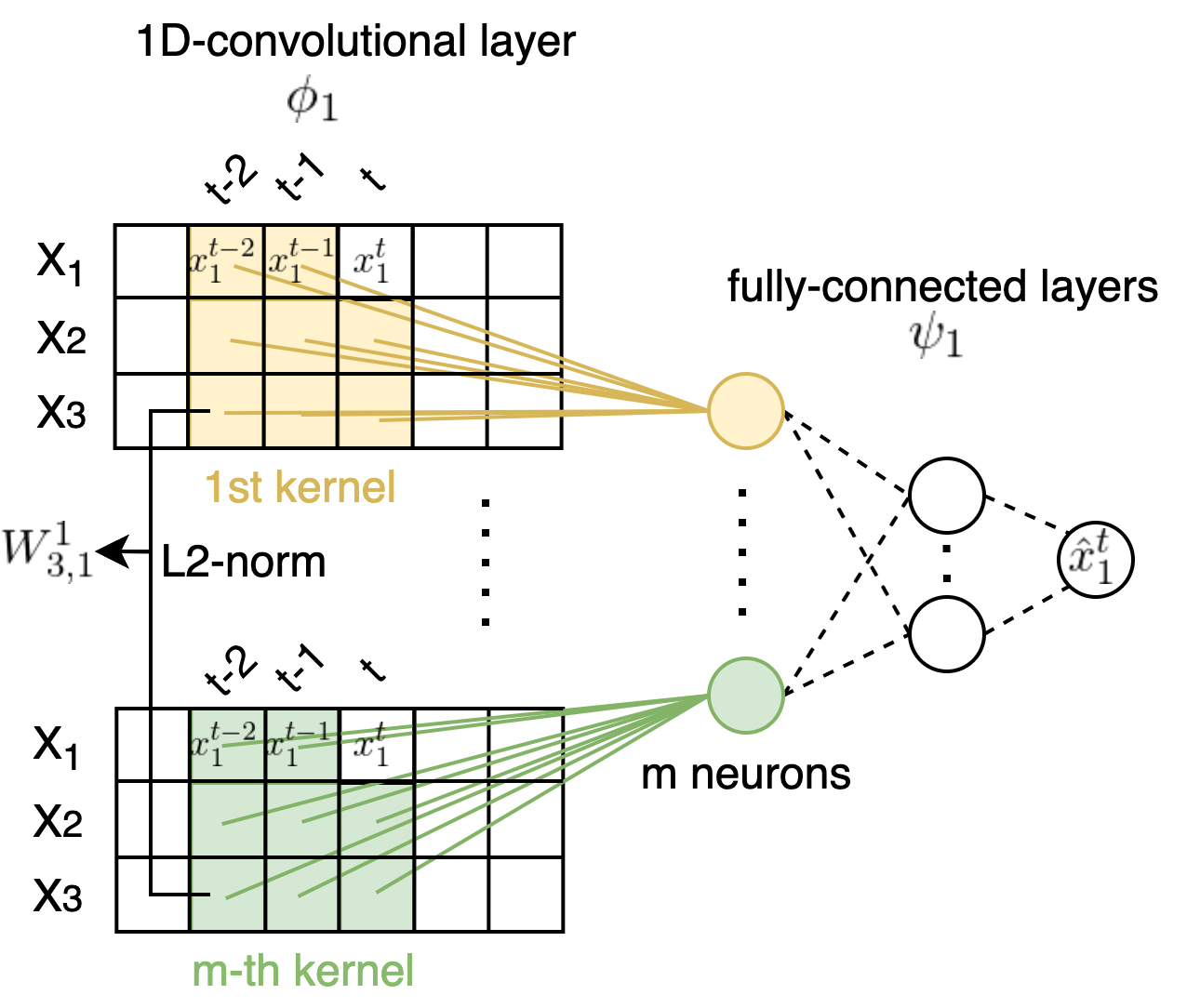}
  \caption{The architecture of NTS-NOTEARS for one child variable $X_1$. The convolutional weights w.r.t. the child variable in the intra-slice $t$ are set to 0. In this example, $K=2$ and $d=3$. For the $j$-th CNN,
  the kernel weights are denoted by $\phi_{j}$, the remaining parameters by $\psi_{j}$, so $\theta_{j}=\{\phi_{j}, \psi_{j}\}$.}
  \label{fig:model_architecture}
\end{figure}

We work with time-series data given by $\{\bm{x}^{t}: t = 1,\ldots, T\}$, where $\bm{x}^{t} \equiv \{x_{1}^{t}, x_{2}^{t}, \ldots, x_{d}^{t}\}$ is a time-indexed $d$-dimensional vector of variables. For categorical variables we use one-hot encoding unless otherwise stated.  
%
%
A DAG $G$ is a directed acyclic graph $(V, E)$ where $V$ represents vertices (i.e. nodes) and $E$ represents 
edges. We assume a one-to-one correspondence between nodes and random variables and treat them interchangeably. Each edge $X_i \rightarrow X_j$ denotes that the variable $X_j$ depends on the value of variable $X_i$.
An edge $X_i^{t} \rightarrow X_j^{t}$ between variables at the same time represents an \textbf{intra-slice} or \textbf{instantaneous} dependency. 
An edge $X_i^{t-k} \rightarrow X_j^{t}$ for $k > 0$ represents an \textbf{inter-slice} or \textbf{lagged} dependency~\citep{pamfil2020dynotears}. 
The structure learning problem is to learn a DBN that captures the dependencies in the data $\{\bm{x}^{t}: t = 1,\ldots, T\}$. 

Following~\citet{pamfil2020dynotears}, 
we assume the underlying data generating process is stationary over time, and can be modelled as a $K$-th order Markov process, where $K$ is a hyperparameter. These are common assumptions that can be found in related works~\citep{runge2020discovering, khanna2019economy, malinsky2018causal, pamfil2020dynotears, hyvarinen2010estimation}.

\paragraph{Temporal CNN Model} Our main contribution is a new nonparametric model of acyclic temporal dependencies between the parent and child variables. We utilize 1D CNNs. A CNN exploits a sequential or grid topology in the input data. For this reason  1D CNNs are used to process sequential sensory and audio data 
~\citep{yildirim2018arrhythmia, jana20201d, guan2019non, li2019speech, abdoli2019end}.
1D CNNs learn local invariant features and aggregate them across the data sequence to learn higher-order sequence features. Higher levels in the CNN aggregate across different lags in a nonlinear trainable way. 

A general MLP does not incorporate data order information. Current MLP-based methods such as cMLP~\citep{Tank_2021} and IDYNO~\citep{gao2022idyno} concatenate the data. Data concatenation with large datasets may cause memory issues and slow down the training speed. Adding positional encoding transformer-style may be an interesting direction for future work, although the quadratic cost of self-attention is a potential concern. Appendix~\ref{section_omitted_methods_from_evaluation} contains further discussion of IDYNO.

We train $d$ CNNs jointly where the $j$-th CNN predicts the expectation of the target variable $X_j^t$ at each time step $t \geq K+1$ given preceding and instantaneous input variables:
\[\EX[X_{j}^{t}| \mathit{PA}(X_{j}^{t}) ] = \mathit{CNN}_{j}(\{\bm{X}^{t-k}:1\leq k \leq K\}, \bm{X}_{-j}^{t})\]
where $\mathit{PA}(X_{j}^{t})$ denotes the parents of $X_{j}^{t}$ that are defined by the trained CNNs (see next paragraph). Here $K$ is a hyperparameter denoting the maximum lag (order), so the input for predicting variable $X_{j}^{t}$ comprises all preceding variables up to the maximum lag, and all variables at the same time step other than $X_{j}$. 
The parameters of the CNN for child variable $X_{j}$ are denoted $\theta_{j}$. Figure~\ref{fig:model_architecture} illustrates our CNN architecture. 
The first layer of each CNN is a 1D convolution layer with $m$ kernels, stride equal to $1$ and no padding. The shape of each convolutional kernel is $ d \times (K+1)$ where the last column $K+1$ represents intra-slice connections. 

\paragraph{From Local CNNs to Model Weights.} Given $d$ parametrized CNNs, we adapt the NOTEARS-MLP approach~\citep{zheng2020learning} to derive a weighted adjacency matrix $W$ that defines a graph structure. Let $\phi_{i,j}^{k}
\subset \theta_{j}$ denote the $m$ kernel weight parameters for input variable $X_{i}^{k}$ in the {\em first} convolutional layer of the $j$-th CNN. 
Each entry $W^{k}_{ij}$ in the weighted adjacency matrix $W$ represents the dependency strength of a directed edge from variable $X_i^{k}$ to variable $X_j^{K+1}$. The estimated dependency strength of an edge between two variables is the $L^2$-norm of their kernel weights:
\begin{equation}\label{CNN_surrogate_formula}
    W^{k}_{ij} = ||\phi_{i,j}^{k}||_{L^2} \mbox{ for } k=1,\ldots,K+1
\end{equation}

Finally, weight thresholds $W^{k}_{\textit{thres}}$ are applied to each time step $k$ to prune edges with weak dependency strengths and define the parent set of each variable.

\paragraph{Expressive Power} We next show that extracting model weights from the first CNN layer involves no loss of expressive power (cf. \citet{zheng2020learning}). 

Say that a function $g(X)$ is independent of input $X$ if and only if $|| \frac{\partial g(X)}{\partial X}||_{L^2} = 0$.
We provide the following theorem characterizing the set of 1D CNNs that are independent of $X_i^k$ (see Appendix~\ref{proof_independence} for the proof).
\begin{theorem}\label{theorem_independence}
Let $F$ be the class of 1D CNNs that are independent of $X_i^k$ and $F_0$ be the class of 1D CNNs such that the $i$-th kernel parameters in the $k$-th column of the $m$ first-layer CNN kernels are all zeros. Then, $F=F_0$.
\end{theorem}

\section{TRAINING OBJECTIVE}\label{section_training_objective}
The training objective comprises four components for local functions: 1) Matching the observed child values given the parents.
2) A sparsity penalty for the CNN weights.
3) A regularization term for all parameters. 
4) A cyclicity penalty to drive the induced weights to define an acyclic graph.

Let $\mathcal{L}$ denote the least-squares loss, $\phi_{j}^{k}$ be the concatenation of the $\phi_{i,j}^{k}$ vectors, and $\theta = \{ \theta_{1}, \ldots,\theta_{d}\}$. The constrained training objective function is defined as:
\begin{gather*}
    \min_{\theta} F(\theta) \\
    \textit{subject to } h(W^{K+1})=0
\end{gather*}
where
\begin{align*}
& F(\theta) = \frac{1}{T-K} \cdot \\
& \sum_{t=K+1}^{T} \sum_{j=1}^{d} \mathcal{L}(X_{j}^{t}, \mathit{CNN}_{\theta_{j}}(\{\bm{X}^{t-k}:1\leq k \leq K\}, \bm{X}_{-j}^{t})) \\ 
& + \sum_{k=1}^{K+1} \lambda_1^{k} \cdot {|| \phi_{j}^{k} ||}_{L^1} + \frac{1}{2}  \lambda_2 \cdot {|| \theta_{j} ||}_{L^2}
\end{align*}
\begin{equation*}
h(W^{K+1}) = \mathit{tr}(e^{W^{K+1} \circ W^{K+1}})-d = 0
\end{equation*}
$tr(A)$ and $e^{A}$ are the trace and matrix exponential of matrix $A$, respectively, and $\circ$ is element-wise product. The function $h$ enforces the acyclicity constraint among intra-slice dependencies~\citep{zheng2020learning}.

The augmented Lagrangian converts the constrained optimization problem to an unconstrained optimization problem. Hence, the actual (unconstrained) training objective function is:
\begin{equation}\label{optimization_formula}
\min_{\theta} F(\theta) + \frac{\rho}{2} \cdot (h(W^{K+1}))^{2} + \alpha \cdot h(W^{K+1})
\end{equation}

If a variable $X_{i}^{k}$ is predictive for the target variable $X_{j}^{t} \mbox{ for } t \geq k$, minimizing~\eqref{optimization_formula} will push $\phi_{i,j}^{k}$ away from $0$. Otherwise, the sparsity penalty and regularization 
will push $\phi_{i,j}^{k}$ towards $0$. If the acyclicity constraint is violated, some parameters in $\phi^{K+1}$ will also be pushed towards $0$ to satisfy the acyclicity constraint. 

We use the L-BFGS-B algorithm~\citep{byrd1995limited, zhu1997algorithm} to optimize the unconstrained objective~\eqref{optimization_formula}.
L-BFGS (and its variants) are commonly used in works related to NOTEARS, with or without neural networks~\citep{zheng2018dags,zheng2020learning,pamfil2020dynotears,ng2020role,yuan2011identifiability}.

\section{FROM PRIOR KNOWLEDGE TO OPTIMIZATION CONSTRAINTS} \label{section_prior_knowledge}


Allowing prior knowledge is often necessary for real-world applications~\citep{shimizu2011directlingam}.
Adding prior knowledge about the ground-truth graph into the learning process increases not only the accuracy but also the speed of learning, since the number of parameters that need to be learned is reduced. 
A useful kind of prior knowledge is specifying a possible range for the dependency weights $W_{ij}^{k}$ between two variables at a fixed lag. For example, specifying $W_{ij}^{k} = 0$ forbids an edge; specifying $W_{ij}^{k} \geq W^{k}_{\textit{thres}}$ requires an edge~\citep{ramsey2018tetrad}. Various DBN structure learning methods make the Granger assumption that there are no intra-slice dependencies (see Section~\ref{section_related_work}). By adding prior knowledge forbidding such edges, 
NTS-NOTEARS can leverage this assumption when valid. We show how such prior knowledge can be represented in our temporal CNN learning method, through the L-BFGS-B formulation. 

According to Equation~\eqref{CNN_surrogate_formula}, the estimated dependency strength $W_{ij}^{k}$ on an edge is equal to the $L^2$-norm of the corresponding CNN kernel parameters. 
Let $b$ denote a dependency strength as prior knowledge specified by user, $m$ be the number of kernels of the convolutional layer of each CNN, and $\bar{b}$ be the translated optimization constraints. Each $b$ is scaled in the following way before being applied to the L-BFGS-B algorithm:
\begin{equation}\label{constraints_translation}
\bar{b} = \sqrt{\frac{{b}^2}{m}}
\end{equation}

The L-BFGS-B algorithm is a second-order memory-efficient nonlinear optimization algorithm that allows bound constraints on parameters. The algorithm allows users to define which sets of parameters are free and which are constrained. For each constrained parameter, the users provide a lower bound and/or an upper bound. 
The bound constraints allow us to integrate prior knowledge into NTS-NOTEARS as follows.  

Let $\theta=\{\dot{\theta}, \bar{\theta}\}$ where $\dot{\theta}$ denote free parameters and $\bar{\theta}$ denote constrained parameters with lower bounds $l$ and upper bounds $u$, where the bounds are translated from prior knowledge according to Equation~\eqref{constraints_translation}. The objective function~\eqref{optimization_formula} becomes
\[ \min_{\dot{\theta} \mathit{, } l_1 \leq \bar{\theta}_1 \leq u_1 \mathit{, } l_2 \leq \bar{\theta}_2 \leq u_2 \mathit{, } \dots } F(\theta) + \frac{\rho}{2} (h(W^{K+1}))^{2} + \alpha h(W^{K+1}) \]
For example, to forbid an edge from $x_1$ in the most recent lag to $x_3$, the following parameter constraint can be applied
\[ \min_{\dot{\theta} \mathit{, } 0 \leq \phi_{1,3}^{K} \leq 0 } F(\theta) + \frac{\rho}{2} (h(W^{K+1}))^{2} + \alpha h(W^{K+1}) \]
The minimum bound for NTS-NOTEARS is $0$, because we take the $L^2$-norm of the parameters in Equation~\eqref{CNN_surrogate_formula} and the estimated dependency strengths on edges are non-negative. Similarly, to require an edge from $x_1$ in the earliest lag to $x_3$, the following parameter constraint can be applied
\[ \min_{\dot{\theta} \mathit{, } l \leq \phi_{1,3}^{1} } F(\theta) + \frac{\rho}{2} (h(W^{K+1}))^{2} + \alpha h(W^{K+1}) \]
where $l$ is a positive number with $l \geq W^1_{\textit{thres}}$. 



\begin{figure*}[ht]
  \centering
  \begin{subfigure}{0.15\textwidth}
      \centering
      \includegraphics[width=\textwidth]{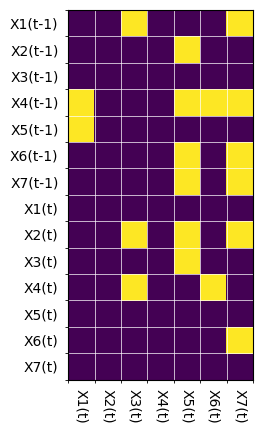}
      \caption{Ground-truth DBN}
      \label{dag:truth}
  \end{subfigure}
  \hfill
  \begin{subfigure}{0.15\textwidth}
      \centering
      \includegraphics[width=\textwidth]{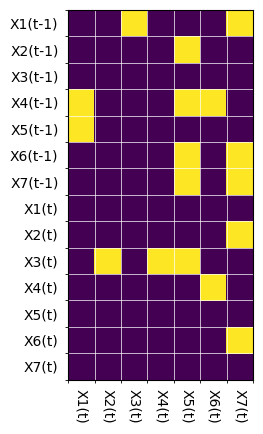}
      \caption{No prior knowledge}
      \label{dag:no-prior}
  \end{subfigure}
  \hfill
  \begin{subfigure}{0.15\textwidth}
    \centering
    \includegraphics[width=\textwidth]{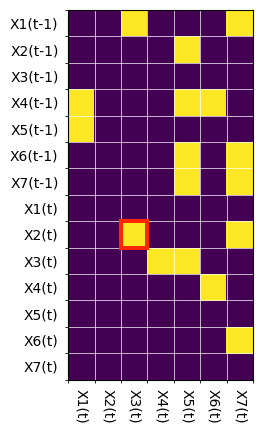}
    \caption{Require $X_2^t \rightarrow X_3^t$.}
    \label{dag:required}
  \end{subfigure}
  \hfill
  \begin{subfigure}{0.15\textwidth}
      \centering
      \includegraphics[width=\textwidth]{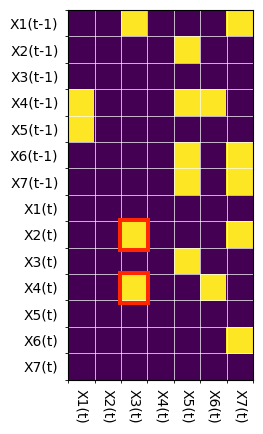}
      \caption{Forbid $X_3^t \rightarrow X_2^t$}
      \label{dag:forbidden}
  \end{subfigure}
  \caption{Each row represents a parent variable and each column represents a child variable. For instance, the yellow top-right cell represents the edge $X_1^{t-1} \rightarrow X_7^{t}$. Figures~\ref{dag:no-prior}--\ref{dag:forbidden} show DBNs learned by NTS-NOTEARS with various types of prior knowledge. ~\eqref{dag:truth} The random ground-truth DBN. ~\eqref{dag:no-prior} Without prior knowledge, the method recovers most of the true edges except $X_2^t \rightarrow X_3^t$ (reversed), $X_4^t \rightarrow X_3^t$ (reversed), $X_2^t \rightarrow X_5^t$ (missed),
  and $X_4^{t-1} \rightarrow X_7^t$ (missed). 
  The structural Hamming distance (SHD) between the ground-truth graph and the estimated graph is 4. ~\eqref{dag:required} By simply adding a lower bound constraint $ W^2_{\textit{thres}} \leq W_{2,3}^{2} $ that requires the edge $X_2^t \rightarrow X_3^t$, the method recovers the true edge $X_2^t \rightarrow X_3^t$. The SHD is reduced to 3.
 ~\eqref{dag:forbidden} With the edge $X_3^t \rightarrow X_2^t$ forbidden, the method recovers not only the true edge $X_2^t \rightarrow X_3^t$ that is directly relevant to the provided prior knowledge but also another true edge $X_4^t \rightarrow X_3^t$ that is not directly relevant. With this prior knowledge alone, the SHD is reduced to 2.
  }
  \label{fig:color_map_prior_knowledge_example}
\end{figure*}


Figure~\ref{fig:color_map_prior_knowledge_example} illustrates the benefits of having prior knowledge. We apply NTS-NOTEARS  to a simulated ground-truth DBN containing 2 time steps and 7 nodes per time step.
Please see the caption for a detailed explanation. It shows that providing prior knowledge via optimization constraints may help to recover edges that are not explicitly encoded by the prior knowledge.

\section{EVALUATION} \label{section_evaluation}

All the experiments were performed on a computer equipped with an Intel Core i7-6850K CPU at 3.60GHz, 32GB memory and an Nvidia GeForce GTX 1080Ti GPU.  
All datasets are normalized to have mean $0$ and standard deviation $1$ to remove patterns in marginal variance~\citep{reisach2021beware}.

\paragraph{Comparison Methods}
We compare NTS-NOTEARS with several recent structure learning methods: TCDF~\citep{nauta2019causal}, DYNOTEARS~\citep{pamfil2020dynotears} 
and PCMCI+ with GPDC nonlinear CI test~\citep{runge2020discovering}. Note that PCMCI+ outputs a CPDAG with undirected edges. We evaluate it favourably by counting the undirected edges as correctly oriented regardless of the ground-truth edge direction.
We follow the closely related DYNOTEARS work and report {\em F1-scores as our main metric for comparing learned graphs to ground-truth graphs.}  Results for other metrics (SHD, precision and recall) are reported in the appendix.


\subsection{Simulated Data}\label{section_simulated}

We generate 48 synthetic parametrized DBN models and then evaluate the DBN structure learned against data sampled from each ground-truth model. For generating synthetic DAGs, we follow~\citet{pamfil2020dynotears}. For sampling from a model, we extend the simulator\footnote{https://github.com/xunzheng/notears} provided by NOTEARS~\citep{zheng2020learning,zheng2018dags} to temporal data. Figure~\ref{fig:simulation_flowchart} in the appendix shows how the simulated data is generated.





{\em DBN Generation.} 
The random ground-truth DAGs are generated based on either the Erdos-Renyi (ER) scheme~\citep{newman2018network} or the Barabasi-Albert (BA) scheme~\citep{barabasi1999emergence} by varying the number of nodes $(K+1) \times d$ and mean out-degrees. Given a random ground-truth DAG, the data is simulated based on one of the following three identifiable nonlinear structural causal models (SCMs): additive noise models (ANM)~\citep{peters2017elements}, additive index models (AIM)~\citep{yuan2011identifiability, alquier2013sparse}
and generalized linear models with Poisson distribution (GLM-Pois)~\citep{park2019high}. 
The data simulated with GLM-Pois is discrete, the data simulated by the other models is continuous with Gaussian noise. The SCM parameters are generated by uniform sampling from a closed interval following~\citet{zheng2020learning, pamfil2020dynotears}. The number of lags is set to $3$ in the ground-truth models. Please see Appendix \ref{section_simulation_details} for more simulation details.

{\em Data Generation.}
For each ground-truth DBN, we generate training data with two sequence lengths $T \in \{200, 1000\}$. 
We create validation data sets as follows. For graph sizes $\{20, 40, 60, 80\}$, reference DBNs are generated using BA, AIM, intra-slice mean out-degree equal to 2, and inter-slice mean out-degree equal to 1. Then sample from each reference DBN two sequences, one for each length $T \in \{200, 1000\}$, with number of lags $=3$.

\paragraph{Hyperparameters} 
For each method and each sequence length, we select hyperparameters with a grid search that maximizes the F1-score, averaged over the validation sets for each sequence length. These hyperparameters are used as defaults for all synthetic datasets. Performance can be further improved by tuning hyperparameters to each dataset through cross-validation. However, using default hyperparameters supports assessing the general approach, as noted by~\citet{ng2020role, zheng2020learning}. Please see Appendix~\ref{section_hyperparameters_simulated} for hyperparameter values.

\begin{figure*}[h]
  \centering
  \begin{subfigure}{0.49\textwidth}
    \centering
    \includegraphics[width=0.49\textwidth, height=0.35\textwidth]{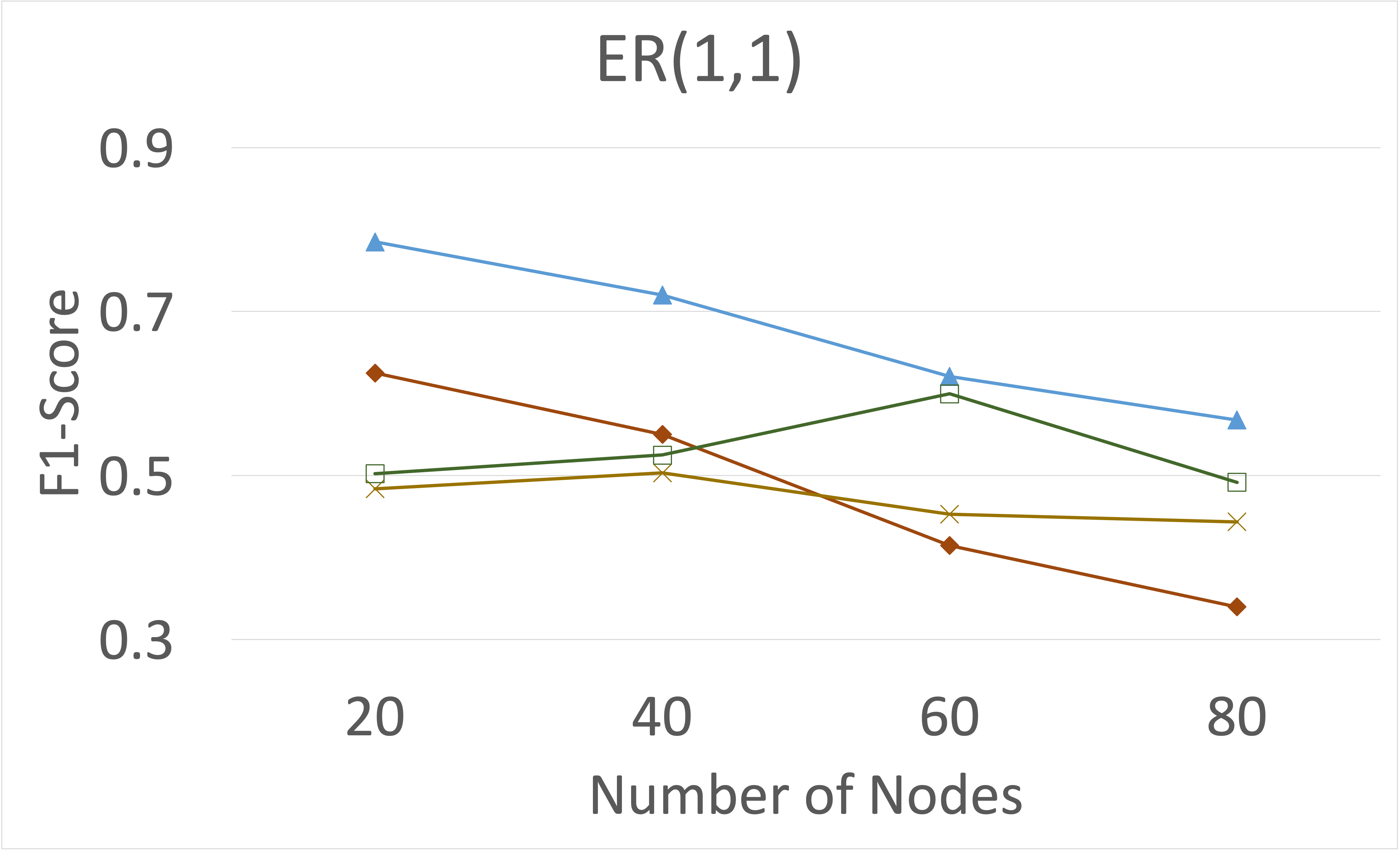}
    \includegraphics[width=0.49\textwidth, height=0.35\textwidth]{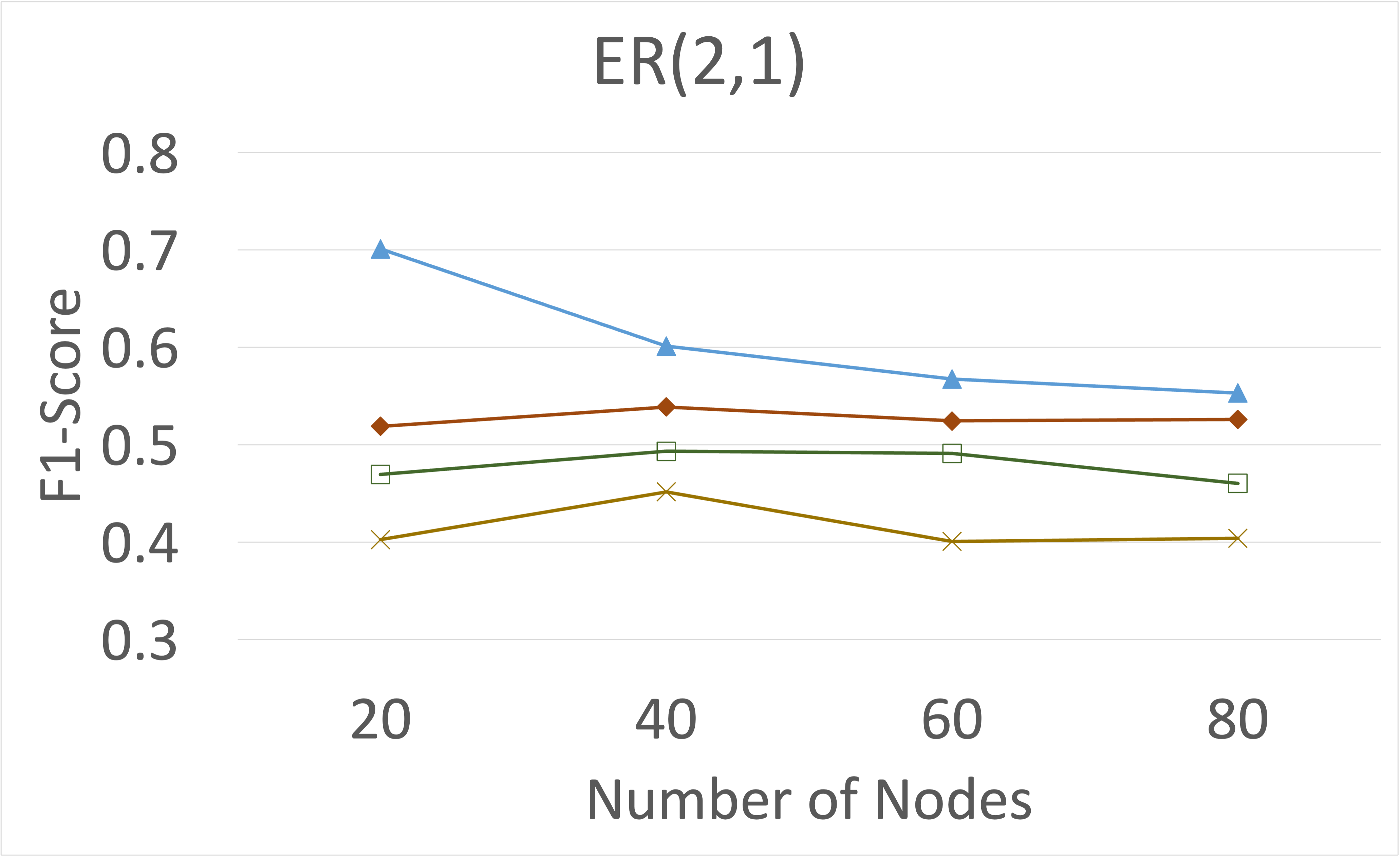}
    \includegraphics[width=0.49\textwidth, height=0.35\textwidth]{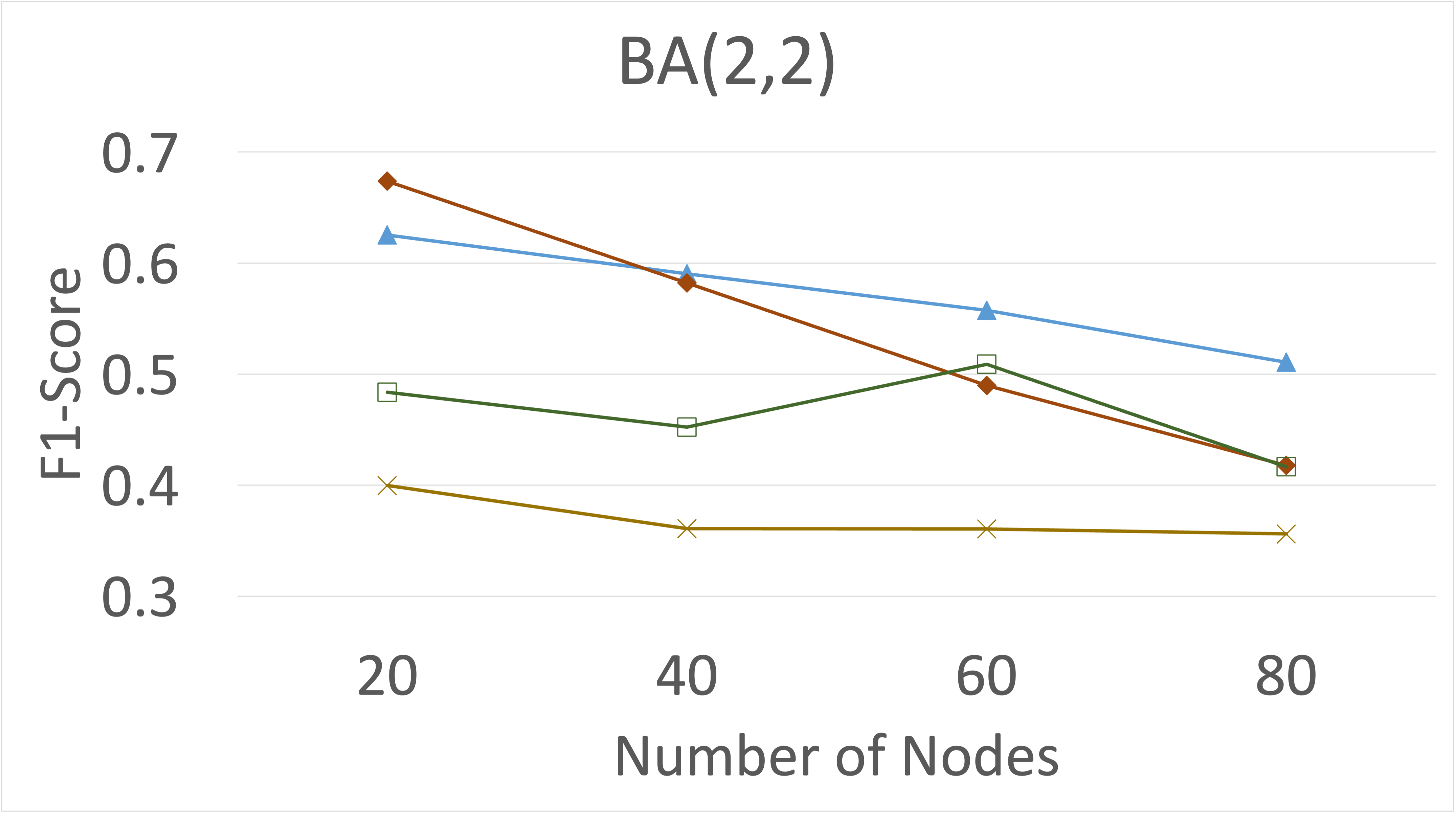}
    \includegraphics[width=0.49\textwidth, height=0.35\textwidth]{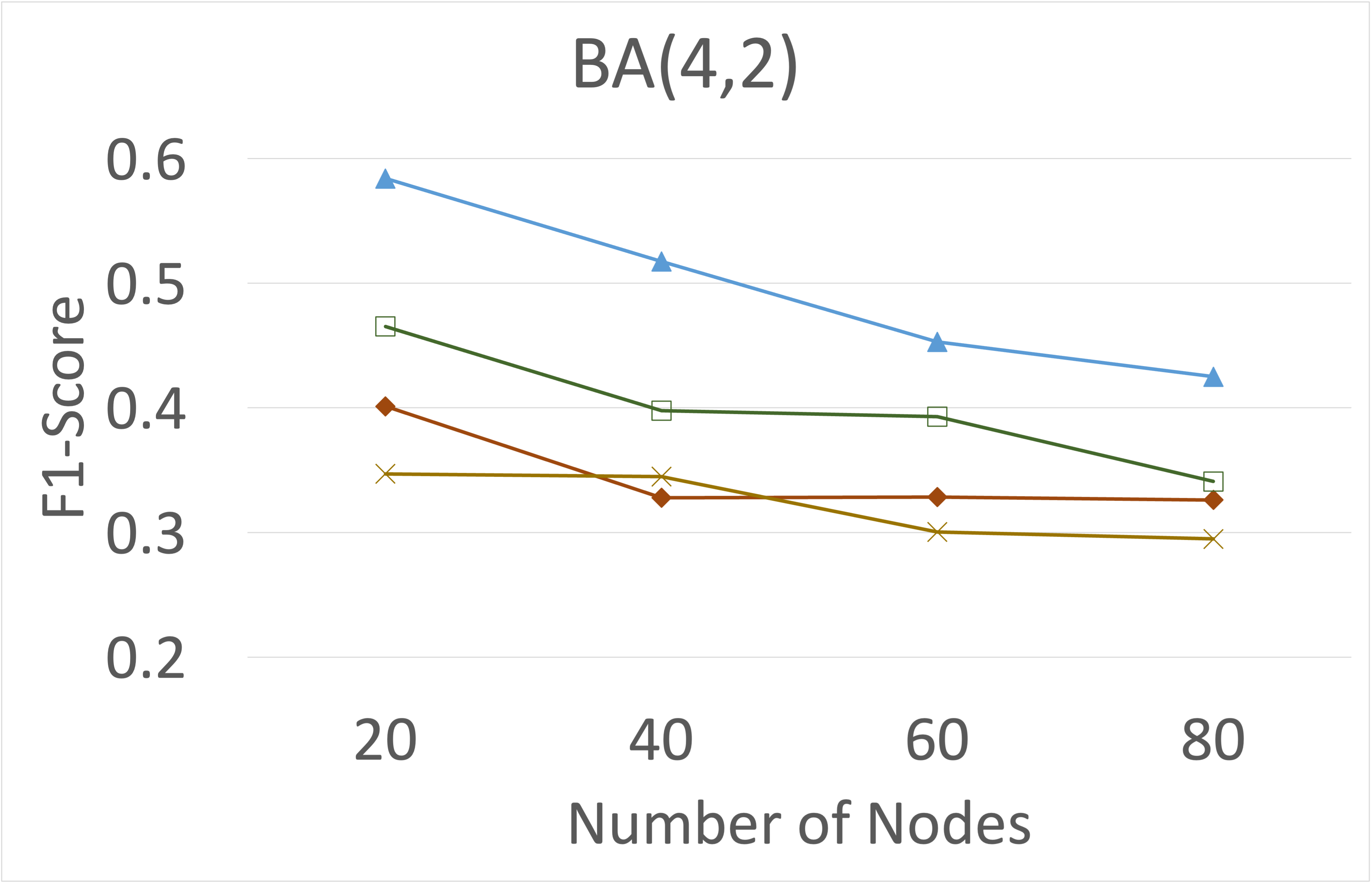}
    \caption{Additive Index Model, $T=200$}
  \end{subfigure}
  \hfill
  \centering
  \begin{subfigure}{0.49\textwidth}
    \centering
    \includegraphics[width=0.49\textwidth, height=0.35\textwidth]{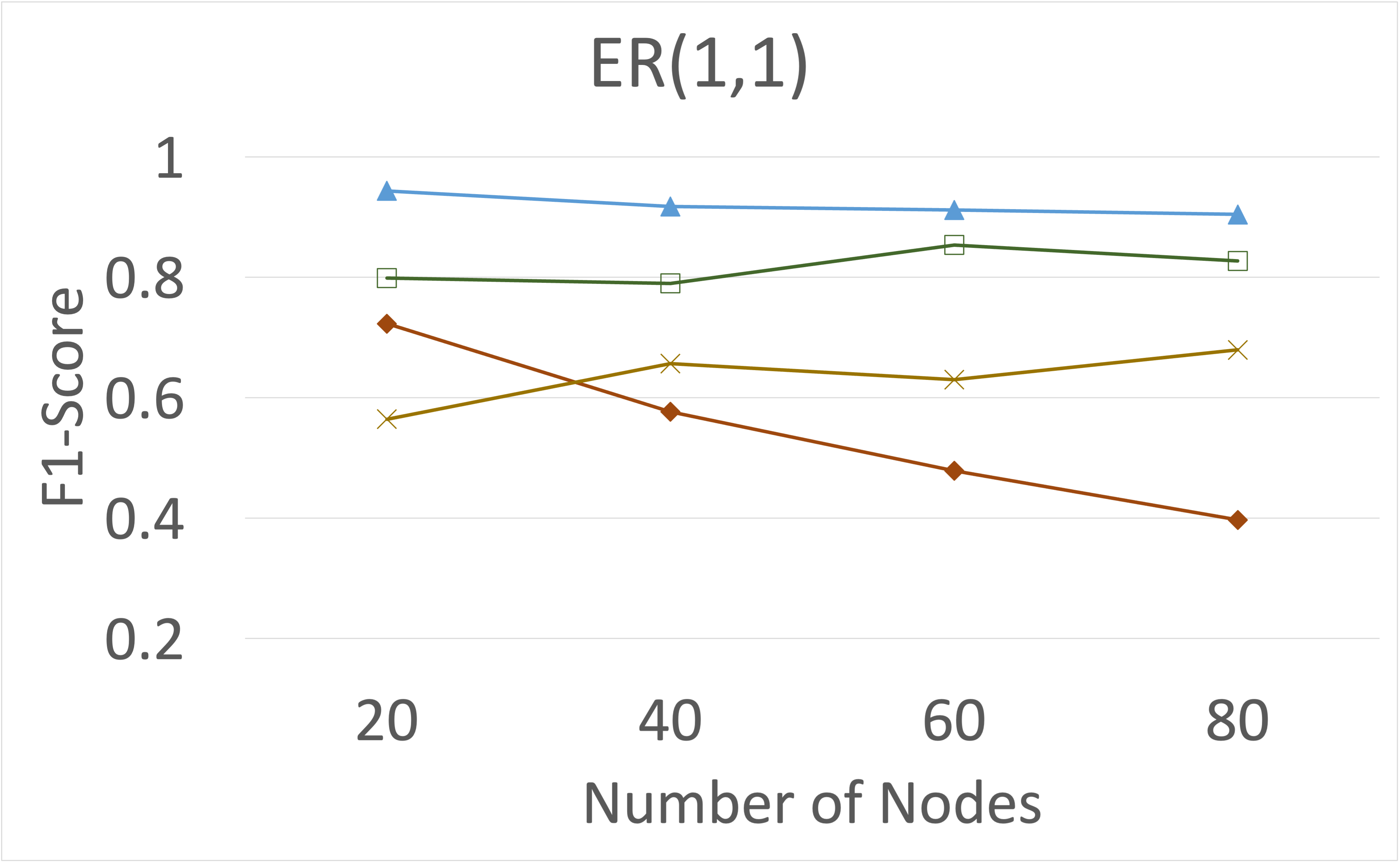}
    \includegraphics[width=0.49\textwidth, height=0.35\textwidth]{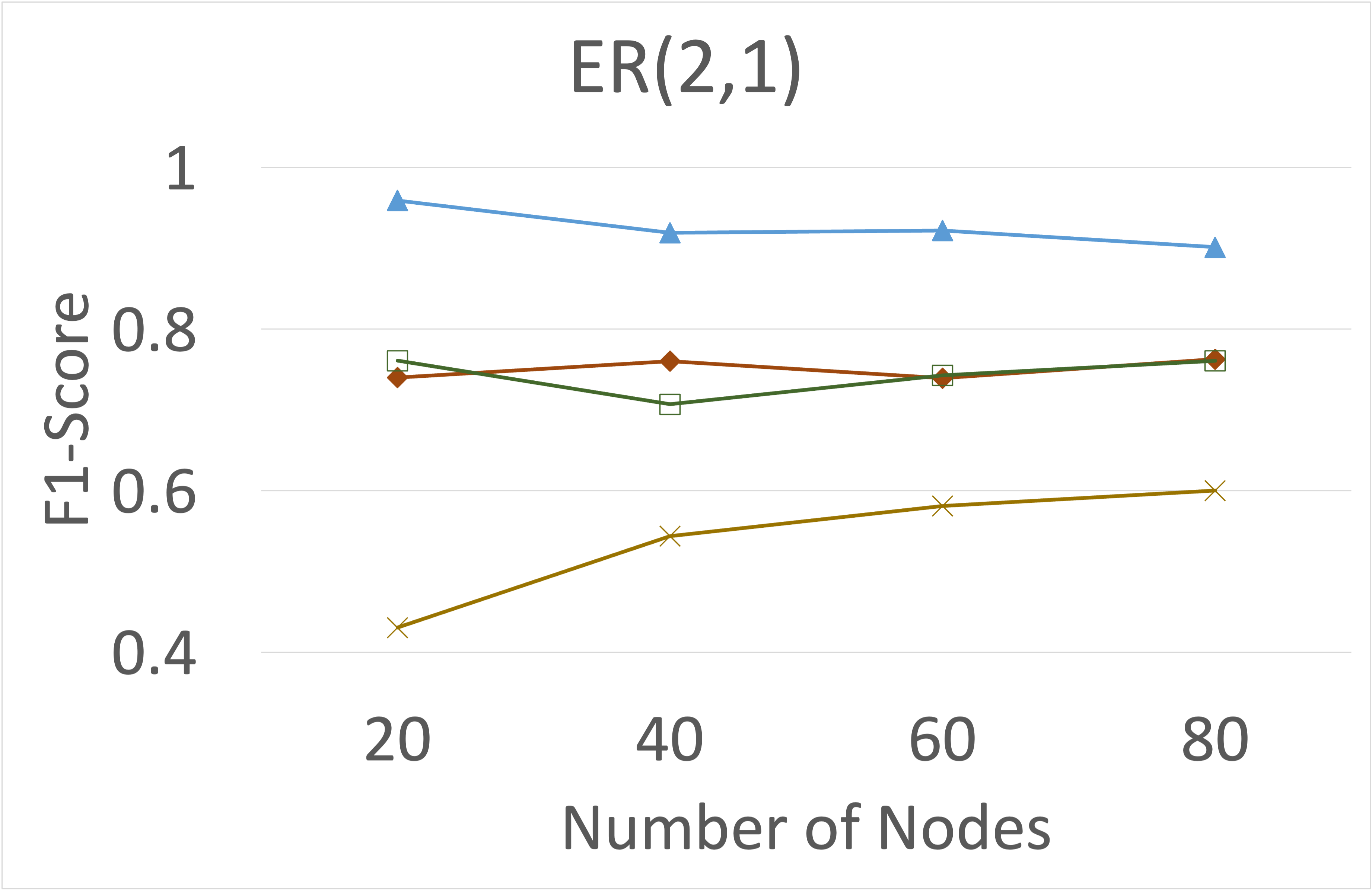}
    \includegraphics[width=0.49\textwidth, height=0.35\textwidth]{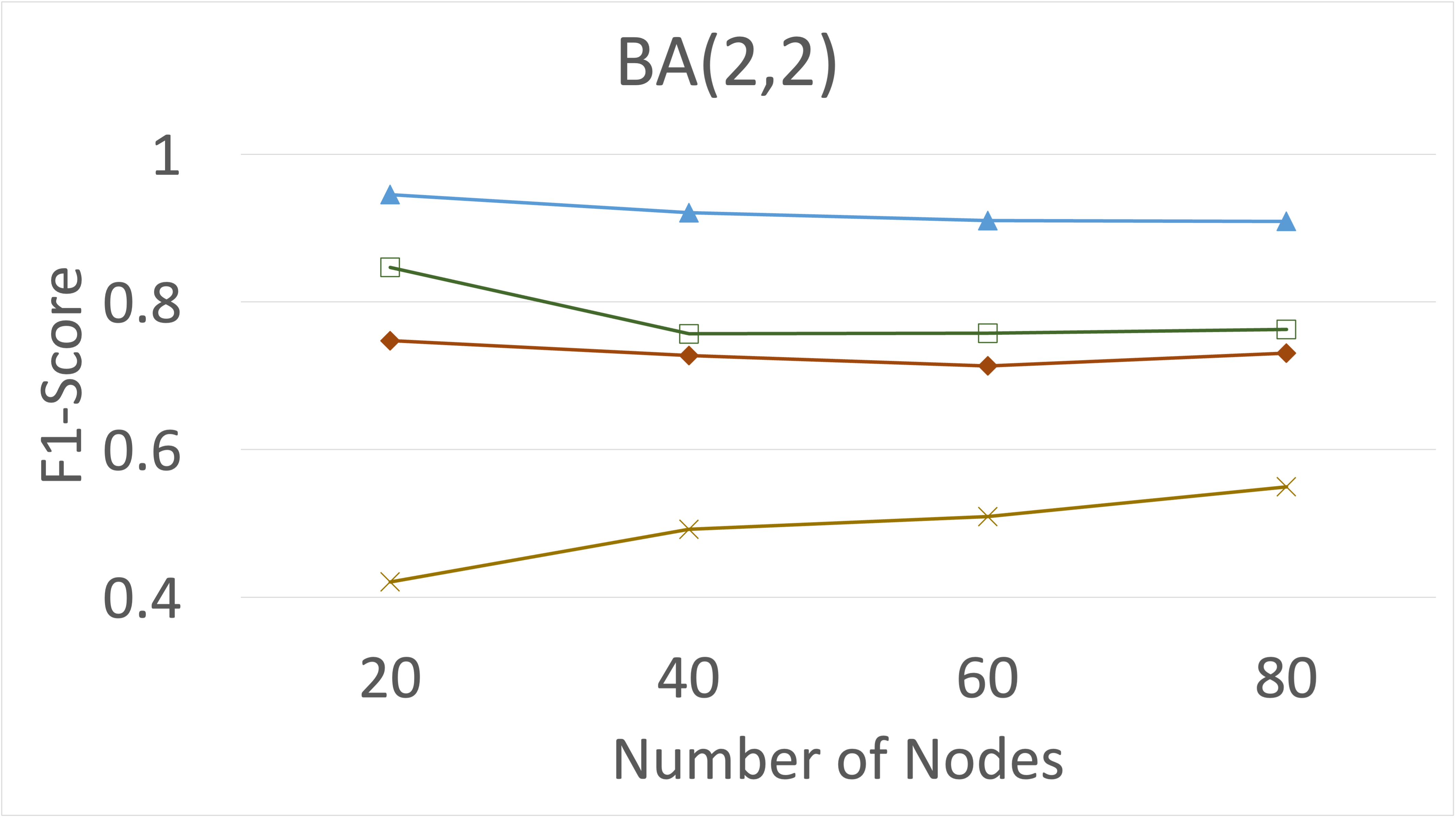}
    \includegraphics[width=0.49\textwidth, height=0.35\textwidth]{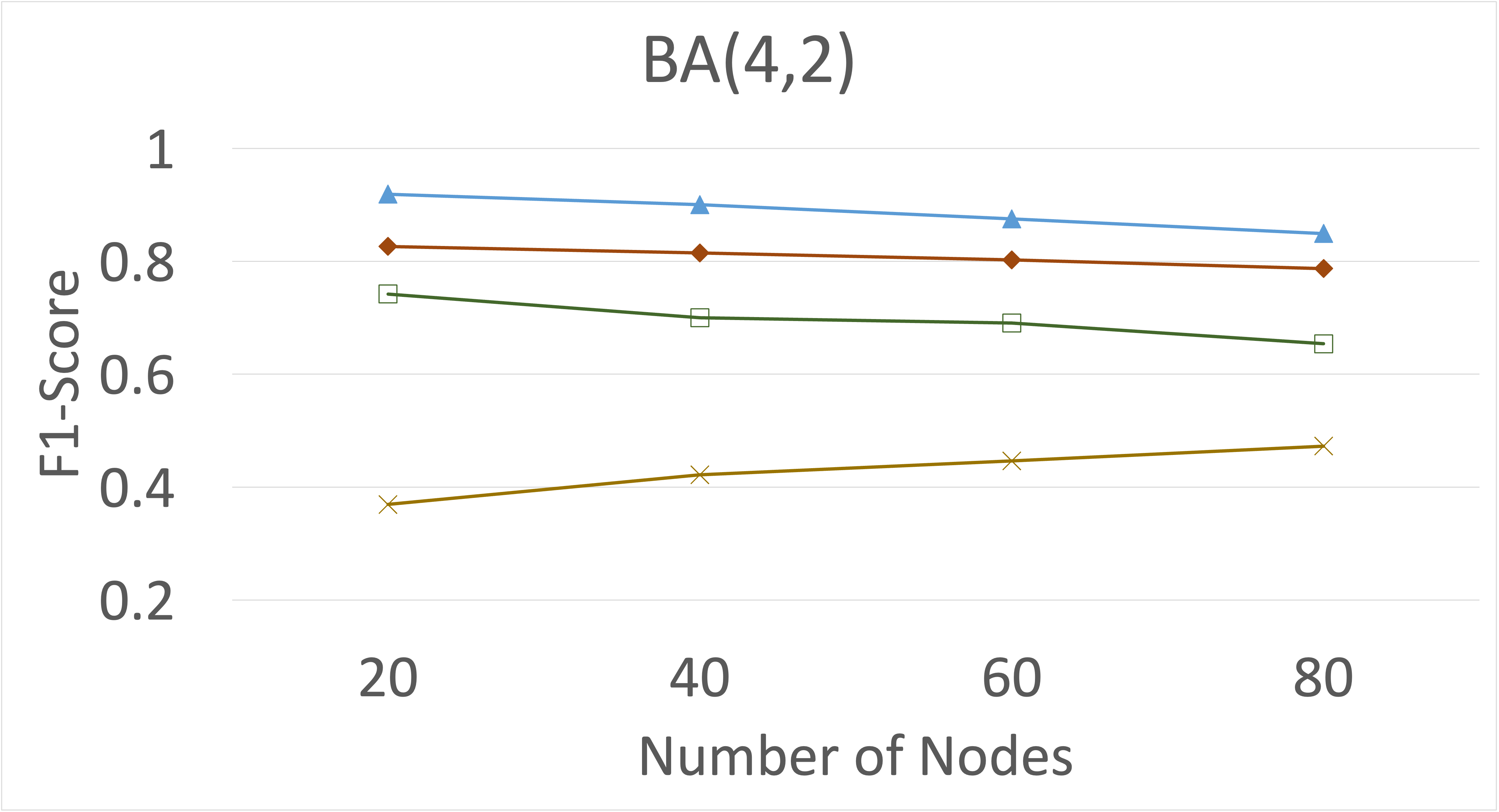}
    \caption{Additive Index Model, $T=1000$}
  \end{subfigure}
  \centering
  \begin{subfigure}{0.49\textwidth}
    \centering
    \includegraphics[width=0.49\textwidth, height=0.35\textwidth]{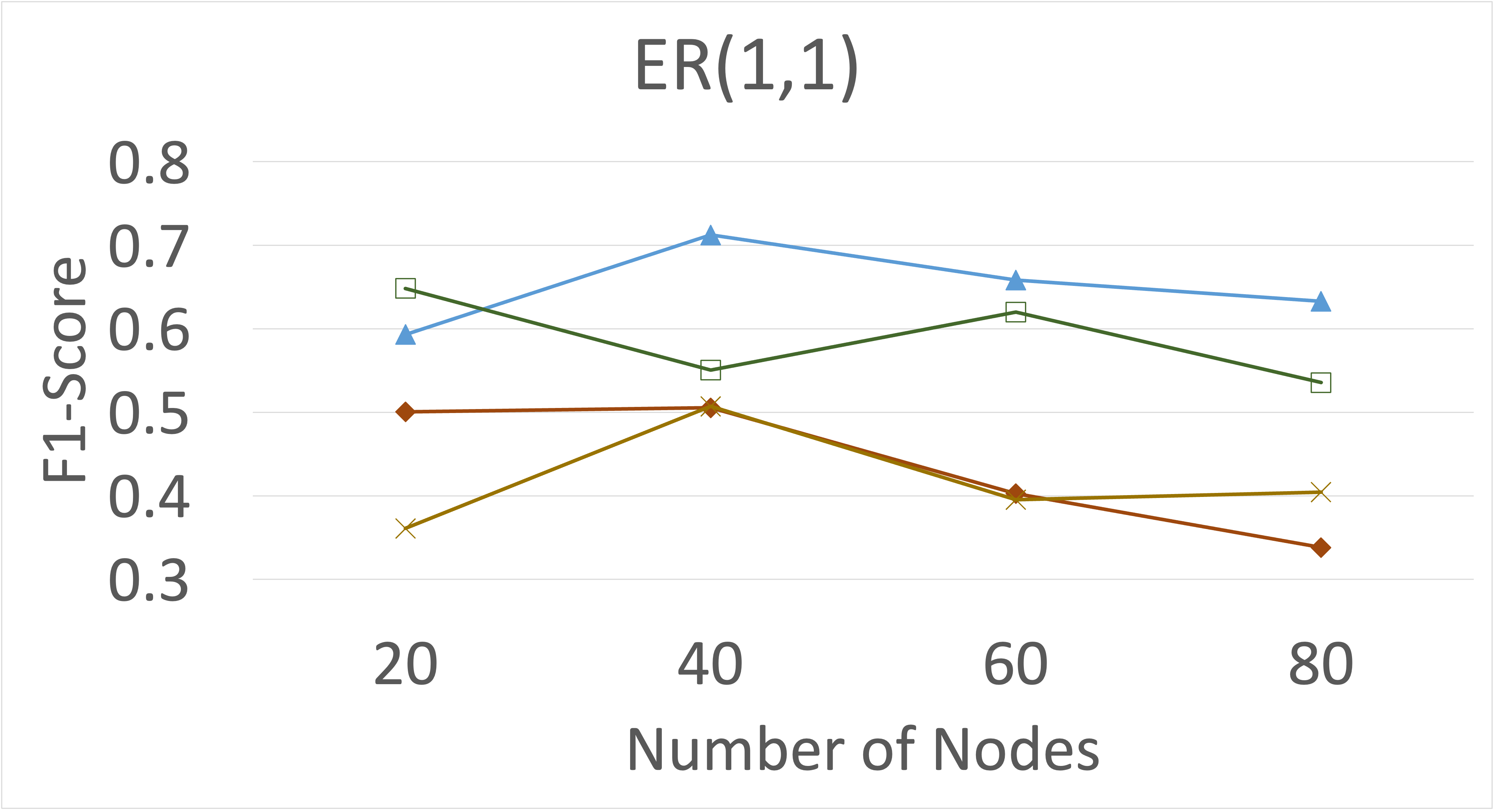}
    \includegraphics[width=0.49\textwidth, height=0.35\textwidth]{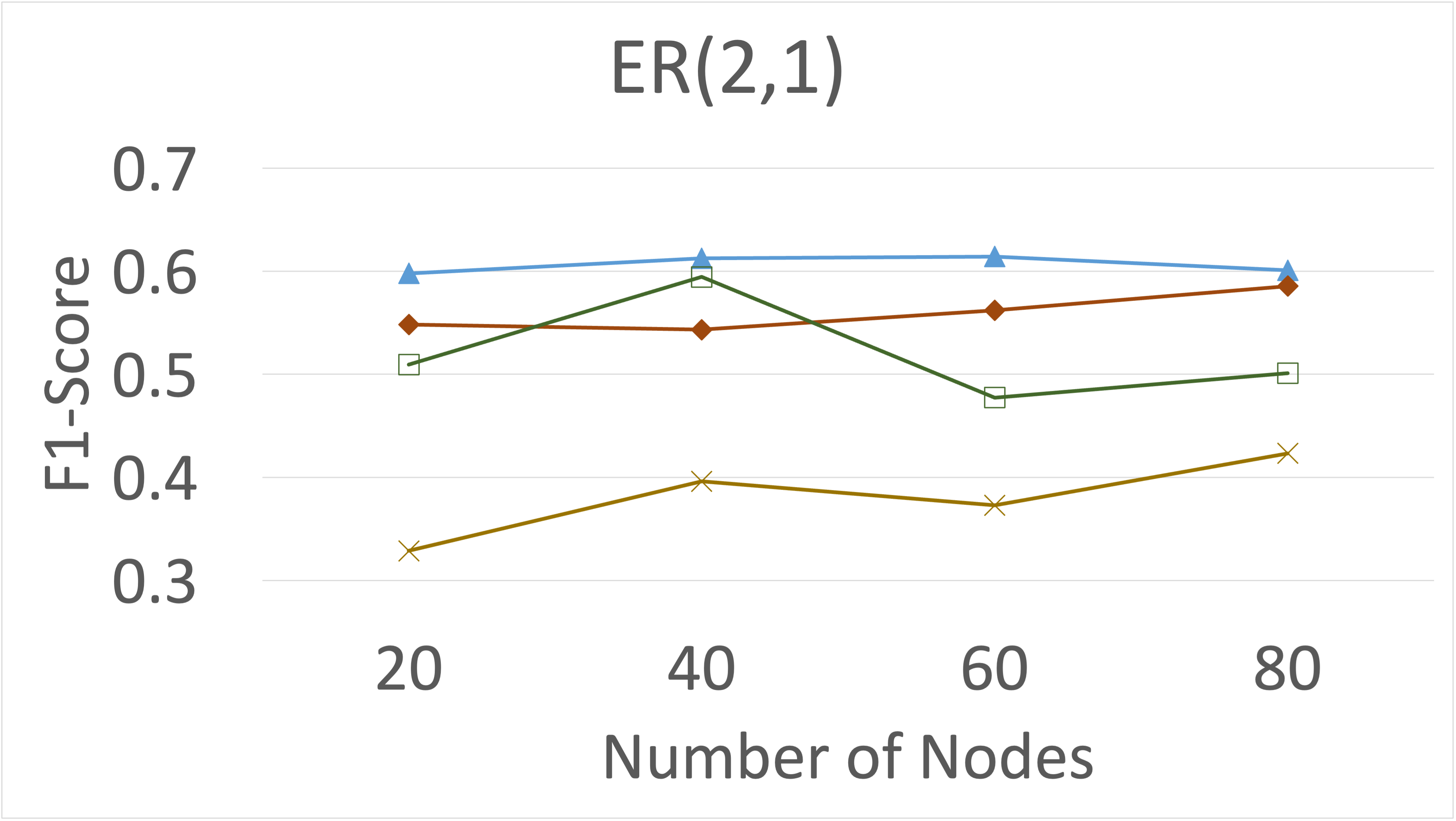}
    \includegraphics[width=0.49\textwidth, height=0.35\textwidth]{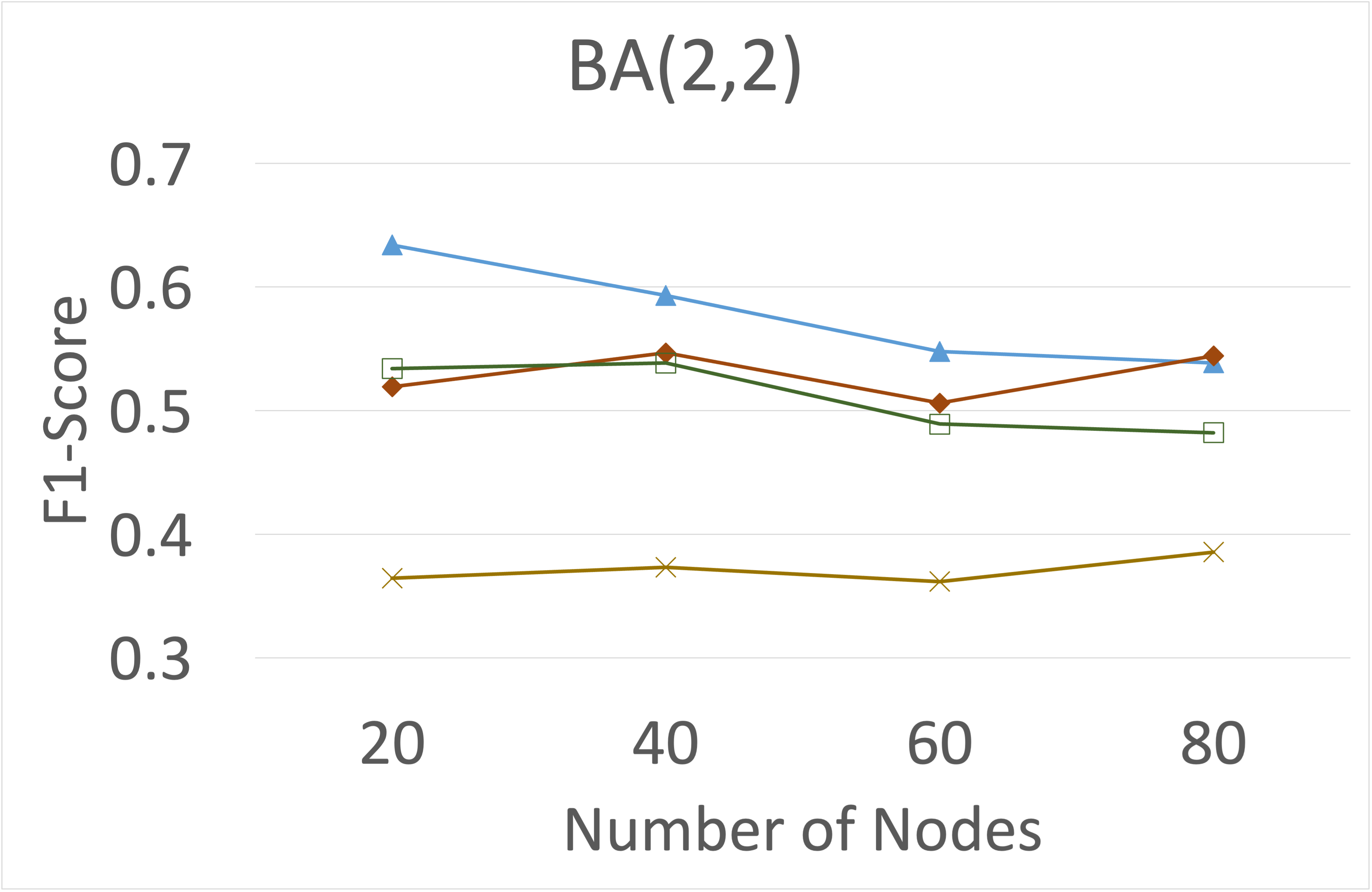}
    \includegraphics[width=0.49\textwidth, height=0.35\textwidth]{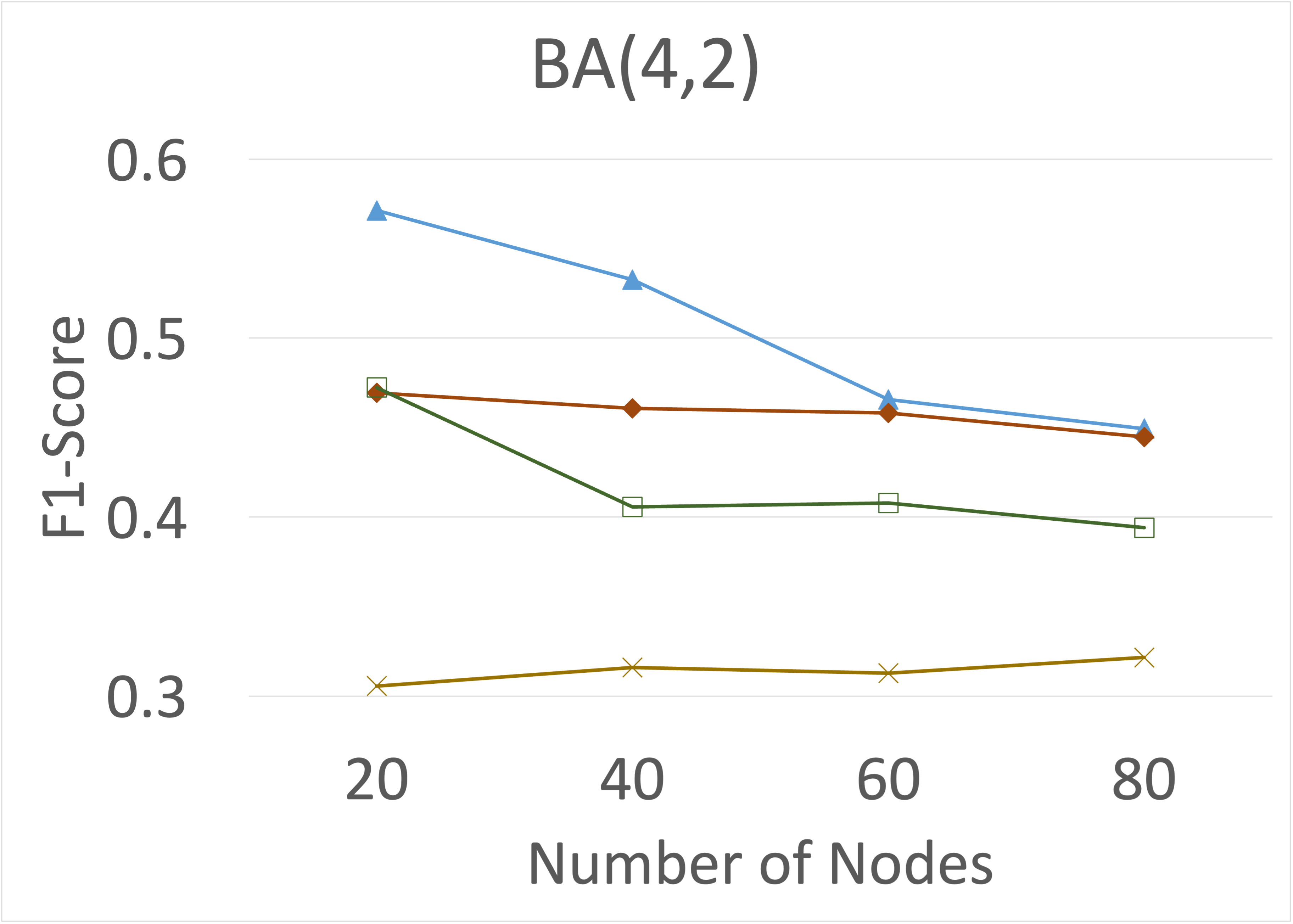}
    \caption{Additive Noise Model, $T=200$}
  \end{subfigure}
  \hfill
  \centering
  \begin{subfigure}{0.49\textwidth}
    \centering
    \includegraphics[width=0.49\textwidth, height=0.35\textwidth]{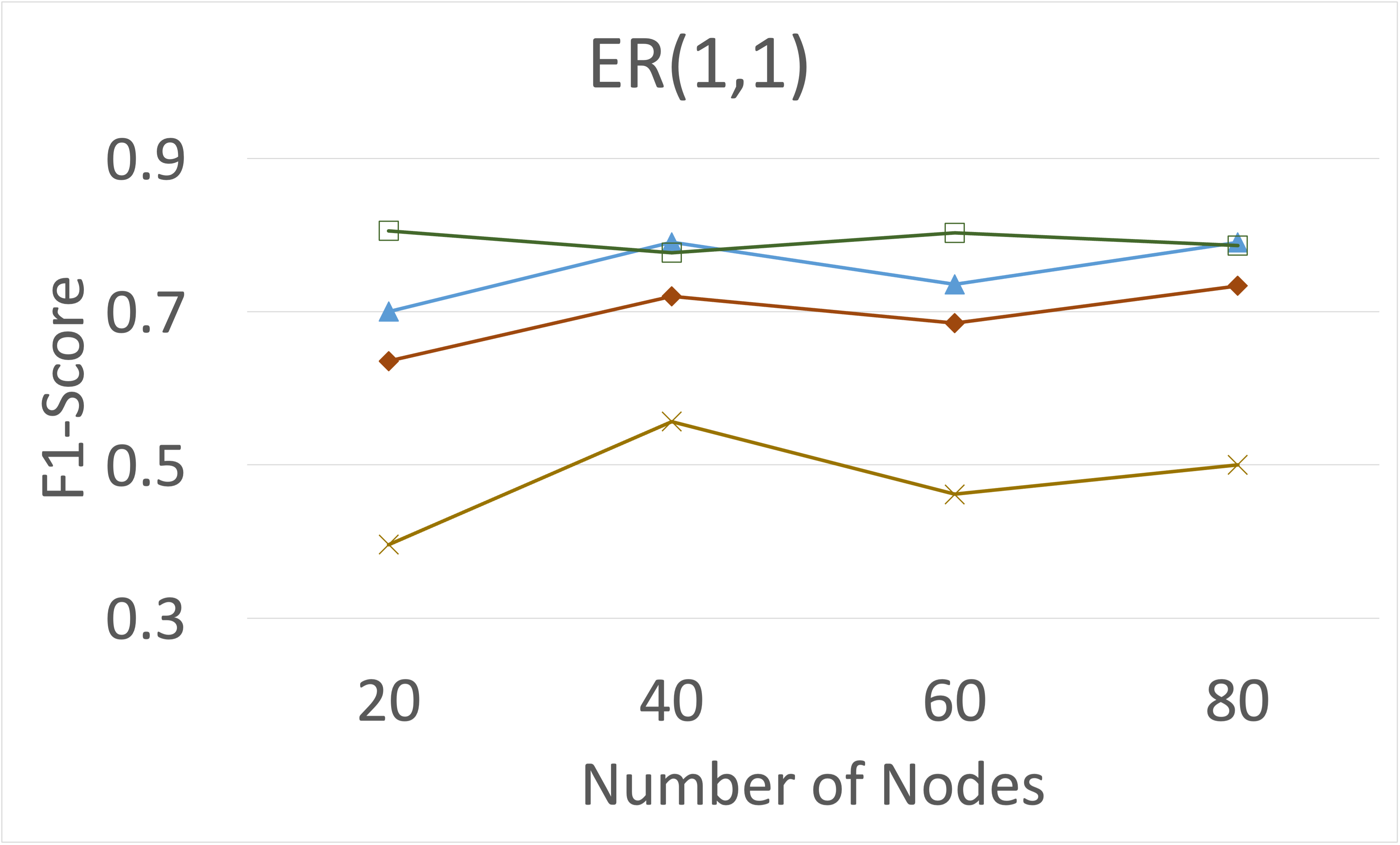}
    \includegraphics[width=0.49\textwidth, height=0.35\textwidth]{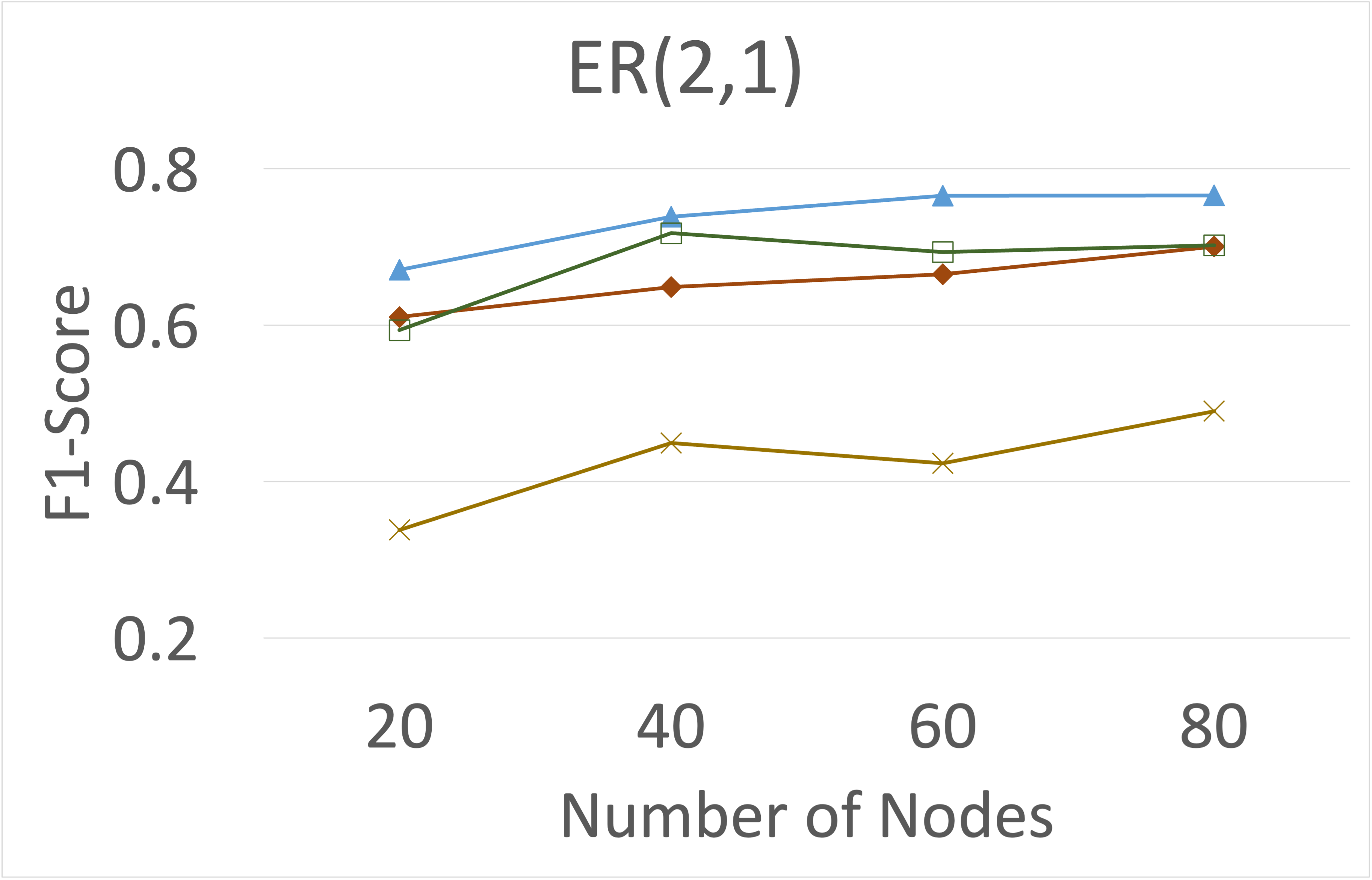}
    \includegraphics[width=0.49\textwidth, height=0.35\textwidth]{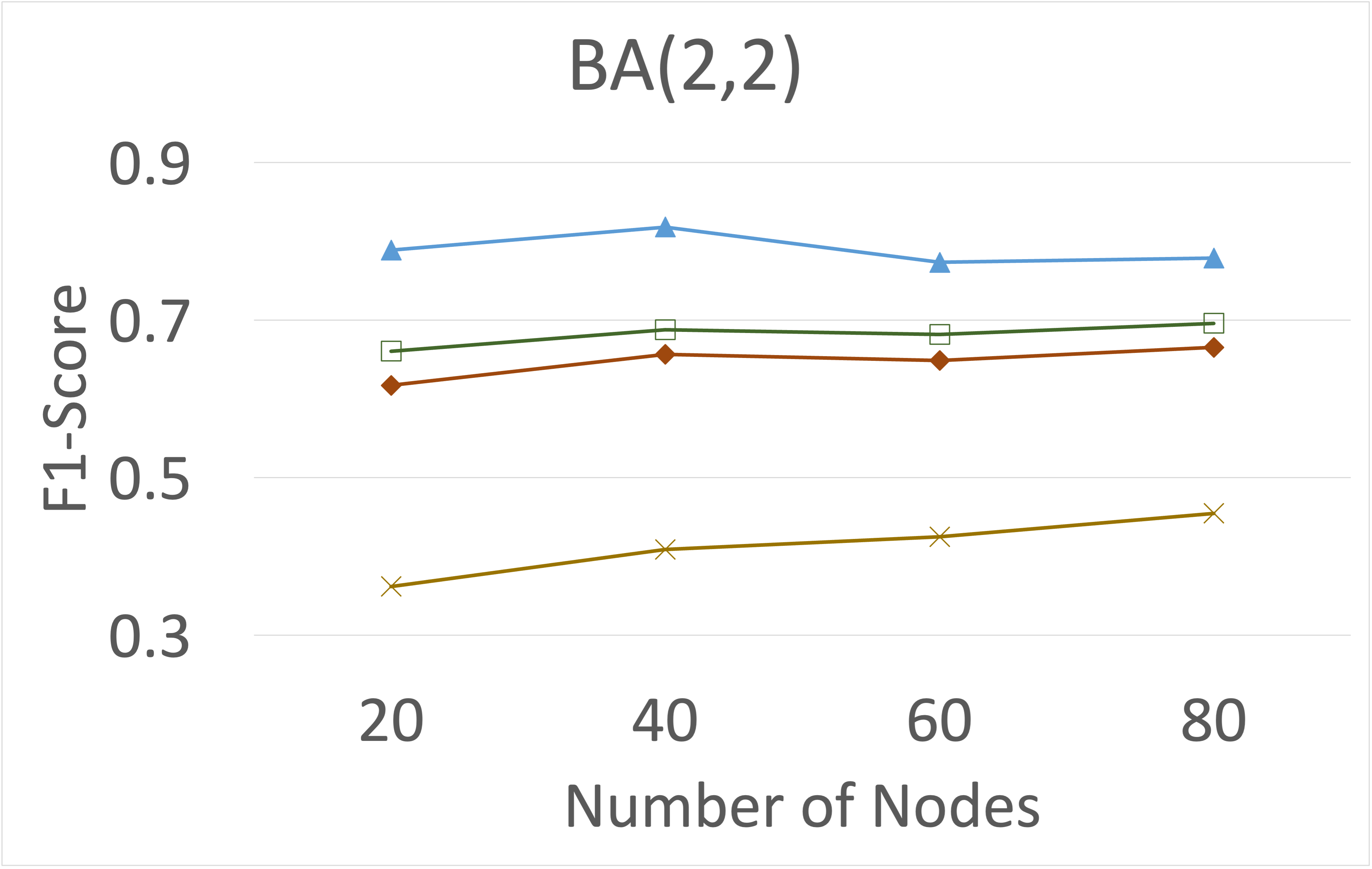}
    \includegraphics[width=0.49\textwidth, height=0.35\textwidth]{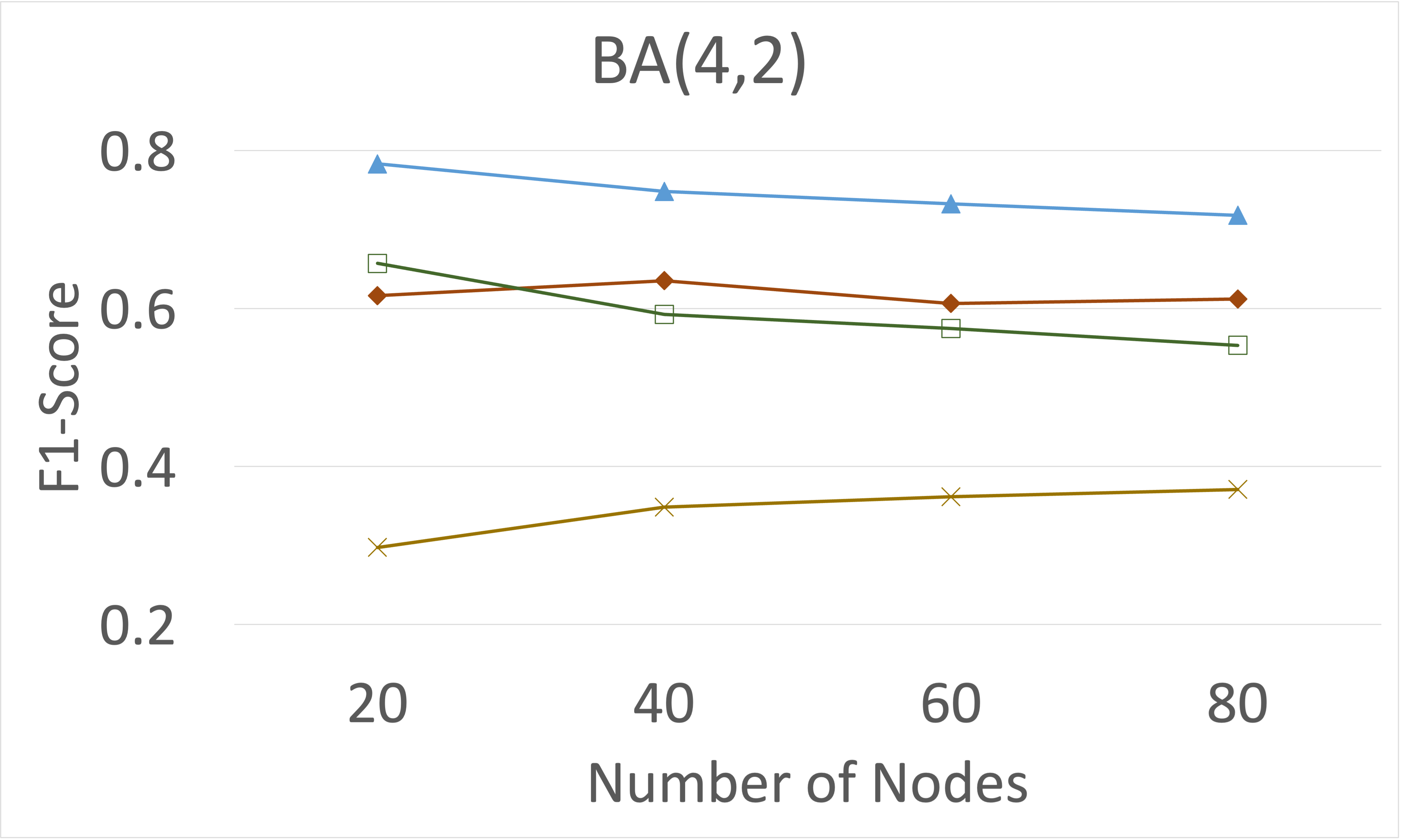}
    \caption{Additive Noise Model, $T=1000$}
  \end{subfigure}
  \centering
  \begin{subfigure}{0.49\textwidth}
    \centering
    \includegraphics[width=0.49\textwidth, height=0.35\textwidth]{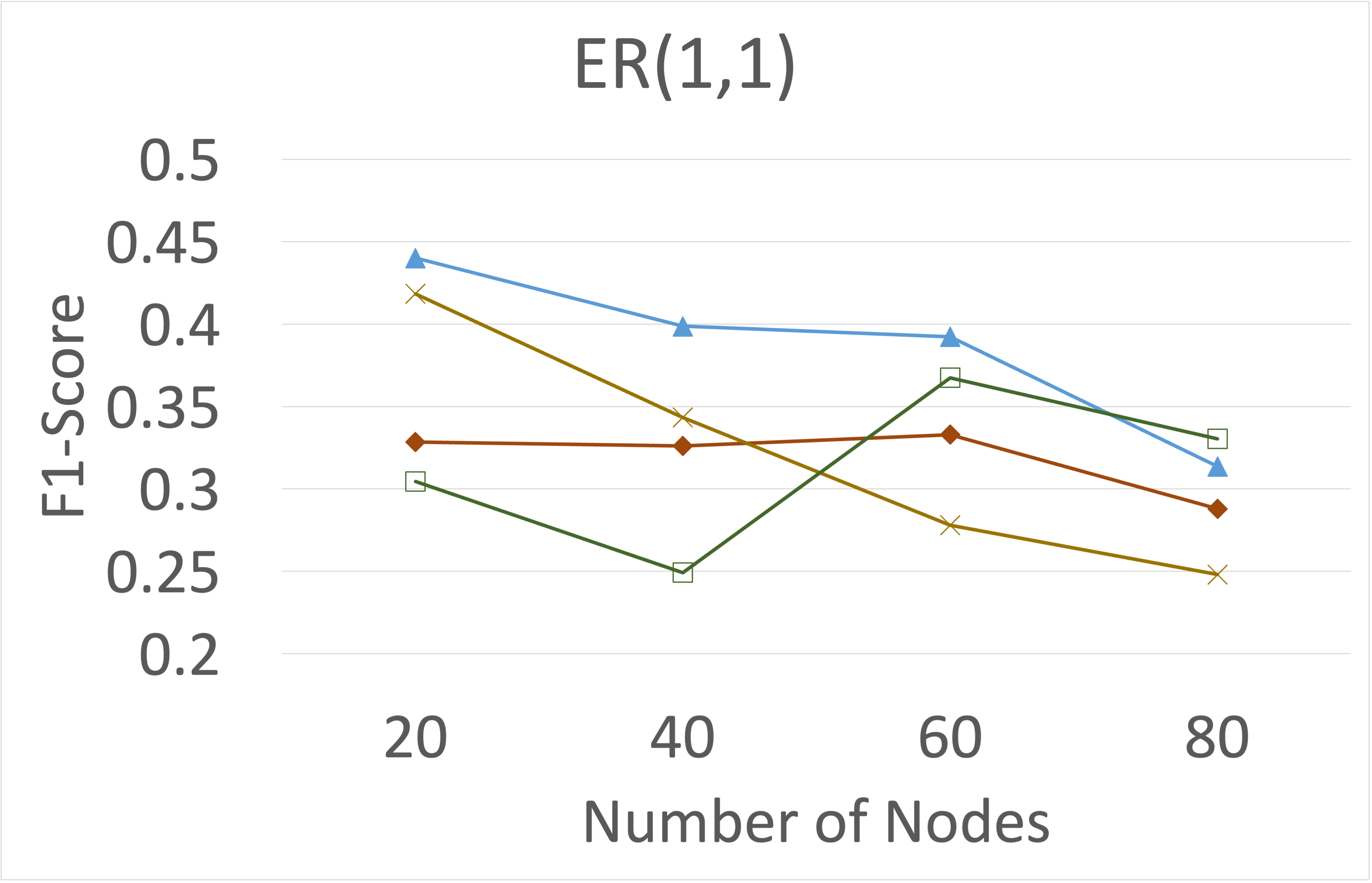}
    \includegraphics[width=0.49\textwidth, height=0.35\textwidth]{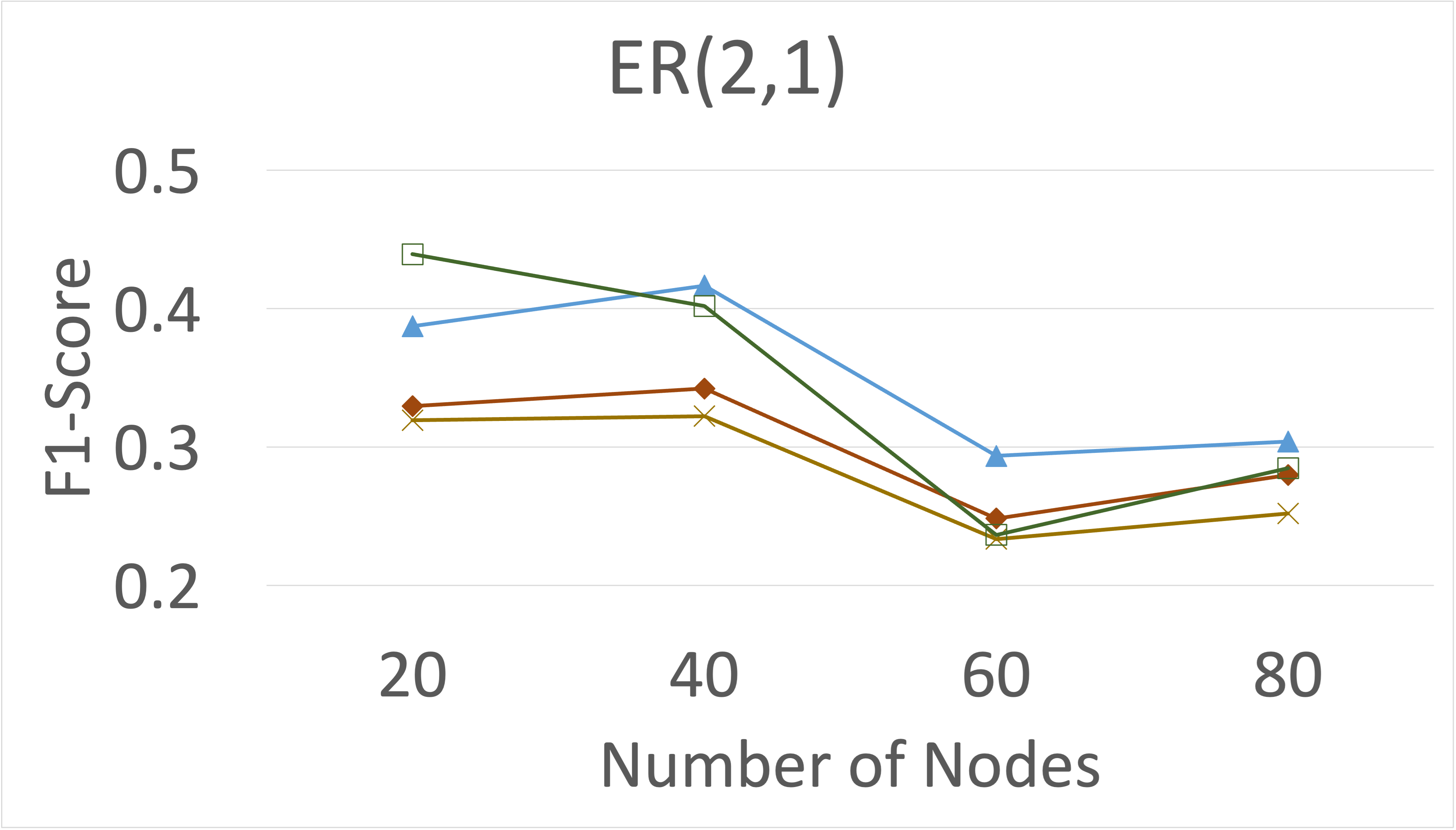}
    \includegraphics[width=0.49\textwidth, height=0.35\textwidth]{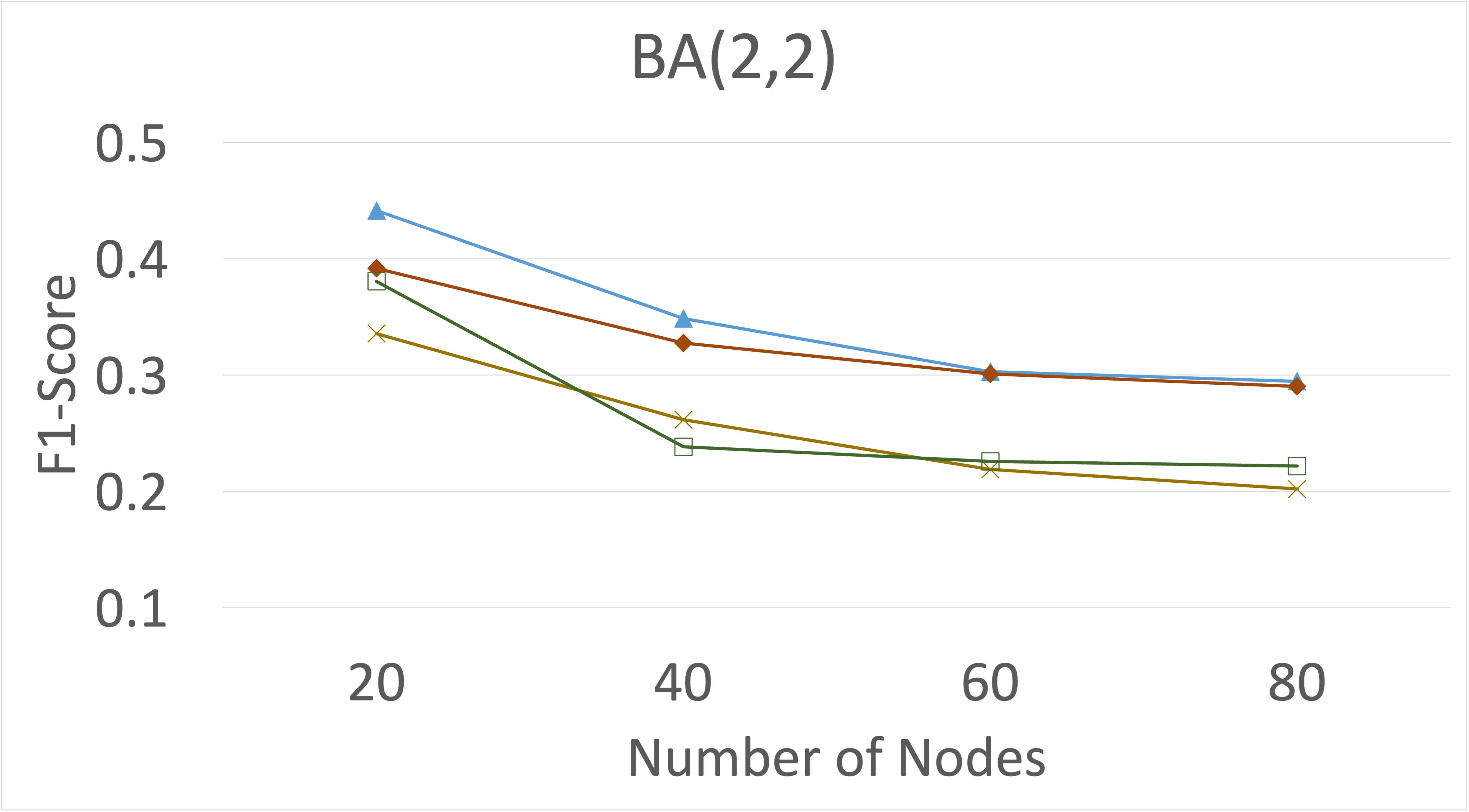}
    \includegraphics[width=0.49\textwidth, height=0.35\textwidth]{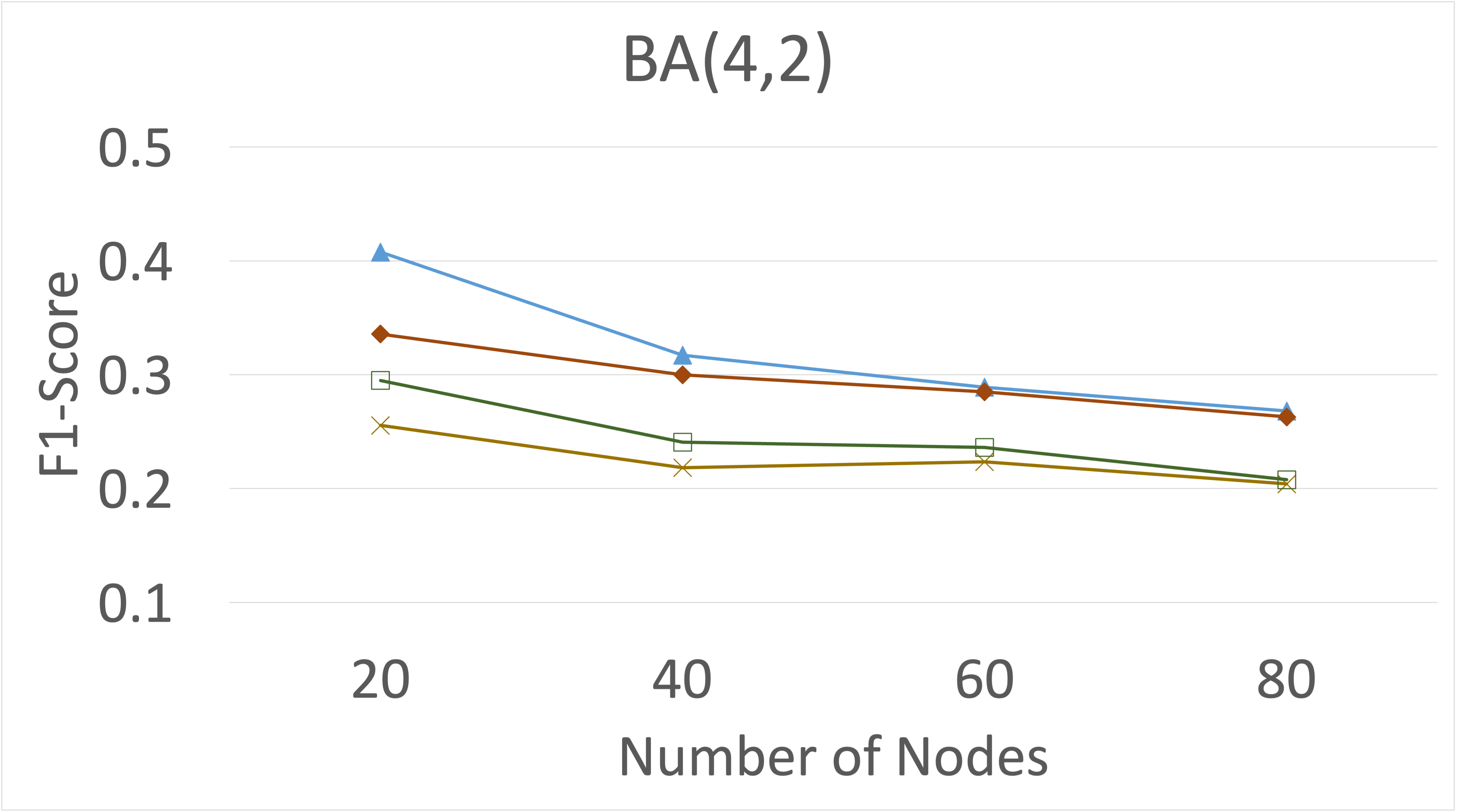}
    \caption{Generalized Linear Model with Poisson
Distribution, $T=200$}
  \end{subfigure}
  \hfill
  \centering
  \begin{subfigure}{0.49\textwidth}
    \centering
    \includegraphics[width=0.49\textwidth, height=0.35\textwidth]{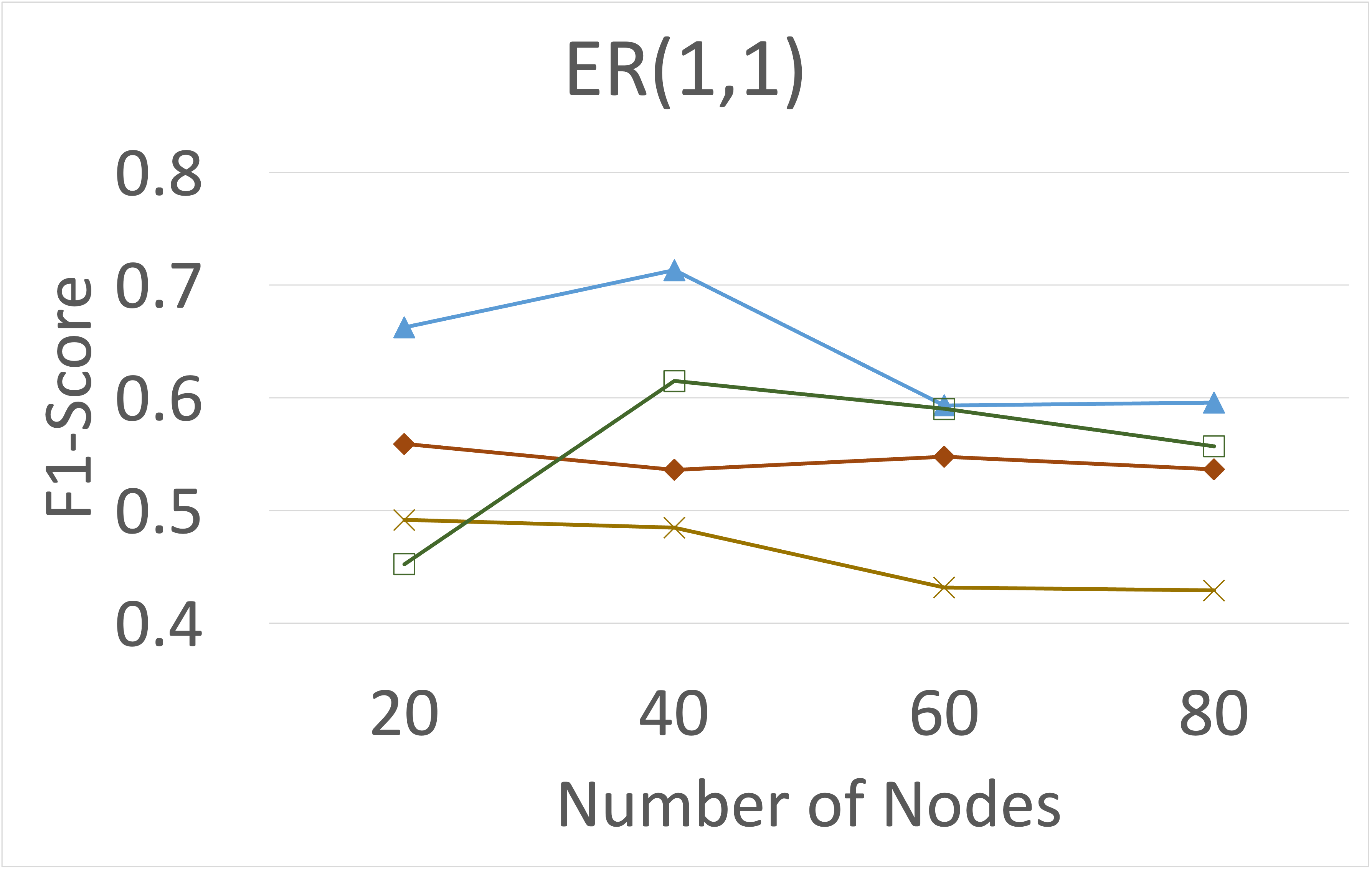}
    \includegraphics[width=0.49\textwidth, height=0.35\textwidth]{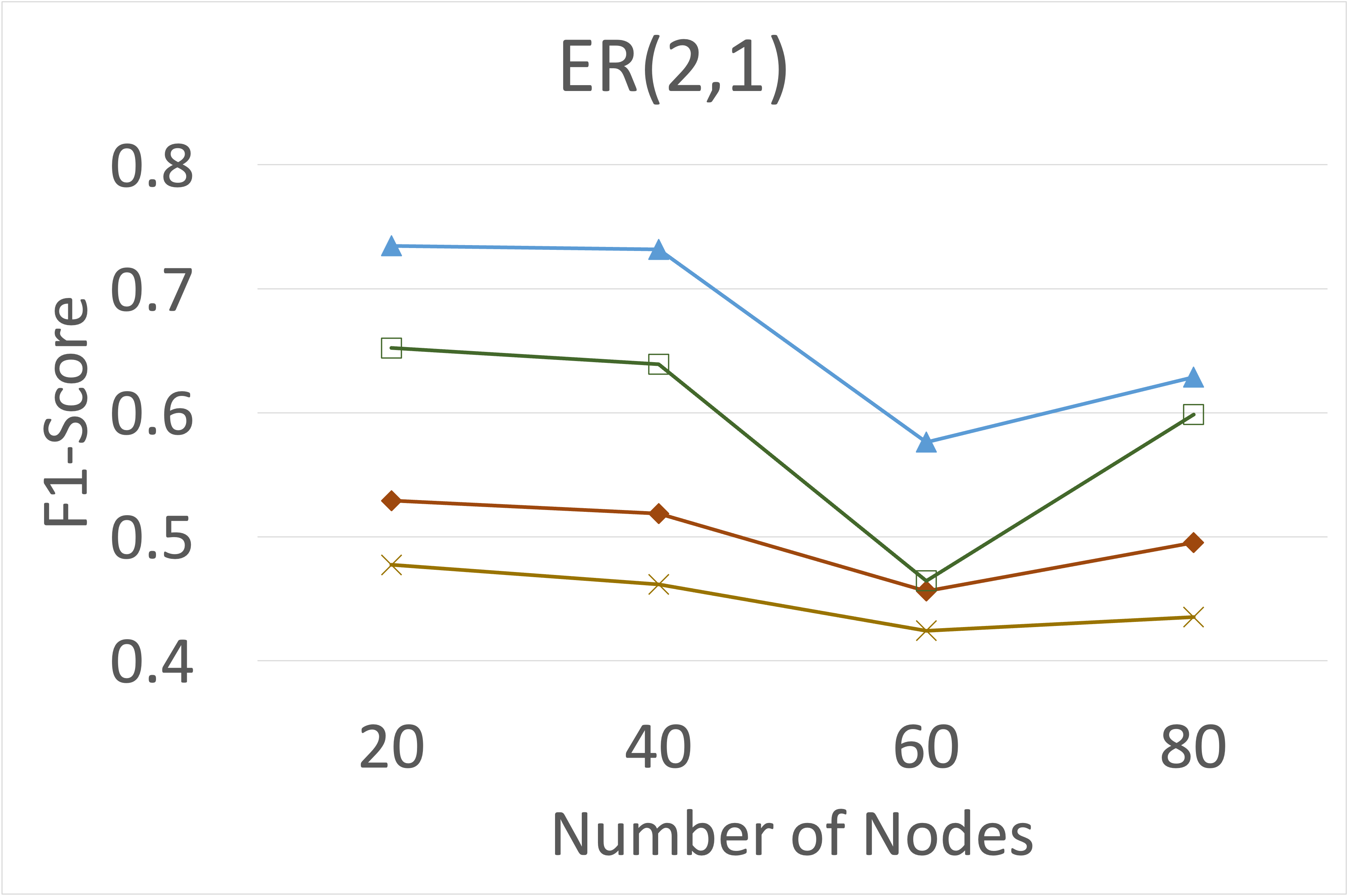}
    \includegraphics[width=0.49\textwidth, height=0.35\textwidth]{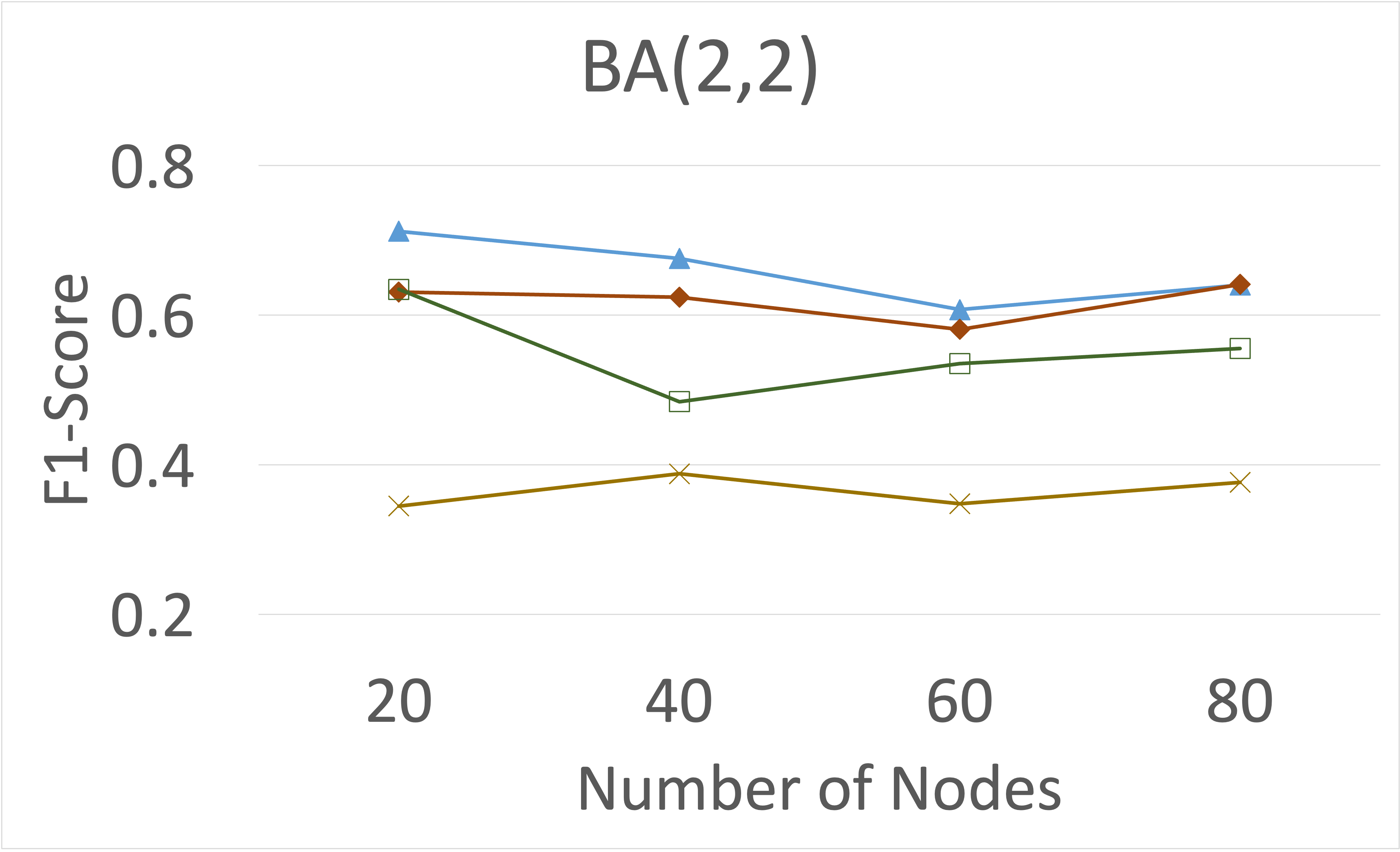}
    \includegraphics[width=0.49\textwidth, height=0.35\textwidth]{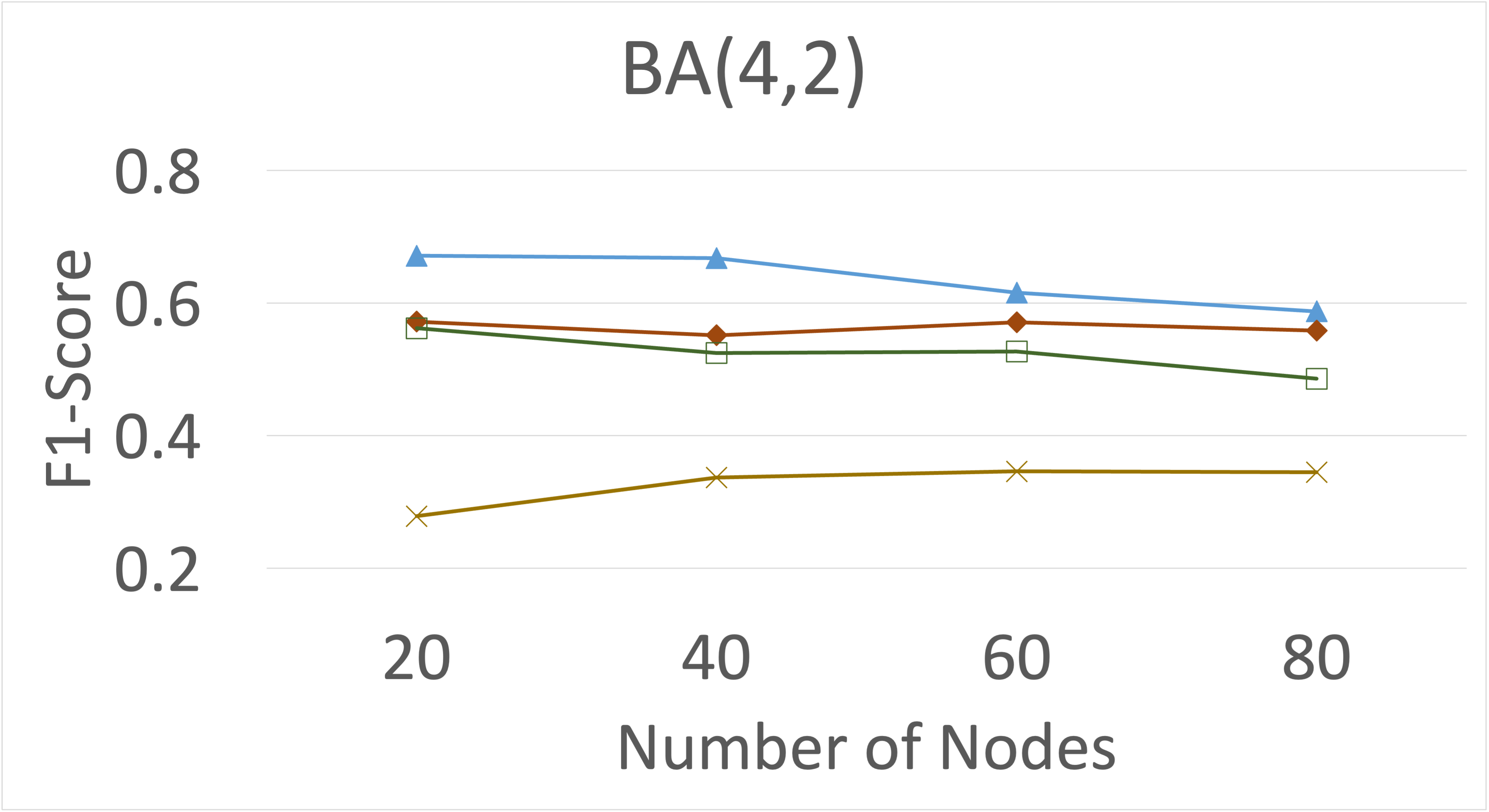}
    \caption{Generalized Linear Model with Poisson
Distribution, $T=1000$}
  \end{subfigure}
  \centering
  \begin{subfigure}{1.0\textwidth}
    \centering
    \includegraphics[width=0.5\textwidth, height=0.03\textwidth]{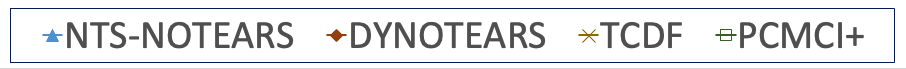}
  \end{subfigure}
  \caption{\textbf{Mean F1-scores} over 10 datasets for each setting with simulated data. Higher F1-score is better. The number of lags $= 3$. ER(2,1) denotes that the ground-truth DAGs are sampled using ER scheme with an intra-slice mean out-degree equal to 2 and inter-slice mean out-degree equal to 1. NTS-NOTEARS achieves the highest F1-scores in the vast majority of the settings.}
  \label{fig:experiments_simulated_f1_score}
\end{figure*}

\paragraph{Results} 
Figure~\ref{fig:experiments_simulated_f1_score} shows the F1-score depending on
sequence length $T$, SCM, mean out-degrees and graph size.
All methods pass the sanity check of improving with more data. 
{\em NTS-NOTEARS achieves the highest F1-scores in 20 out of 24 settings.}
The score of NTS-NOTEARS is much better than that of TCDF, which shows the strength of the NOTEARS approach of defining edge weights over the attention-based approach used by TCDF. 
On other metrics (e.g., SHD), we also find that NTS-NOTEARS scores the best; please see Appendix~\ref{section_simulated_more_results} for detailed results.


\subsubsection{Running Time}\label{section_running_time}

\begin{figure}[ht]
  \centering
  \includegraphics[width=0.5\textwidth]{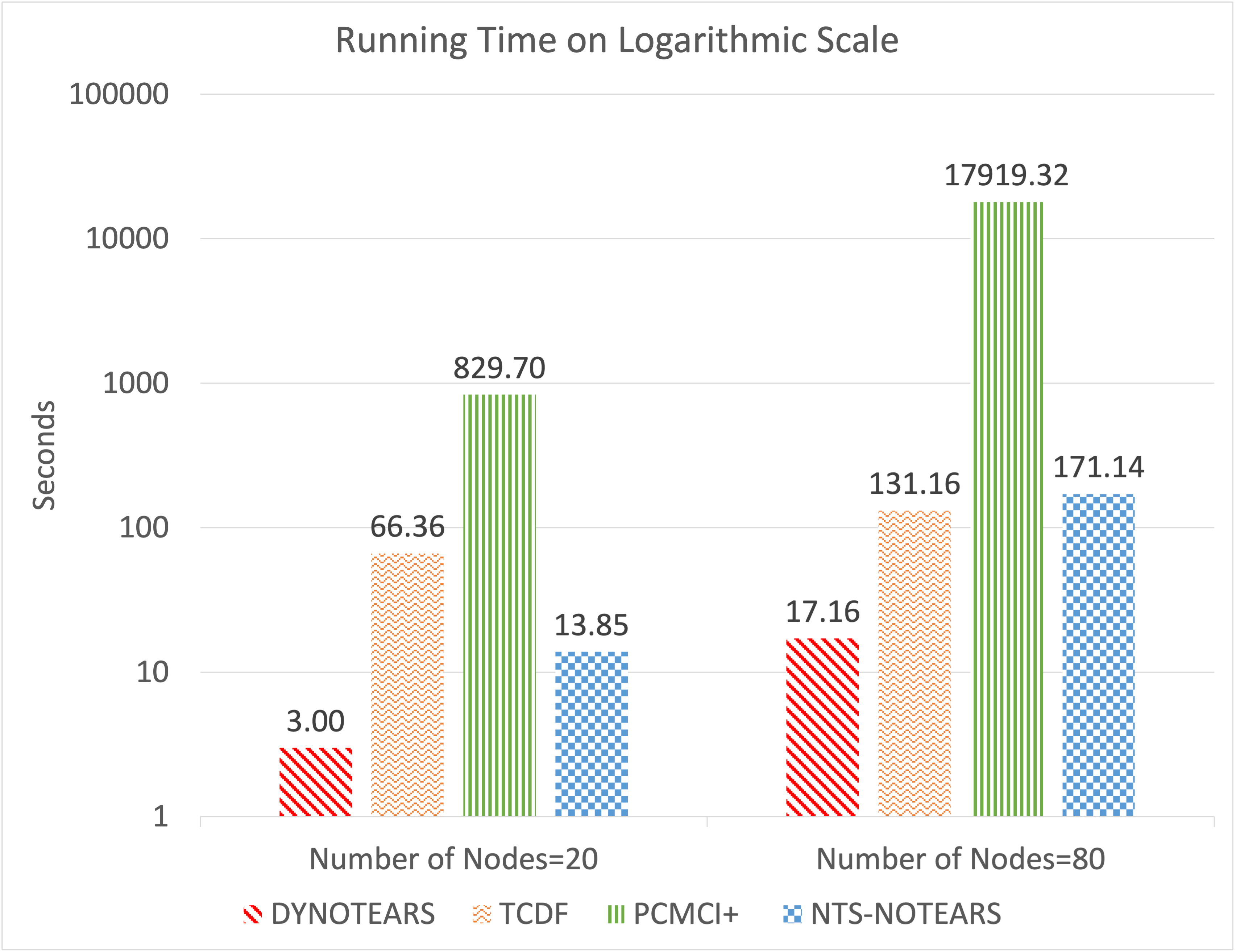}
  \caption{The average running time over 10 datasets measured in seconds with additive noise model, $K=3$, $T=1000$ and ER(2,1). The heights of bars are on a logarithmic scale. }
  \label{fig:running_time}
\end{figure}

Figure~\ref{fig:running_time} compares the average running time of the evaluation methods over 10 datasets. The neural networks and the GPDC CI test are accelerated by the same GPU. 
The linear method DYNOTEARS is the fastest. The constraint-based method PCMCI+ with the nonlinear GPDC CI test is substantially slower than the neural-based methods such as TCDF and NTS-NOTEARS. NTS-NOTEARS therefore offers a sweet spot trade-off between speed and learning performance. 


\subsection{Benchmark Data: Lorenz 96 \& fMRI}\label{section_lorenz_fmri}

Lorenz 96~\citep{lorenz1996predictability} and fMRI~\citep{smith2011network}
are two common benchmarks to evaluate causal discovery algorithms with nonlinear time-series data and nonlinear Granger causality algorithms~\citep{nauta2019causal, monti2020causal, marcinkevics2021interpretable, Tank_2021, khanna2019economy}.
The Lorenz 96 model is popular in climate science as a testbed for chaotic behaviors~\citep{schneider2017earth}. The data follows the nonlinear dynamics given by:
\[ \frac{d{x^{t+1}_i}}{dt} = (x^t_{i+1}-x^t_{i-2}) \cdot x^t_{i-1} - x^t_i+F \]
where F controls the chaoticity of the system. Similar to previous work~\citep{marcinkevics2021interpretable,khanna2019economy}, we consider two settings where $F \in \{10, 40\}$. The fMRI benchmark contains rich, realistic simulated blood-oxygen-level-dependent time-series for modelling brain networks. Each node in the network represents a region of interest in the brain. 
The ground-truth DAG in each benchmark has 2 time steps 
(see Figure~\ref{fig:lorenz96_fmri_color_maps} in the appendix).


In Table~\ref{tab:lorenz96_fmri_f1_score}, we evaluate the methods and report their mean F1-scores and standard errors (SE) with 5 datasets sampled from the Lorenz 96 benchmark where each dataset has $d=20$ and $T=500$, and 10 datasets sampled from the fMRI benchmark where each dataset has $d=5$ and $T\in\{200,1200,5000\}$.
Please see Appendix~\ref{section_hyperparameters_benchmarks} for hyperparameter values. 
\emph{NTS-NOTEARS achieves the best F1-scores with both benchmarks, by a margin of more than 10\%.} NTS-NOTEARS also achieves the best scores for other metrics (please see Appendix~\ref{section_benchmark_more_results}).

Similar to~\citet{marcinkevics2021interpretable, khanna2019economy}, in Table~\ref{tab:lorenz96_fmri_auc} we also report the area under the receiver operating characteristic curve (AUROC) by varying the hyperparameters of each method. {\em NTS-NOTEARS achieves the best AUROC with both benchmarks.}

\begin{table}[h]
\centering
\caption{\textbf{Mean F1-scores ($\pm$ SE)} computed with Lorenz 96 and fMRI benchmarks.}
    \begin{tabular}{l|c|c}
    \toprule
    Method & Lorenz 96 & fMRI \\
    \midrule
    DYNOTEARS & 0.855 ($\pm$ 0.016) & 0.475 ($\pm$ 0.020) \\
    \midrule
    TCDF & 0.459 ($\pm$ 0.017) & 0.347 ($\pm$ 0.059) \\
    \midrule
    PCMCI+ & 0.637 ($\pm$ 0.028) & 0.502 ($\pm$ 0.045) \\
    \midrule
    NTS-NOTEARS & \textbf{0.996 ($\pm$ 0.002)} & \textbf{0.628 ($\pm$ 0.023)} \\
    \bottomrule
    \end{tabular}
    \label{tab:lorenz96_fmri_f1_score}
\end{table}

\begin{table}[h]
\centering
\caption{\textbf{AUROC} computed with Lorenz 96 and fMRI benchmarks by varying the hyperparameters of each evaluation method.}
    \begin{tabular}{l|c|c}
    \toprule
    Method & Lorenz 96 & fMRI \\
    \midrule
    DYNOTEARS & 0.788 & 0.708 \\
    \midrule
    TCDF & 0.585 & 0.612 \\
    \midrule
    PCMCI+ & 0.706 & 0.743 \\
    \midrule
    NTS-NOTEARS & \textbf{0.811} & \textbf{0.749} \\
    \bottomrule
    \end{tabular}
    \label{tab:lorenz96_fmri_auc}
\end{table}

\subsection{Real-World Ice Hockey Data}\label{section_ice_hockey}

We apply NTS-NOTEARS to real-world data collected by Sportlogiq from ice hockey games in the 2018-2019 NHL season. The dataset contains a mixture of continuous, binary and categorical variables. The ground-truth distribution of each variable is unspecified. 
Please see Appendix~\ref{section_ice_hockey_appendix} for data description.
Since the play restarts after a goal is scored (i.e. face-off), we incorporate the prior knowledge that forbids edges coming from \textit{goal(t)} or \textit{goal(t-1)}.
Because DYNOTEARS does not provide a way to incorporate prior knowledge, we manually remove any outgoing edges coming from \textit{goal(t)} or \textit{goal(t-1)}. We set the DYNOTEARS hyperparameters so that both methods produce the same number of edges for comparability (see Appendix~\ref{section_ice_hockey_appendix}). The estimated DBNs capture many meaningful relationships between variables. An interesting question to ask in ice hockey is ``what contributes to a goal?''~\citep{sun2020cracking, 
schulte2017apples}.
 By identifying the parent nodes of \textit{goal(t)} in the DBN estimated by NTS-NOTEARS, we can answer the question: the preceding shot, the duration of the shot, the distance between the shot and the net (i.e. \textit{xAdjCoord(t-1)}), the manpower situation and the velocity of the puck are important for scoring a goal. However, due to nonlinearity, DYNOTEARS fails to identify several goal contributors such as the duration of the shot, the distance between the shot and the net (i.e. \textit{xAdjCoord(t-1)}), and the manpower situation. NTS-NOTEARS captures them all. Please see Figure~\ref{fig:ice_hockey_colourmaps} in the appendix for the learned DBNs.

\section{CONCLUSION}\label{section_conclusion}

This paper described NTS-NOTEARS for learning nonparametric DBNs, a score-based structure learning method using 1D CNNs for time-series data, either with or without prior knowledge of dependencies. The learned DBNs capture both inter-slice and intra-slice dependencies.
The system is user-friendly in that it supports both continuous and discrete data, and does not require  knowledge of independence tests or parametric data generation models. We showed how to adapt the NOTEARS continuous optimization strategy~\citep{zheng2018dags} for 1D CNNs, which allows us to learn intra-slice edges with an acyclicity constraint. Based on simulated data and standard benchmarks, we show the superior DBN structure learning quality and running speed of NTS-NOTEARS compared to several comparison methods, and demonstrate the advantage of providing prior knowledge using optimization constraints. We also apply the NTS-NOTEARS to a complex real-world sports dataset that contains a mixture of continuous and discrete variables without knowing the ground-truth underlying data distribution. 
A next step for future work is to extend NTS-NOTEARS to causal modelling, in particular to causal graph learning in the presence of latent confounders. 

\subsubsection*{Acknowledgements}
This work was supported by a Discovery Grant for Oliver Schulte from the Natural Sciences and Engineering Research Council of Canada. 

\clearpage
\bibliography{references}

\appendix
\onecolumn

\section{OTHER METHODS FOR TIME SERIES DATA}\label{section_omitted_methods_from_evaluation} 

The comparison methods in our experiments come from the same model class as NTS-NOTEARS: temporal graphs with both intra-slice and inter-slice dependencies. In this section we discuss other methods for time series data, with an emphasis on neural methods (see Table~\ref{tab:method_comparison}).

\paragraph{Models without intra-slice edges.}
The neural method cMLP~\citep{Tank_2021} does not have the capability to learn intra-slice edges. On datasets where the ground-truth model comprises only inter-slice edges, its performance is competitive with NTS-NOTEARS (e.g., worse F1-score on Lorenz, better F1-score on fMRI). 
GVAR~\citep{marcinkevics2021interpretable} also does not learn intra-slice edges, and does not explicitly model different lags.
Economy-SRU~\citep{khanna2019economy} does not estimate edges over multiple lags.

\paragraph{Other Methods with Intra-Slice Edges.}

We conducted experiments using linear methods such as VAR-LINGAM~\citep{hyvarinen2010estimation} and PCMCI+ with linear CI test - partial correlation test (ParCorr)~\citep{runge2019detecting}. We exclude their results because their performance is poor for the nonlinear datasets in our experiments. 

Among constraint-based methods, LPCMCI~\citep{gerhardus2020high} focuses on latent confounders and outputs a PAG.
With nonlinear CI test, LPCMCI is computationally too expensive to be compared when the number of nodes is large.
CMIknn and CMIsymb~\citep{runge2019detecting} are nonlinear CI tests based on conditional mutual information. Although the two CI tests are nonparametric, they are computationally expensive when the number of nodes is large, which makes them infeasible to be included in the experiments. 

IDYNO~\citep{gao2022idyno} matches NTS-NOTEARS in terms of the properties listed in Table~\ref{tab:method_comparison}. Their work was published after we  posted the arxiv version of this paper. The main difference is that IDYNO focuses on interventional data, with only one experiment on observational data with linear relationships. Other differences include the following:

\begin{enumerate}
    \item IDYNO uses a generalized linear model. We use a nonparametric CNN to exploit the sequence topology. 
    \item Like DYNOTEARS, IDYNO must concatenate the data. So the original data with dimension $T \times d$ becomes $T \times d \times (K+1)$. Larger data size may cause memory problem as well as slowing down training. 
The CNN allows NTS-NOTEARS to use the original data without data concatenation.
\item IDYNO uses 3 MLPs for each target variable. So there are 3 times more neural nets to train compared to NTS-NOTEARS.
\end{enumerate}


As no code is available for IDYNO, we used our own implementation\footnote{https://github.com/xiangyu-sun-789/IDYNO\_reproduce}. 
We used the same methodology described in the paper to evaluate IDYNO with the Lorenz 96 and fMRI benchmarks.
IDYNO is less accurate than NTS-NOTEARS, by about a factor of 2 in F1-score and SHD. IDYNO is also slower than NTS-NOTEARS. For example, with $d=5$, $K=1$ and $T \in \{200, 1200, 5000\}$ in the fMRI benchmark, IDYNO is about 7 times slower than NTS-NOTEARS. IDYNO is even slower with larger $K$, because it means more data to concatenate and to put in the memory.

\section{PROOF OF THEOREM~\ref{theorem_independence}}\label{proof_independence}

\begin{proof}
To show $F=F_0$, we will show that $F_0 \subseteq F$ and $F \subseteq F_0$.

We have:
\begin{gather*}
F = \{ f | f(X) = \mathit{CNN}(X; C^{(1)}, \dots, C^{(h_c)}, A^{(1)}, \dots, A^{(h_a)}), f \text{ is independent of } X_i^k \} \\
\end{gather*}
and 
\begin{gather*}
F_0 = \{ f | f(X) = \mathit{CNN}(X; C^{(1)}, \dots, C^{(h_c)}, A^{(1)}, \dots, A^{(h_a)}), C_{i,b}^{(1),k} = 0, \forall b=\{1, \dots, m\} \} \\
\end{gather*}
where $C^{(u)}$ is the kernel weights on the $u$-th CNN layer, $C_{i,b}^{(1),k}$ is the first-layer kernel weights in the $b$-th kernel connecting to input variable $X_i^k$, and $A^{(u)}$ is the weights on the $u$-th MLP layer. The bias terms are omitted as they do not affect the proof.

Also, 
\begin{gather*}
\mathit{CNN}(X; C^{(1)}, \dots, C^{(h_c)}, A^{(1)}, \dots, A^{(h_a)}) = \sigma(A^{(h_a)} \ast \sigma( \dots A^{(1)} \ast \sigma(C^{(h_c)} \circ \sigma(\dots \sigma(C^{(1)} \circ X))))) \\
\end{gather*}
where $\ast$ is matrix product, $\circ$ is the convolution operation of two matrices and $\sigma$ is the activation functions.

(1) To show $F_0 \subseteq F$: 

For any $f_0 \in F_0$, we have $f_0(X) = \mathit{CNN}(X; C^{(1)}, \dots, C^{(h_c)}, A^{(1)}, \dots, A^{(h_a)})$ where $C_{i,b}^{(1),k} = 0$ for all $ b=\{1, \dots, m\}$. Therefore, $C^{(1)} \circ X$ is independent of $X_i^k$. Therefore, $f_0(X) = \sigma(A^{(h_a)} \ast \sigma( \dots A^{(1)} \ast \sigma(C^{(h_c)} \circ \sigma(\dots \sigma(C^{(1)} \circ X)))))$ is also independent of $X_i^k$. Hence, $f_0 \in F$.

(2) To show $F \subseteq F_0$: 

For any $f \in F$, we have $f(X) = \mathit{CNN}(X; C^{(1)}, \dots, C^{(h_c)}, A^{(1)}, \dots, A^{(h_a)})$ and $f$ is independent of $X_i^k$. Let $\tilde{X}$ be identical to $X$ except $\tilde{X}_{i}^k = 0$. $f$ is independent of $X_i^k$, similarly, $f$ is independent of $\tilde{X}_i^k$. Therefore,
\begin{equation}
\begin{gathered}\label{CNN_proof_2}
f(X) = f(\tilde{X}) = \mathit{CNN}(\tilde{X}; C^{(1)}, \dots, C^{(h_c)}, A^{(1)}, \dots, A^{(h_a)}) \\
= \sigma(A^{(h_a)} \ast \sigma( \dots A^{(1)} \ast \sigma(C^{(h_c)} \circ \sigma(\dots \sigma(C^{(1)} \circ \tilde{X}))))) \\
\end{gathered}
\end{equation}

Let $\tilde{C}^{(1)}$ be identical to $C^{(1)}$ except $\tilde{C}_{i,b}^{(1),k} = 0$ for all $b = \{ 1, \dots, m\}$. Let $C_{(b)}^{(1)}$ be the first-layer kernel weights of the $b$-th kernel. We have:
\begin{gather*}
C_{(b)}^{(1)} \circ \tilde{X} = \sum_{k'=1}^{K+1} \sum_{i'=1}^{d} C_{i',b}^{(1),k'} \cdot \tilde{X}_{i'}^{k'} \\
= (\sum_{k' \neq k} \sum_{i' \neq i} C_{i',b}^{(1),k'} \cdot \tilde{X}_{i'}^{k'}) + (\sum_{i' \neq i} C_{i',b}^{(1),k} \cdot \tilde{X}_{i'}^k) + (\sum_{k' \neq k} C_{i,b}^{(1),k'} \cdot \tilde{X}_{i}^{k'}) + C_{i,b}^{(1),k} \cdot \tilde{X}_{i}^k \\
= (\sum_{k' \neq k} \sum_{i' \neq i} C_{i',b}^{(1),k'} \cdot X_{i'}^{k'}) + (\sum_{i' \neq i} C_{i',b}^{(1),k} \cdot X_{i'}^k) + (\sum_{k' \neq k} C_{i,b}^{(1),k'} \cdot X_{i}^{k'}) + C_{i,b}^{(1),k} \cdot \tilde{X}_{i}^k \\
= (\sum_{k' \neq k} \sum_{i' \neq i} C_{i',b}^{(1),k'} \cdot X_{i'}^{k'}) + (\sum_{i' \neq i} C_{i',b}^{(1),k} \cdot X_{i'}^k) + (\sum_{k' \neq k} C_{i,b}^{(1),k'} \cdot X_{i}^{k'}) + 0 \\
= (\sum_{k' \neq k} \sum_{i' \neq i} C_{i',b}^{(1),k'} \cdot X_{i'}^{k'}) + (\sum_{i' \neq i} C_{i',b}^{(1),k} \cdot X_{i'}^k) + (\sum_{k' \neq k} C_{i,b}^{(1),k'} \cdot X_{i}^{k'}) + \tilde{C}_{i,b}^{(1),k} \cdot X_i^k \\
= (\sum_{k' \neq k} \sum_{i' \neq i} \tilde{C}_{i',b}^{(1),k'} \cdot X_{i'}^{k'}) + (\sum_{i' \neq i} \tilde{C}_{i',b}^{(1),k} \cdot X_{i'}^k) + (\sum_{k' \neq k} \tilde{C}_{i,b}^{(1),k'} \cdot X_{i}^{k'}) + \tilde{C}_{i,b}^{(1),k} \cdot X_i^k \\
\sum_{k'=1}^{K+1} \sum_{i'=1}^{d} \tilde{C}_{i',b}^{(1),k'} \cdot X_{i'}^{k'} = \tilde{C}_{(b)}^{(1)} \circ X \\
\end{gather*}

Therefore, $C^{(1)} \circ \tilde{X} = \tilde{C}^{(1)} \circ X$. From Equation~\eqref{CNN_proof_2}, we have:
\[ f(X) = \sigma(A^{(h_a)} \ast \sigma( \dots A^{(1)} \ast \sigma(C^{(h_c)} \circ \sigma(\dots \sigma(\tilde{C}^{(1)} \circ X))))) = \mathit{CNN}(X; \tilde{C}^{(1)}, \dots, C^{(h_c)}, A^{(1)}, \dots, A^{(h_a)}) \in F_0 \]

Hence, $f \in F_0$

\end{proof}

\section{SIMULATION DETAILS}\label{section_simulation_details}

\begin{figure}[h]
  \centering
  \includegraphics[scale=0.4]{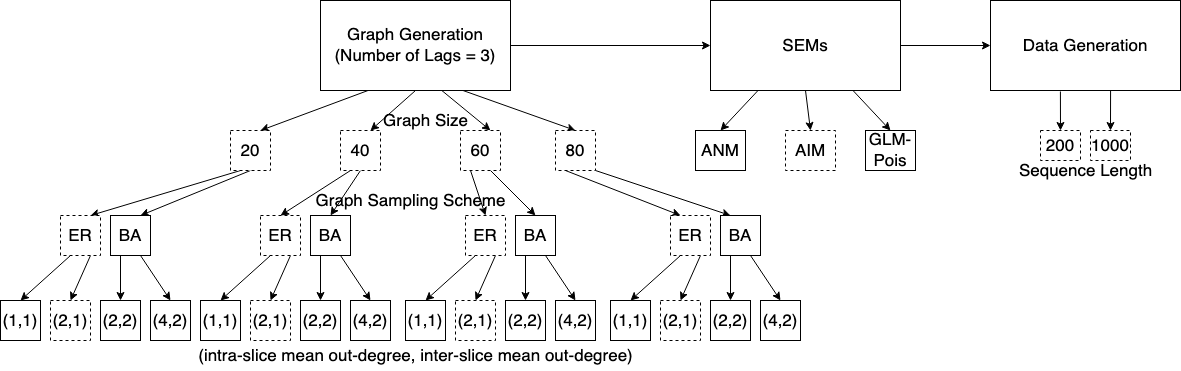}
  \caption{Visualization of the process for generating simulated training data. Besides generating the training data, the dashed boxes also indicate how the validation data was generated. A total of 48 DBNs and 96 training datasets were generated.}
  \label{fig:simulation_flowchart}
\end{figure}

Given a graph generated by either an ER or a BA scheme, we simulate data according to one of the three identifiable SCMs:
\begin{itemize}
  \item Additive Noise Model (ANM)~\citep{peters2017elements}: $X_j^t = f_j(\mathit{PA}(X_j^t) \cdot \theta_1) \cdot \theta_2 + Z_j^t$, where $f_j$ is the sigmoid function.
  \item Additive Index Model (AIM)~\citep{yuan2011identifiability, alquier2013sparse}: $X_j^t = Z_j^t + \sum_{m=1}^{3} h_m(\mathit{PA}(X_j^t) \cdot \theta_m)$, where $h_1=\mathit{tanh}$, $h_2=\mathit{cos}$, $h_3=\mathit{sin}$.
  \item Generalized Linear Model with Poisson Distribution (GLM-Pois)~\citep{park2019high}: $X_j^t = \mathit{Pois}(g_j(\mathit{PA}(X_j^t) \cdot \theta_1) + \phi)$, where $g_j=\mathit{tanh}$ and $\phi$ is sampled uniformly from range $[1, 3]$.
\end{itemize}
$\mathit{PA}(X_j^t)$ denotes the parents of $X_j^t$. Each $Z_j$ is a standard Gaussian noise. Each $\theta$ is sampled uniformly from range $[-2, -0.5] \cup [0.5, 2]$.

\section{METHOD HYPERPARAMETERS FOR SIMULATED DATA}\label{section_hyperparameters_simulated}

Regarding hyperparameter searching, for compatibility purpose with constraint-based baseline PCMCI+~\citep{runge2020discovering}, we use validation sets to find hyperparameter values.
We perform an extensive grid search on the hyperparameters of each method to find the sets of hyperparameters that give the best F1-scores for each method with the validation sets. 

%
%


\begin{itemize}
  
  \item NTS-NOTEARS
  \begin{itemize}
    \item $\bm{\lambda_1} \in \{ \bm{0.01}, \bm{0.001} \}$ for $T \in \{200, 1000\}$, respectively.
    \item $\lambda_2 = 0.05$
    \item $K = \textit{number of lags}$
    \item $m=d$
    \item the number of hidden layers $=1$
    \item $\bm{W}_{\textit{thres}} = \bm{0.3}$
  \end{itemize}


  \item PCMCI+
  \begin{itemize}
    \item CI test: Gaussian process regression plus distance correlation test (GPDC)
    \item $ \tau_{\textit{min}} = 0 $
    \item $ \tau_{\textit{max}} =$ \textit{number of lags}
    \item $ \alpha \in \{0.01, 0.05\}$ for $T \in \{500, 2000\}$, respectively
  \end{itemize}

  \item TCDF
  \begin{itemize}
    \item $\textit{significance} = 0.8$
    \item $\textit{learning rate} = 0.001$
    \item $\textit{epochs} = 1000$
    \item $\textit{levels} = 2$
    \item $\textit{kernel size} = \textit{number of lags} + 1$
    \item $\textit{dilation coefficient} = \textit{number of lags} + 1$
  \end{itemize}


  \item DYNOTEARS
  \begin{itemize}
    \item $ \lambda_{a} = \lambda_{w} = 0.1 $
    \item $p = \textit{number of lags}$
    \item $\textit{weight threshold} = 0.01$
  \end{itemize}
  
  
\end{itemize}

\section{MORE RESULTS WITH SIMULATED DATA}\label{section_simulated_more_results}

\begin{figure*}[ht]
  \centering
  \begin{subfigure}{0.49\textwidth}
    \centering
    \includegraphics[width=0.49\textwidth, height=0.35\textwidth]{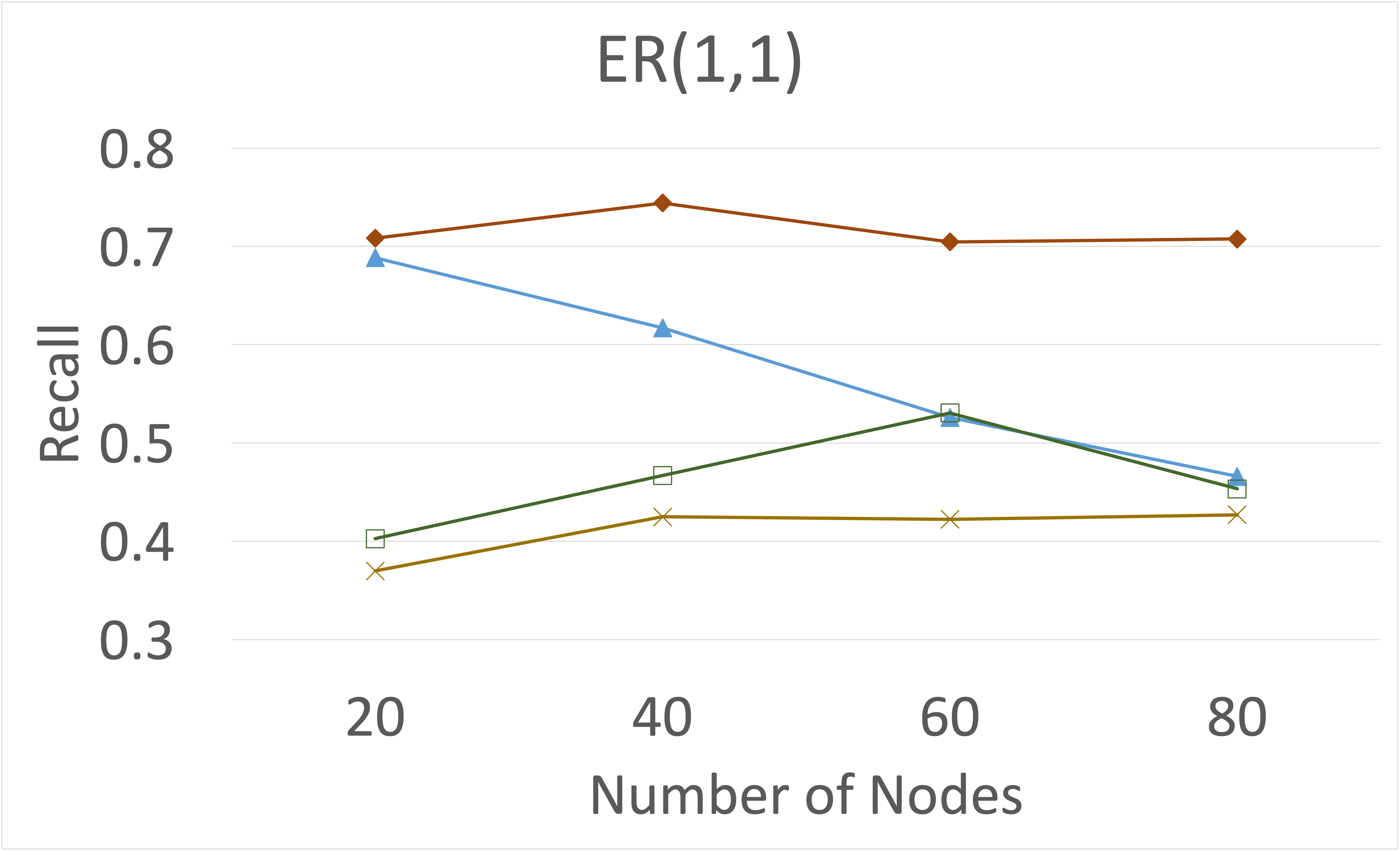}
    \includegraphics[width=0.49\textwidth, height=0.35\textwidth]{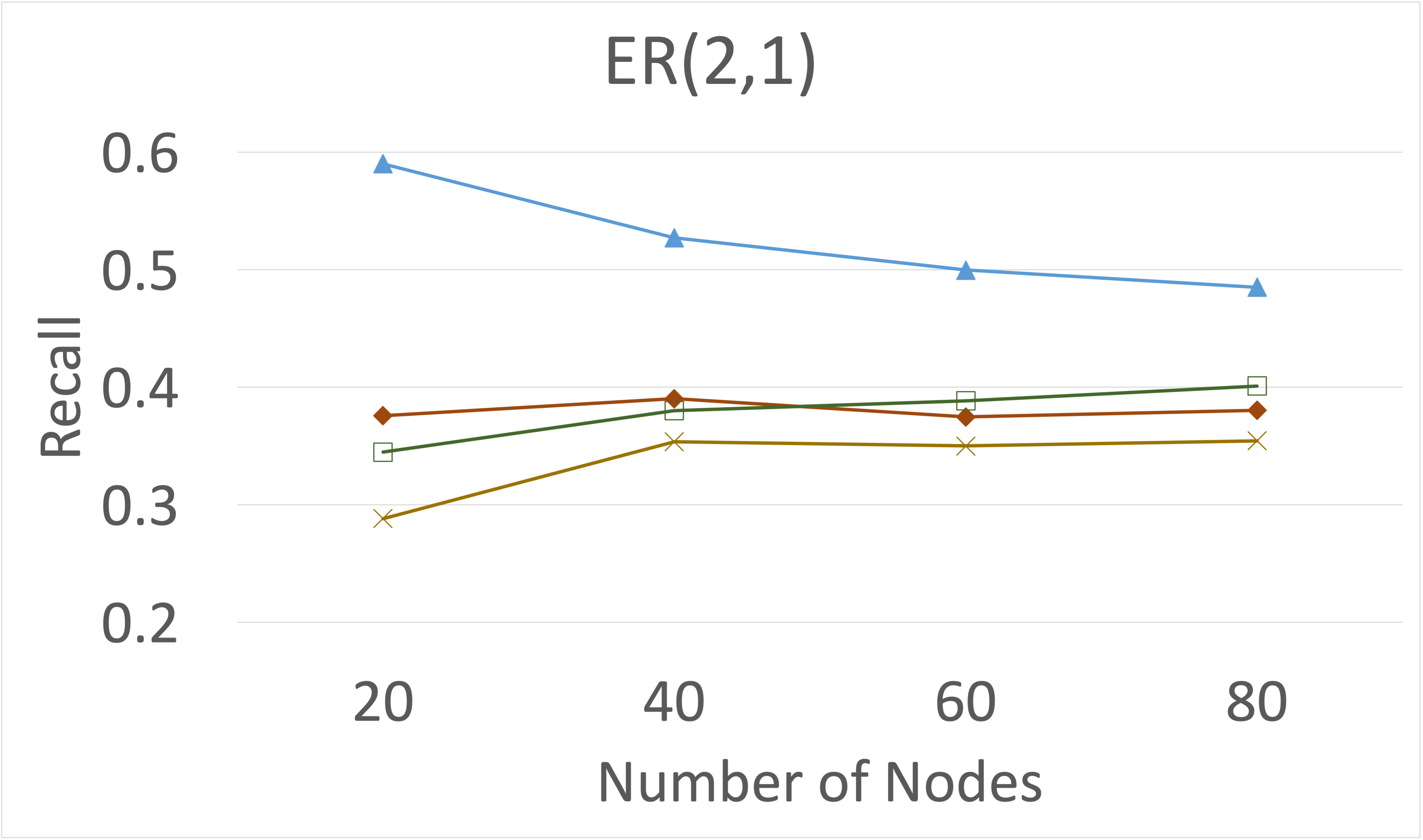}
    \includegraphics[width=0.49\textwidth, height=0.35\textwidth]{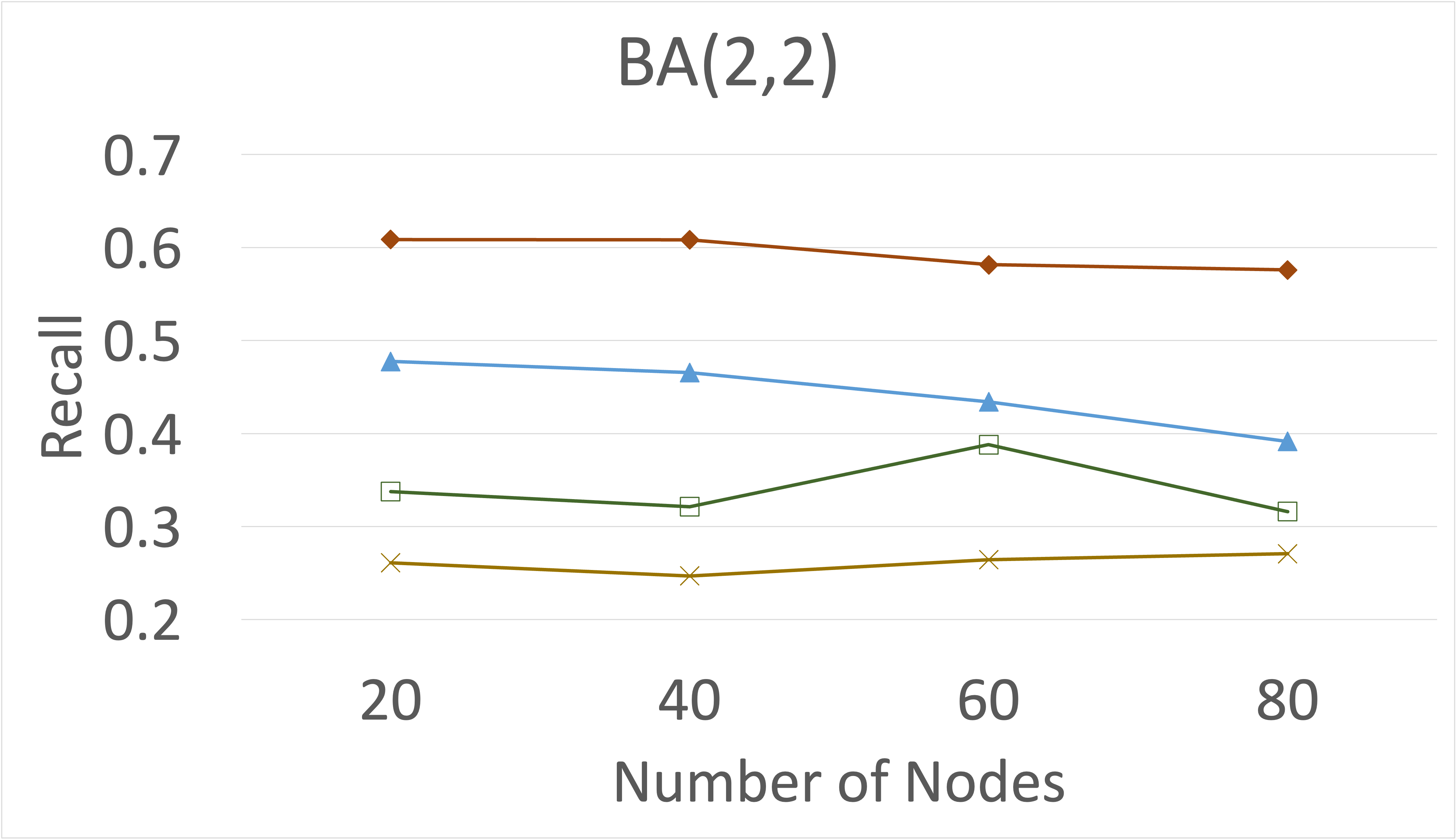}
    \includegraphics[width=0.49\textwidth, height=0.35\textwidth]{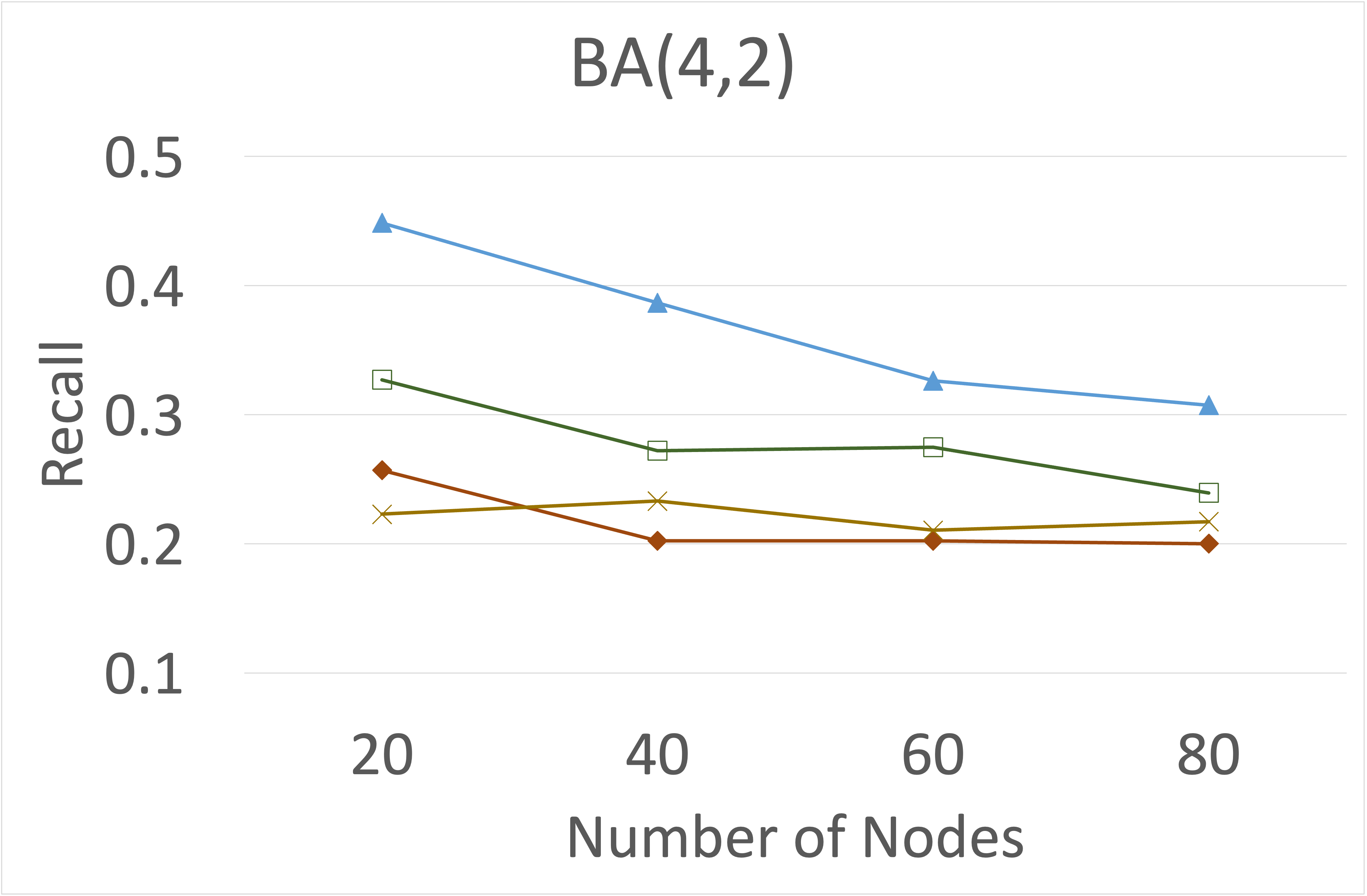}
    \caption{Additive Index Model, $T=200$}
  \end{subfigure}
  \hfill
  \centering
  \begin{subfigure}{0.49\textwidth}
    \centering
    \includegraphics[width=0.49\textwidth, height=0.35\textwidth]{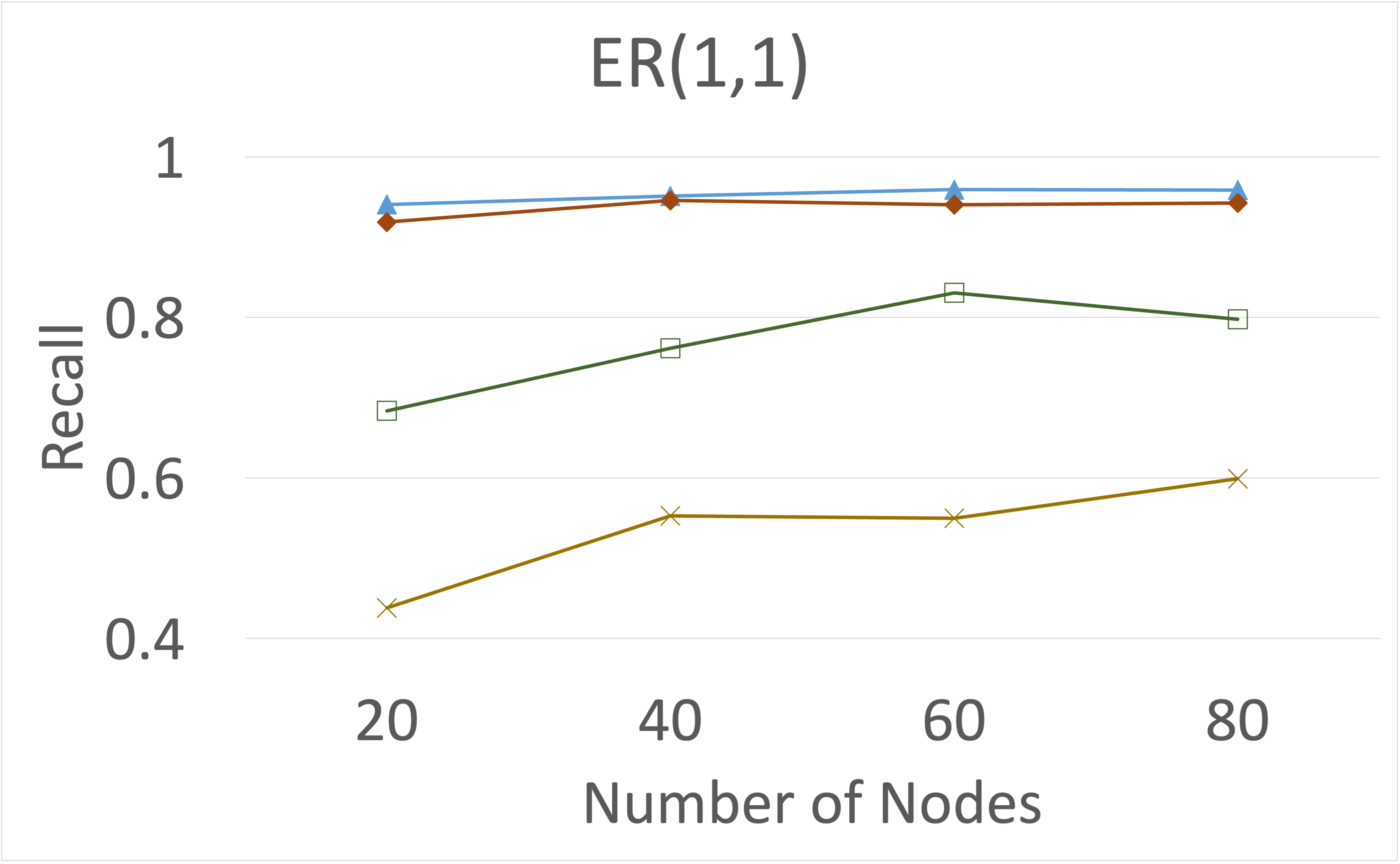}
    \includegraphics[width=0.49\textwidth, height=0.35\textwidth]{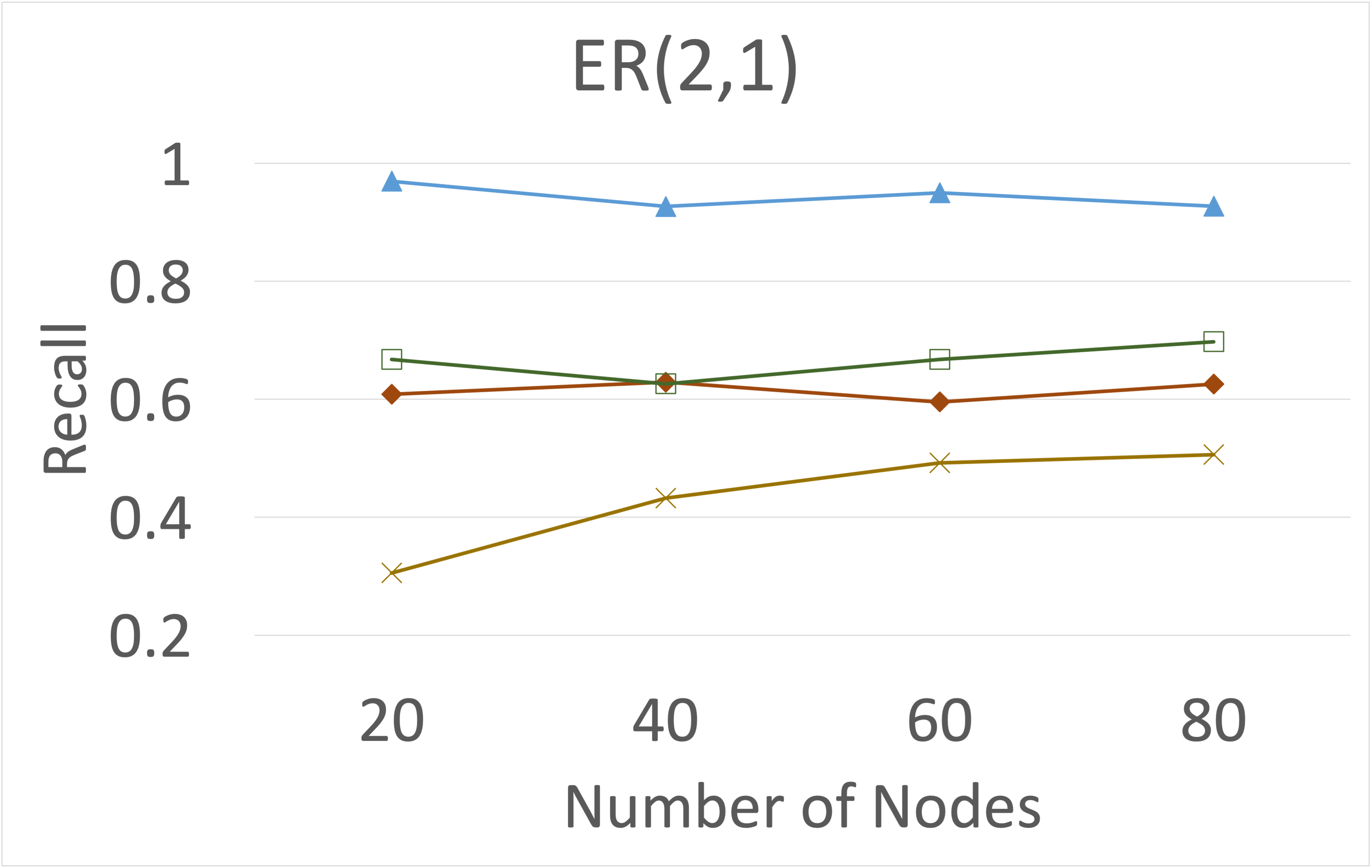}
    \includegraphics[width=0.49\textwidth, height=0.35\textwidth]{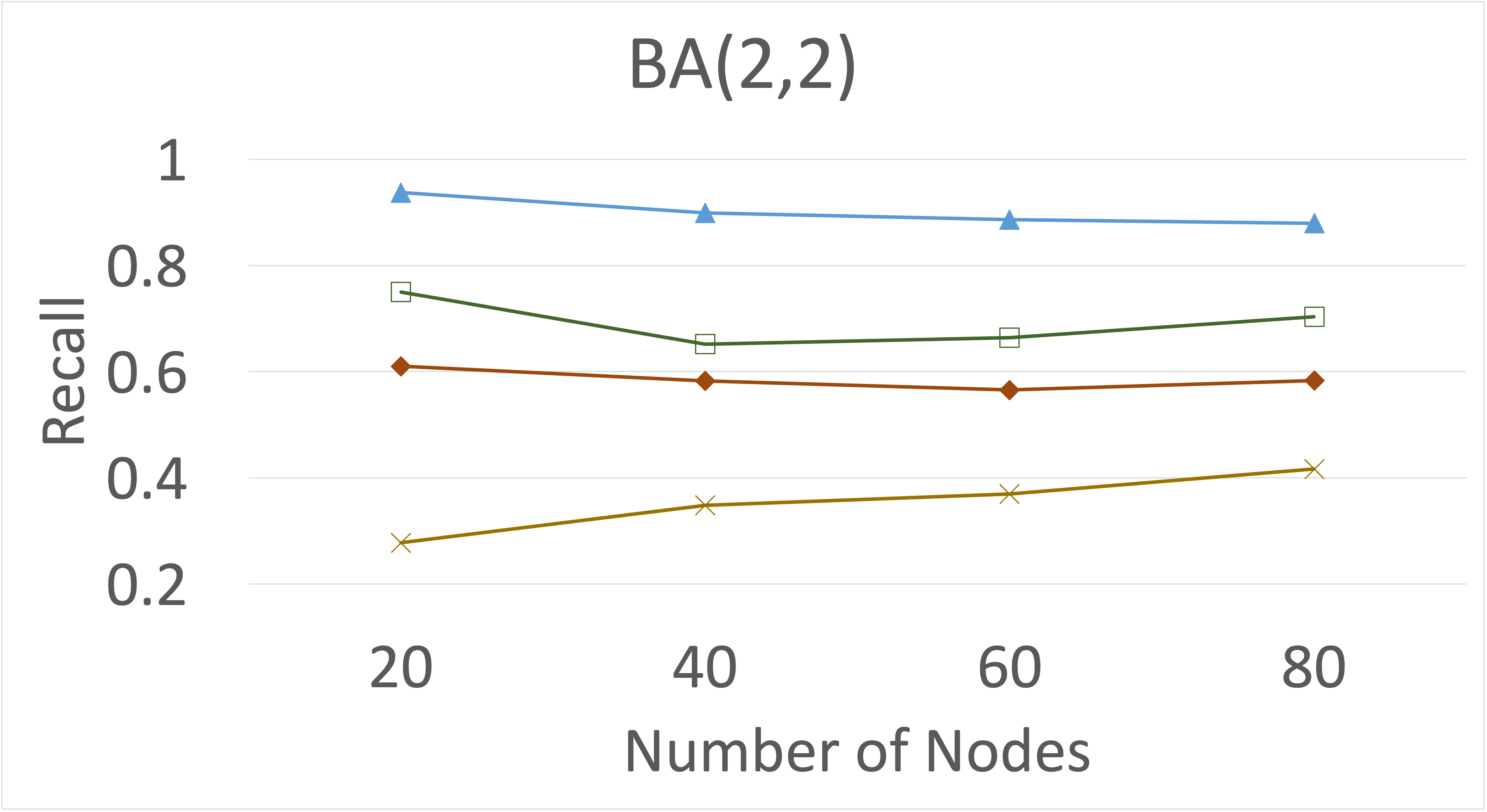}
    \includegraphics[width=0.49\textwidth, height=0.35\textwidth]{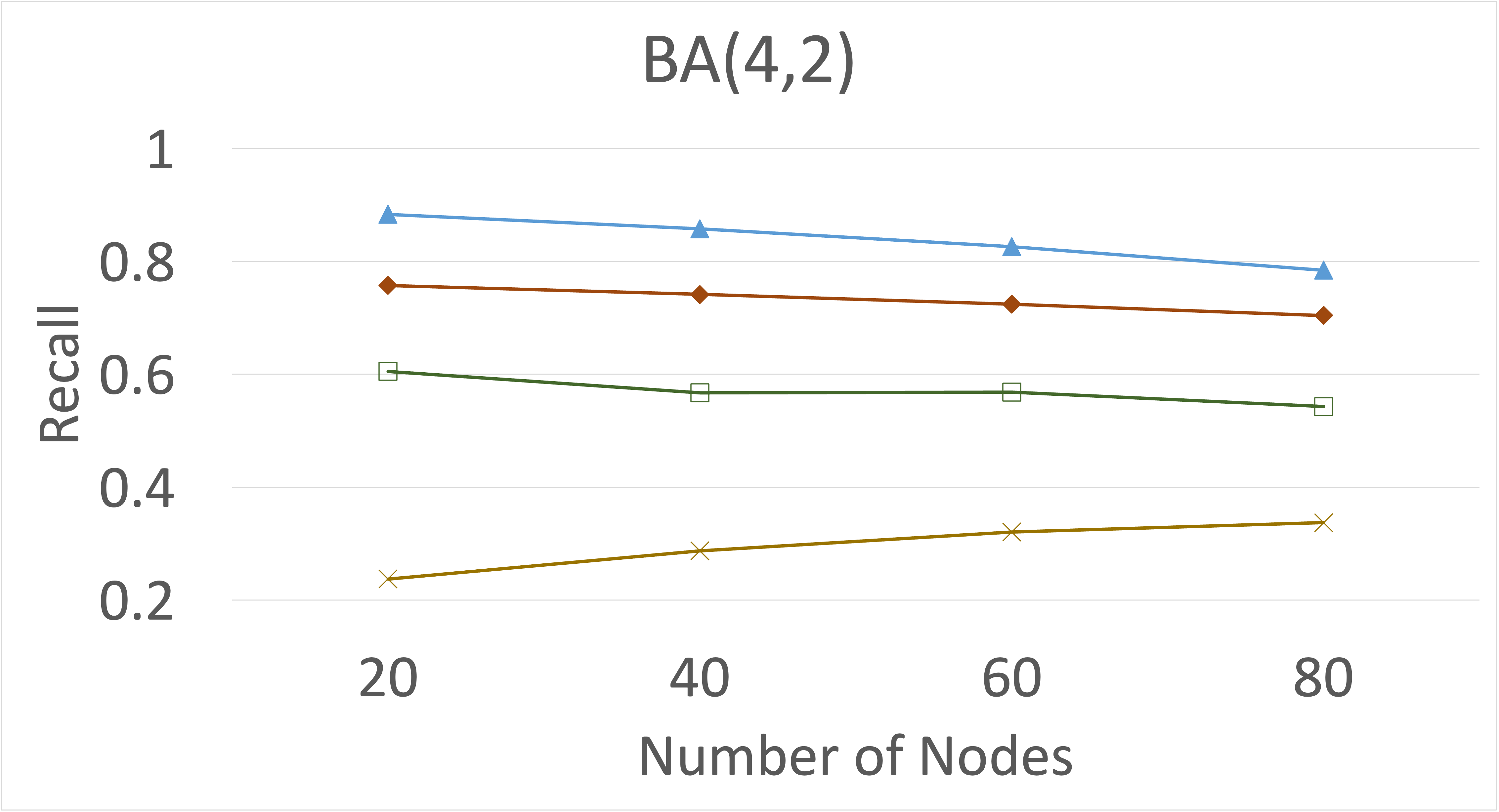}
    \caption{Additive Index Model, $T=1000$}
  \end{subfigure}
  \centering
  \begin{subfigure}{0.49\textwidth}
    \centering
    \includegraphics[width=0.49\textwidth, height=0.35\textwidth]{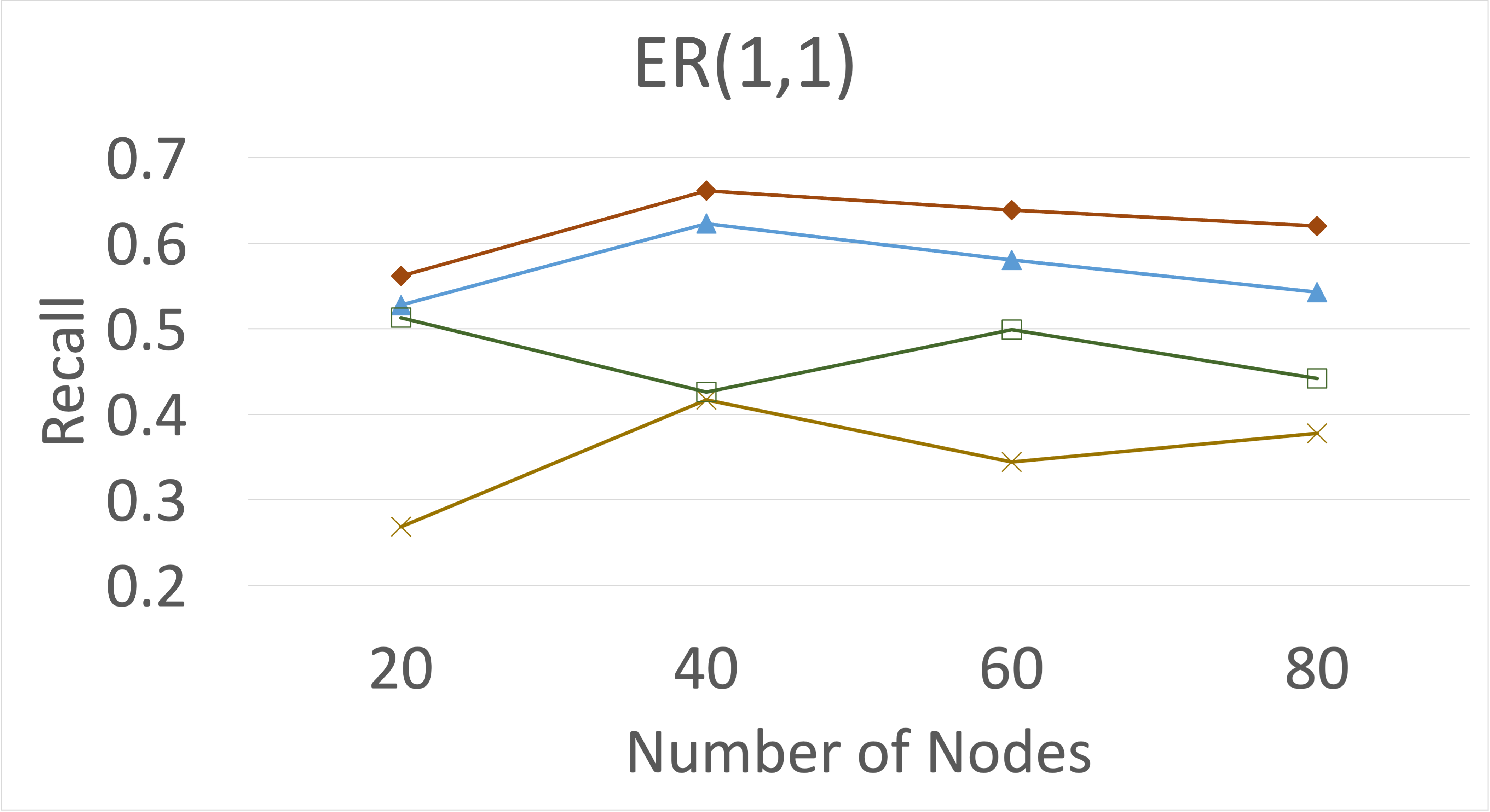}
    \includegraphics[width=0.49\textwidth, height=0.35\textwidth]{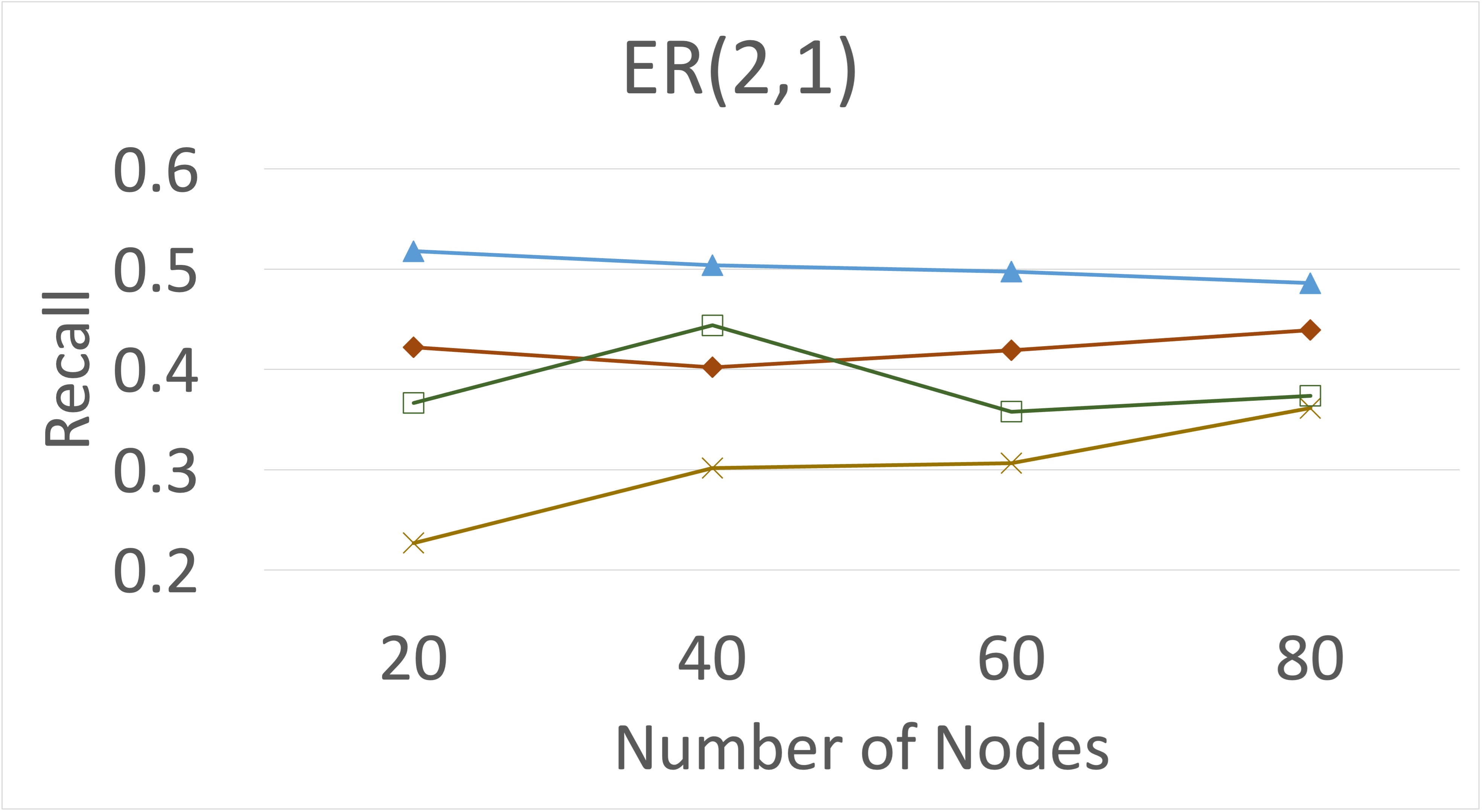}
    \includegraphics[width=0.49\textwidth, height=0.35\textwidth]{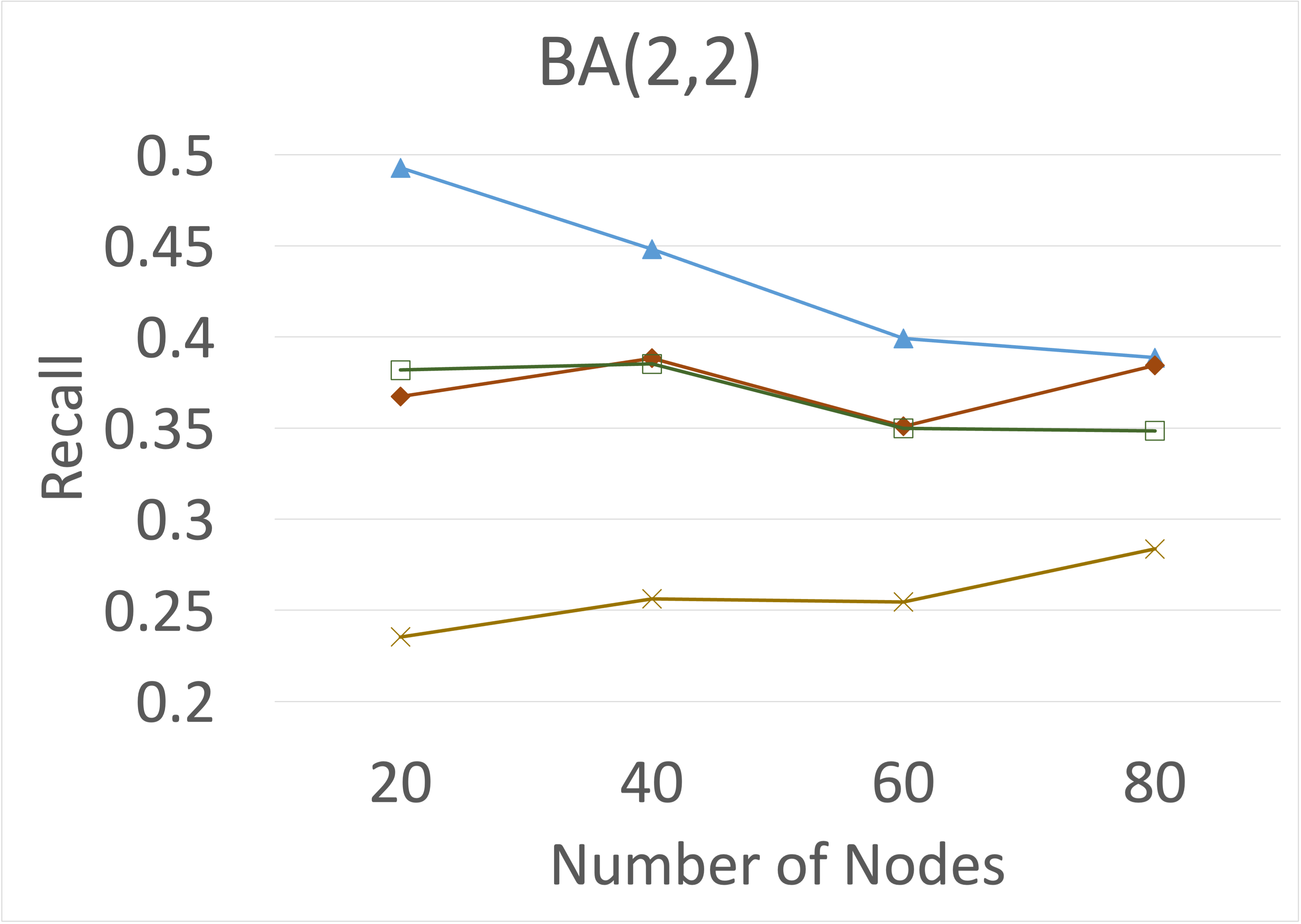}
    \includegraphics[width=0.49\textwidth, height=0.35\textwidth]{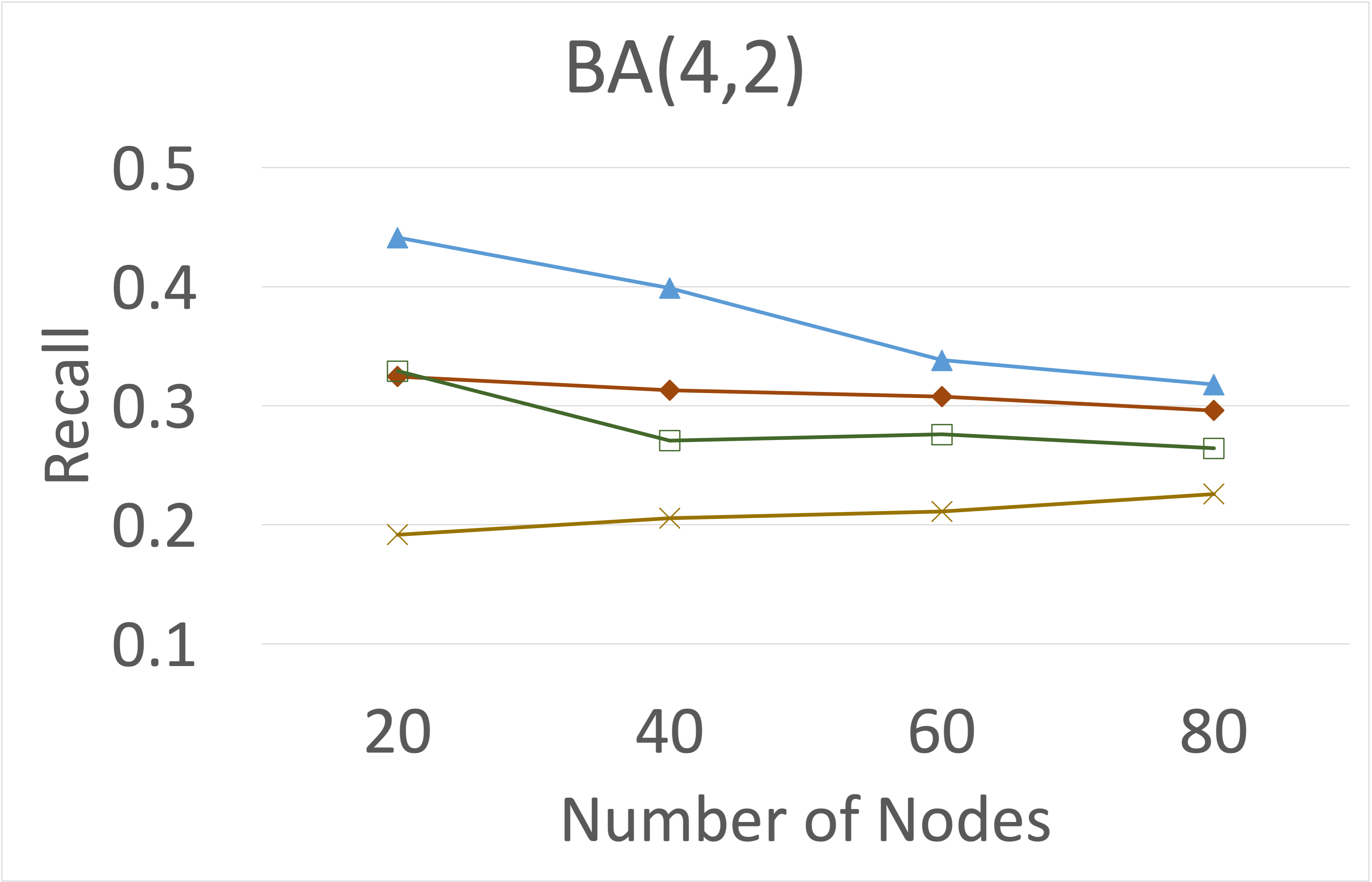}
    \caption{Additive Noise Model, $T=200$}
  \end{subfigure}
  \hfill
  \centering
  \begin{subfigure}{0.49\textwidth}
    \centering
    \includegraphics[width=0.49\textwidth, height=0.35\textwidth]{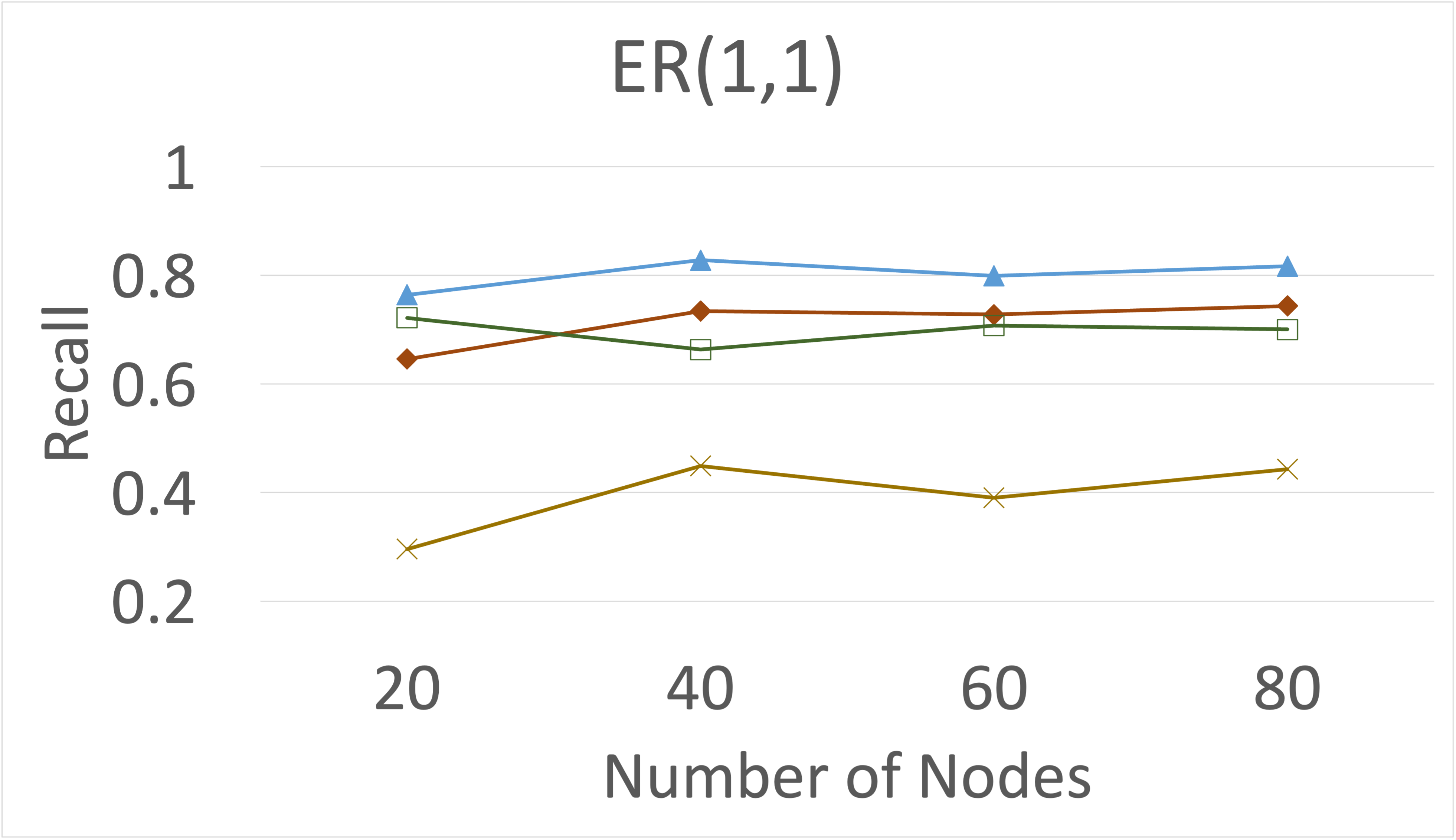}
    \includegraphics[width=0.49\textwidth, height=0.35\textwidth]{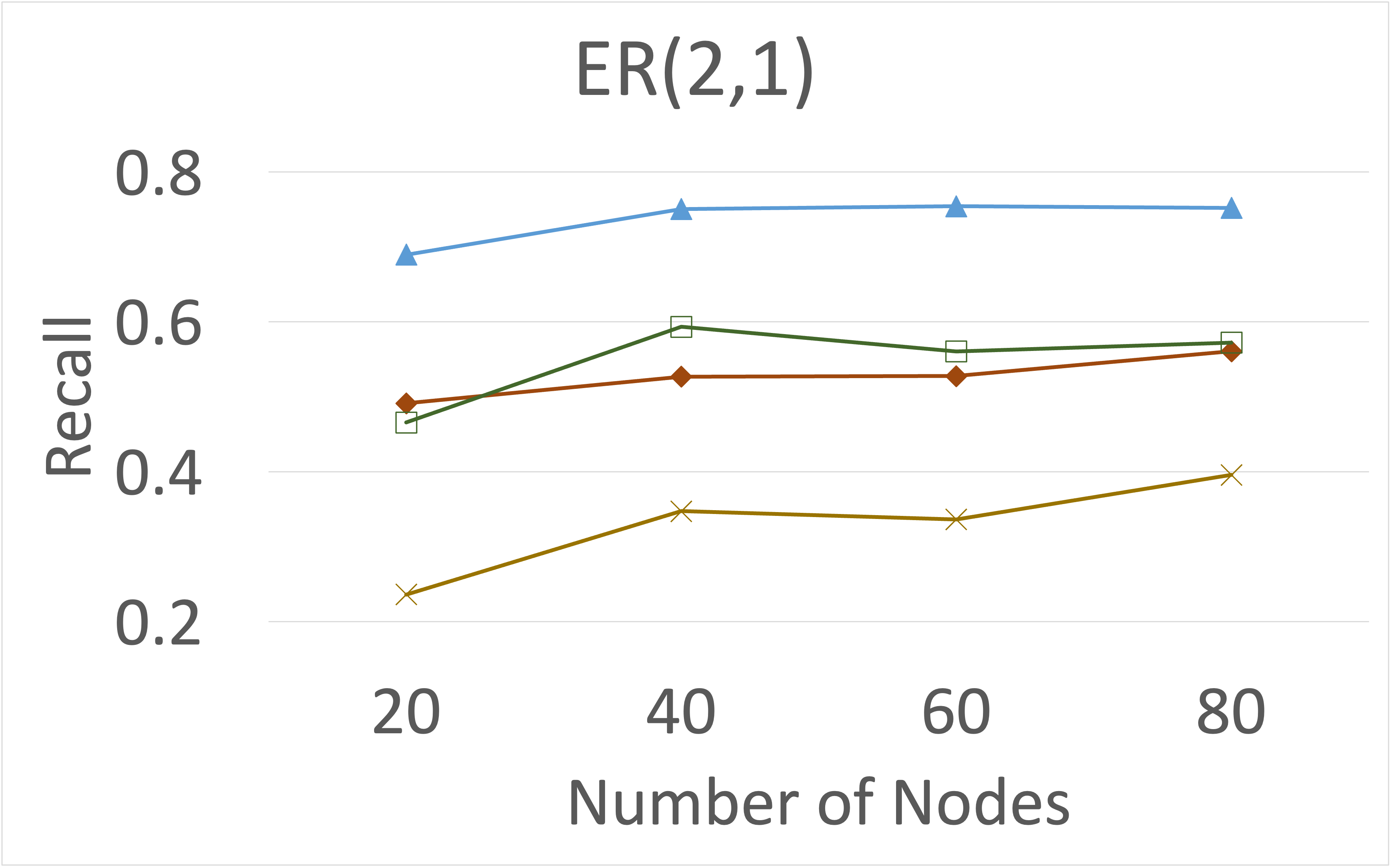}
    \includegraphics[width=0.49\textwidth, height=0.35\textwidth]{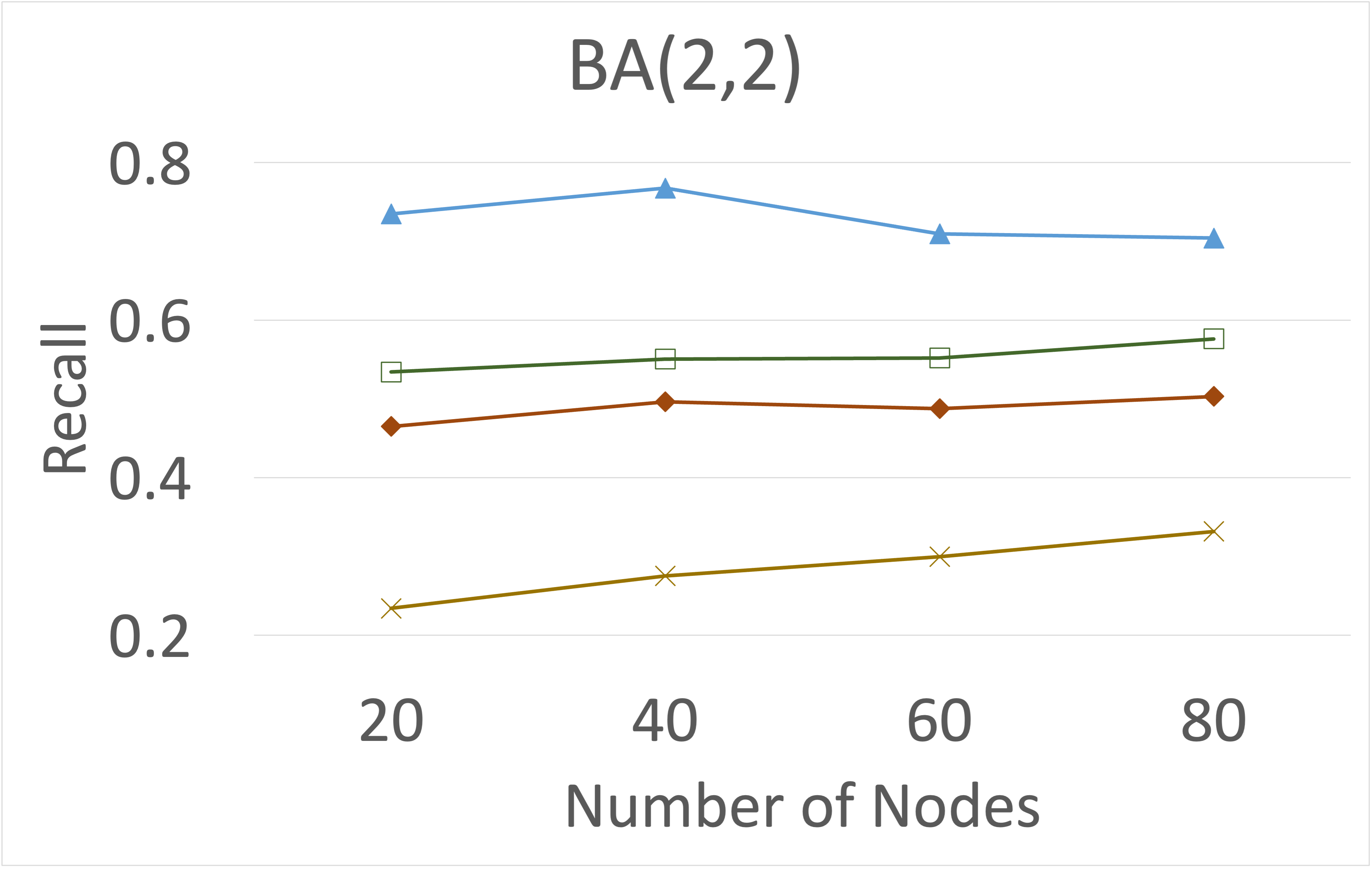}
    \includegraphics[width=0.49\textwidth, height=0.35\textwidth]{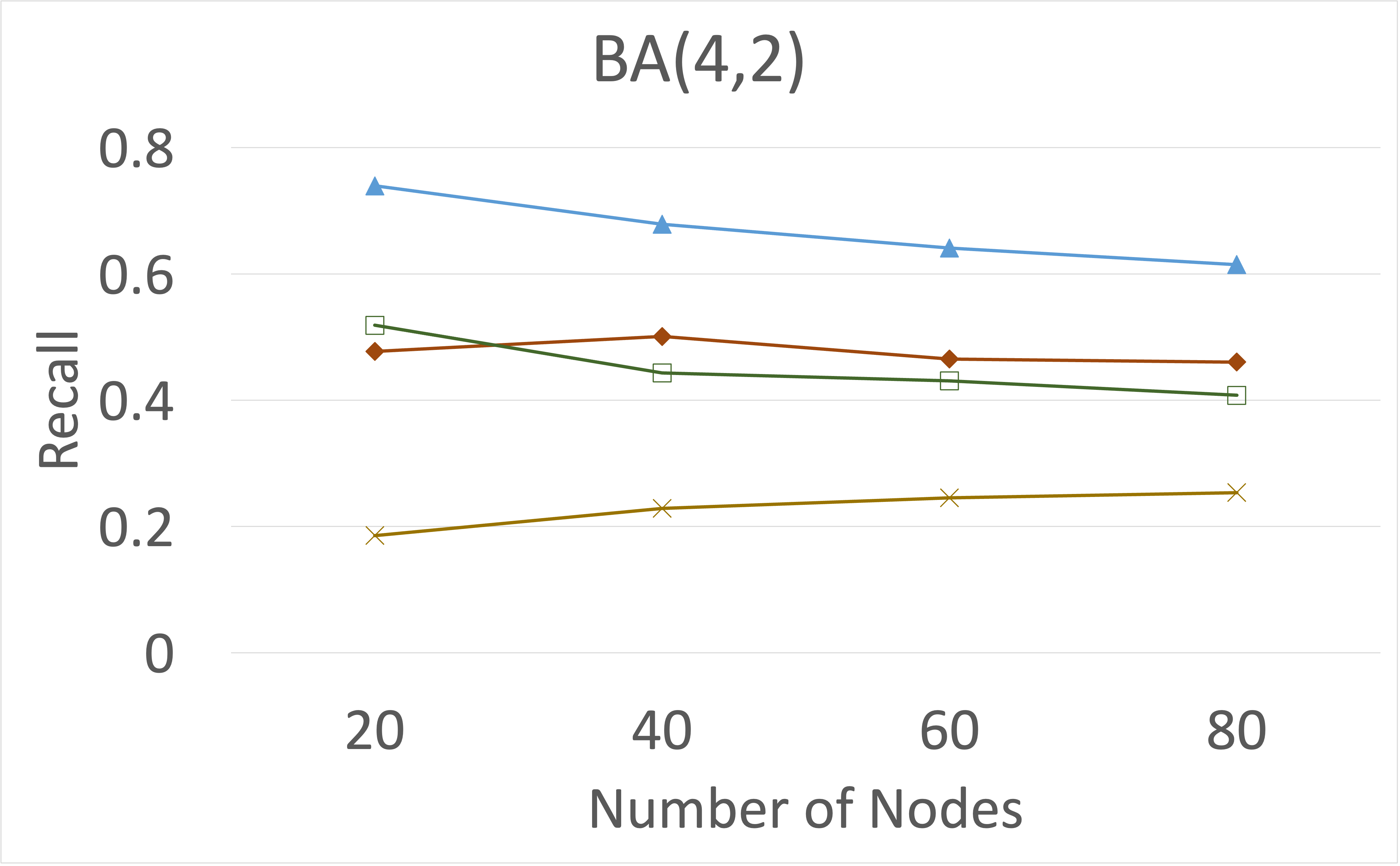}
    \caption{Additive Noise Model, $T=1000$}
  \end{subfigure}
  \centering
  \begin{subfigure}{0.49\textwidth}
    \centering
    \includegraphics[width=0.49\textwidth, height=0.35\textwidth]{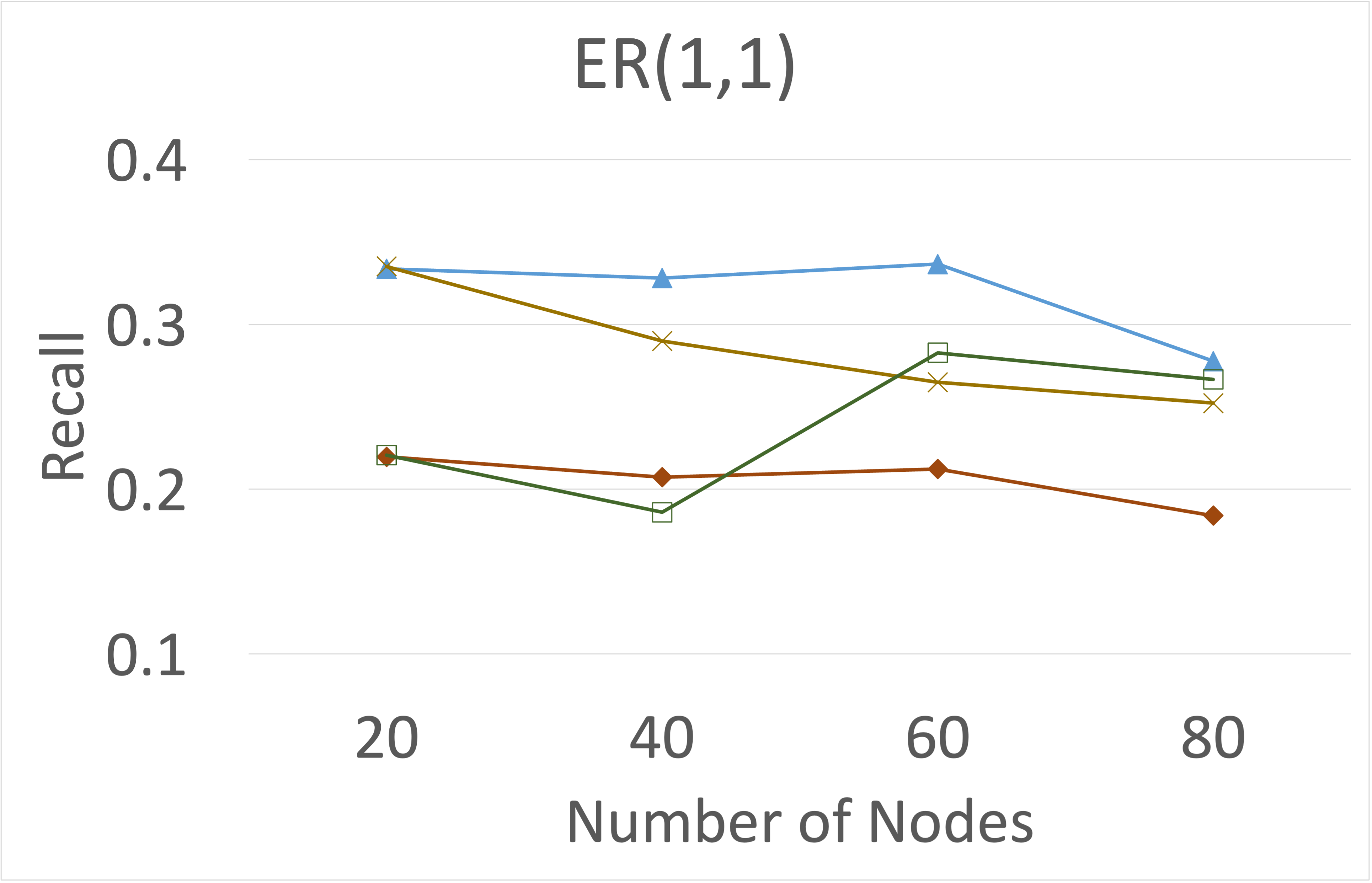}
    \includegraphics[width=0.49\textwidth, height=0.35\textwidth]{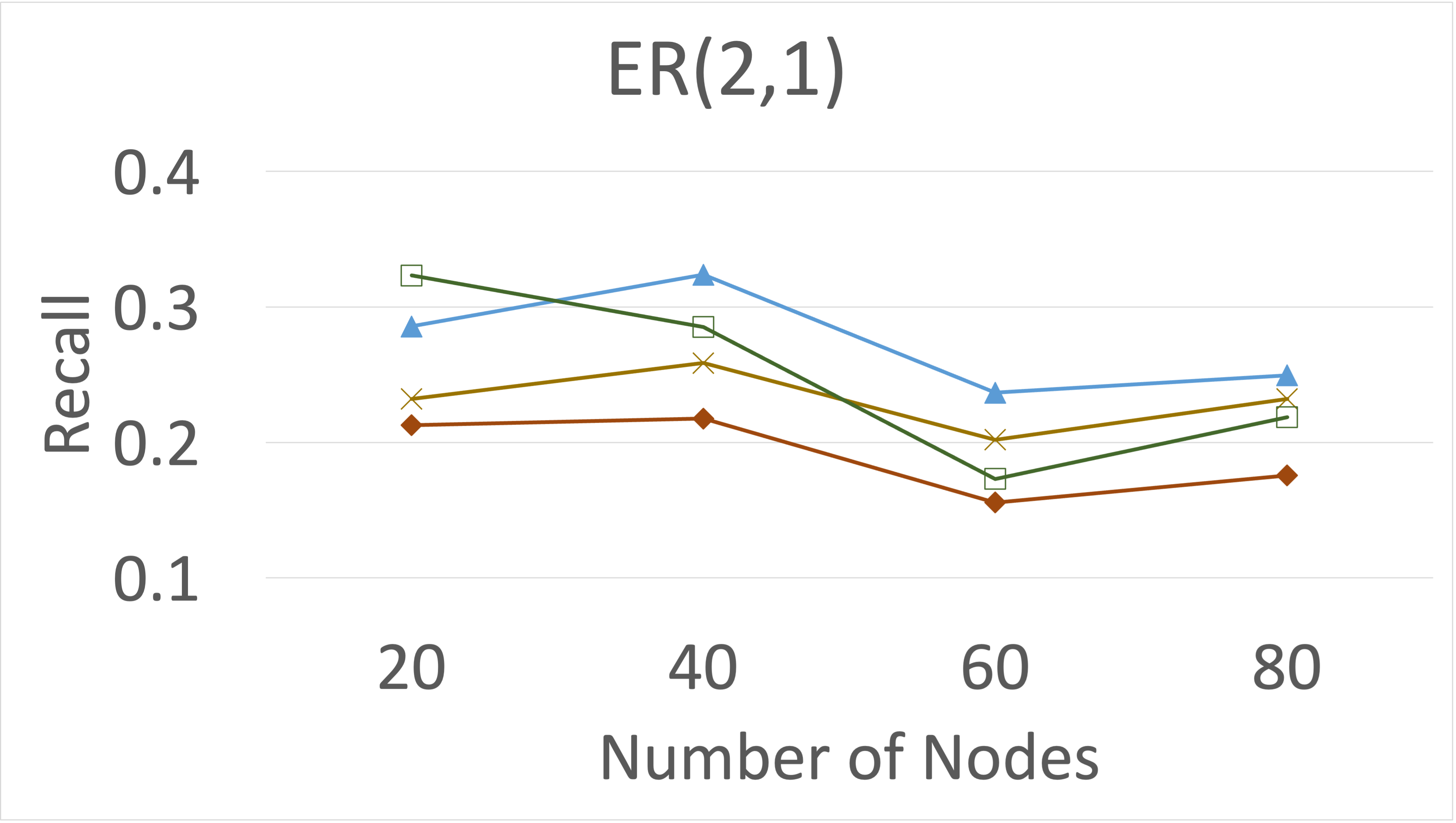}
    \includegraphics[width=0.49\textwidth, height=0.35\textwidth]{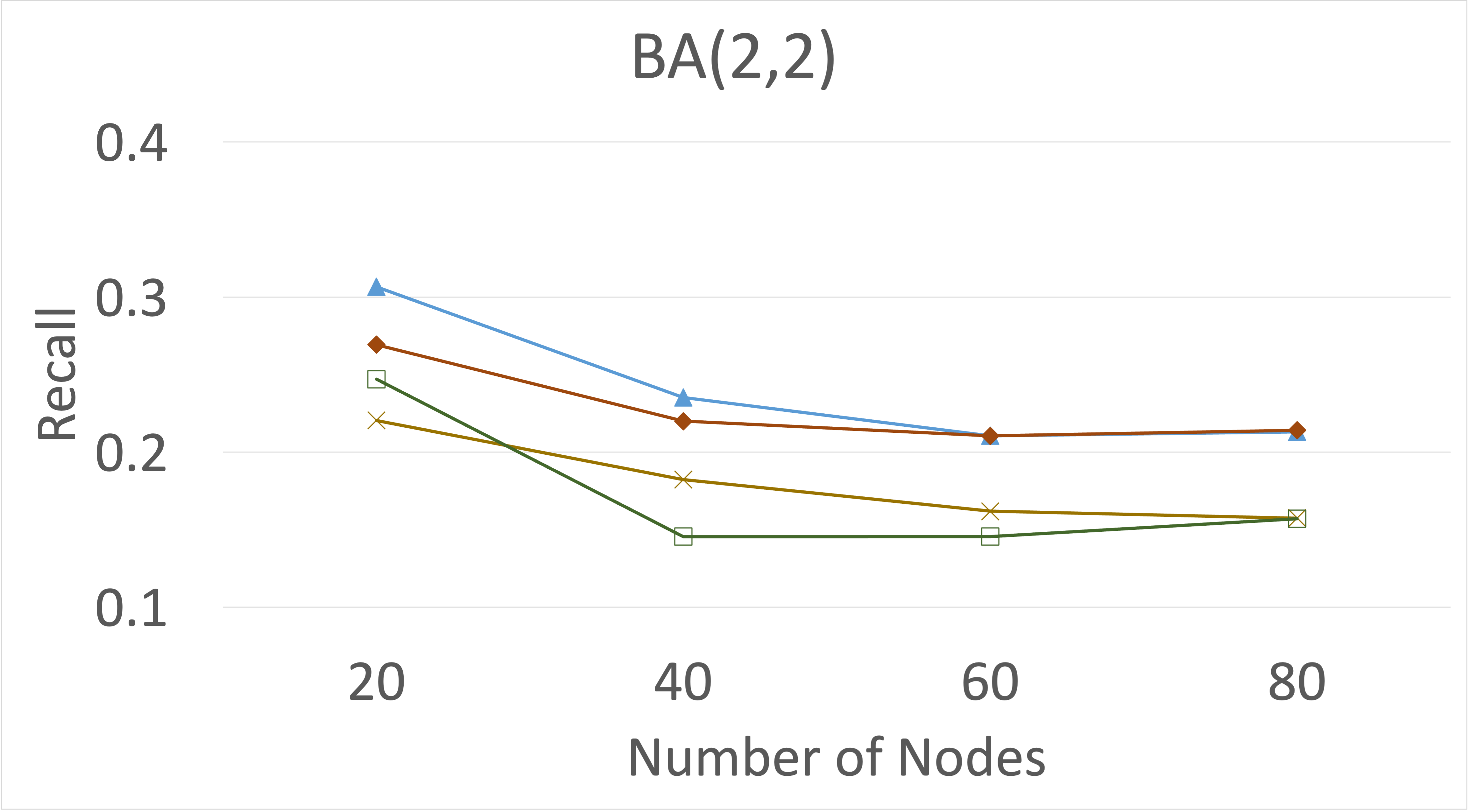}
    \includegraphics[width=0.49\textwidth, height=0.35\textwidth]{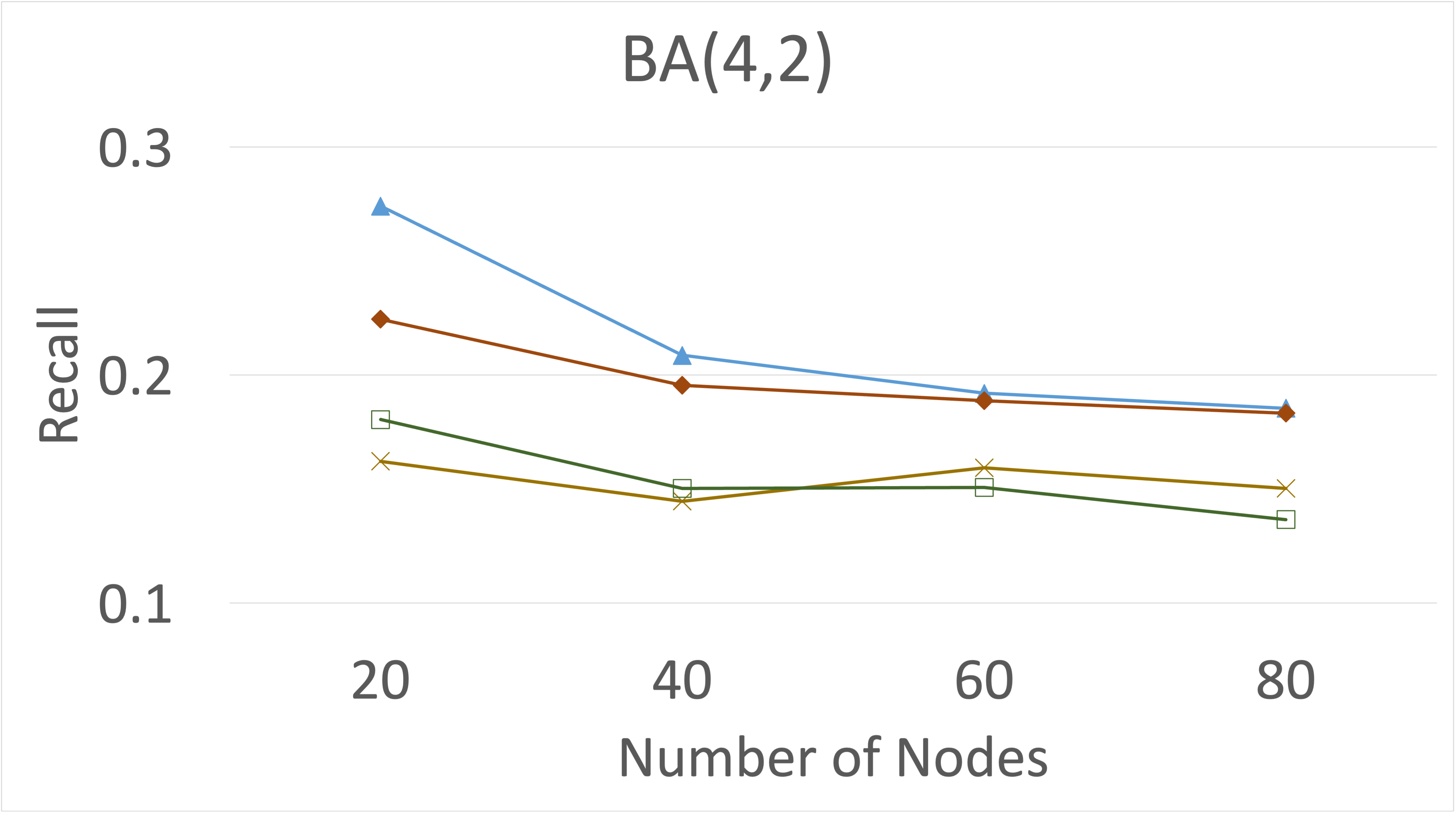}
    \caption{Generalized Linear Model with Poisson
Distribution, $T=200$}
  \end{subfigure}
  \hfill
  \centering
  \begin{subfigure}{0.49\textwidth}
    \centering
    \includegraphics[width=0.49\textwidth, height=0.35\textwidth]{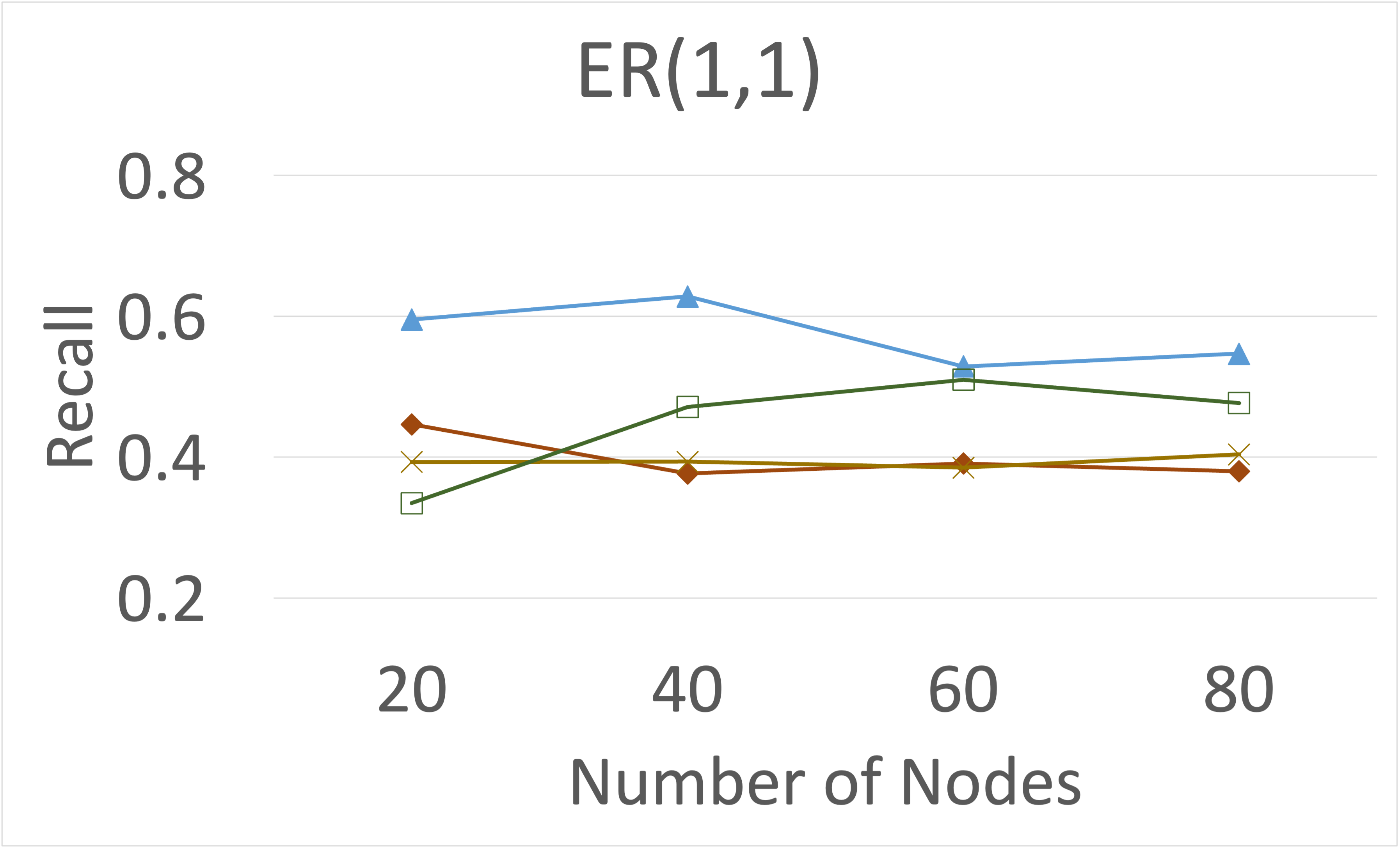}
    \includegraphics[width=0.49\textwidth, height=0.35\textwidth]{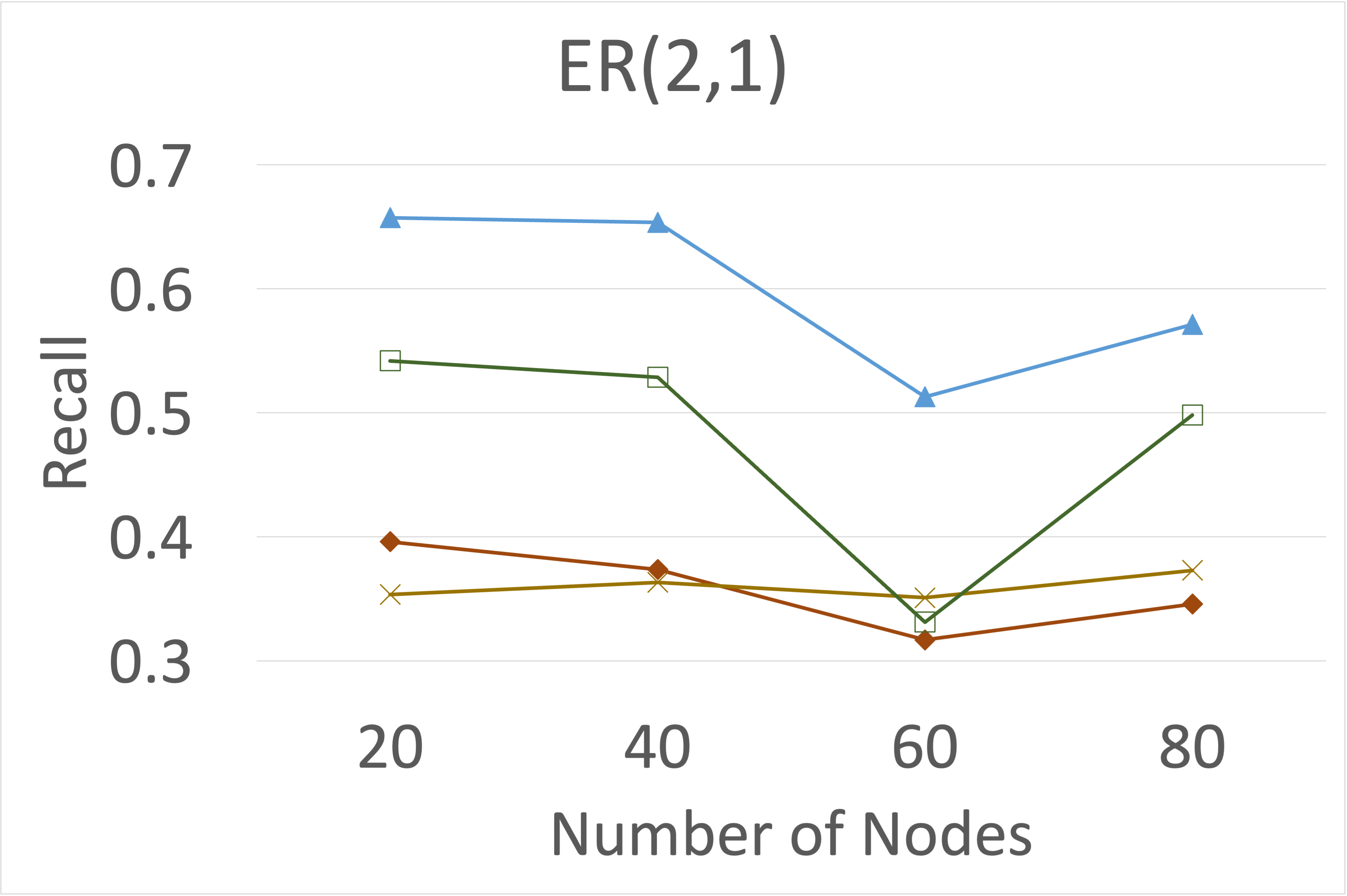}
    \includegraphics[width=0.49\textwidth, height=0.35\textwidth]{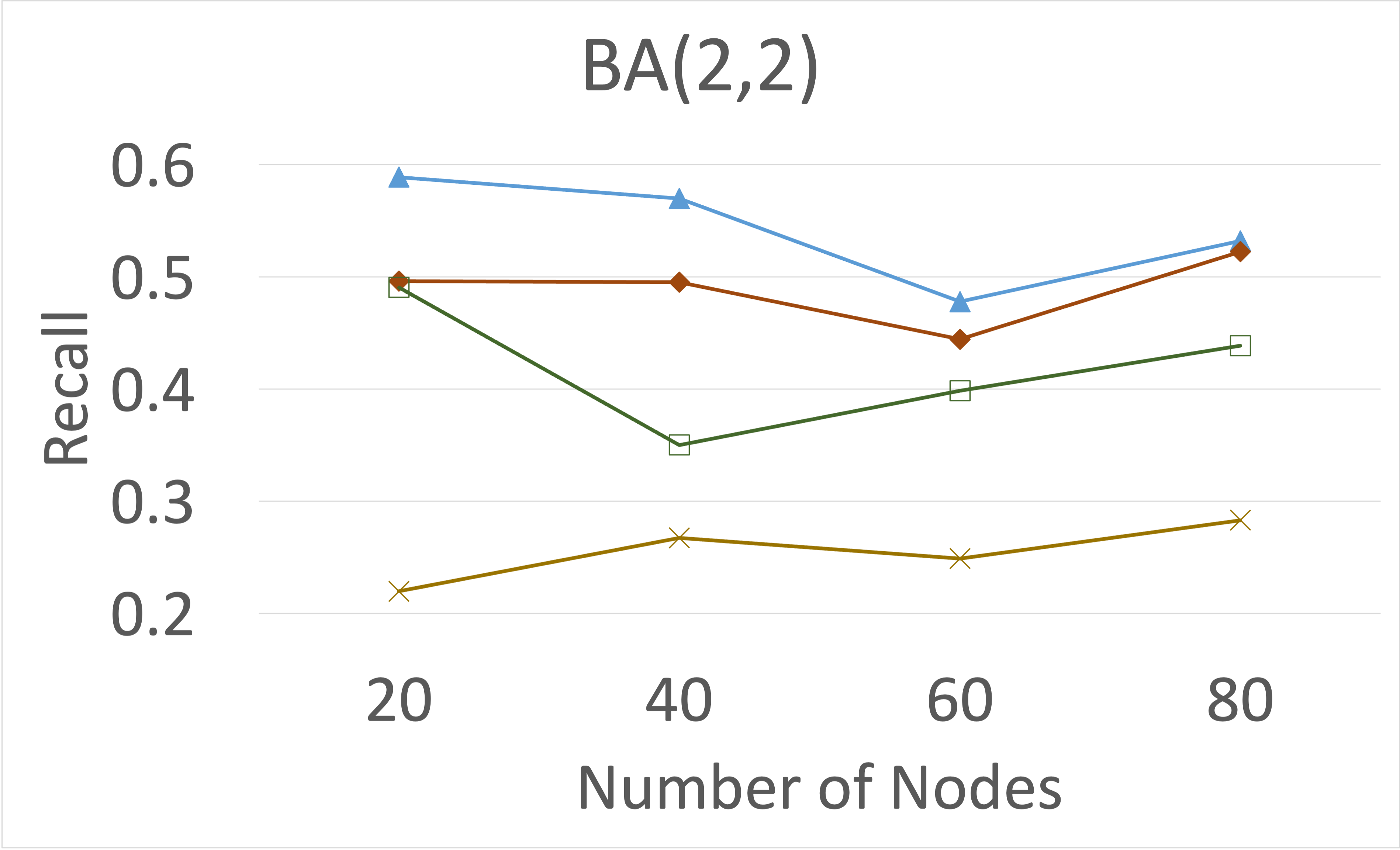}
    \includegraphics[width=0.49\textwidth, height=0.35\textwidth]{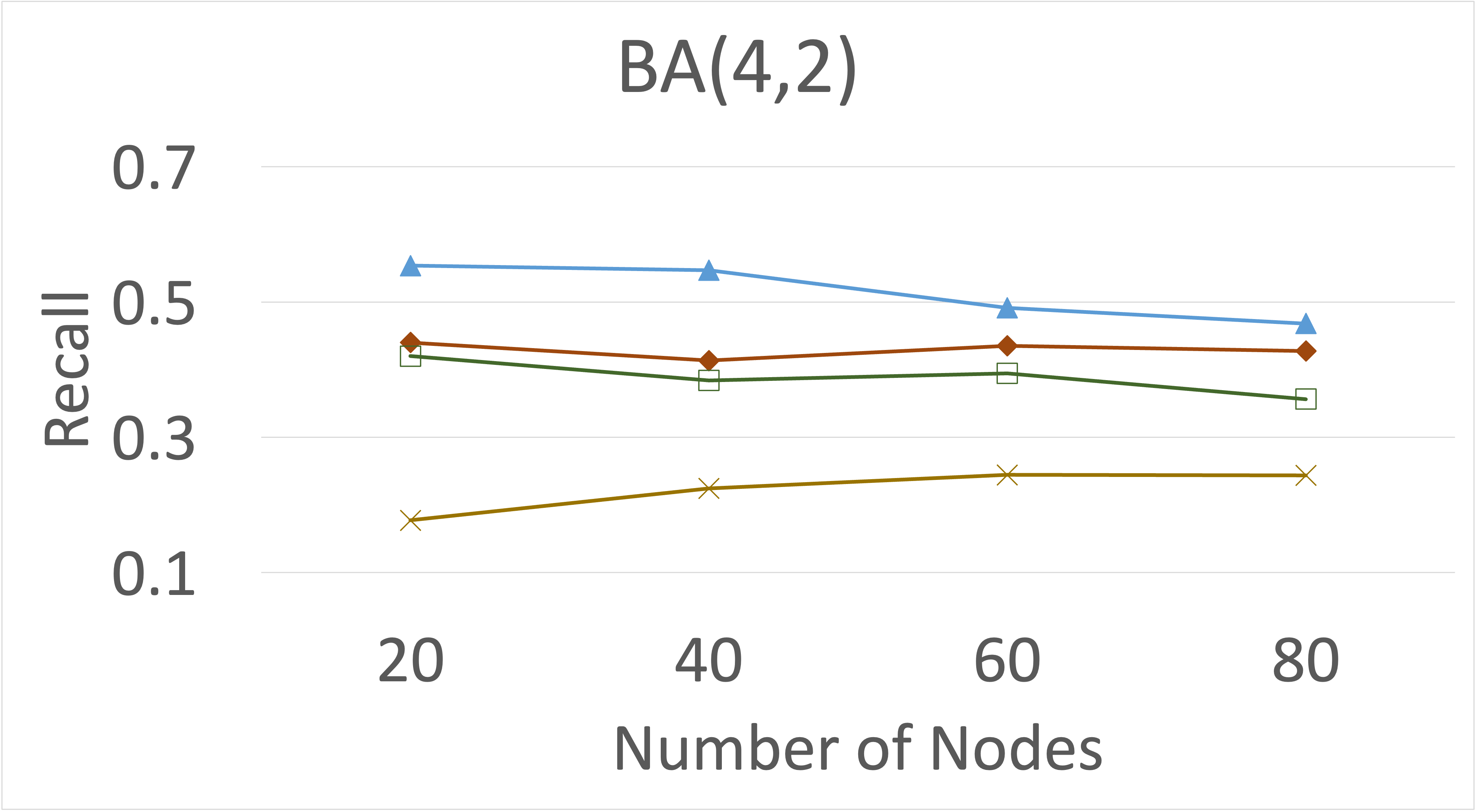}
    \caption{Generalized Linear Model with Poisson
Distribution, $T=1000$}
  \end{subfigure}
  \centering
  \begin{subfigure}{1.0\textwidth}
    \centering
    \includegraphics[width=0.5\textwidth, height=0.03\textwidth]{images/Simulated_Data/legend.png}
  \end{subfigure}
  \caption{\textbf{Mean recalls} over 10 datasets for each setting with simulated data. Higher Recall is better. The number of lags $= 3$.}
  \label{fig:experiments_simulated_recall}
\end{figure*}

\begin{figure*}[ht]
    \centering
    \begin{subfigure}{0.49\textwidth}
      \centering
      \includegraphics[width=0.49\textwidth, height=0.35\textwidth]{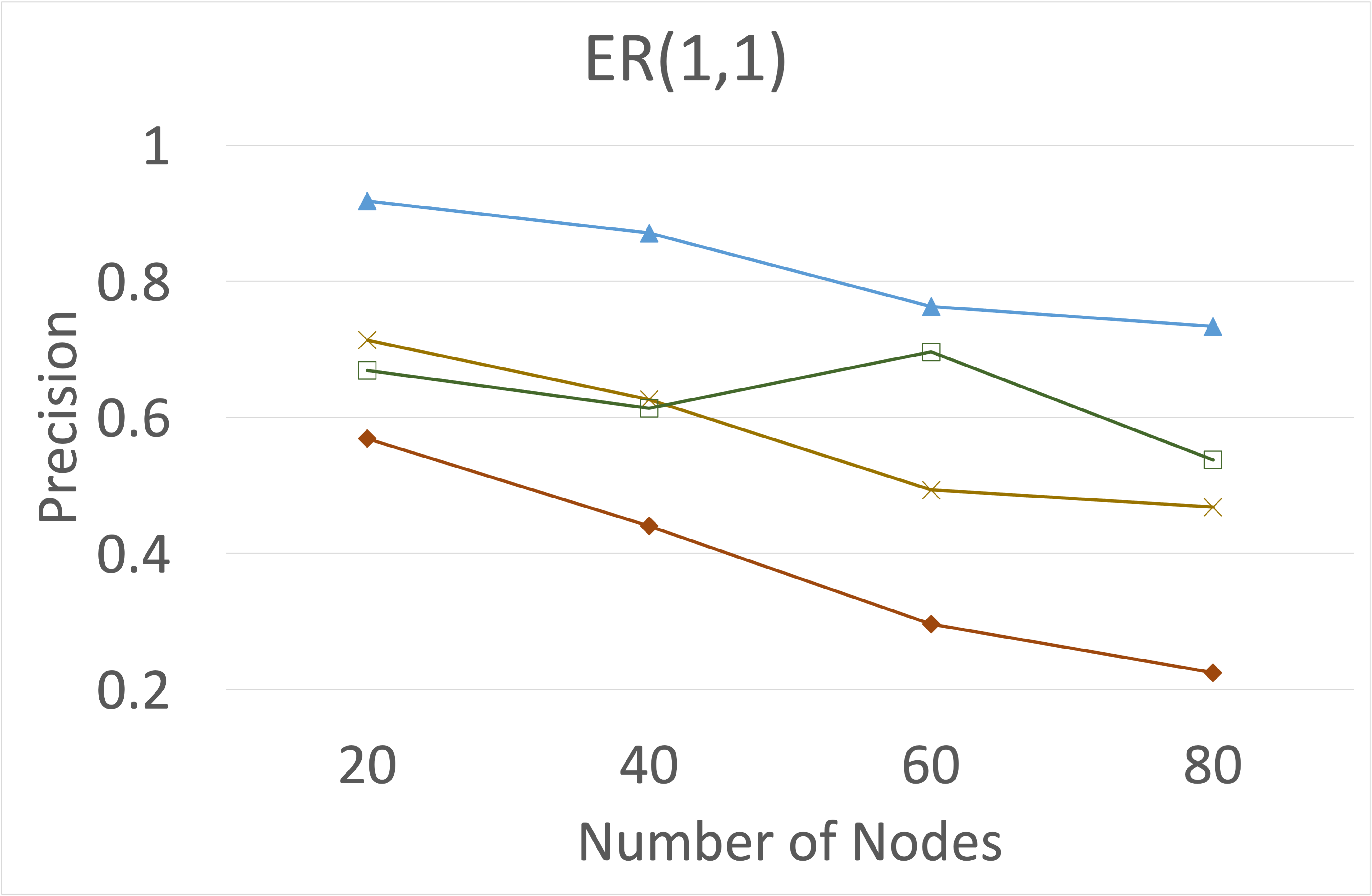}
      \includegraphics[width=0.49\textwidth, height=0.35\textwidth]{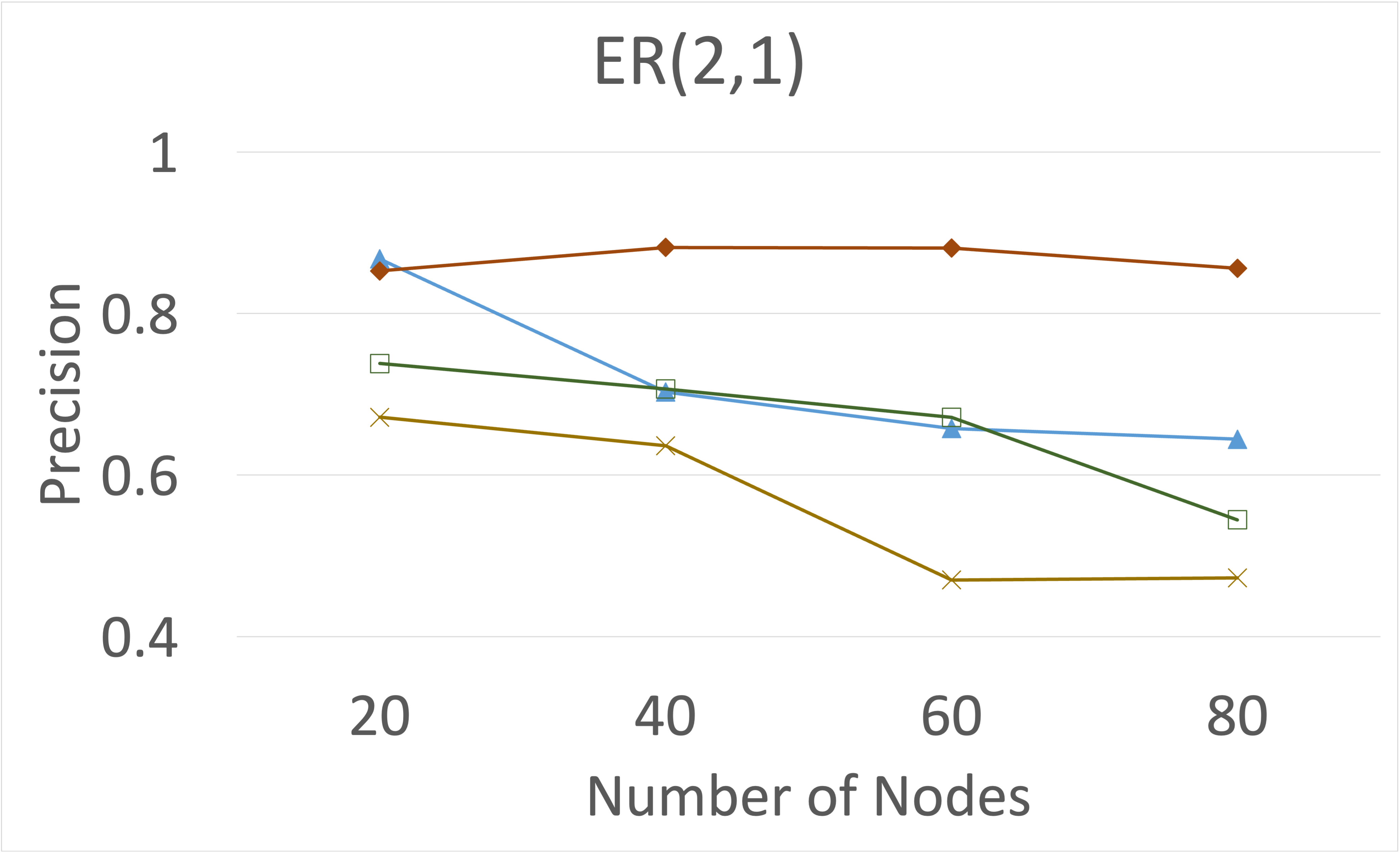}
      \includegraphics[width=0.49\textwidth, height=0.35\textwidth]{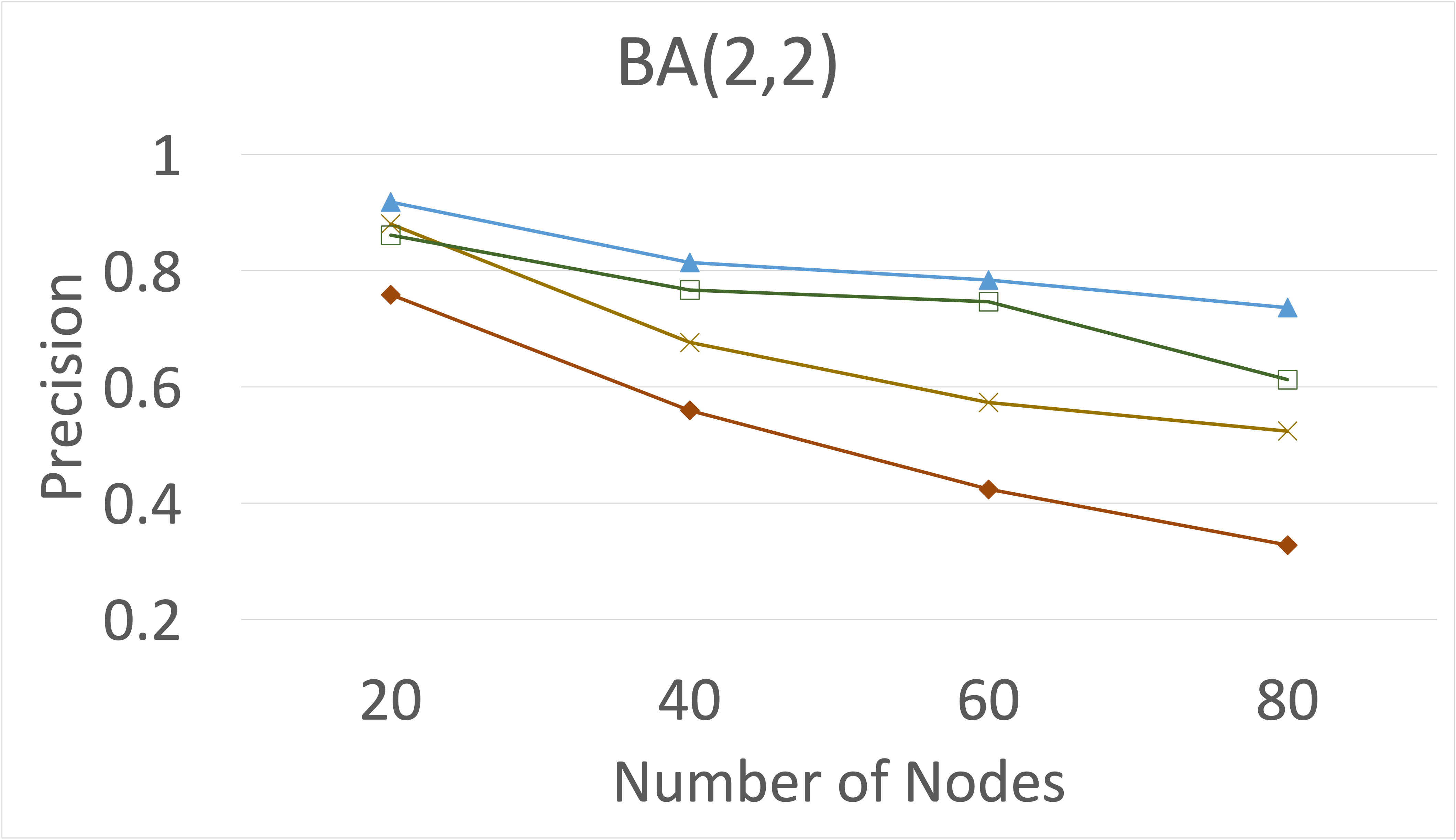}
      \includegraphics[width=0.49\textwidth, height=0.35\textwidth]{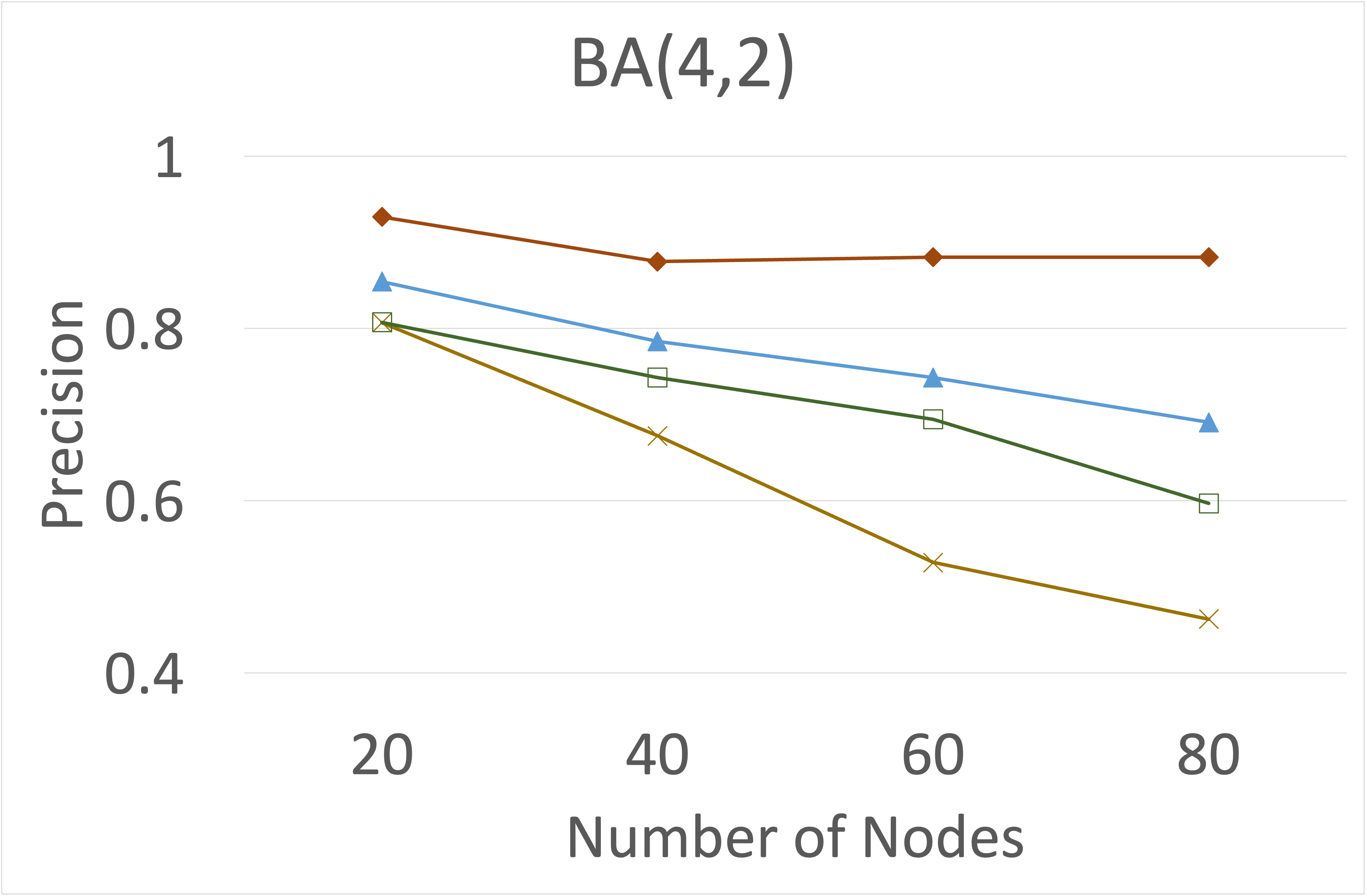}
      \caption{Additive Index Model, $T=200$}
    \end{subfigure}
    \hfill
    \centering
    \begin{subfigure}{0.49\textwidth}
      \centering
      \includegraphics[width=0.49\textwidth, height=0.35\textwidth]{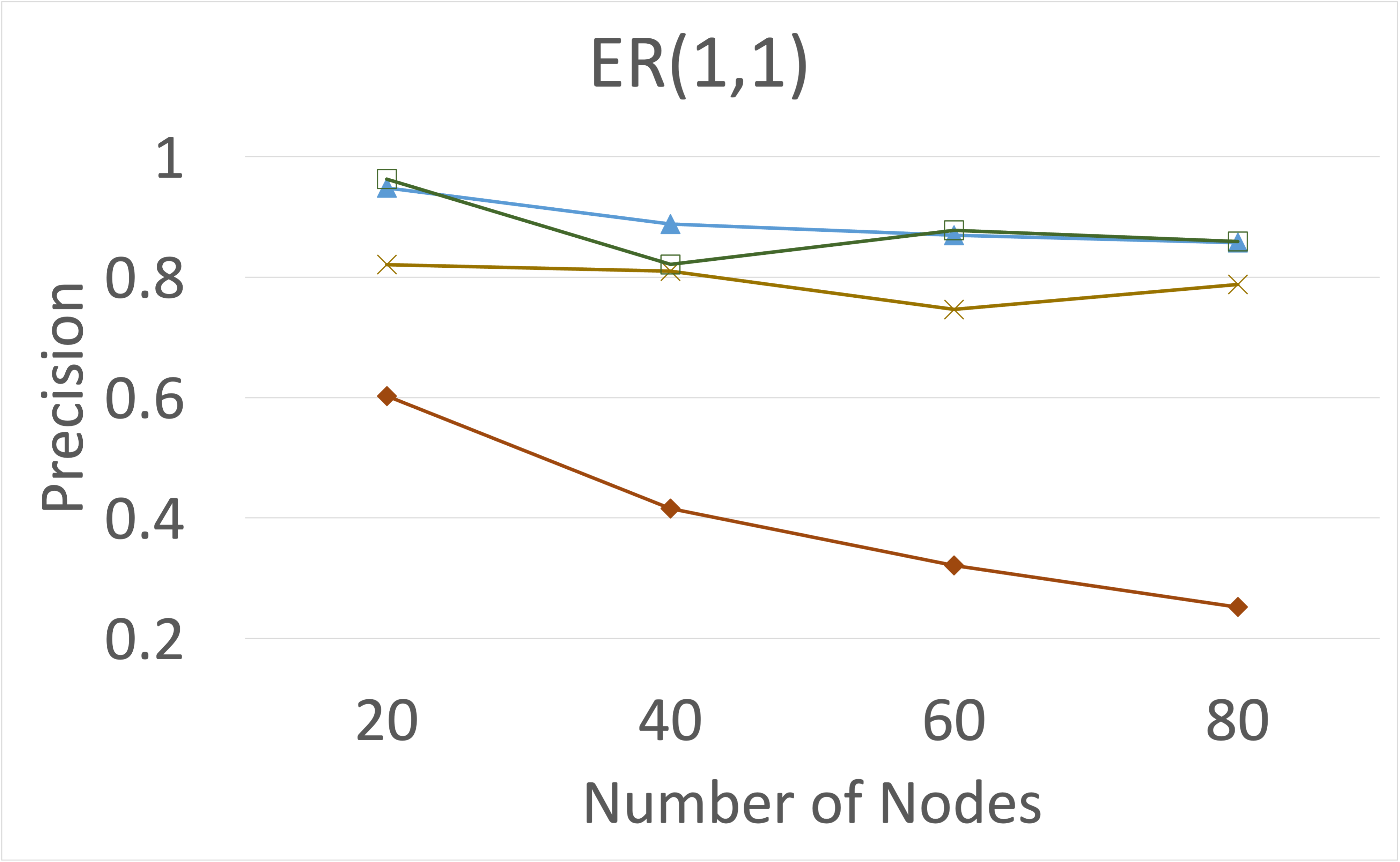}
      \includegraphics[width=0.49\textwidth, height=0.35\textwidth]{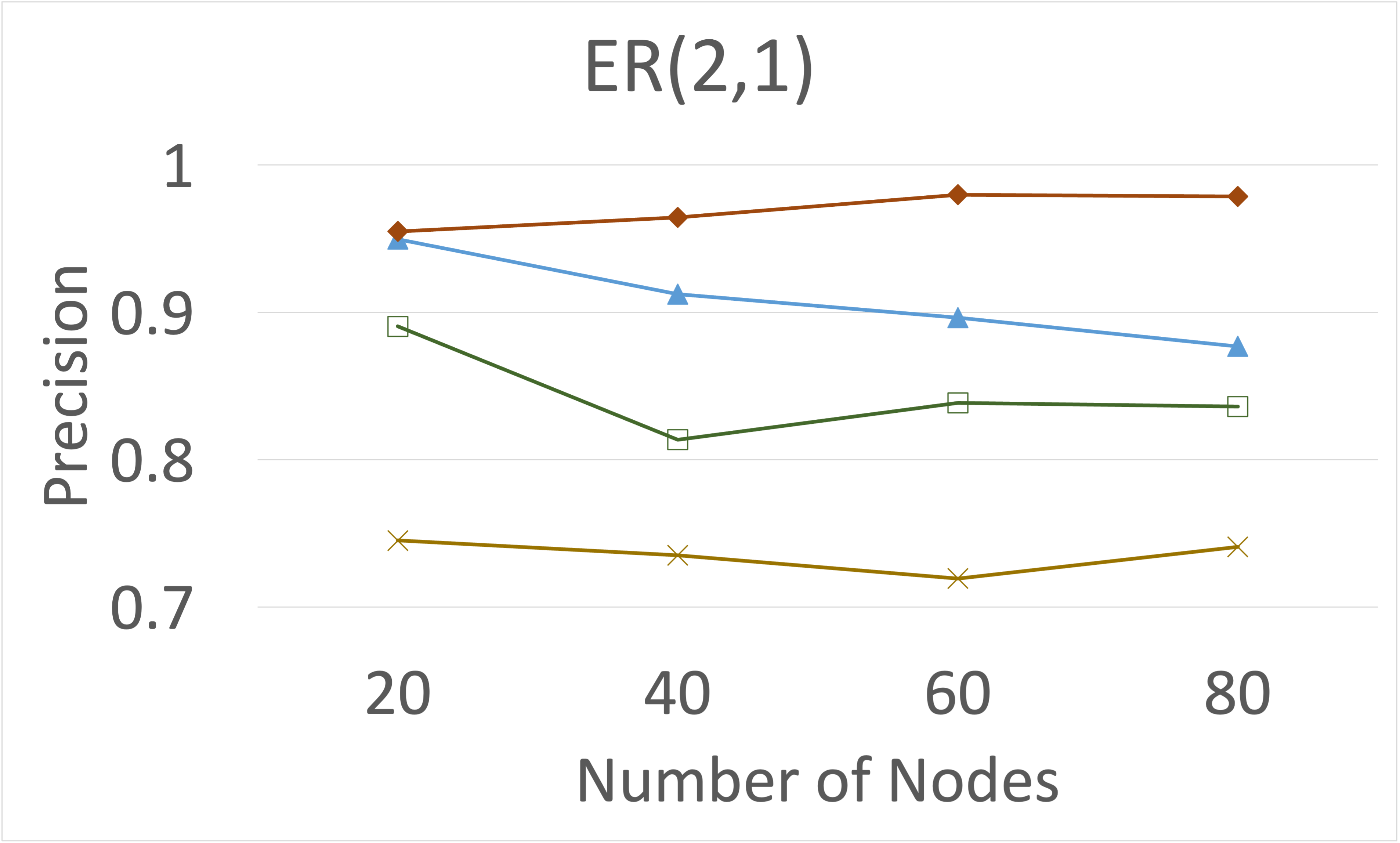}
      \includegraphics[width=0.49\textwidth, height=0.35\textwidth]{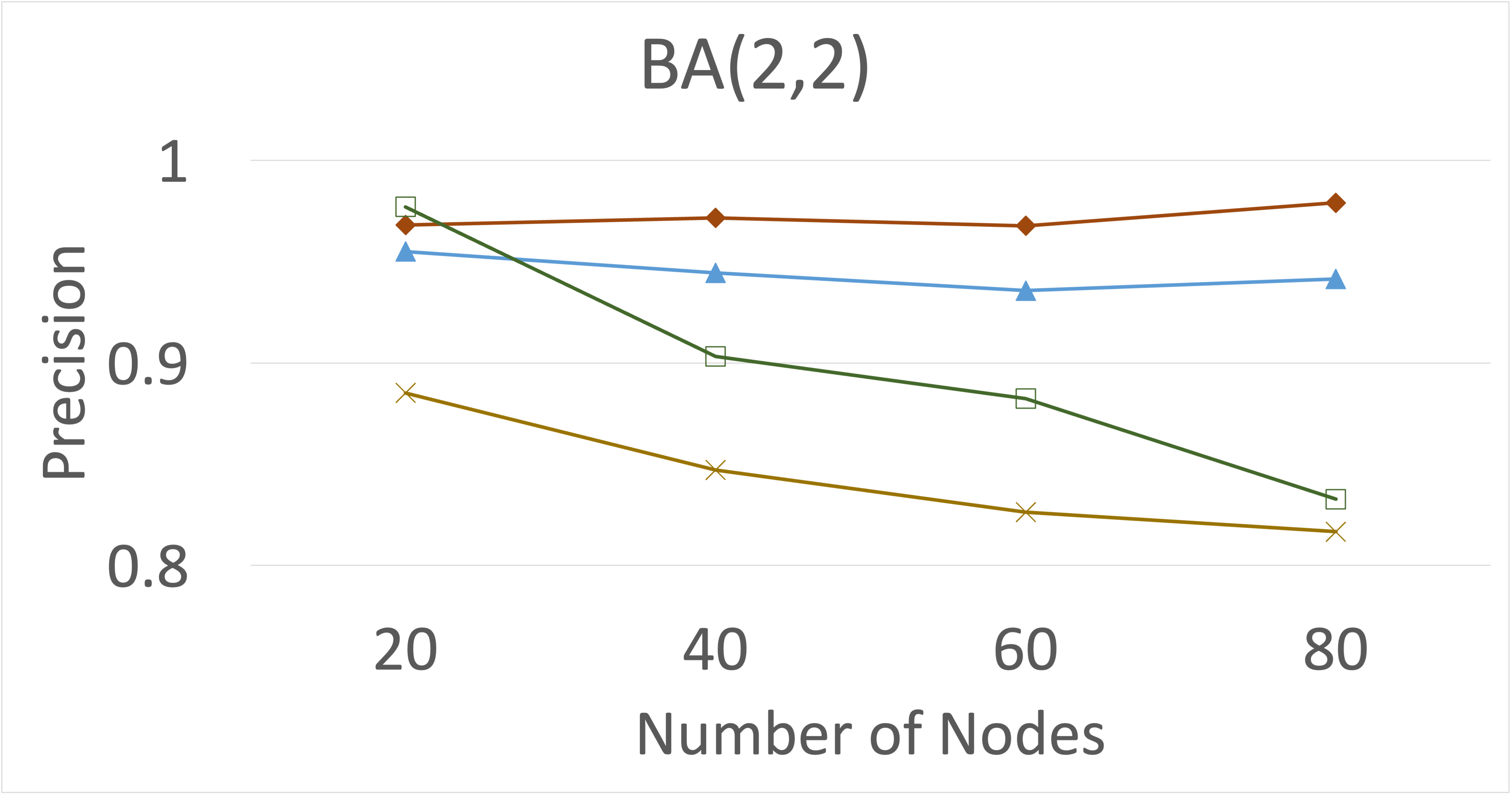}
      \includegraphics[width=0.49\textwidth, height=0.35\textwidth]{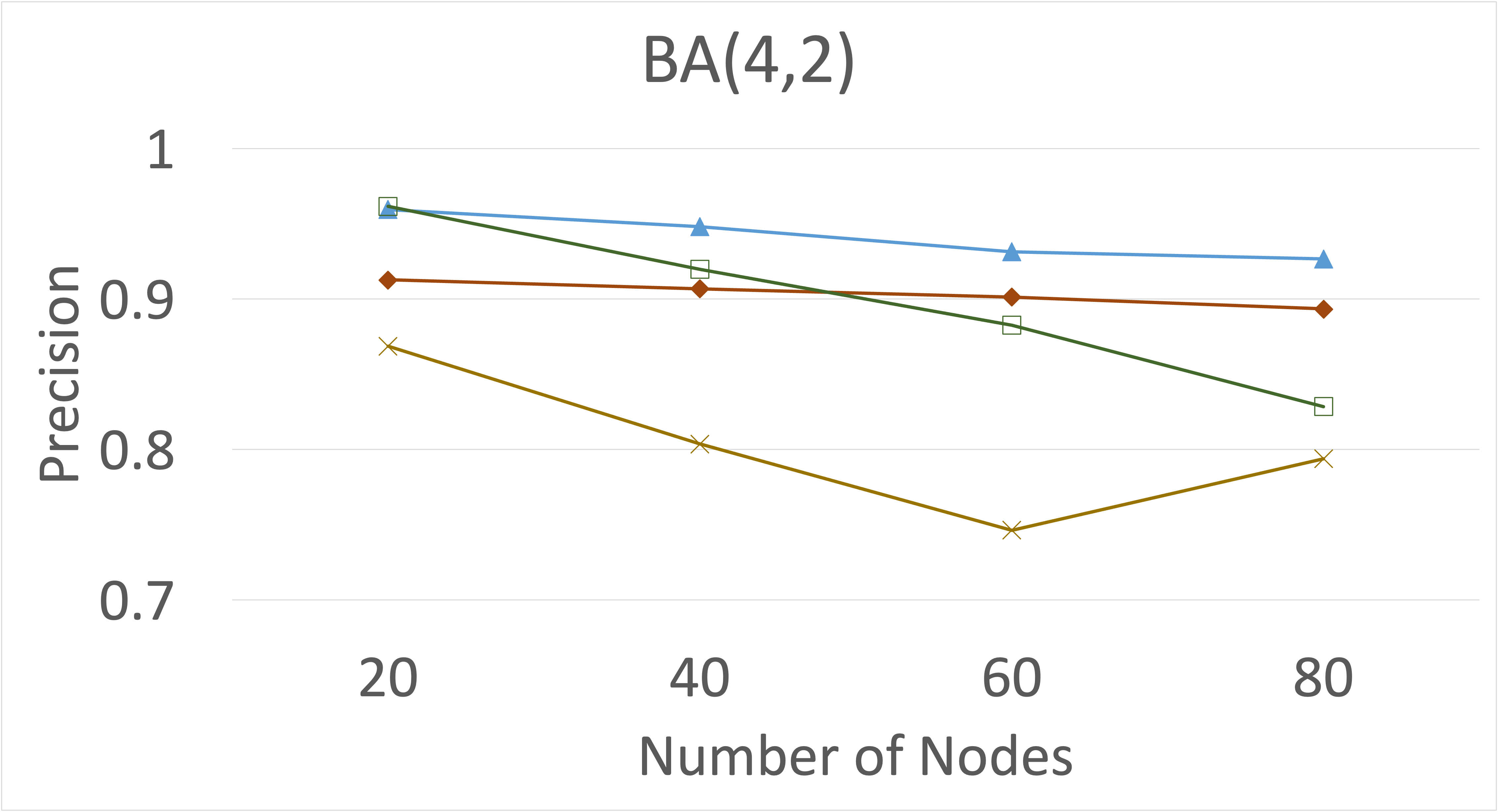}
      \caption{Additive Index Model, $T=1000$}
    \end{subfigure}
    \centering
    \begin{subfigure}{0.49\textwidth}
      \centering
      \includegraphics[width=0.49\textwidth, height=0.35\textwidth]{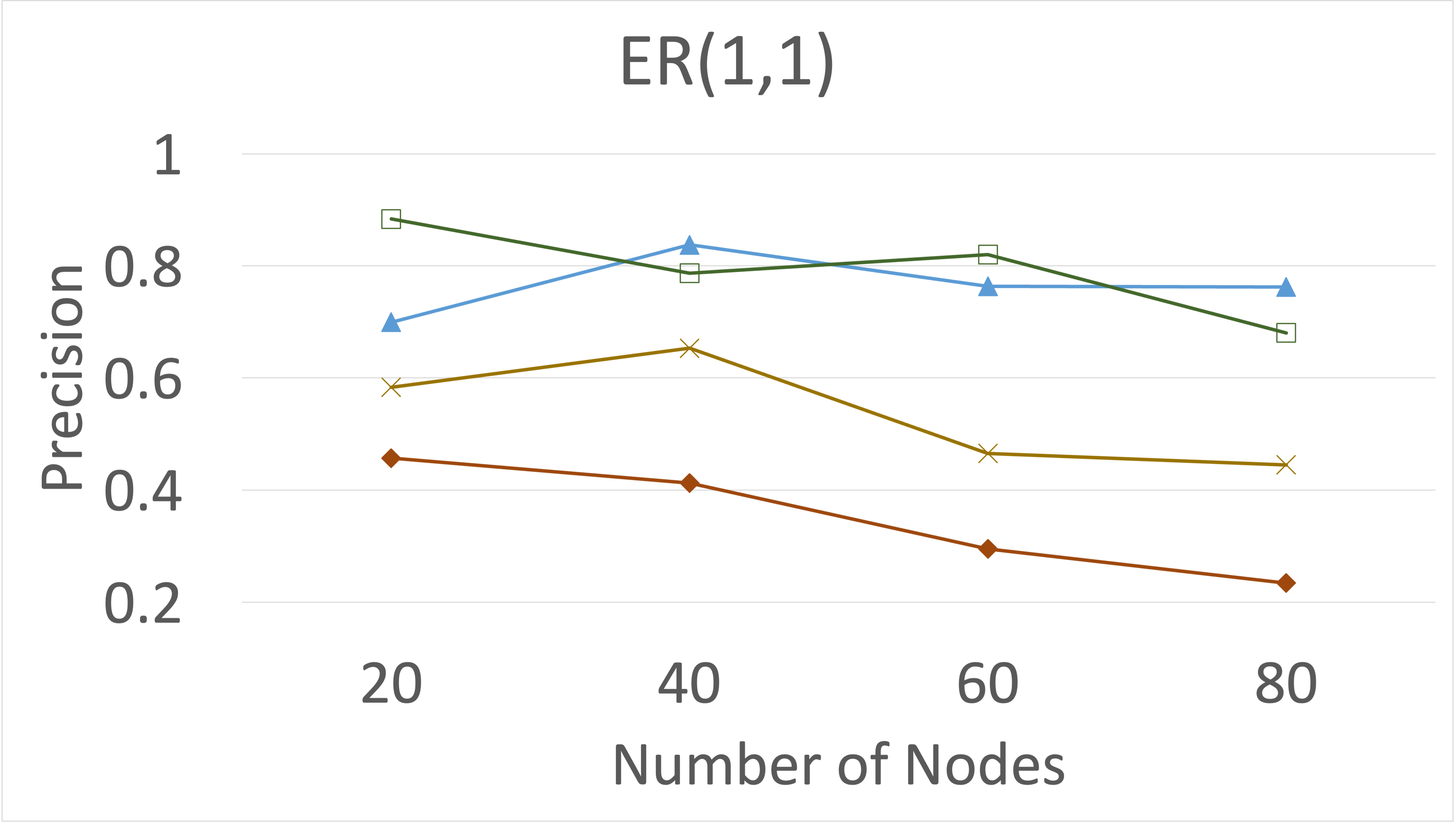}
      \includegraphics[width=0.49\textwidth, height=0.35\textwidth]{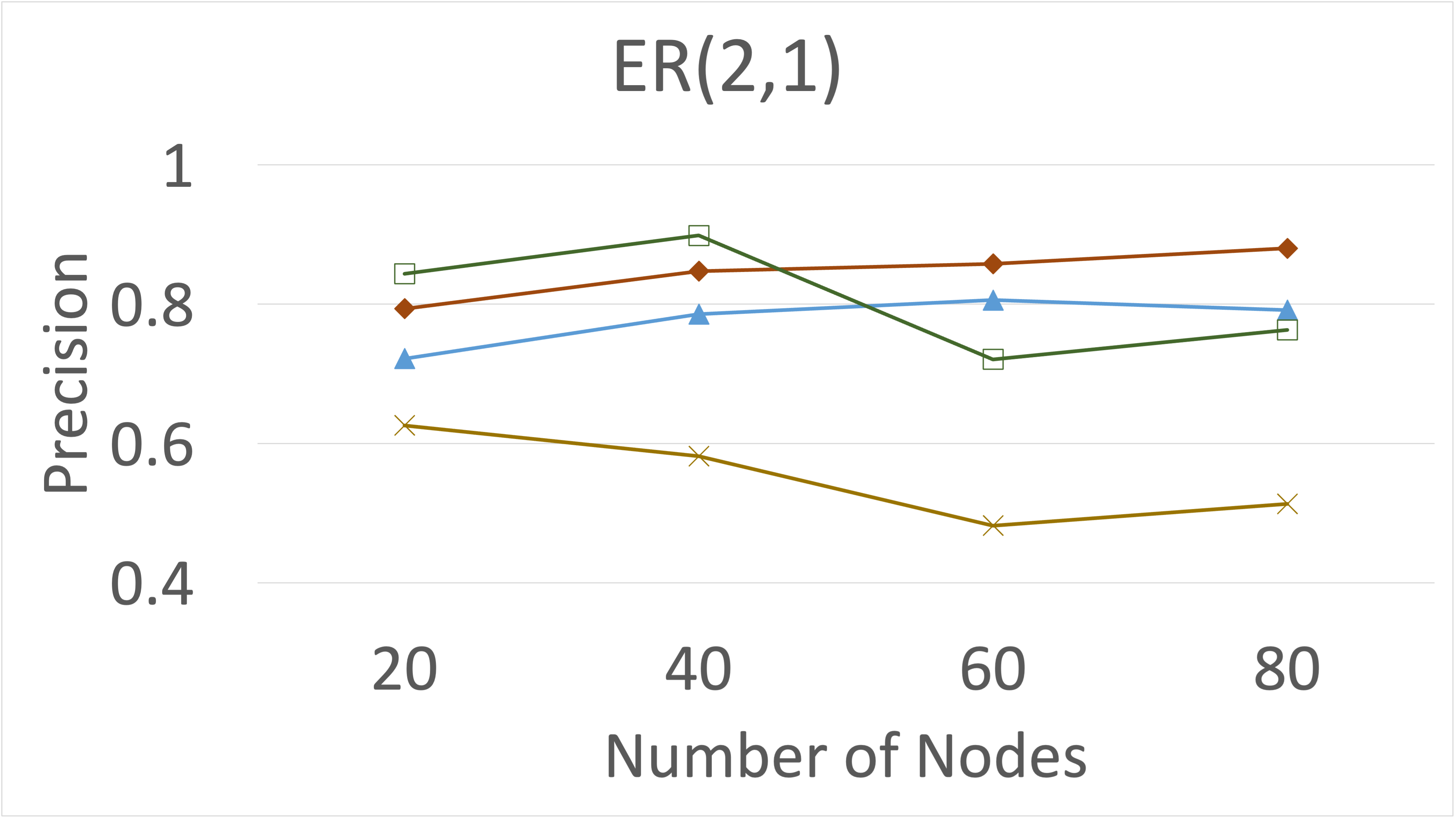}
      \includegraphics[width=0.49\textwidth, height=0.35\textwidth]{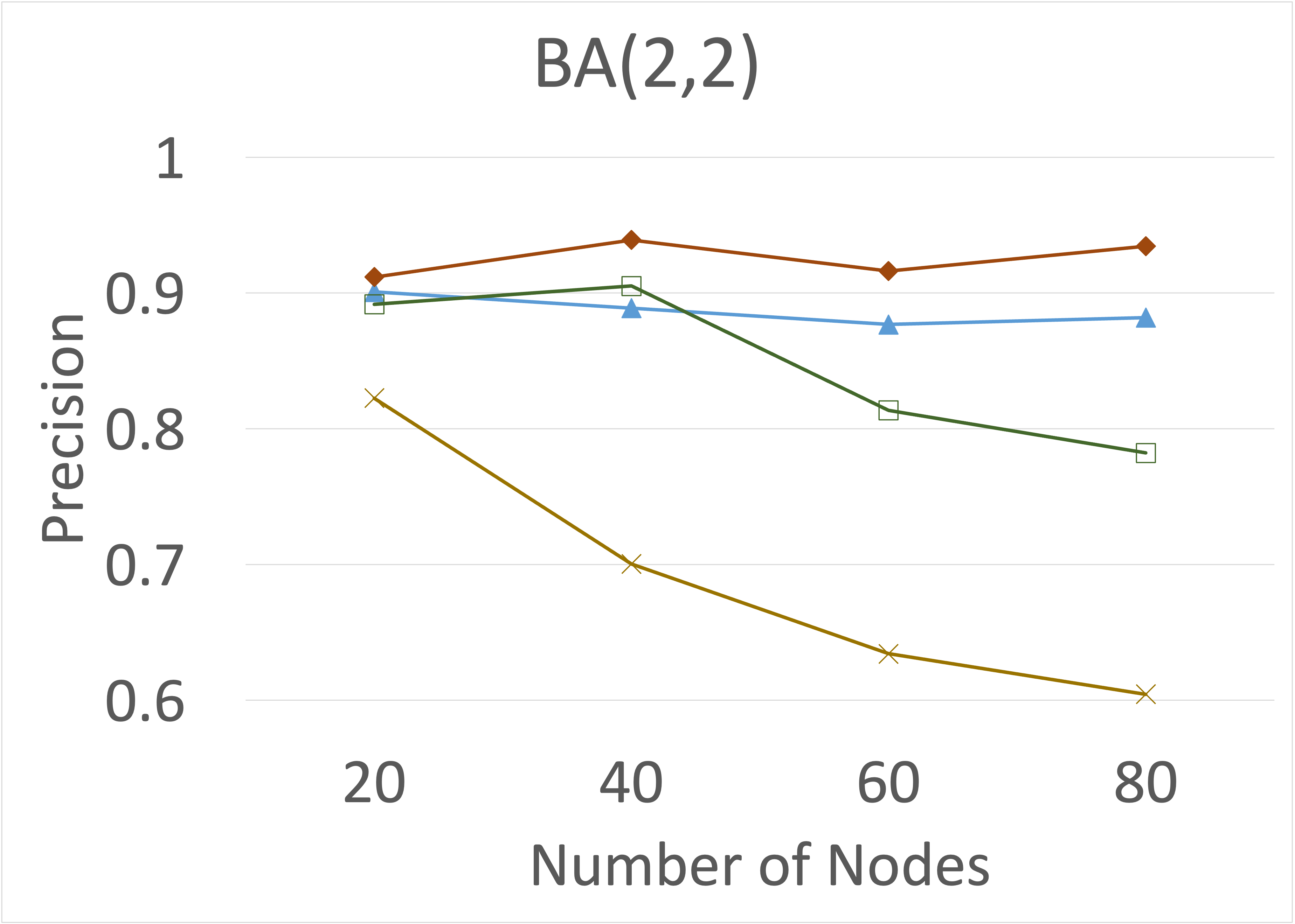}
      \includegraphics[width=0.49\textwidth, height=0.35\textwidth]{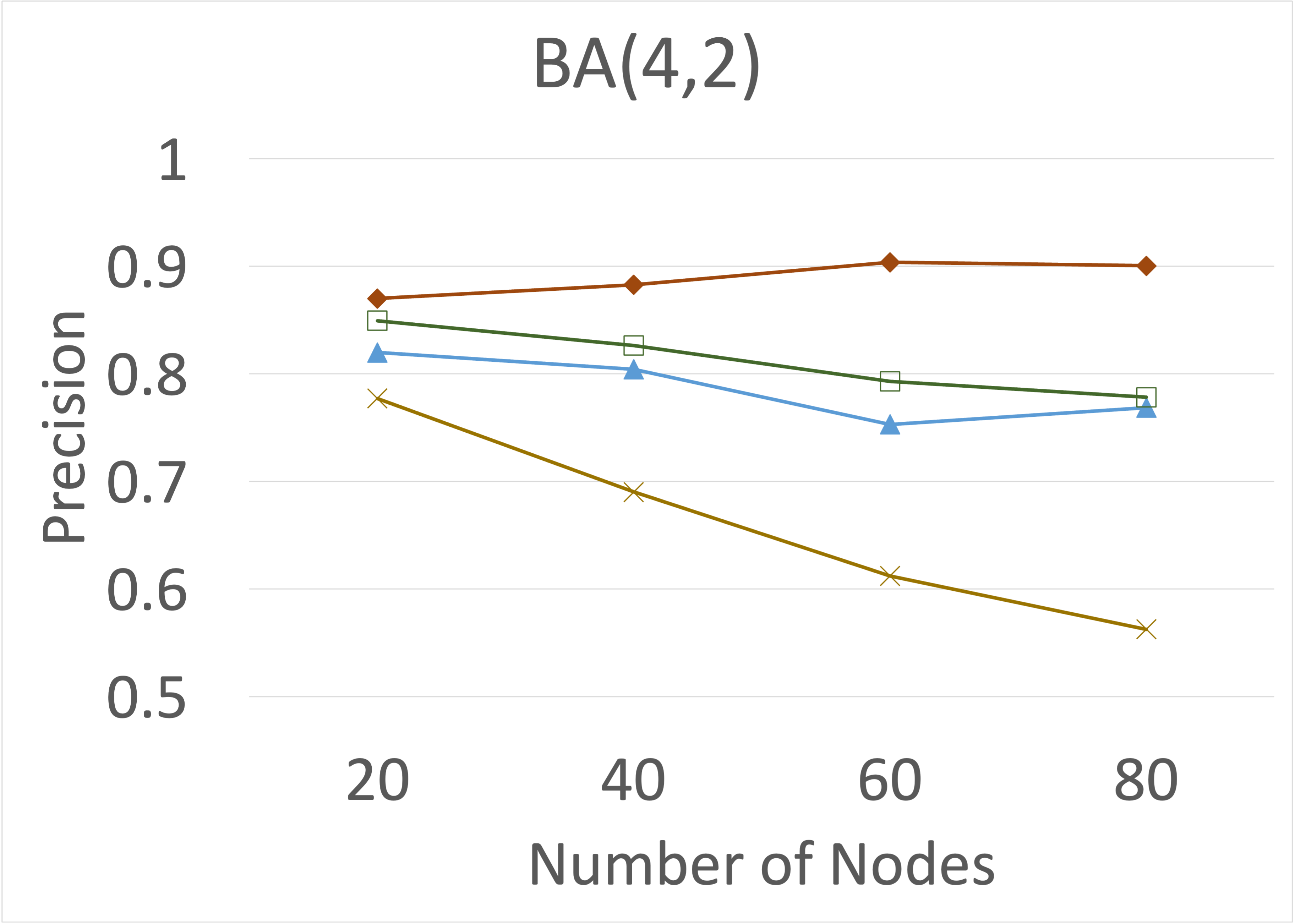}
      \caption{Additive Noise Model, $T=200$}
    \end{subfigure}
    \hfill
    \centering
    \begin{subfigure}{0.49\textwidth}
      \centering
      \includegraphics[width=0.49\textwidth, height=0.35\textwidth]{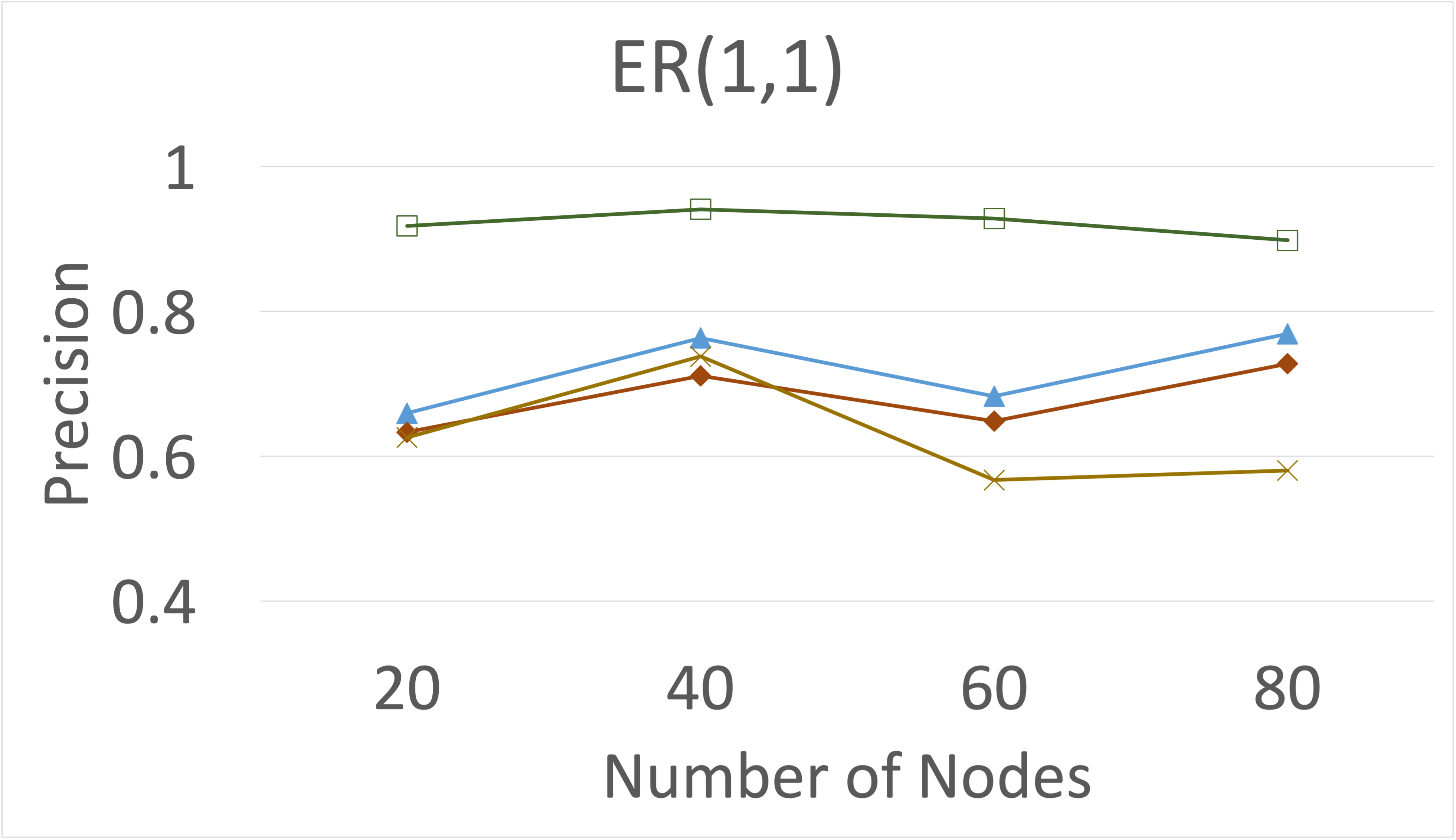}
      \includegraphics[width=0.49\textwidth, height=0.35\textwidth]{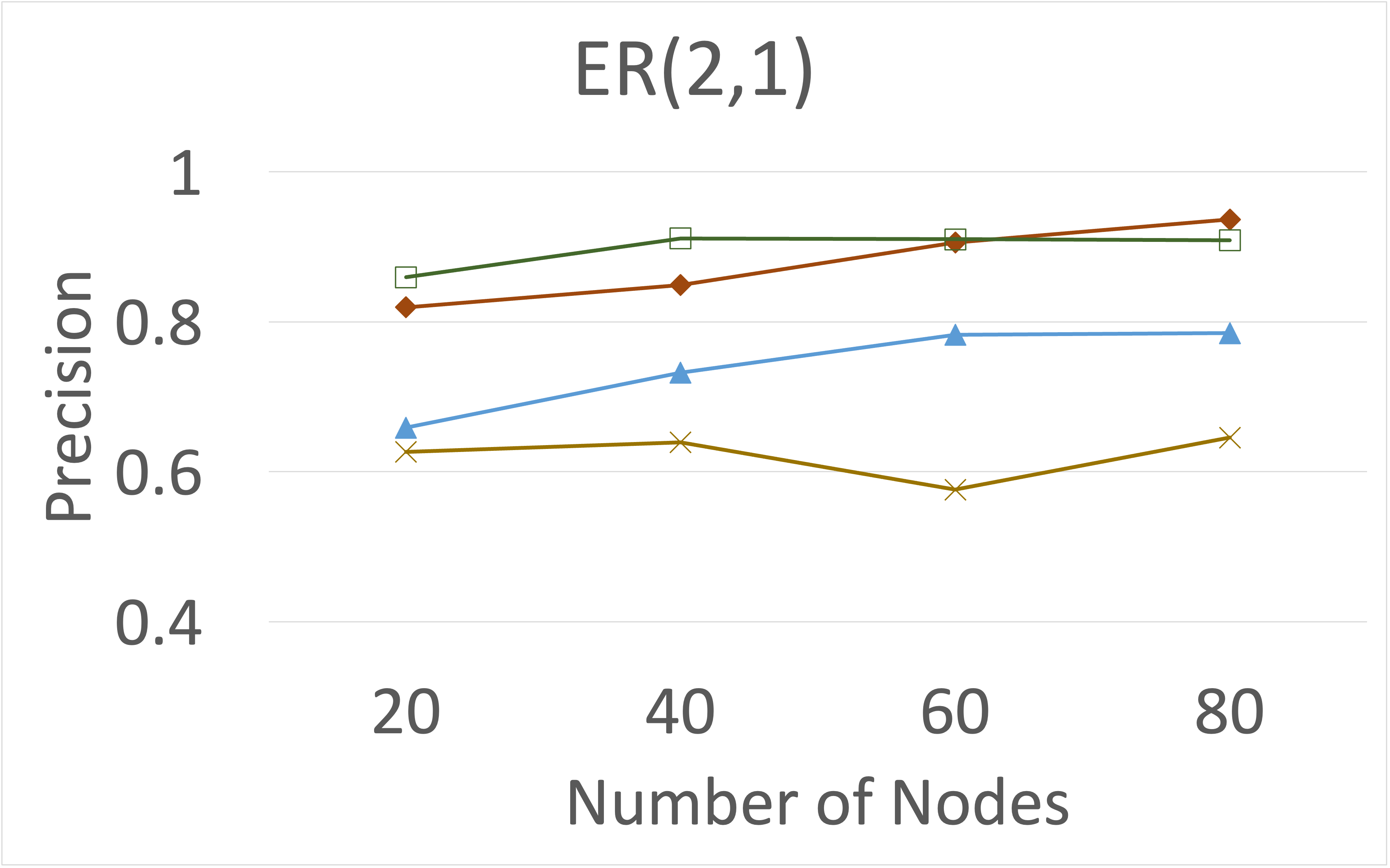}
      \includegraphics[width=0.49\textwidth, height=0.35\textwidth]{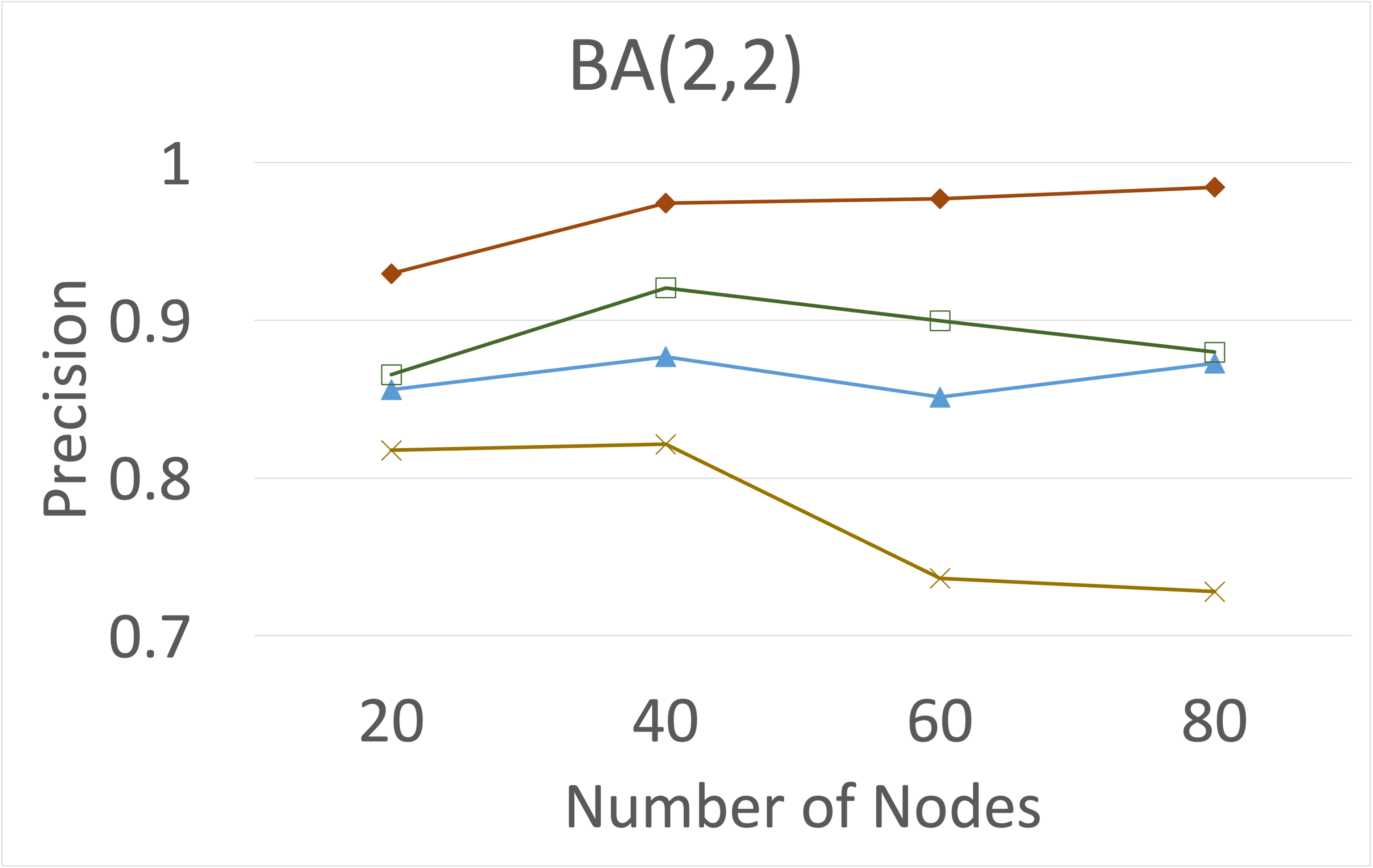}
      \includegraphics[width=0.49\textwidth, height=0.35\textwidth]{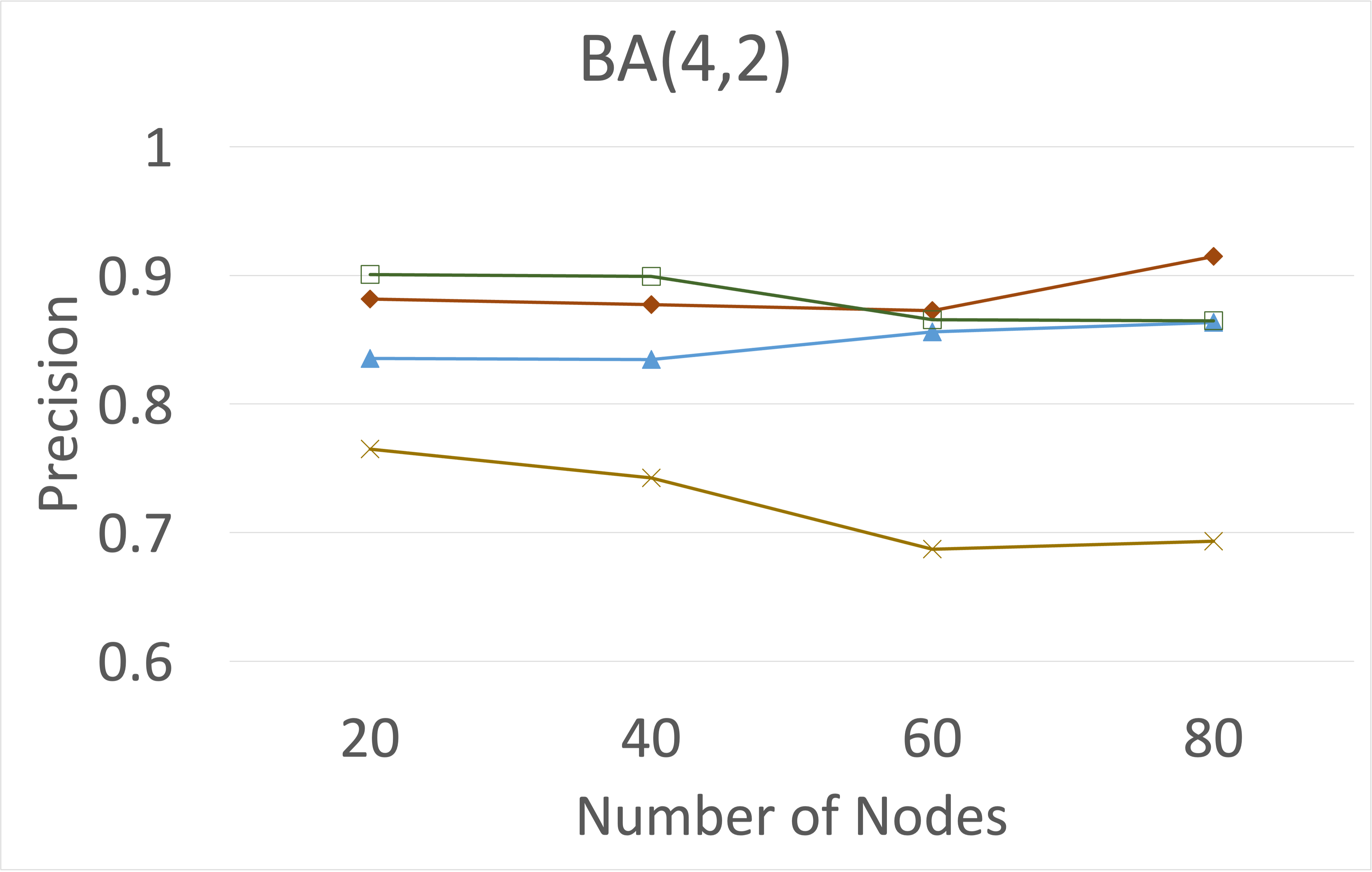}
      \caption{Additive Noise Model, $T=1000$}
    \end{subfigure}
    \centering
    \begin{subfigure}{0.49\textwidth}
      \centering
      \includegraphics[width=0.49\textwidth, height=0.35\textwidth]{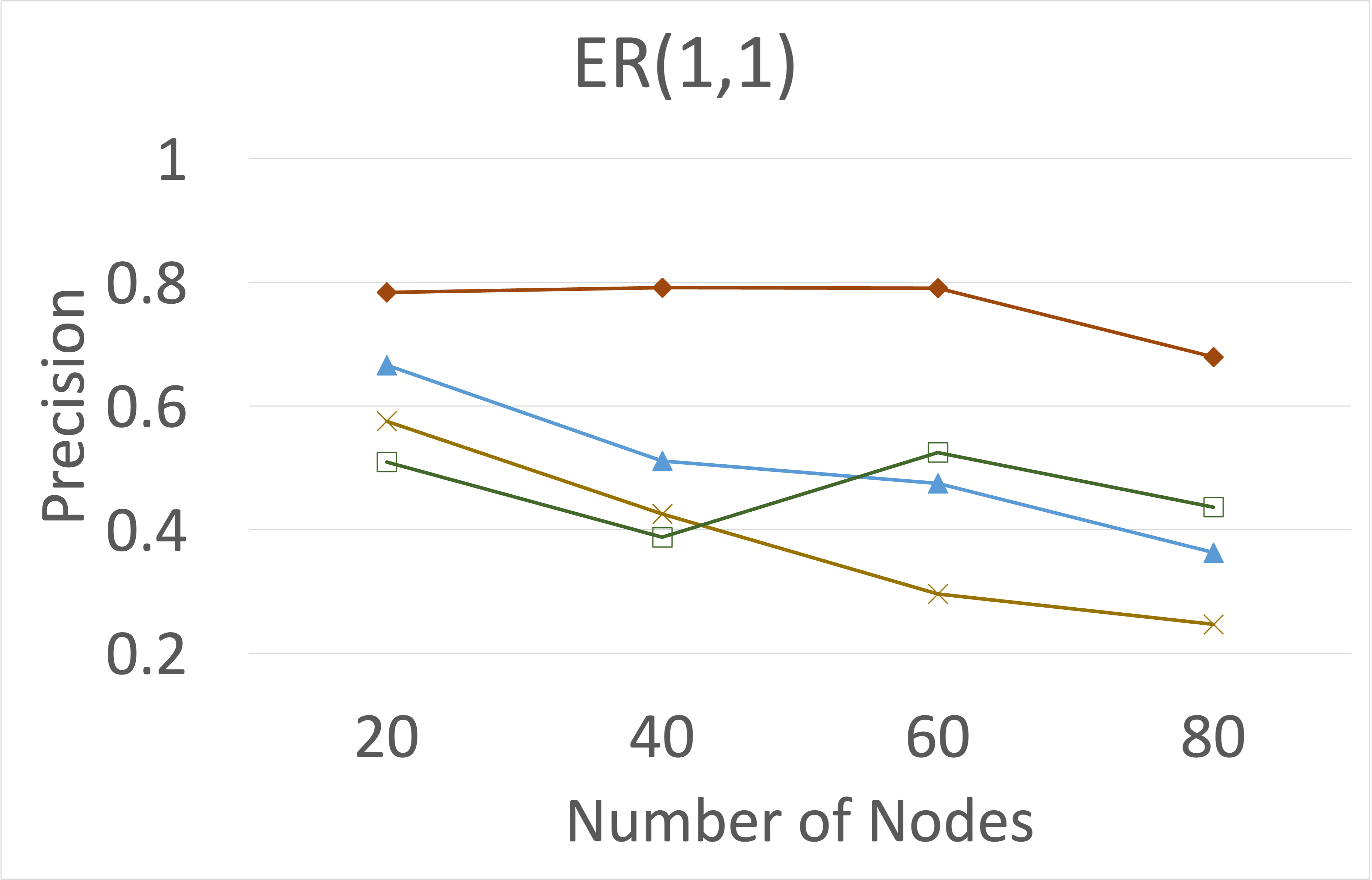}
      \includegraphics[width=0.49\textwidth, height=0.35\textwidth]{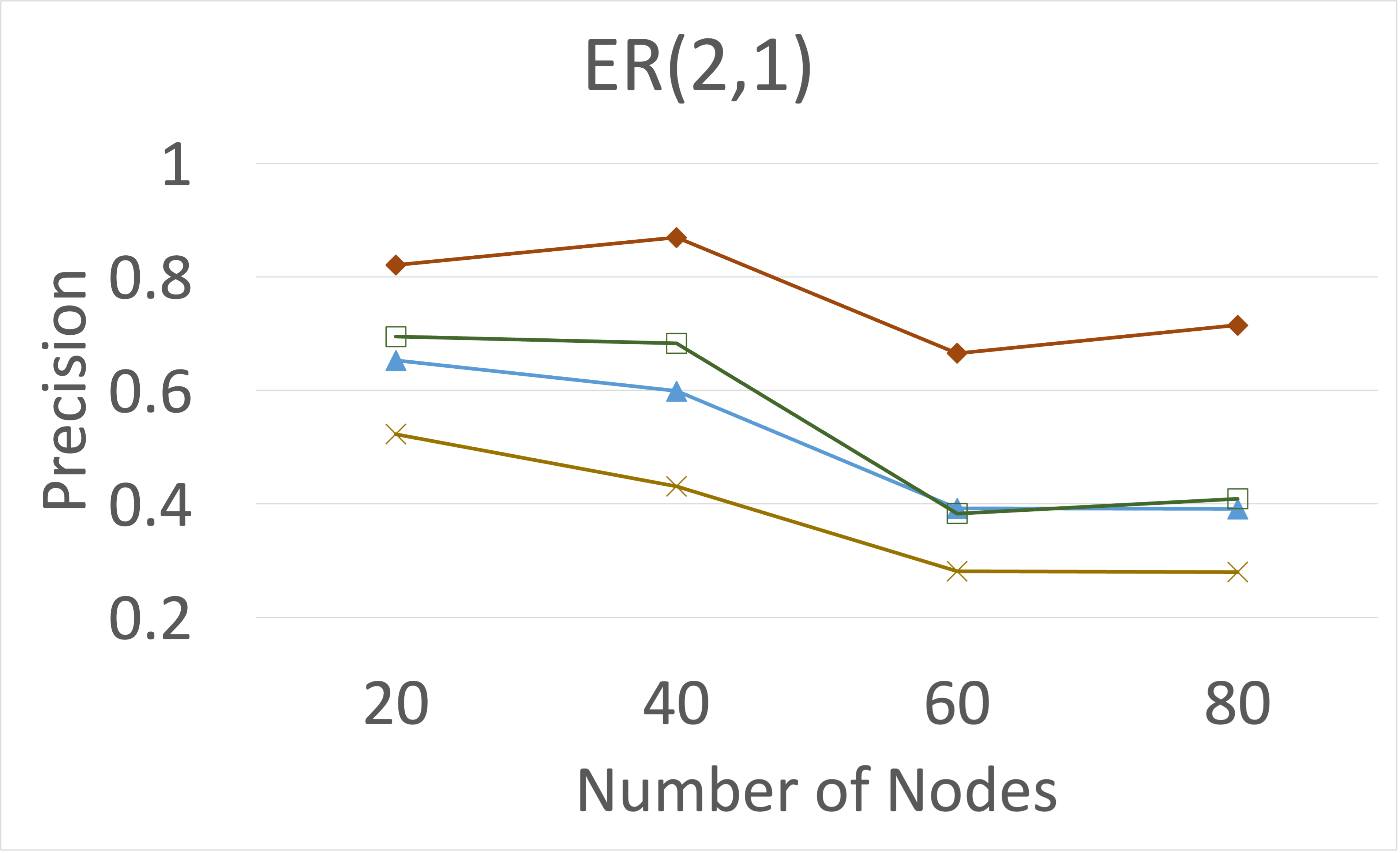}
      \includegraphics[width=0.49\textwidth, height=0.35\textwidth]{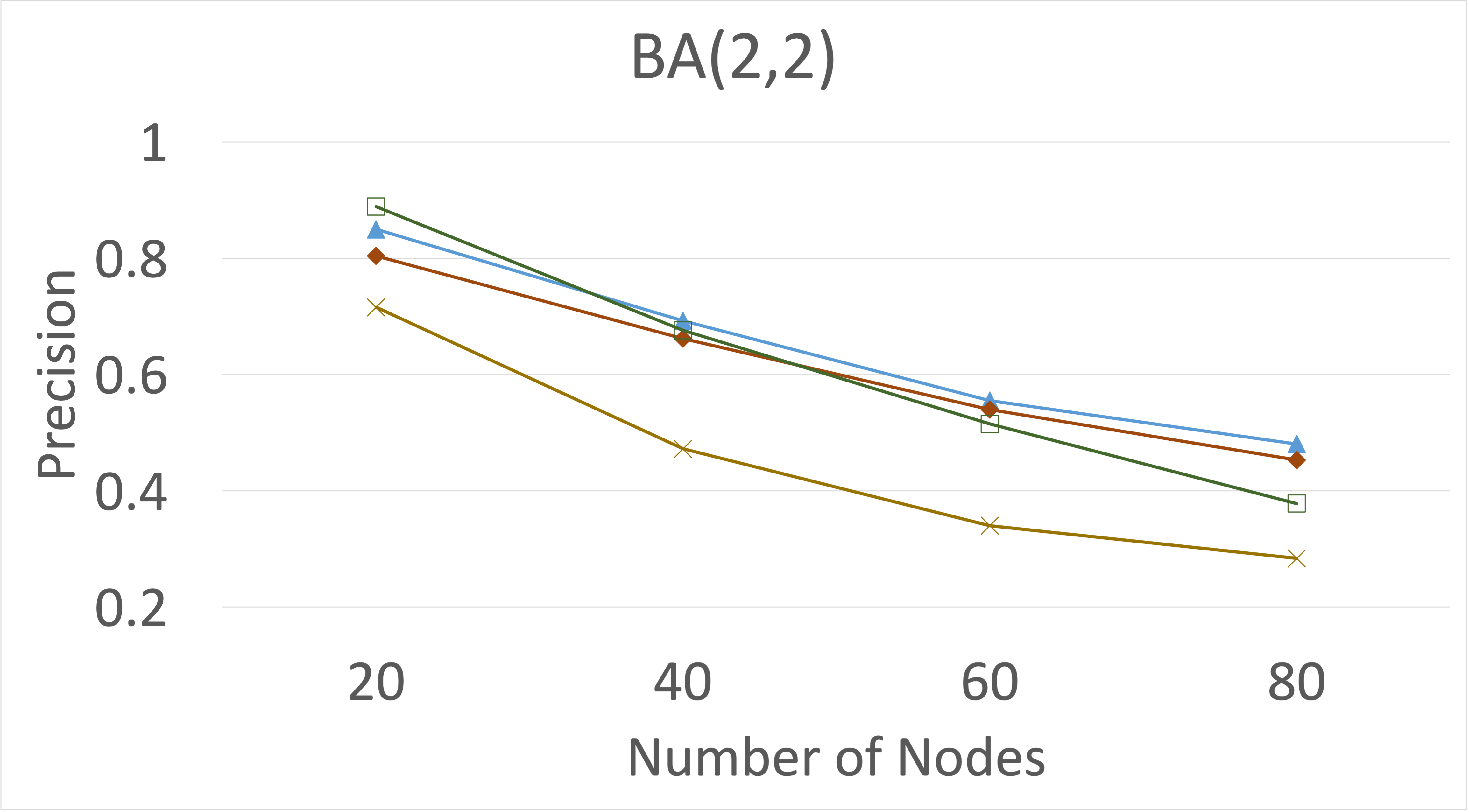}
      \includegraphics[width=0.49\textwidth, height=0.35\textwidth]{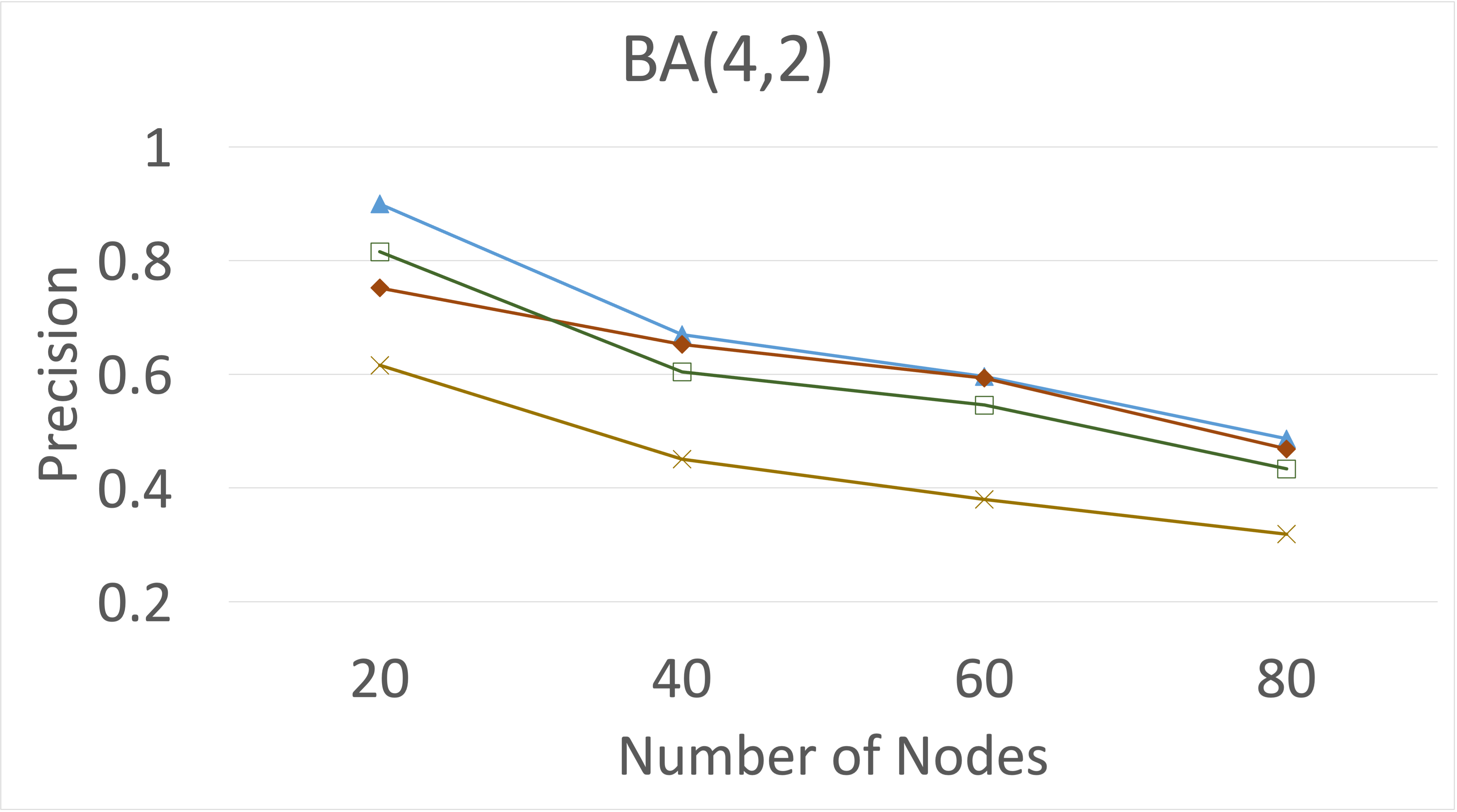}
      \caption{Generalized Linear Model with Poisson
  Distribution, $T=200$}
    \end{subfigure}
    \hfill
    \centering
    \begin{subfigure}{0.49\textwidth}
      \centering
      \includegraphics[width=0.49\textwidth, height=0.35\textwidth]{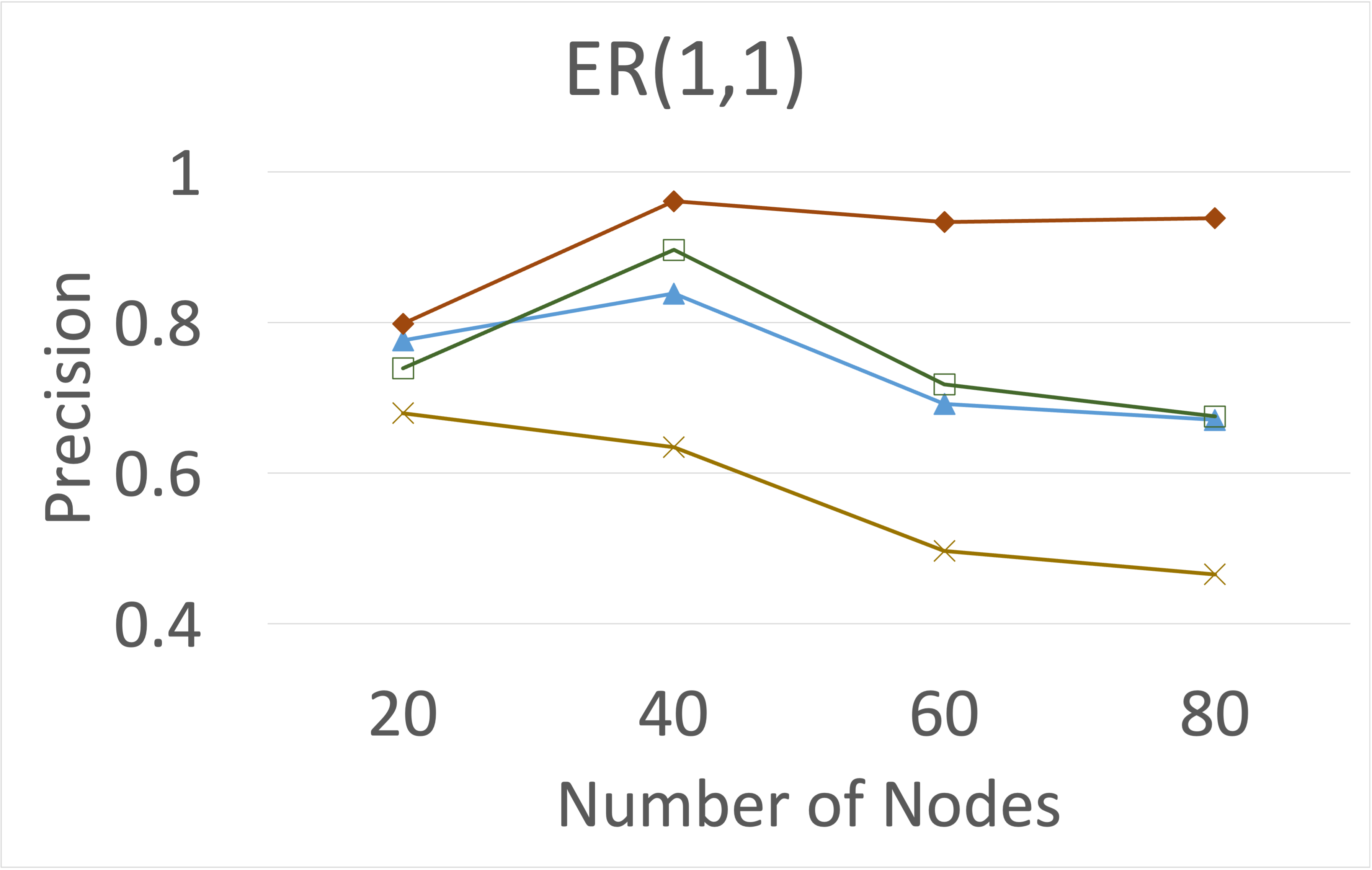}
      \includegraphics[width=0.49\textwidth, height=0.35\textwidth]{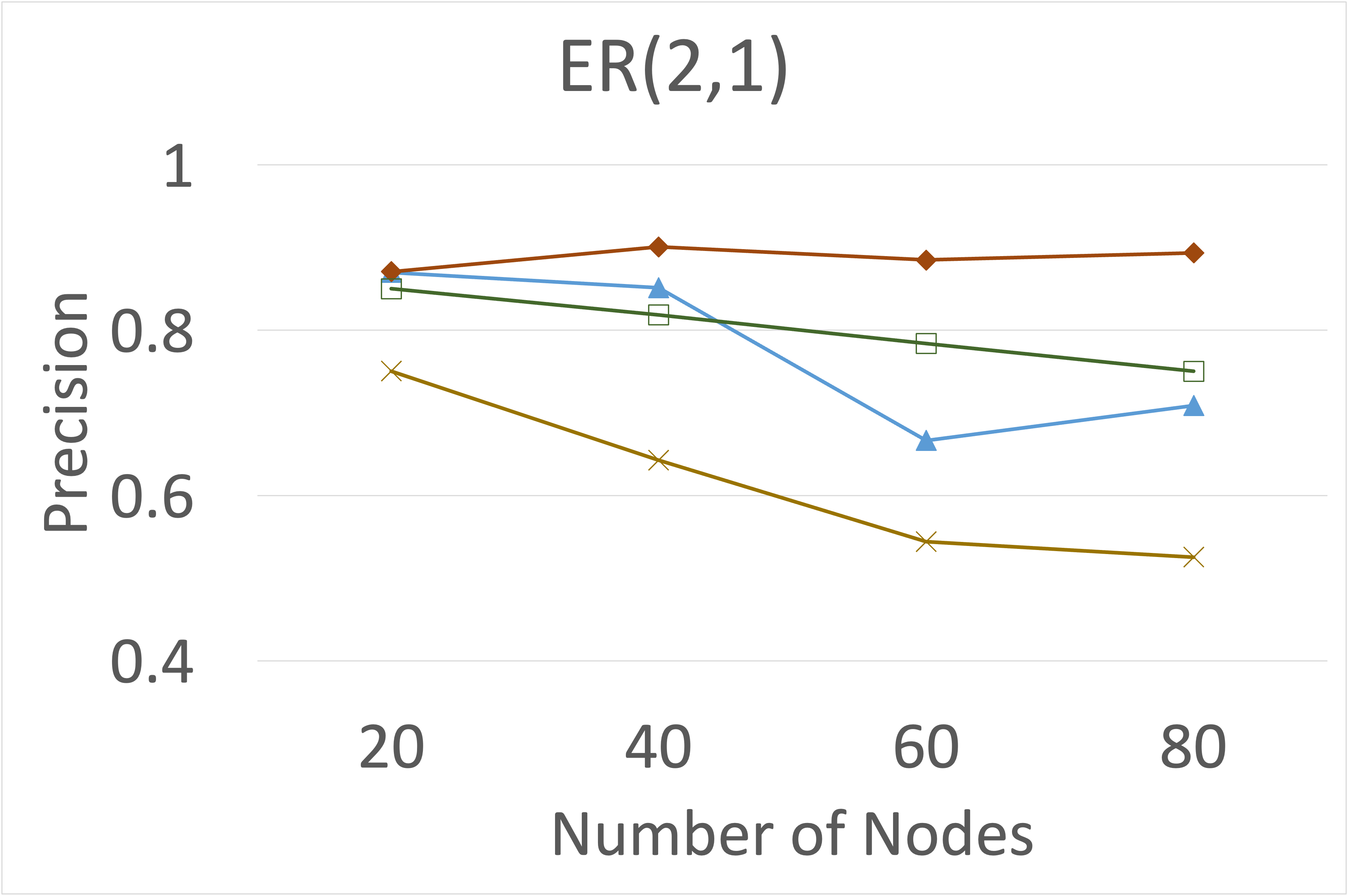}
      \includegraphics[width=0.49\textwidth, height=0.35\textwidth]{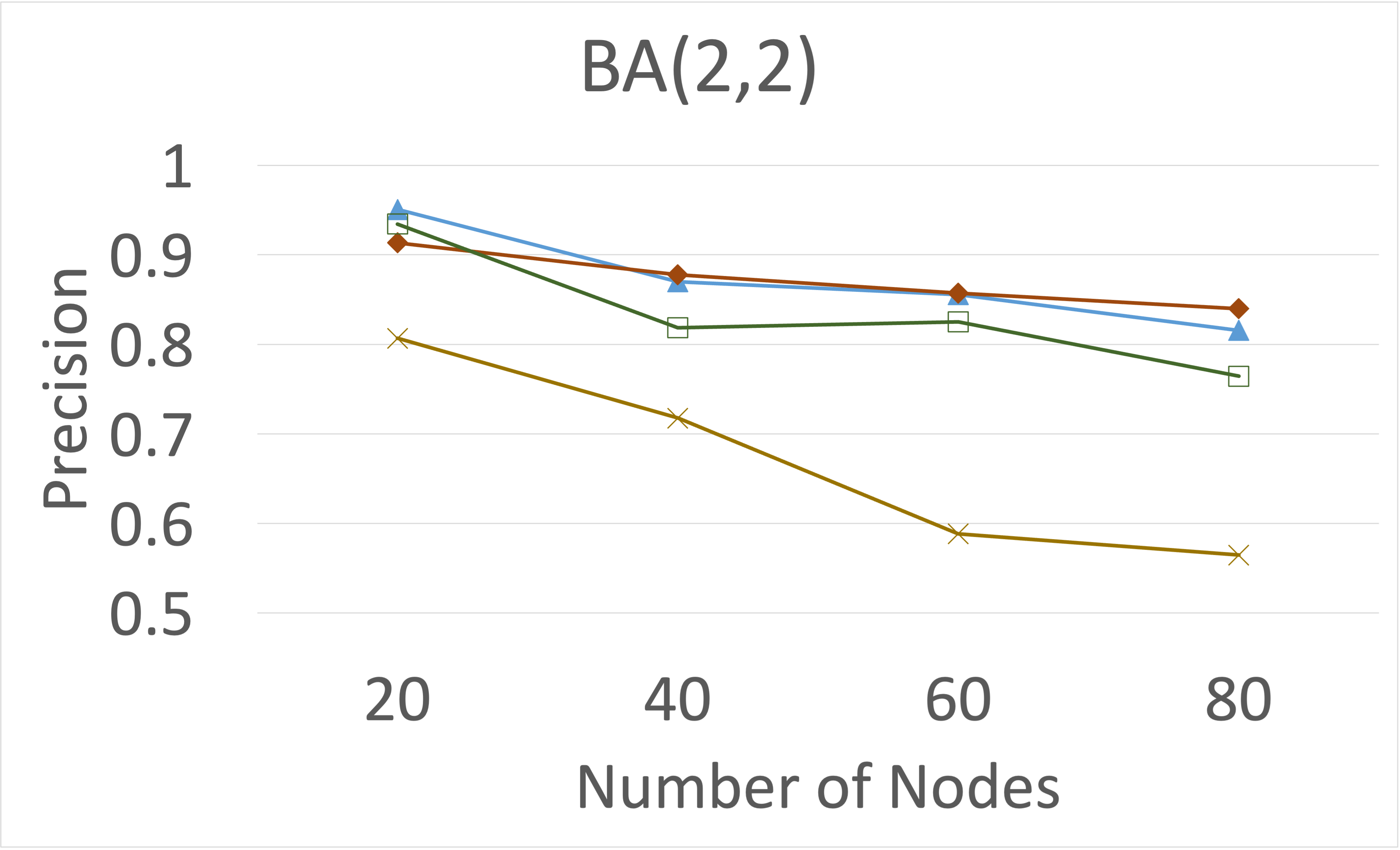}
      \includegraphics[width=0.49\textwidth, height=0.35\textwidth]{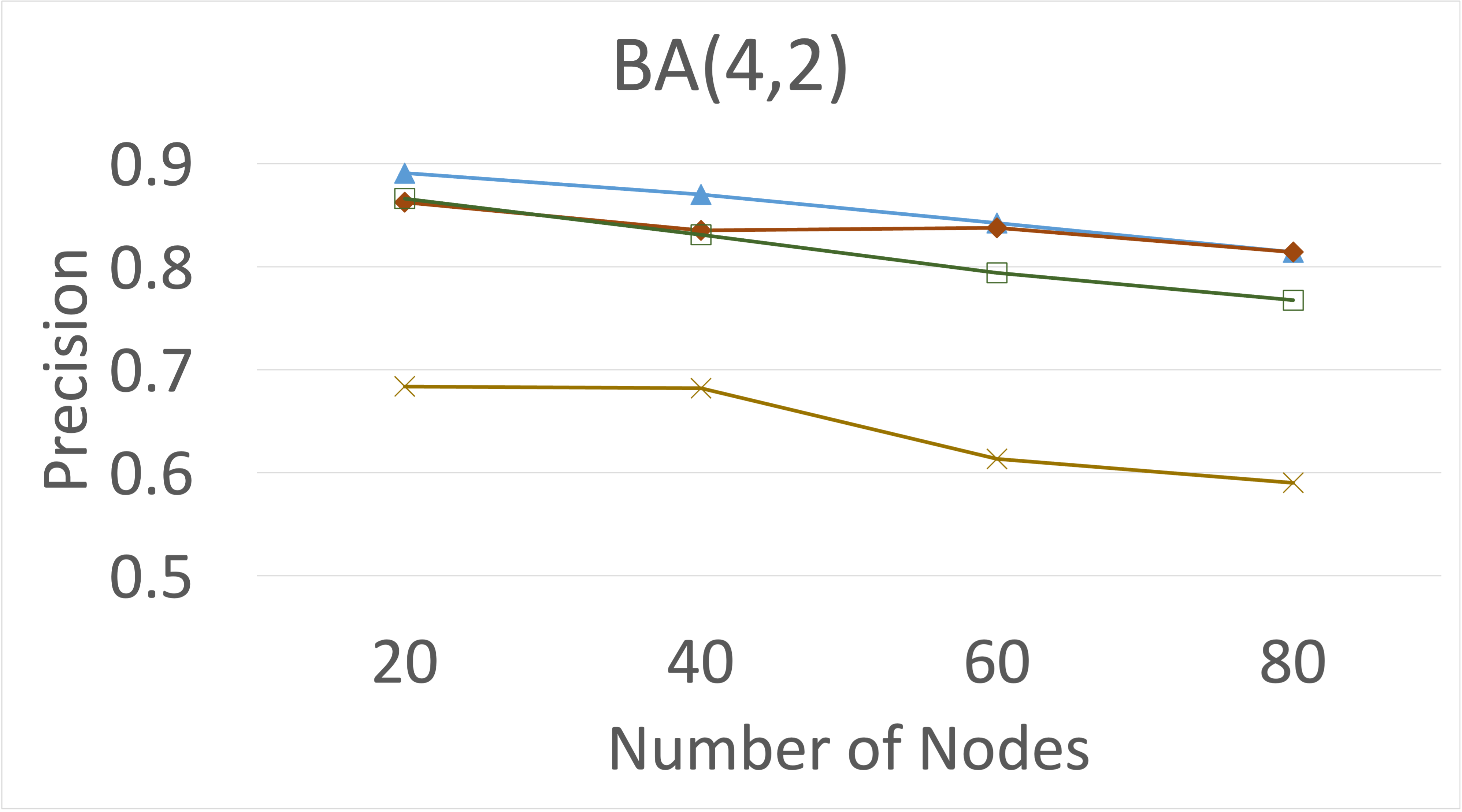}
      \caption{Generalized Linear Model with Poisson
  Distribution, $T=1000$}
    \end{subfigure}
    \centering
    \begin{subfigure}{1.0\textwidth}
      \centering
      \includegraphics[width=0.5\textwidth, height=0.03\textwidth]{images/Simulated_Data/legend.png}
    \end{subfigure}
    \caption{\textbf{Mean precisions} over 10 datasets for each setting with simulated data. Higher Precision is better. The number of lags $= 3$.}
    \label{fig:experiments_simulated_precision}
  \end{figure*}

\begin{figure*}[ht]
  \centering
  \begin{subfigure}{0.49\textwidth}
    \centering
    \includegraphics[width=0.49\textwidth, height=0.35\textwidth]{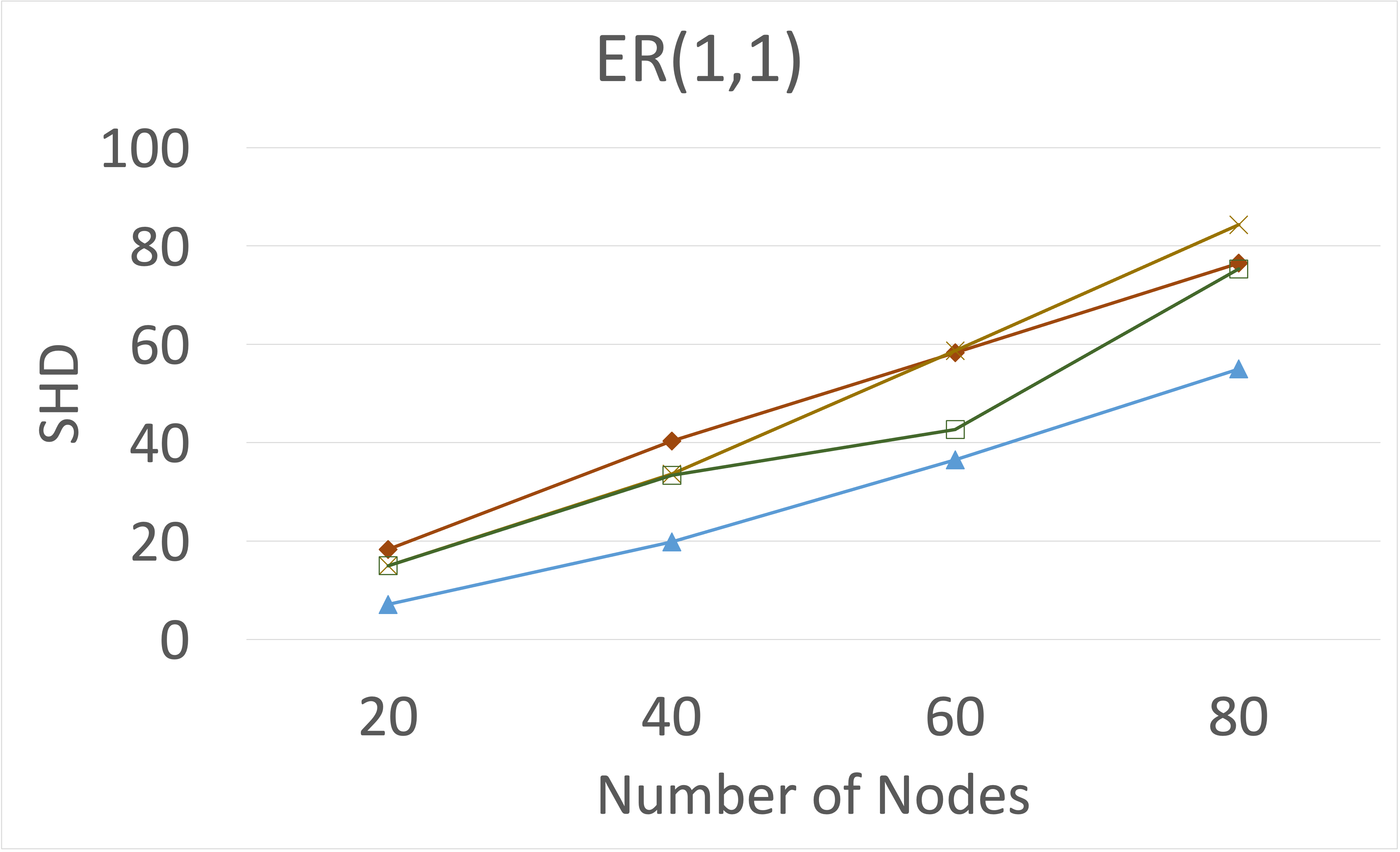}
    \includegraphics[width=0.49\textwidth, height=0.35\textwidth]{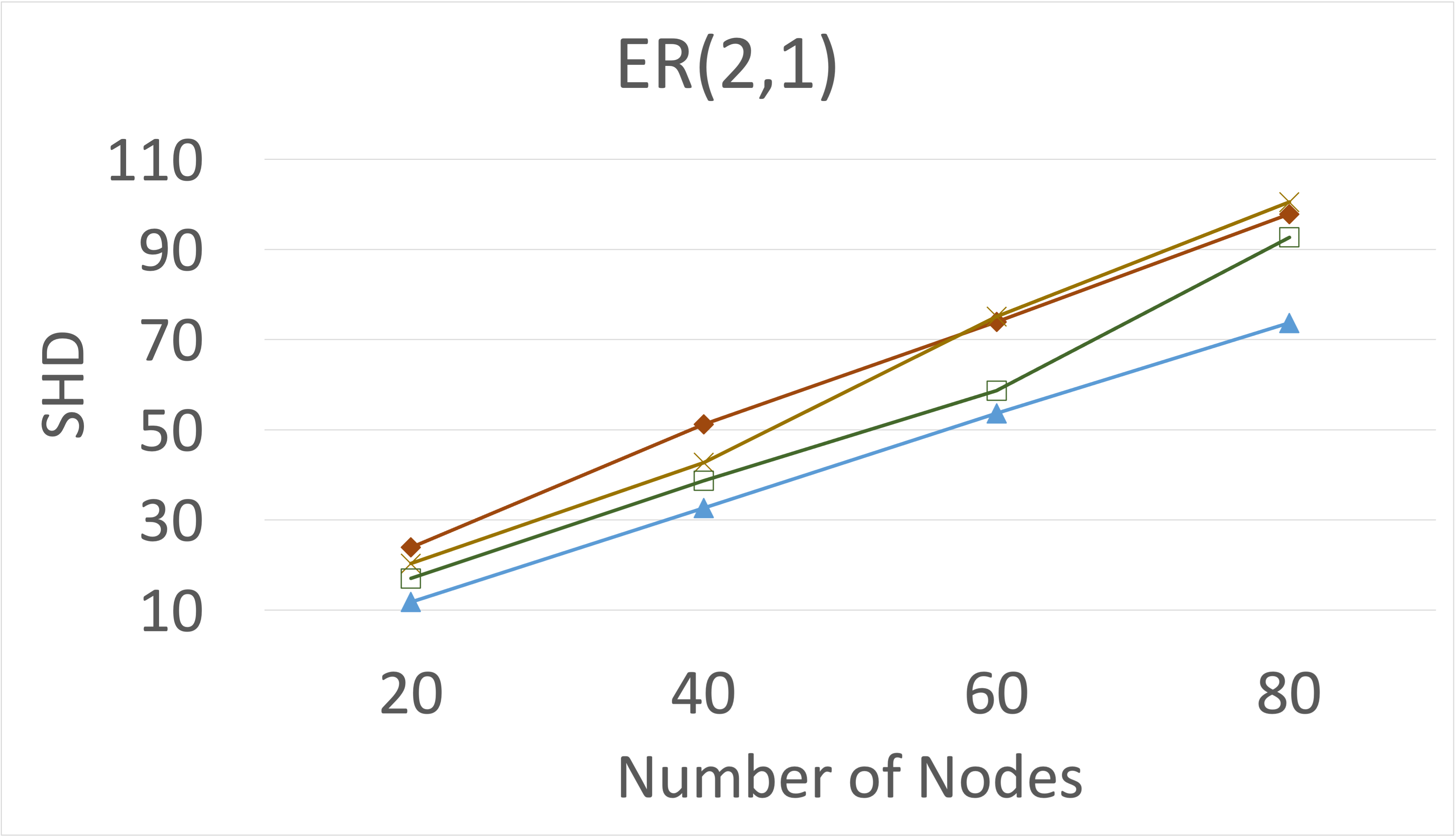}
    \includegraphics[width=0.49\textwidth, height=0.35\textwidth]{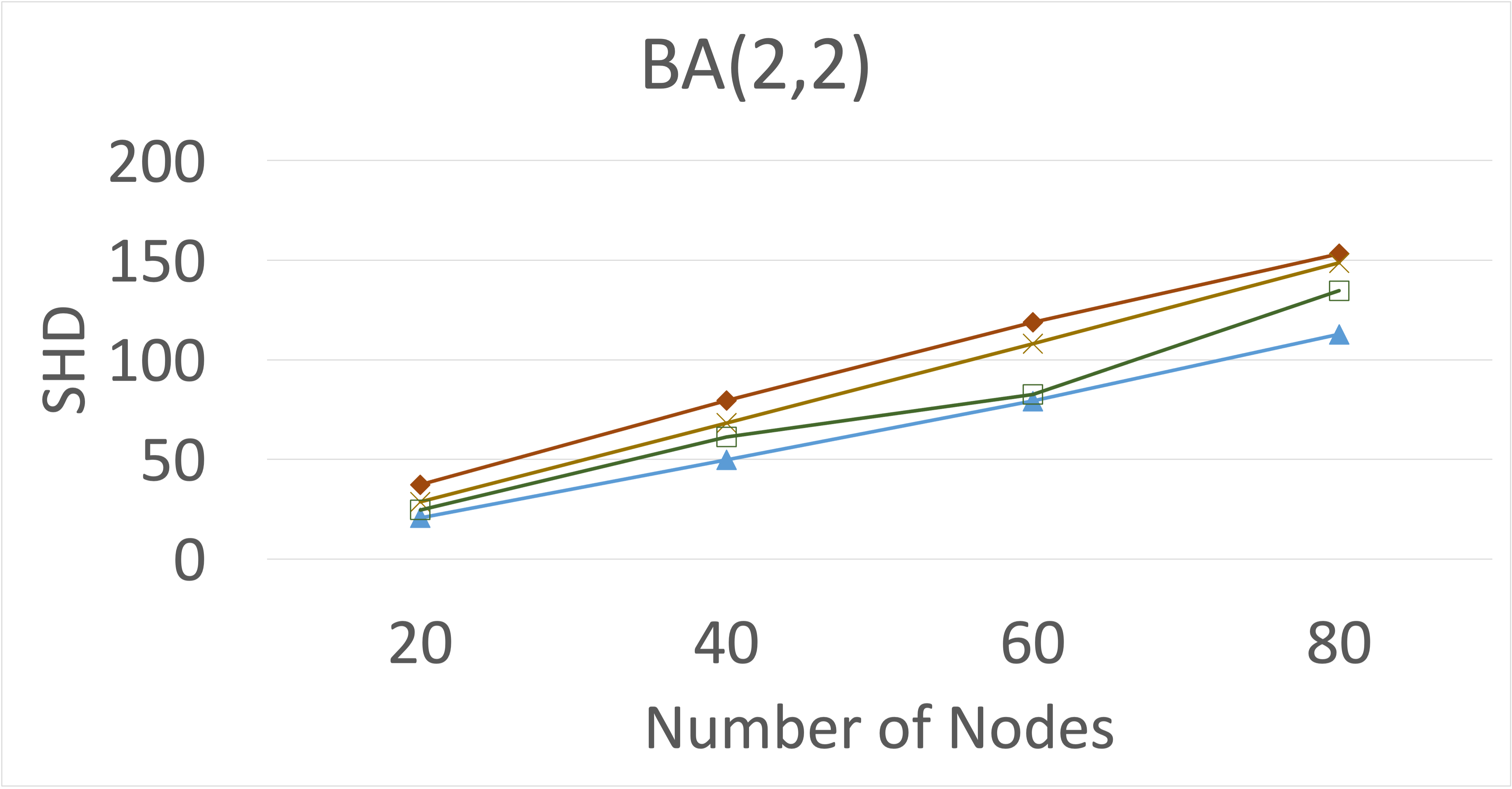}
    \includegraphics[width=0.49\textwidth, height=0.35\textwidth]{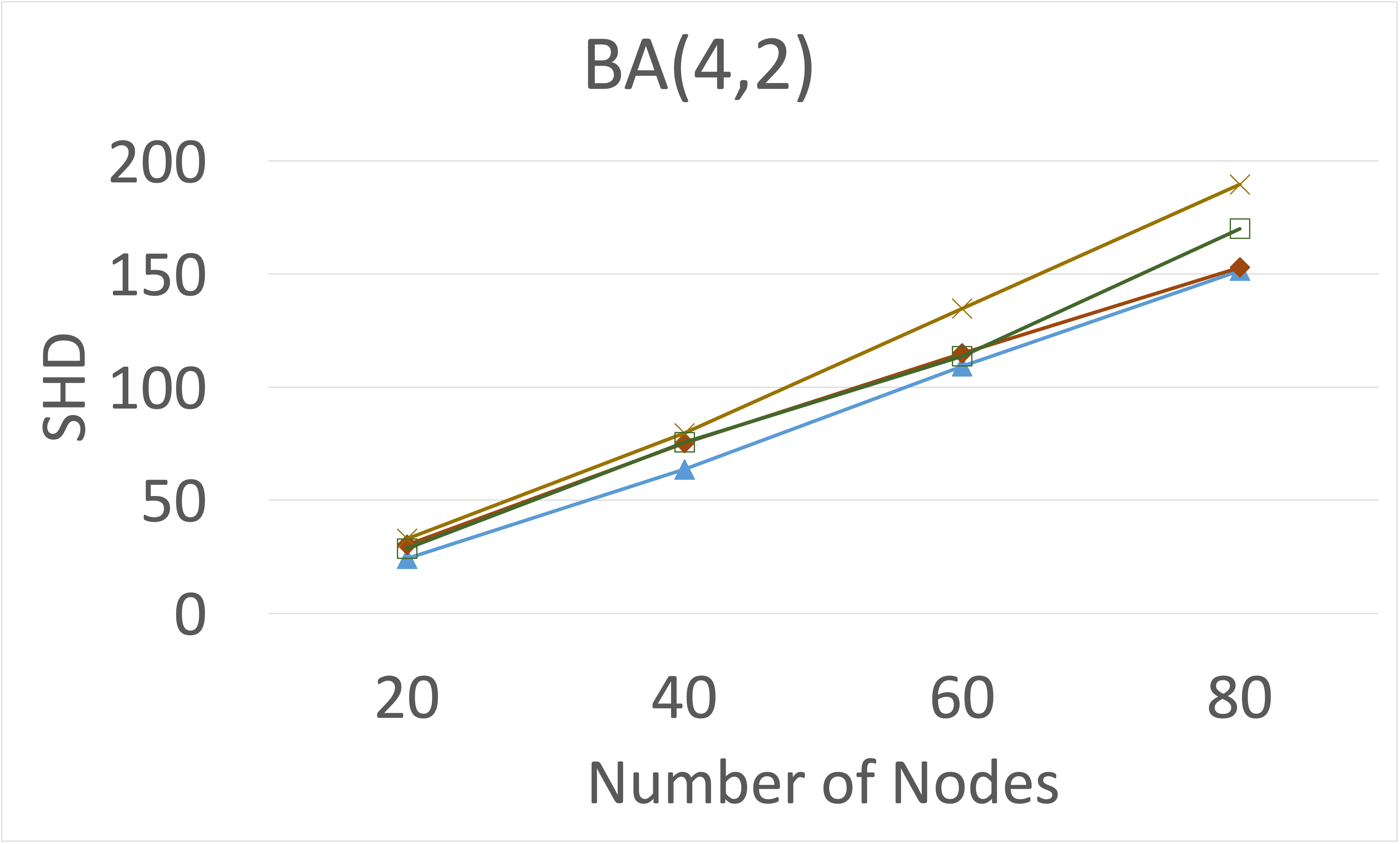}
    \caption{Additive Index Model, $T=200$}
  \end{subfigure}
  \hfill
  \centering
  \begin{subfigure}{0.49\textwidth}
    \centering
    \includegraphics[width=0.49\textwidth, height=0.35\textwidth]{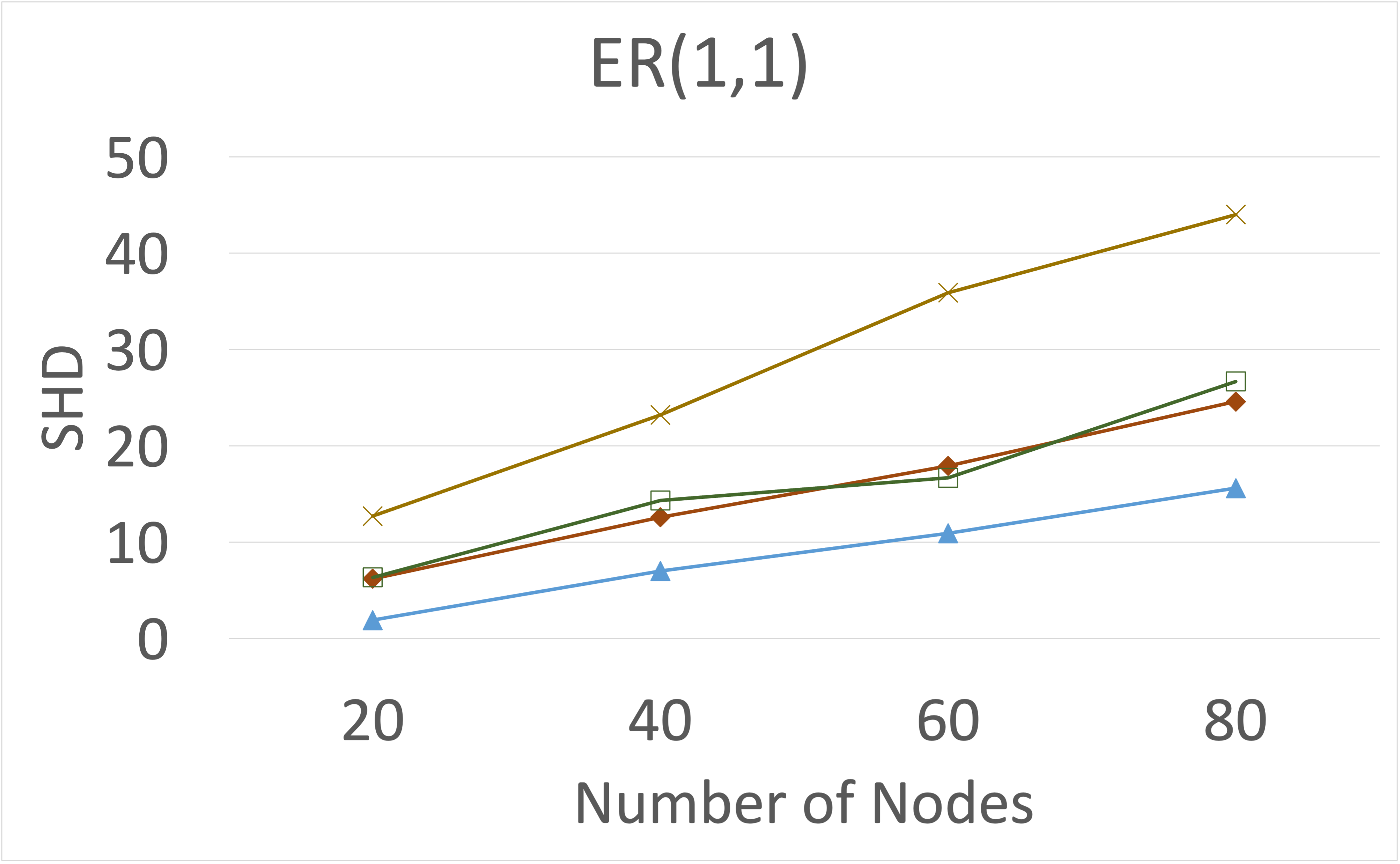}
    \includegraphics[width=0.49\textwidth, height=0.35\textwidth]{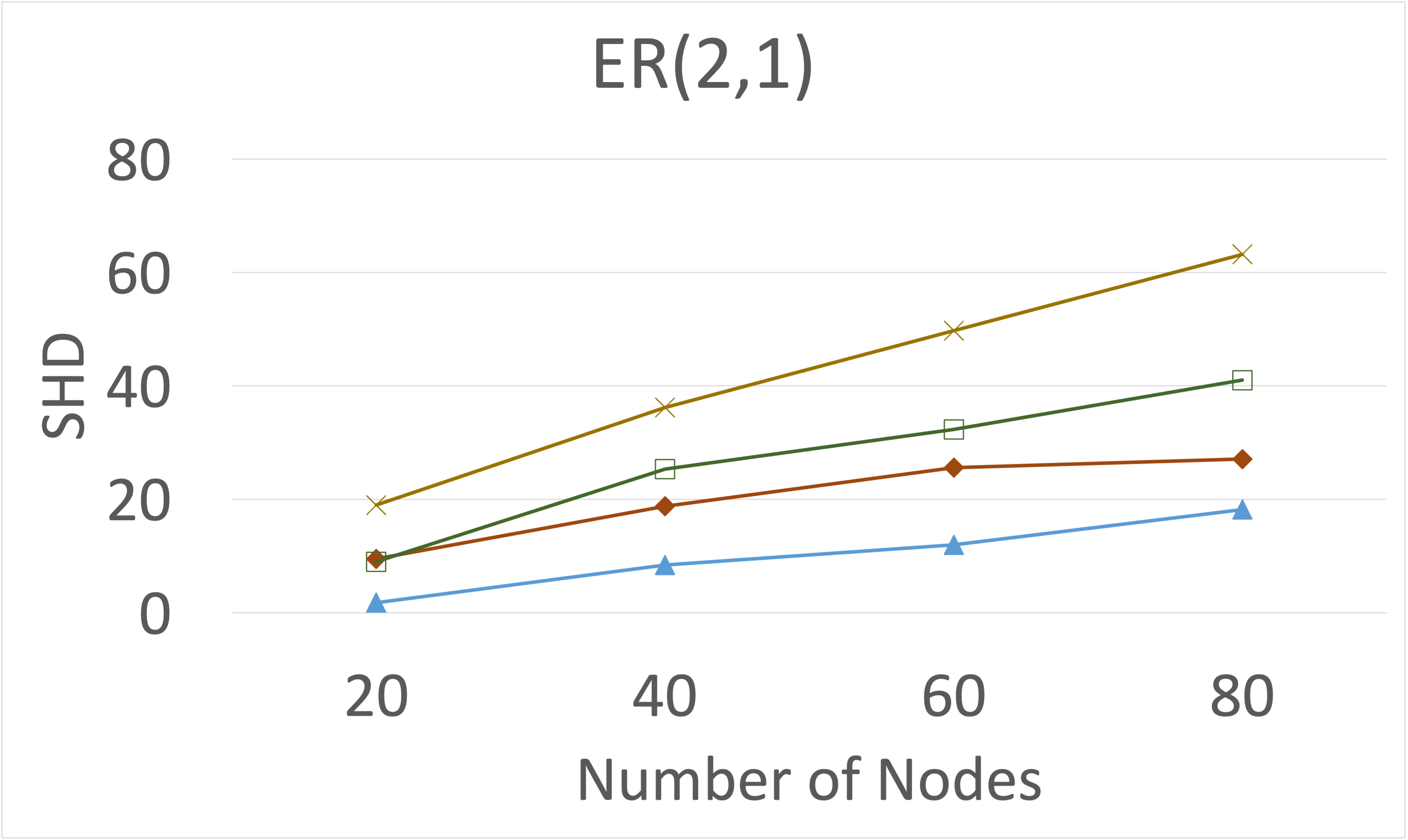}
    \includegraphics[width=0.49\textwidth, height=0.35\textwidth]{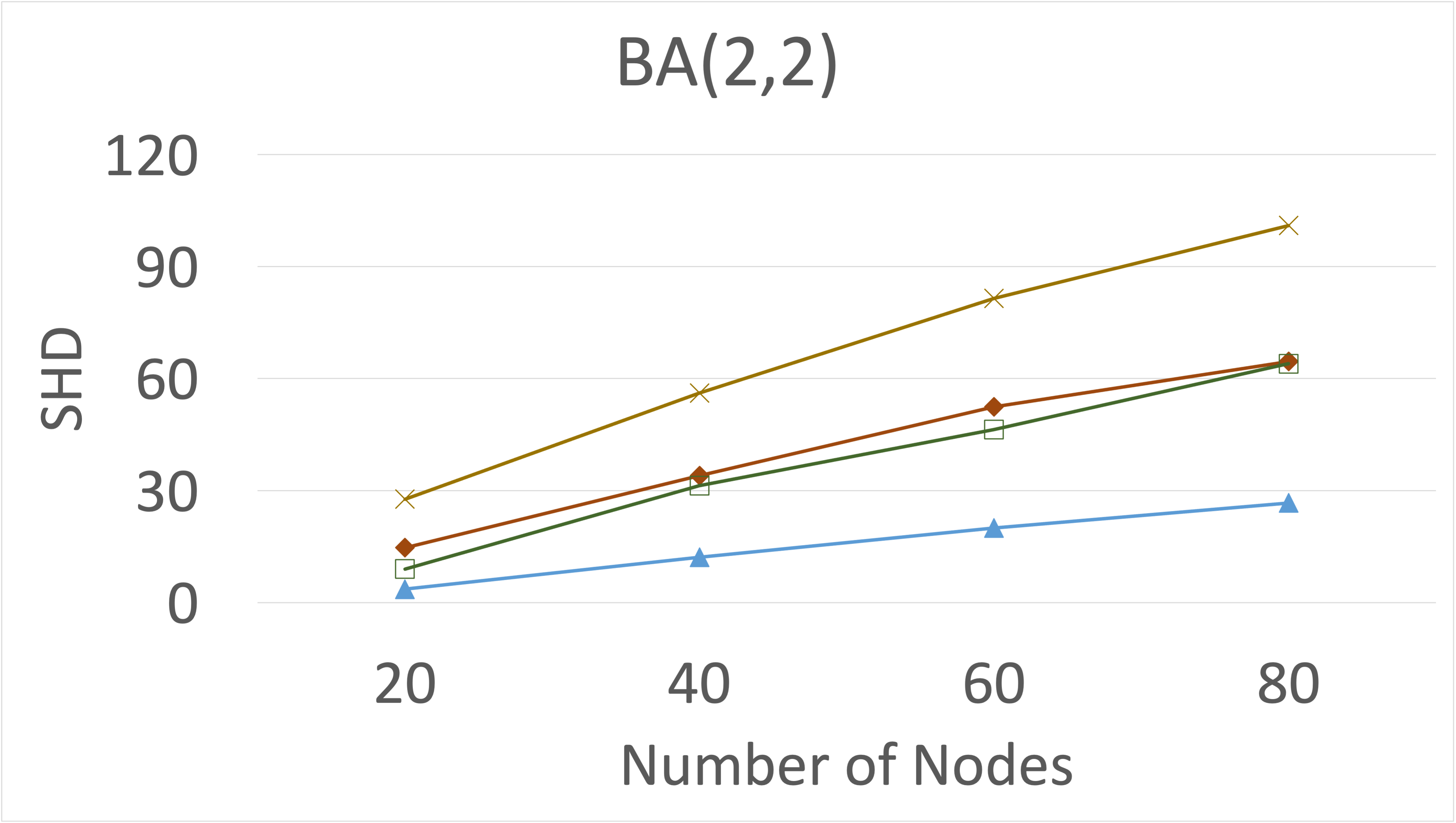}
    \includegraphics[width=0.49\textwidth, height=0.35\textwidth]{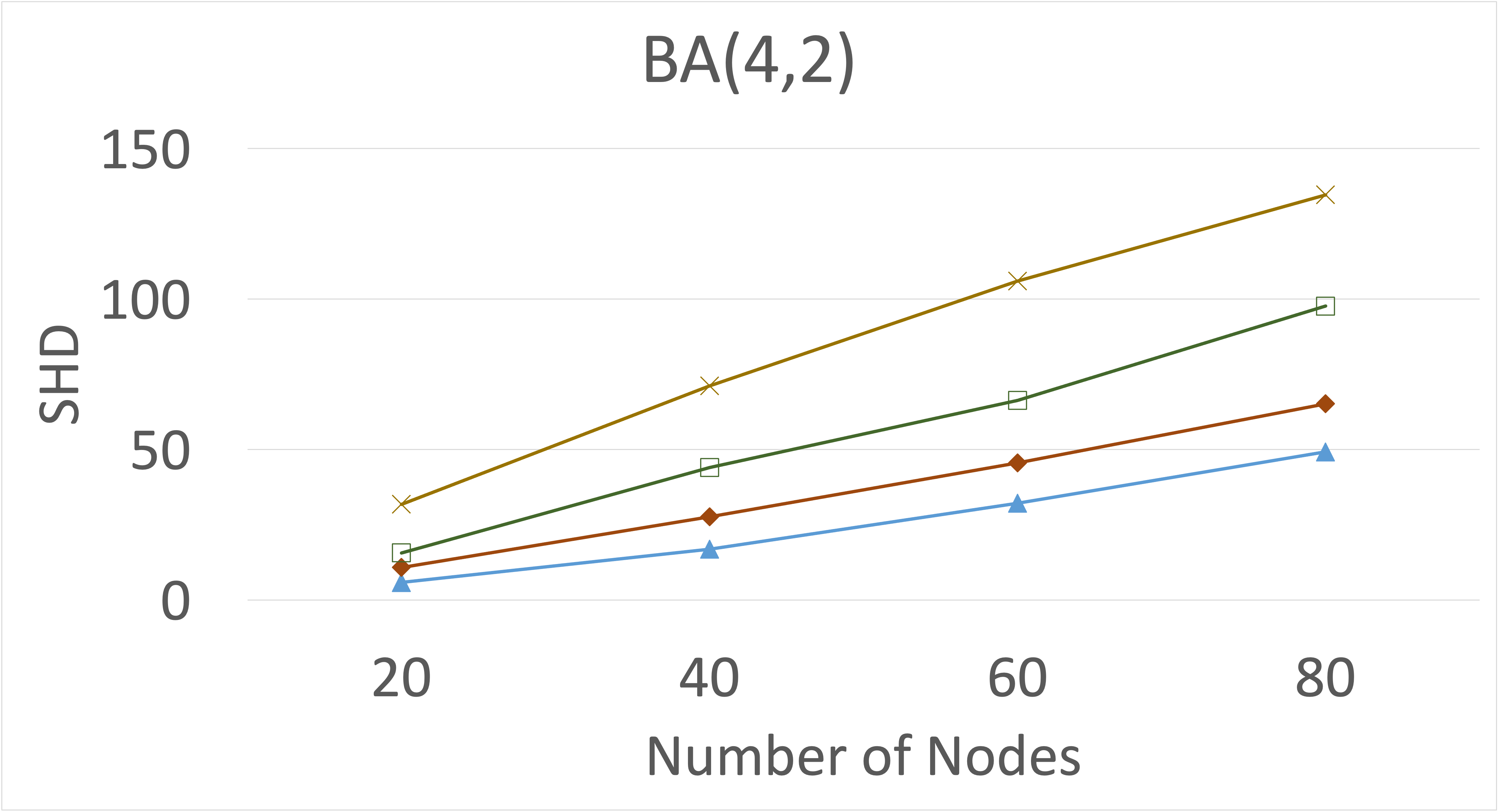}
    \caption{Additive Index Model, $T=1000$}
  \end{subfigure}
  \centering
  \begin{subfigure}{0.49\textwidth}
    \centering
    \includegraphics[width=0.49\textwidth, height=0.35\textwidth]{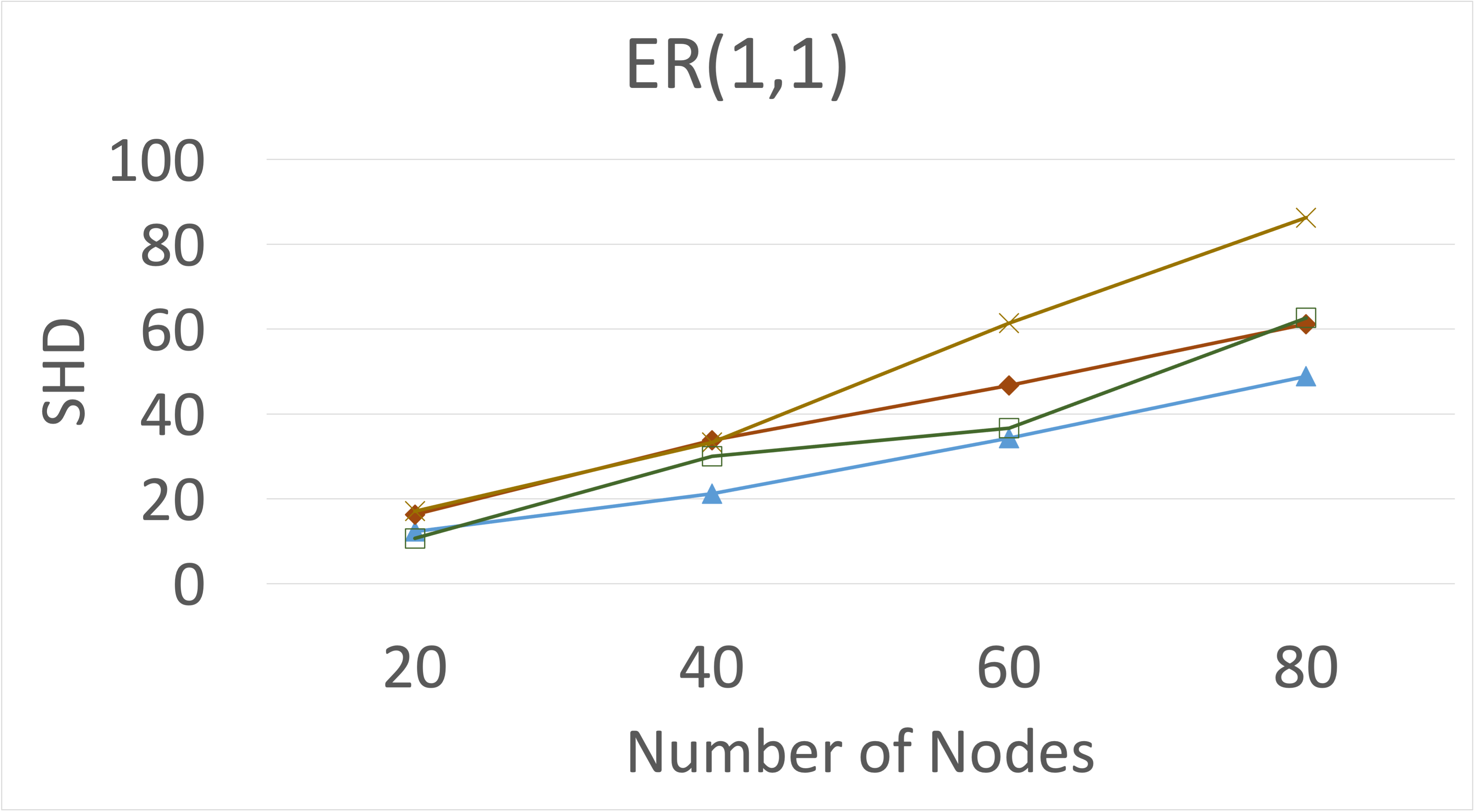}
    \includegraphics[width=0.49\textwidth, height=0.35\textwidth]{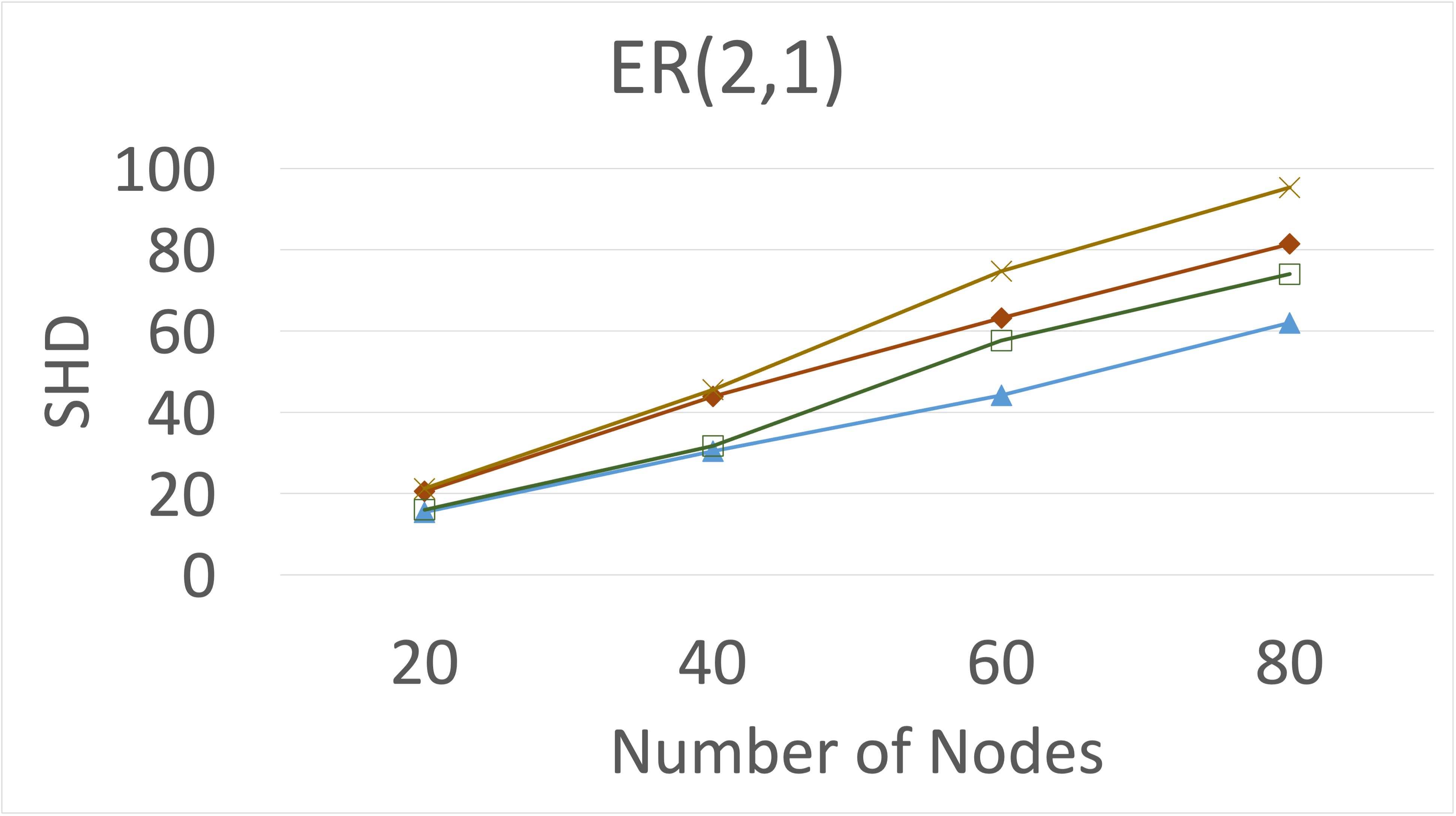}
    \includegraphics[width=0.49\textwidth, height=0.35\textwidth]{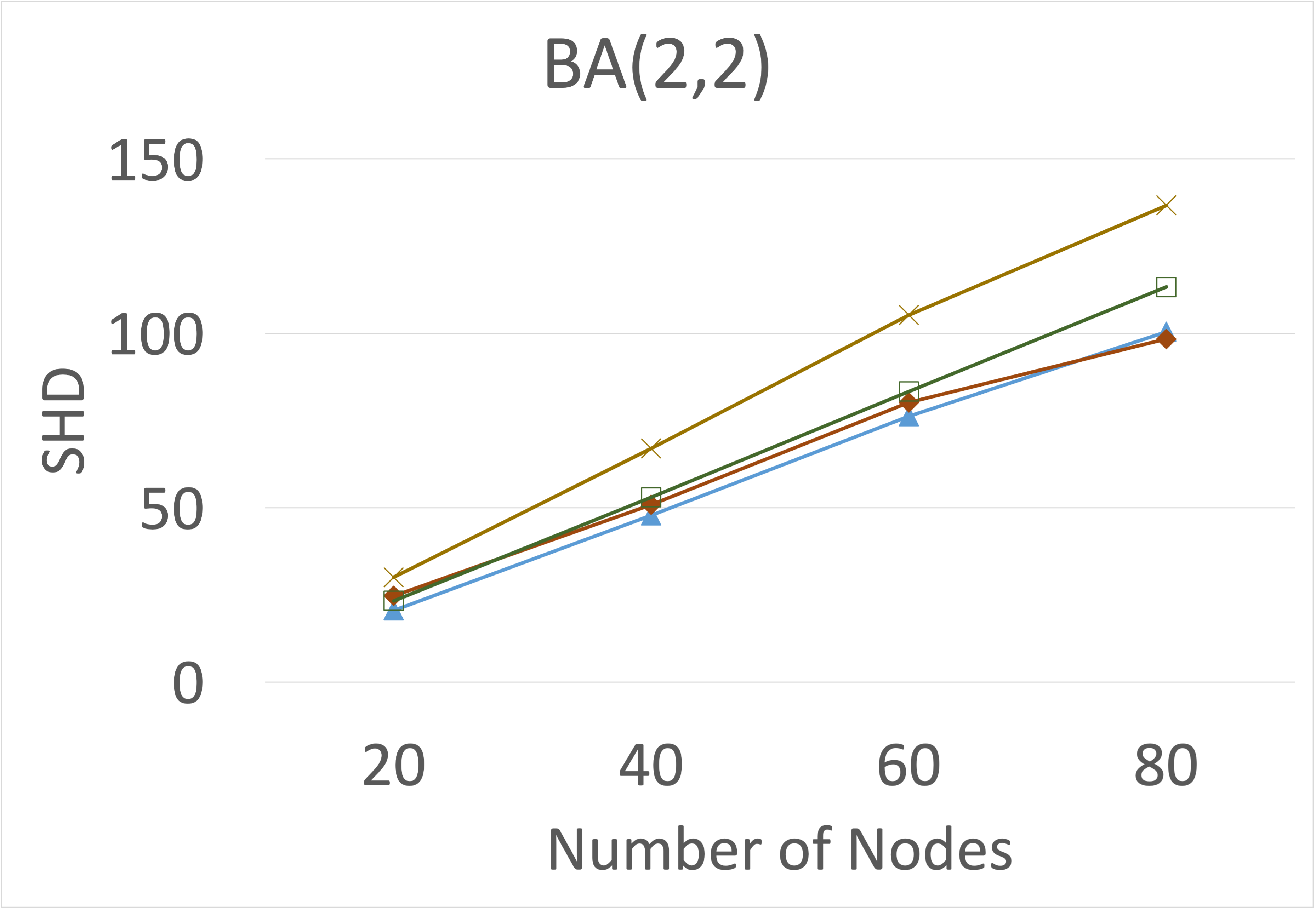}
    \includegraphics[width=0.49\textwidth, height=0.35\textwidth]{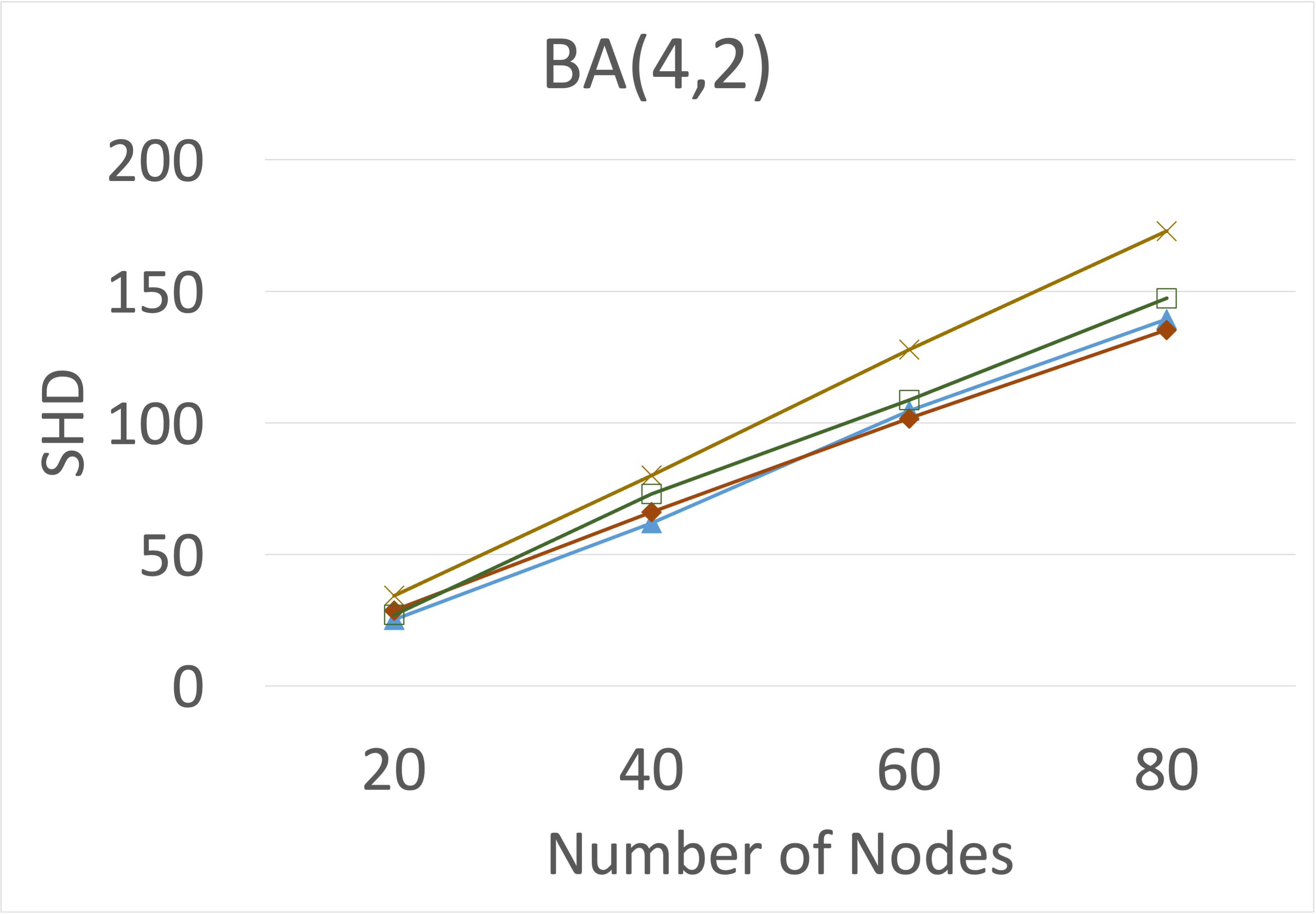}
    \caption{Additive Noise Model, $T=200$}
  \end{subfigure}
  \hfill
  \centering
  \begin{subfigure}{0.49\textwidth}
    \centering
    \includegraphics[width=0.49\textwidth, height=0.35\textwidth]{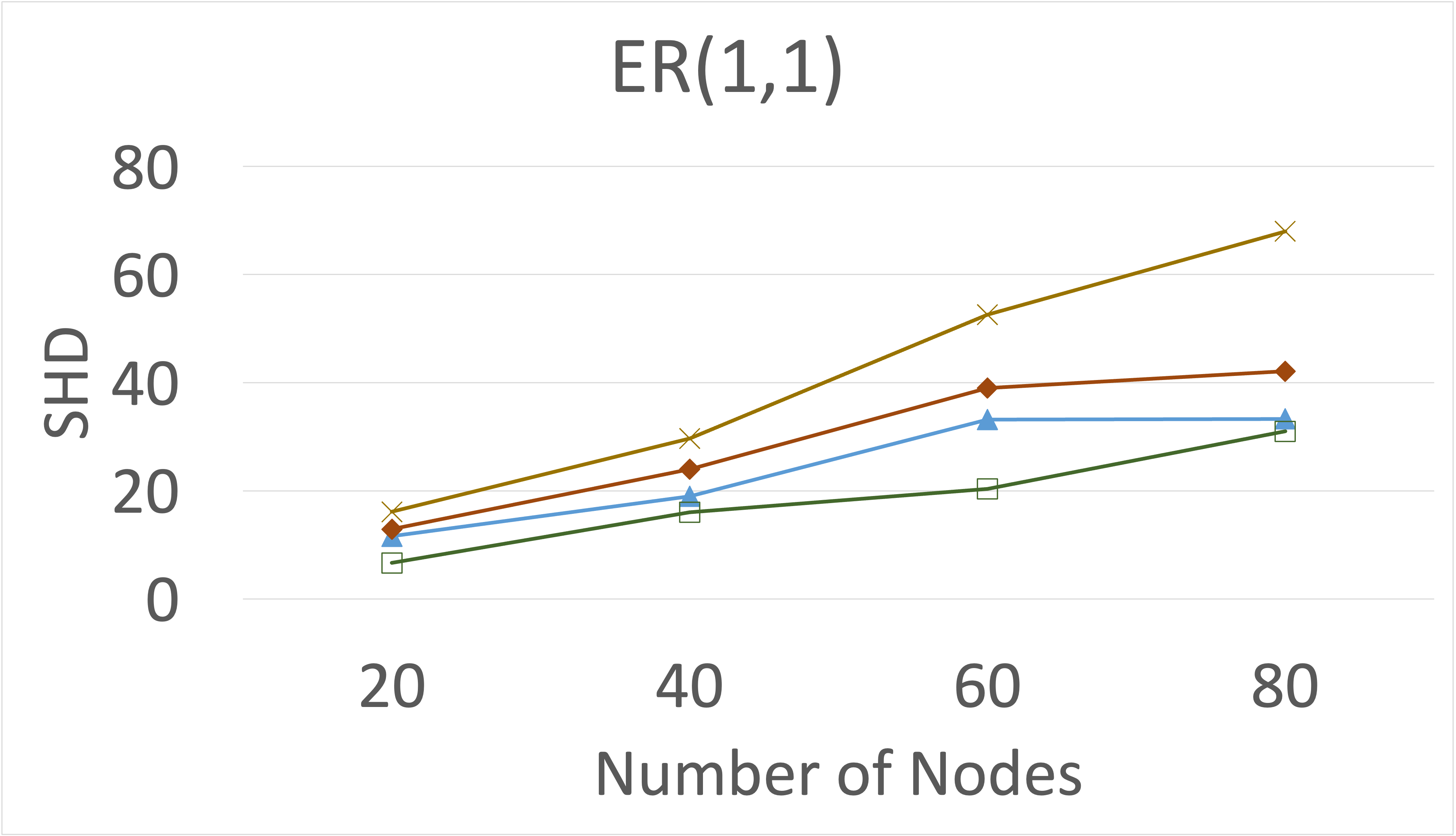}
    \includegraphics[width=0.49\textwidth, height=0.35\textwidth]{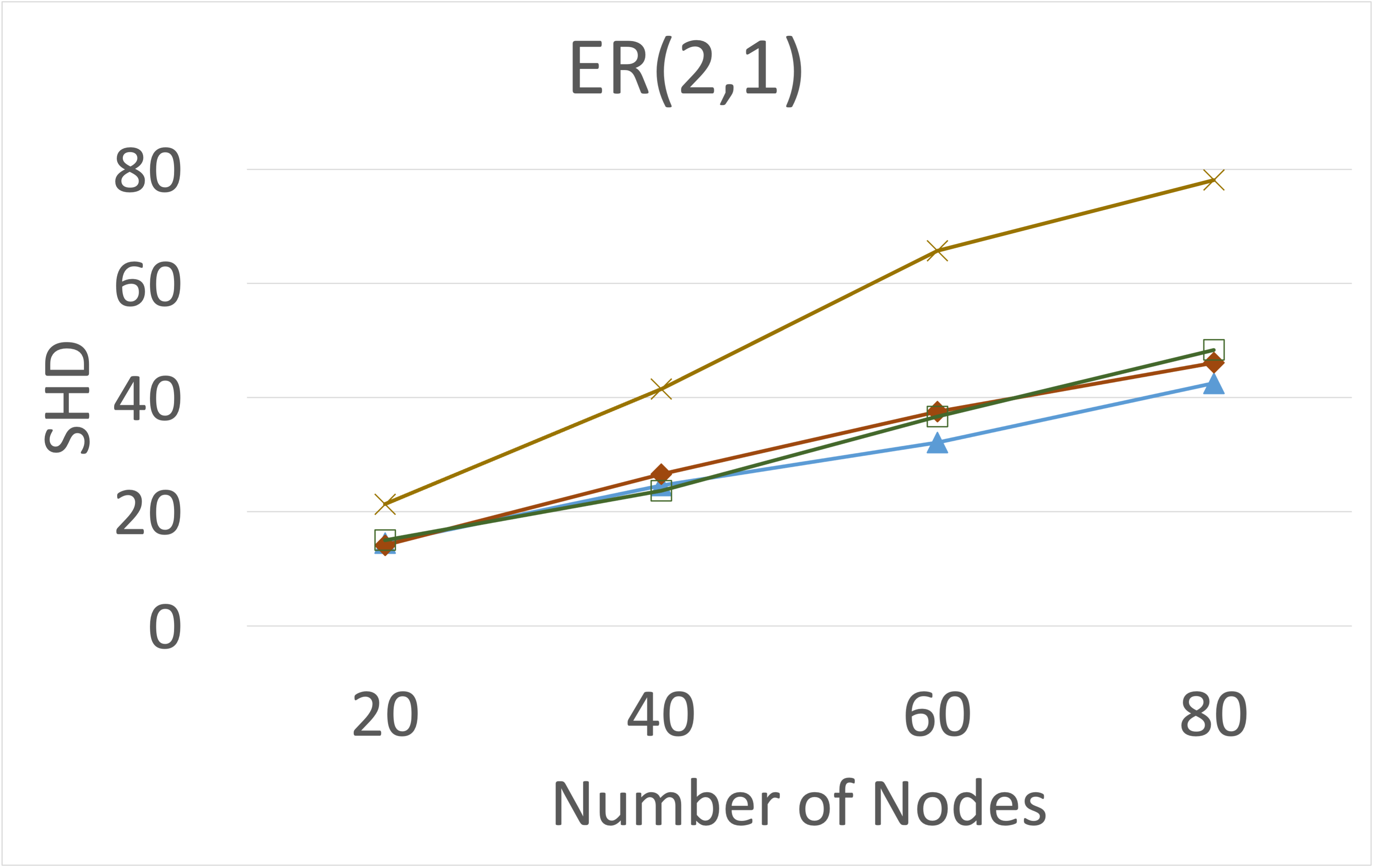}
    \includegraphics[width=0.49\textwidth, height=0.35\textwidth]{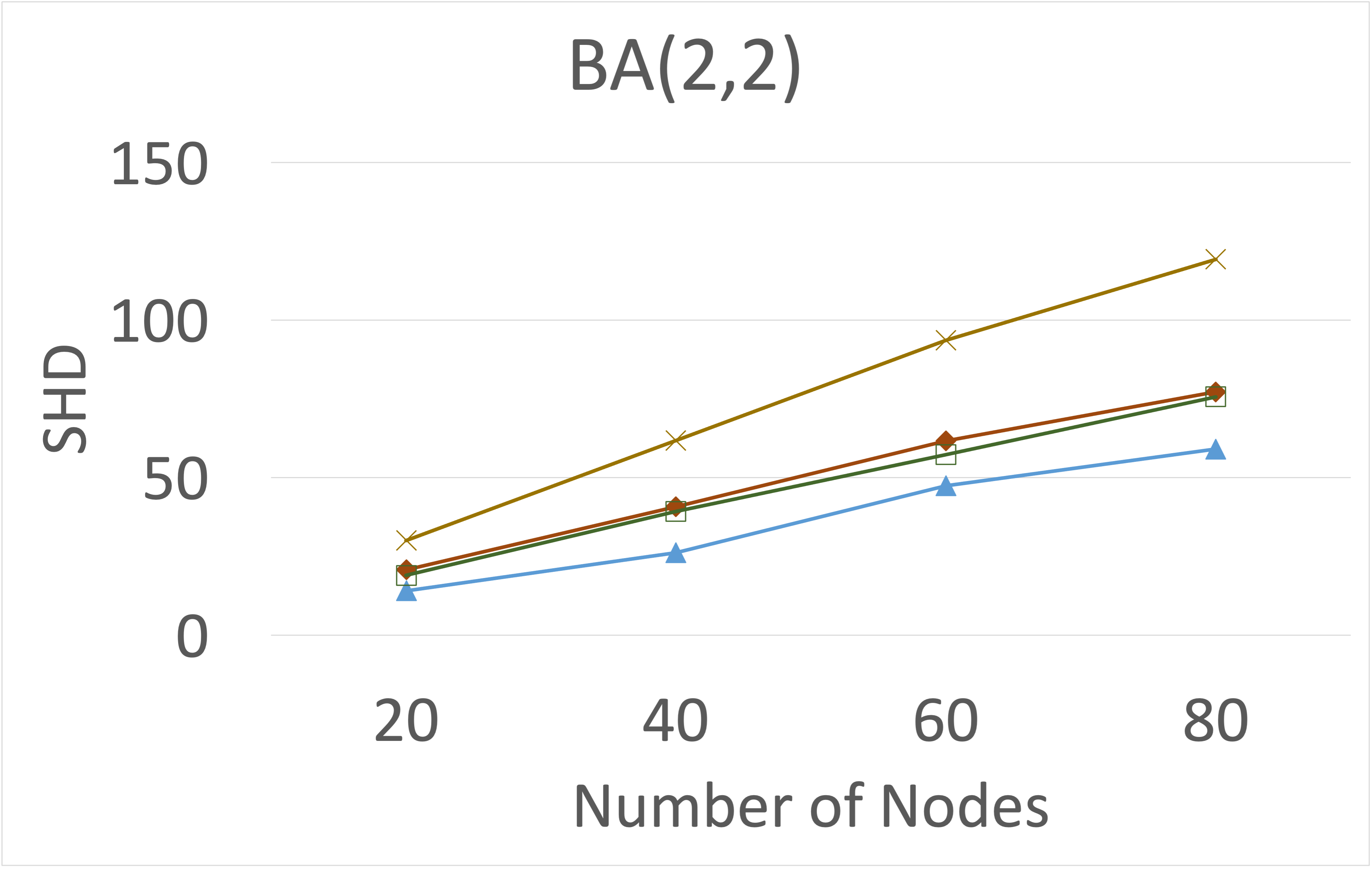}
    \includegraphics[width=0.49\textwidth, height=0.35\textwidth]{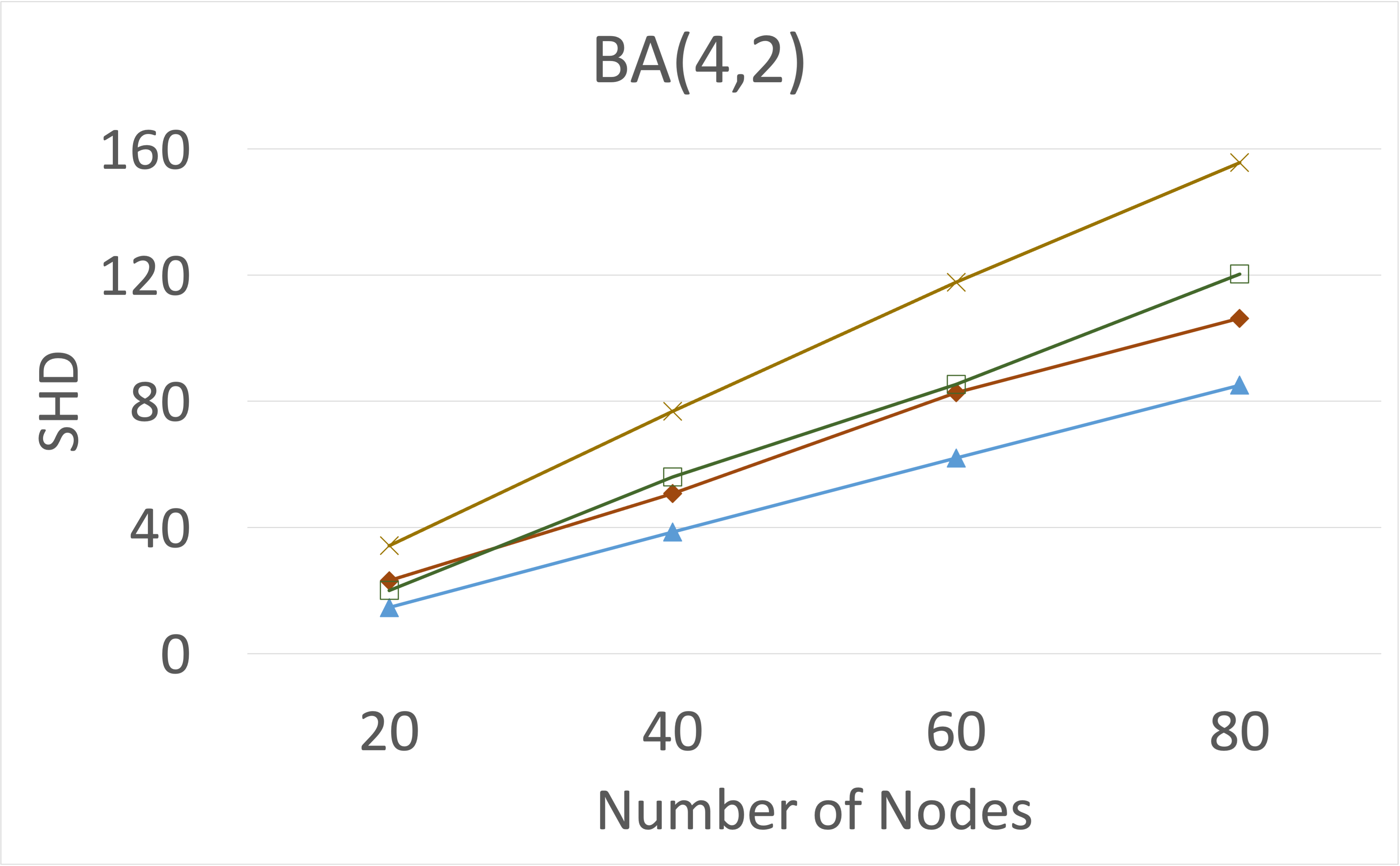}
    \caption{Additive Noise Model, $T=1000$}
  \end{subfigure}
  \centering
  \begin{subfigure}{0.49\textwidth}
    \centering
    \includegraphics[width=0.49\textwidth, height=0.35\textwidth]{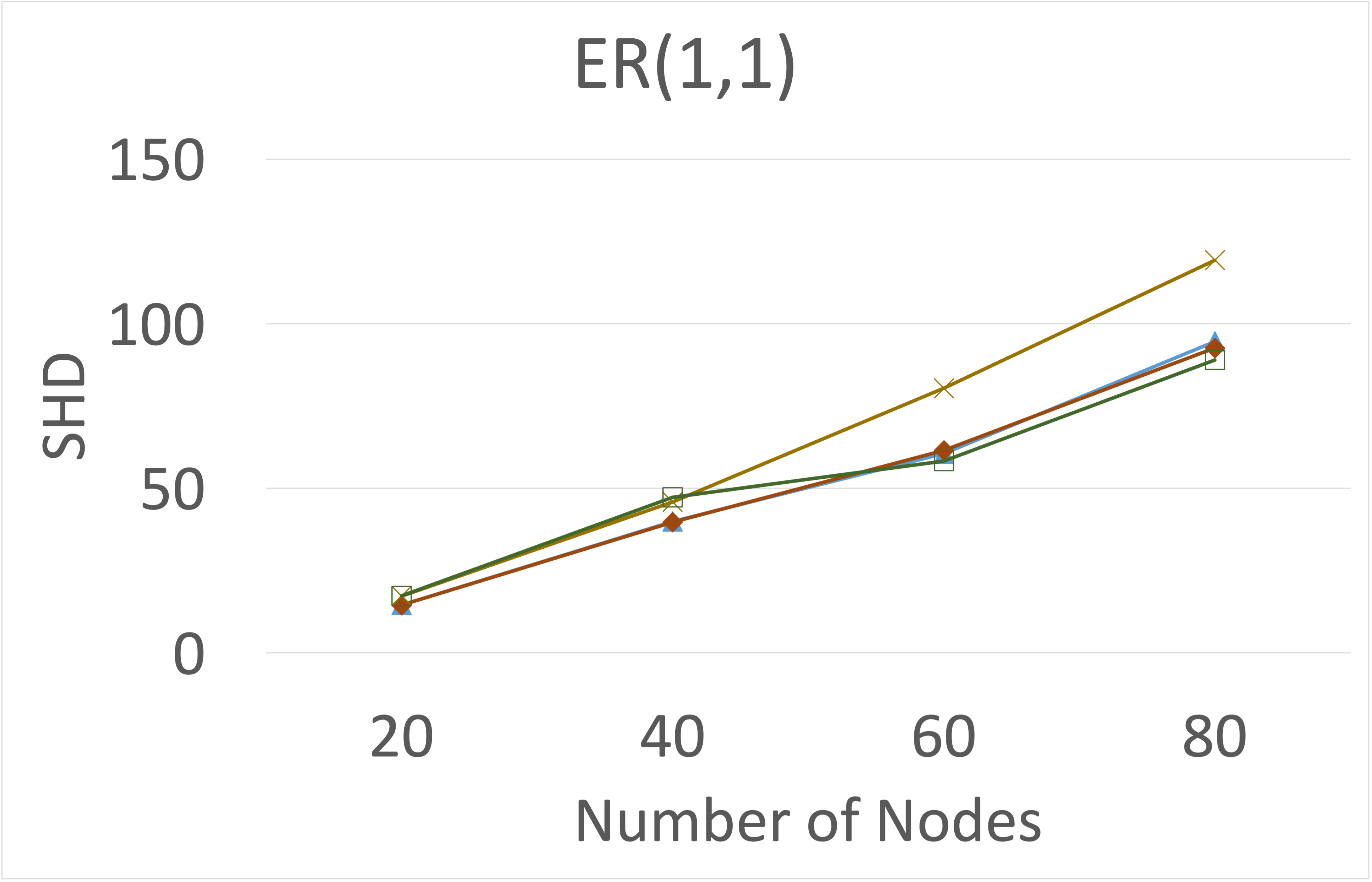}
    \includegraphics[width=0.49\textwidth, height=0.35\textwidth]{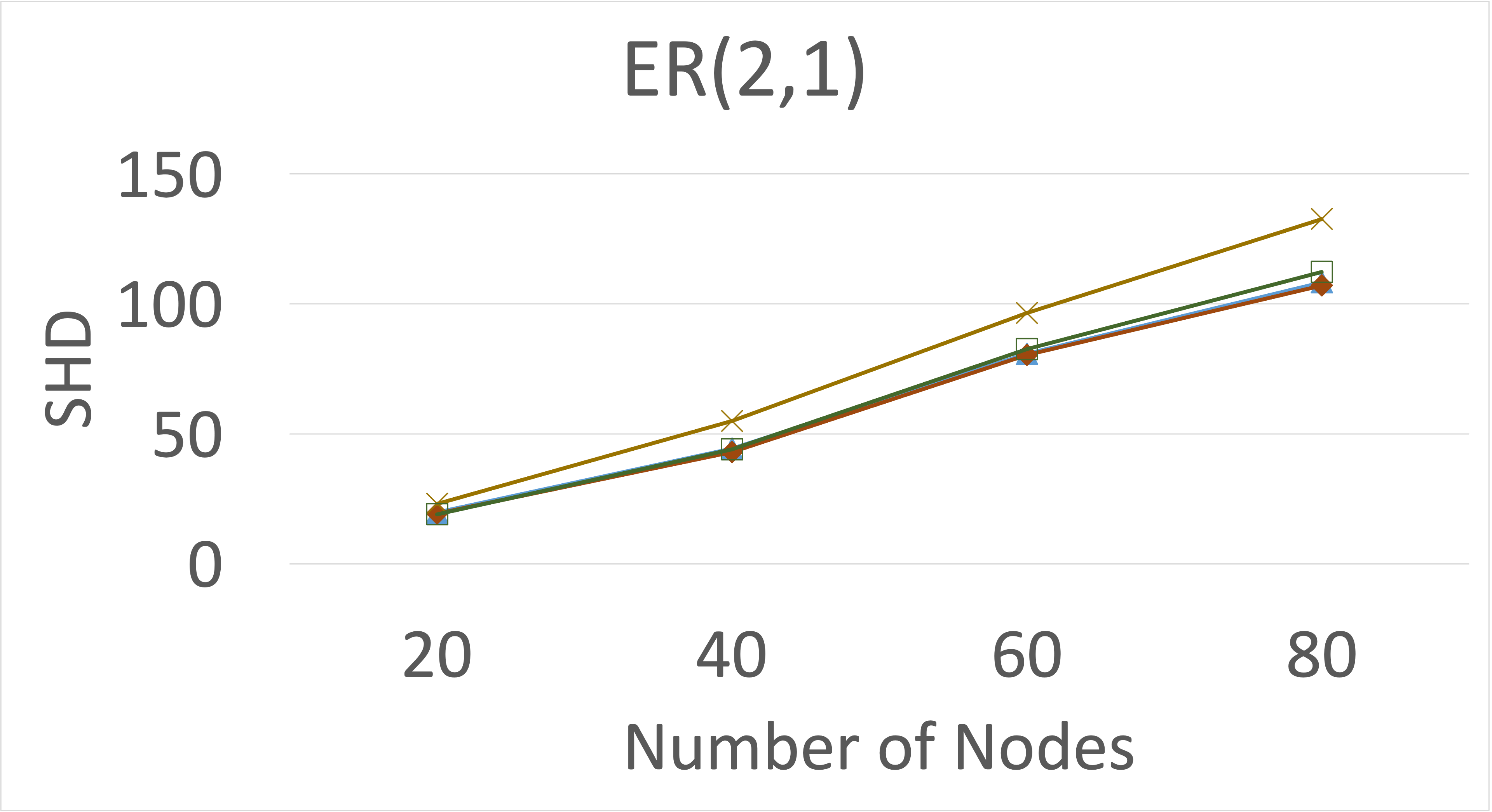}
    \includegraphics[width=0.49\textwidth, height=0.35\textwidth]{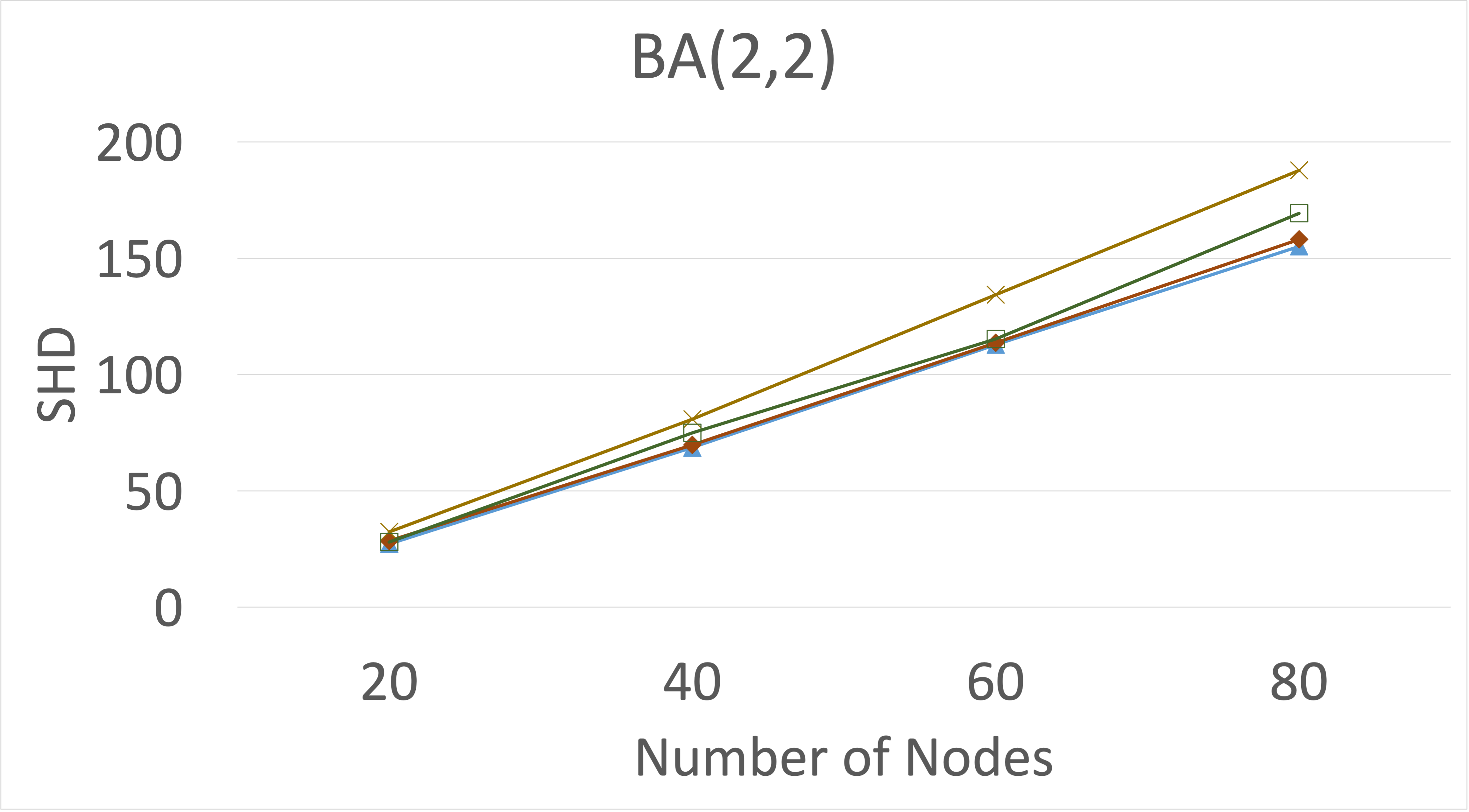}
    \includegraphics[width=0.49\textwidth, height=0.35\textwidth]{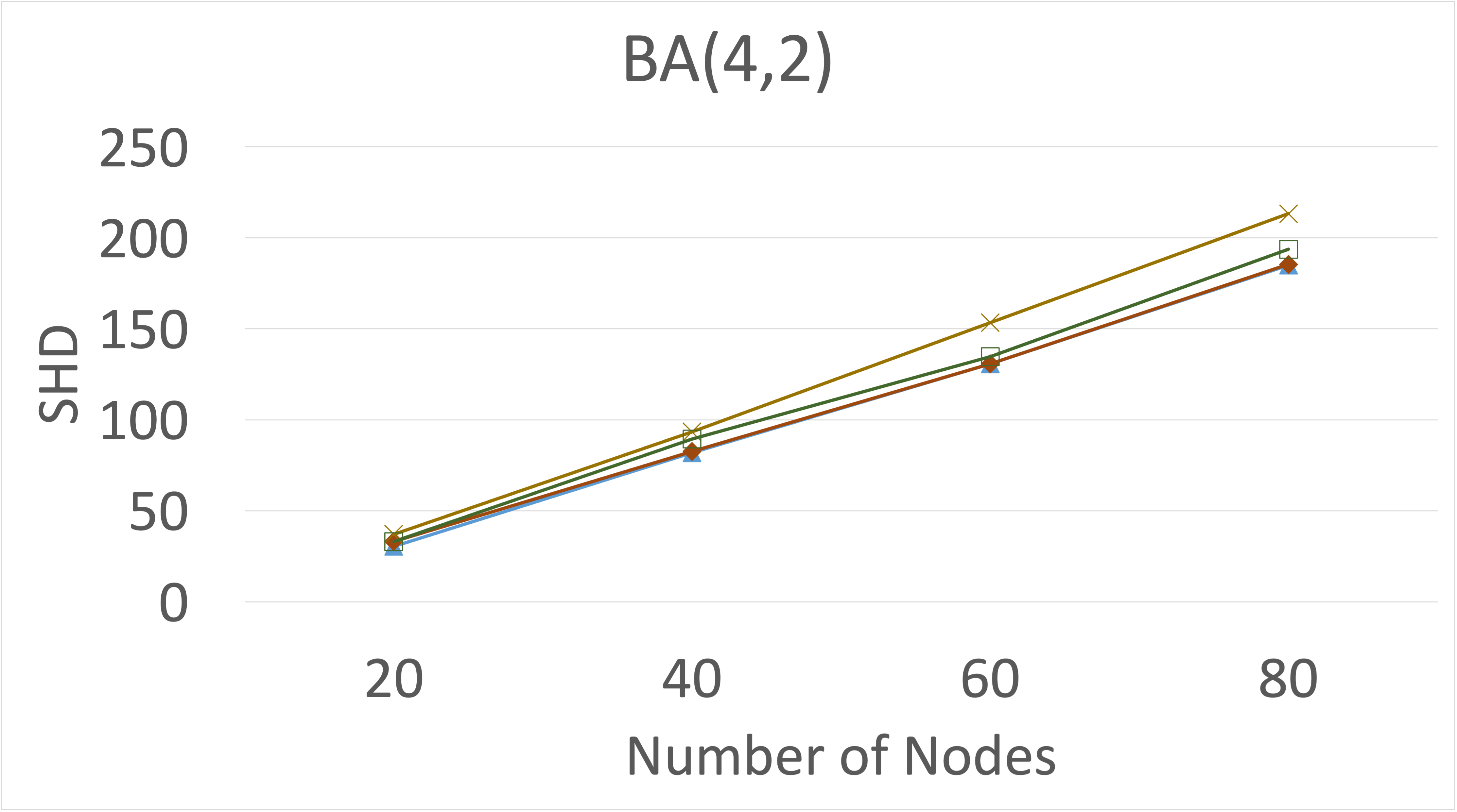}
    \caption{Generalized Linear Model with Poisson
Distribution, $T=200$}
  \end{subfigure}
  \hfill
  \centering
  \begin{subfigure}{0.49\textwidth}
    \centering
    \includegraphics[width=0.49\textwidth, height=0.35\textwidth]{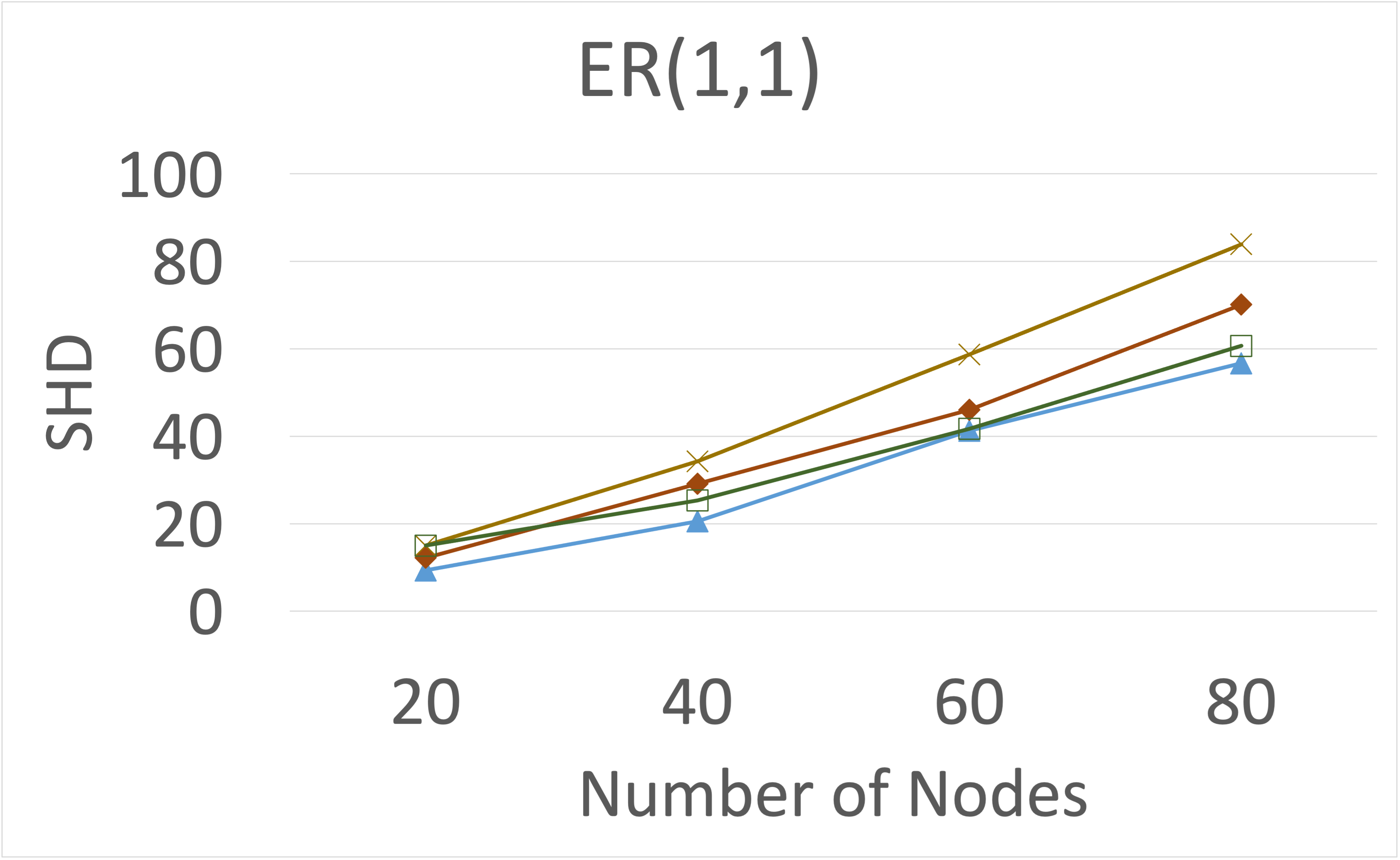}
    \includegraphics[width=0.49\textwidth, height=0.35\textwidth]{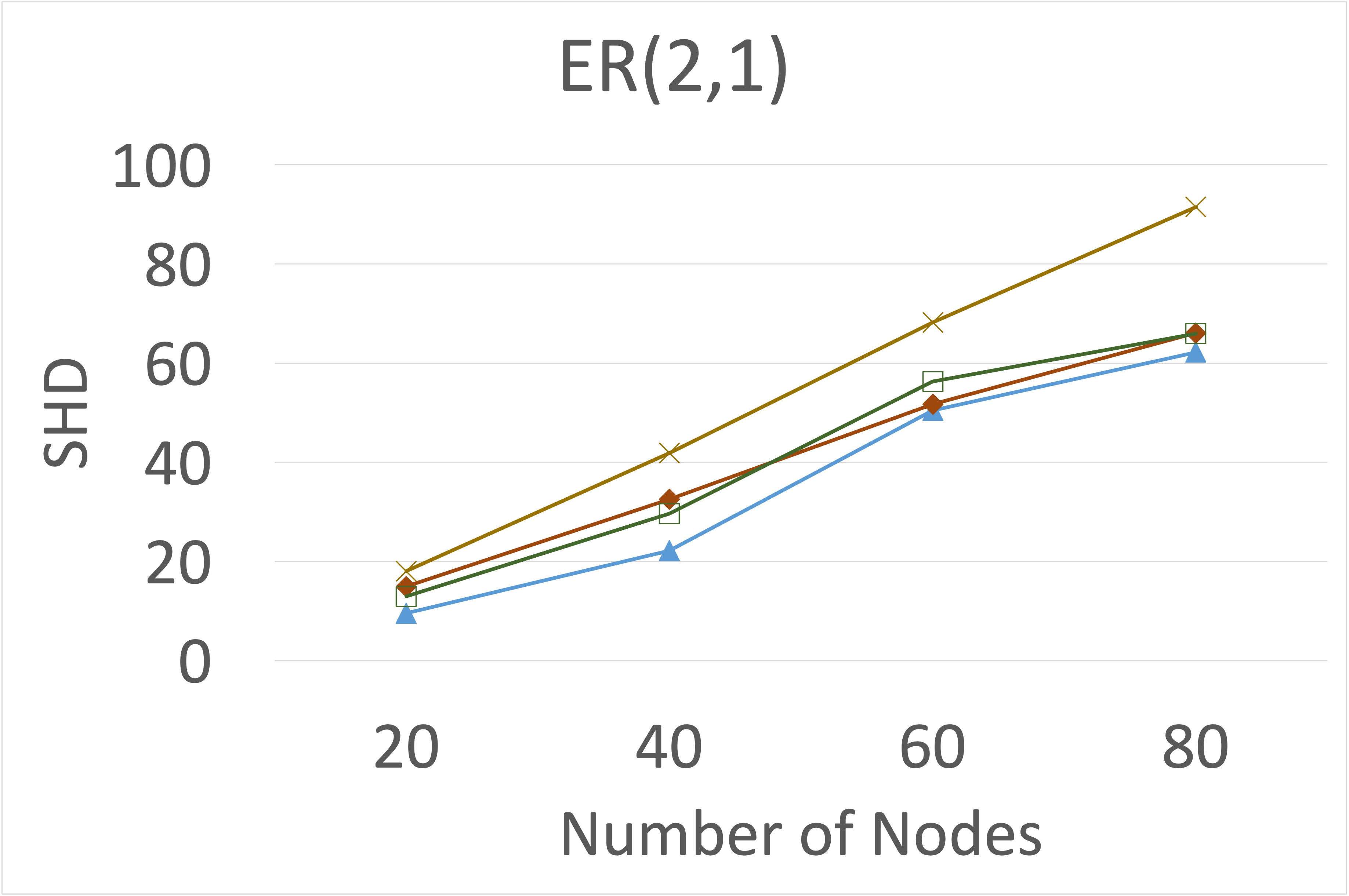}
    \includegraphics[width=0.49\textwidth, height=0.35\textwidth]{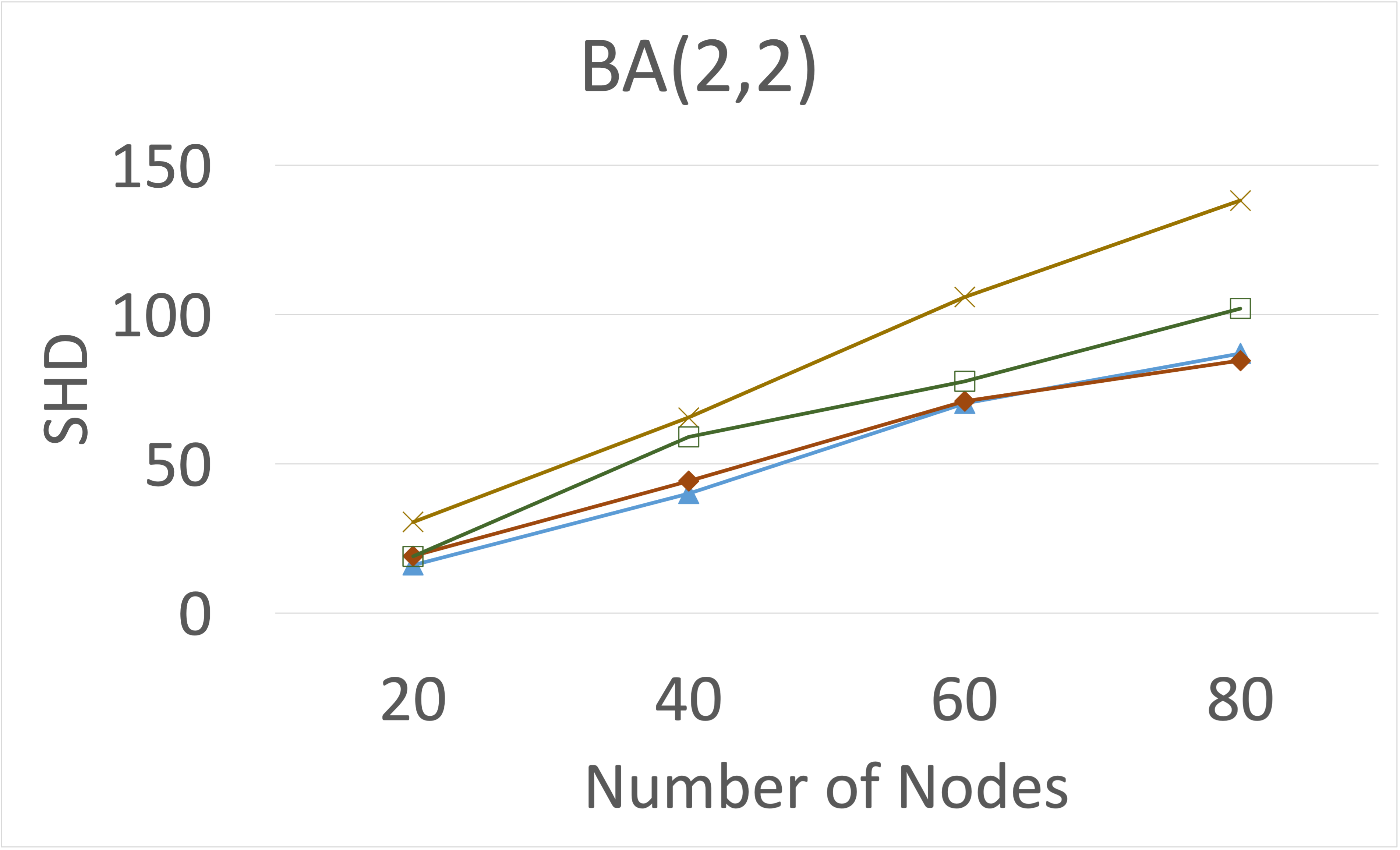}
    \includegraphics[width=0.49\textwidth, height=0.35\textwidth]{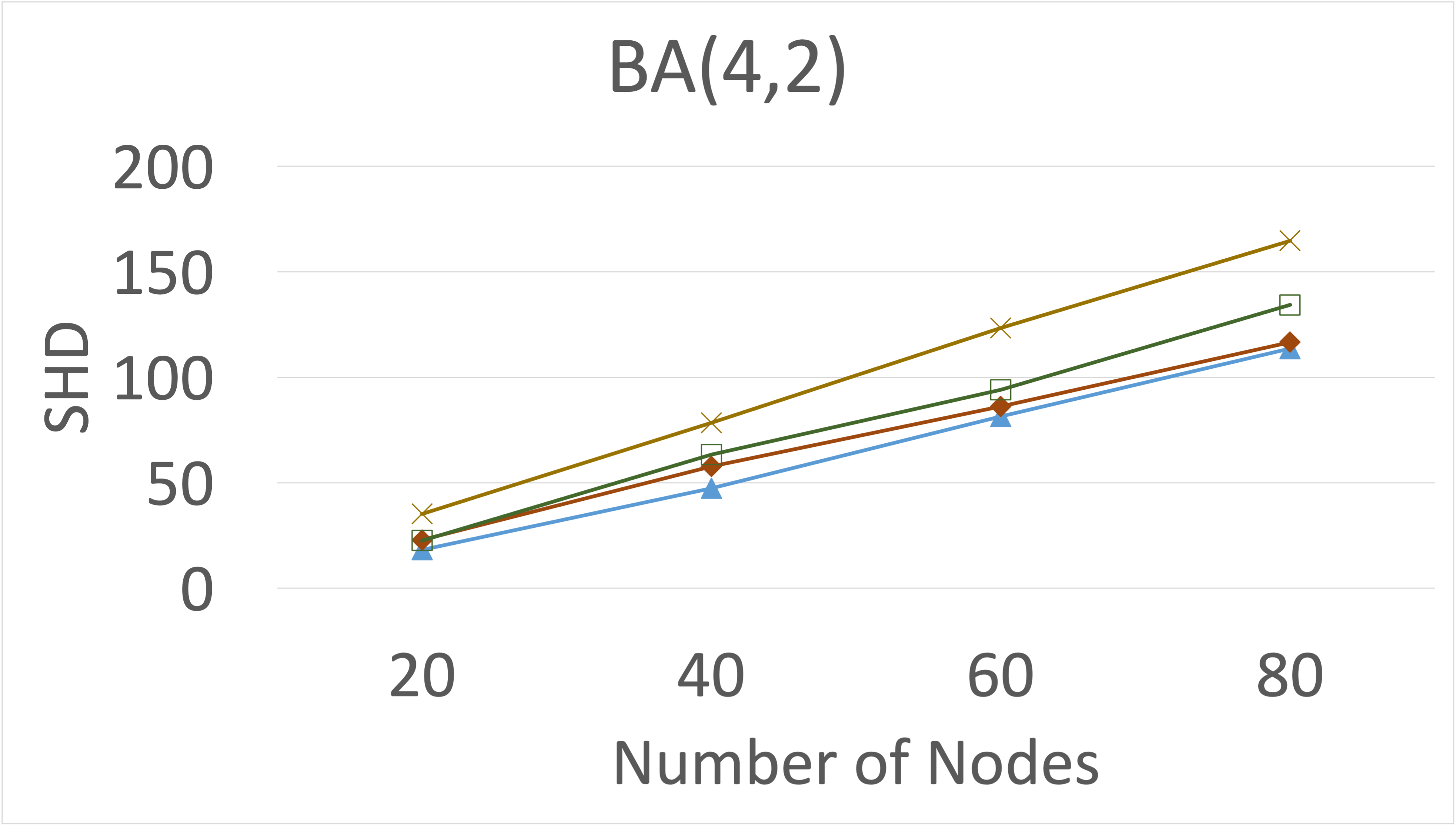}
    \caption{Generalized Linear Model with Poisson
Distribution, $T=1000$}
  \end{subfigure}
  \centering
  \begin{subfigure}{1.0\textwidth}
    \centering
    \includegraphics[width=0.5\textwidth, height=0.03\textwidth]{images/Simulated_Data/legend.png}
  \end{subfigure}
  \caption{\textbf{Mean SHDs} over 10 datasets for each setting with simulated data. Lower SHD is better. The number of lags $= 3$.}
  \label{fig:experiments_simulated_shd}
\end{figure*}

Besides reporting the F1-score in the main article, we also use recall, precision and SHD to evaluate the methods with simulated datasets. Please see Figure~\ref{fig:experiments_simulated_recall},~\ref{fig:experiments_simulated_precision},~\ref{fig:experiments_simulated_shd}. NTS-NOTEARS achieves the best recall and SHD in the vast majority of the settings and is among the top methods in terms of precision.

\section{METHOD HYPERPARAMETERS FOR LORENZ 96 \& FMRI BENCHMARKS}\label{section_hyperparameters_benchmarks}

To find the hyperparameter values for the evaluation methods, we select one dataset from each benchmark as the validation sets, and perform grid search over hyperparameters to maximize the F1-socre. These hyperparameters are used for evaluation with the benchmarks.
For NTS-NOTEARS, a unique set of hyperparameters is used for both benchmarks:

\begin{itemize}
  \item NTS-NOTEARS
    \begin{itemize}
    \item $\lambda_1^1 = 0.001$, $\lambda_1^2 = 0.1$
    \item $\lambda_2 = 0.01$
    \item $K = \textit{number of lags}$
    \item $m = 2d$
    \item the number of hidden layers $=1$
    \item $\bm{W}_{\textit{thres}} = \bm{0.5}$
  \end{itemize}
\end{itemize}


For the Lorenz 96 benchmark:
\begin{itemize}


  \item PCMCI+
  \begin{itemize}
    \item CI test: GPDC
    \item $ \tau_{\textit{min}} = 0 $
    \item $ \tau_{\textit{max}} = \textit{number of lags}$
    \item $ \alpha = 0.005$
  \end{itemize}

  \item TCDF
  \begin{itemize}
    \item $\textit{significance} = 1.5$
    \item $\textit{learning rate} = 0.001$
    \item $\textit{epochs} = 1000$
    \item $\textit{levels} = 3$
    \item $\textit{kernel size} = \textit{number of lags} + 1$
    \item $\textit{dilation coefficient} = \textit{number of lags} + 1$
  \end{itemize}


  \item DYNOTEARS
  \begin{itemize}
    \item $ \lambda_{w} = 0.1 $
    \item $ \lambda_{a} = 0.01 $
    \item $p = \textit{number of lags}$
    \item $\textit{weight threshold} = 0.1$
  \end{itemize}
  
  
\end{itemize}

For the fMRI benchmark:
\begin{itemize}

  \item PCMCI+
  \begin{itemize}
    \item CI test: GPDC
    \item $ \tau_{\textit{min}} = 0 $
    \item $ \tau_{\textit{max}} = \textit{number of lags}$
    \item $ \alpha = 0.001$
  \end{itemize}

  \item TCDF
  \begin{itemize}
    \item $\textit{significance} = 0.5$
    \item $\textit{learning rate} = 0.01$
    \item $\textit{epochs} = 2000$
    \item $\textit{levels} = 2$
    \item $\textit{kernel size} = \textit{number of lags} + 1$
    \item $\textit{dilation coefficient} = \textit{number of lags} + 1$
  \end{itemize}


  \item DYNOTEARS
  \begin{itemize}
    \item $ \lambda_{w} = 0.1 $
    \item $ \lambda_{a} = 0.1 $
    \item $p = \textit{number of lags}$
    \item $\textit{weight threshold} = 0.1$
  \end{itemize}
  

\end{itemize}

\begin{figure}[ht]
  \centering
  \begin{subfigure}{0.20\textwidth}
    \centering
    \includegraphics[width=\textwidth]{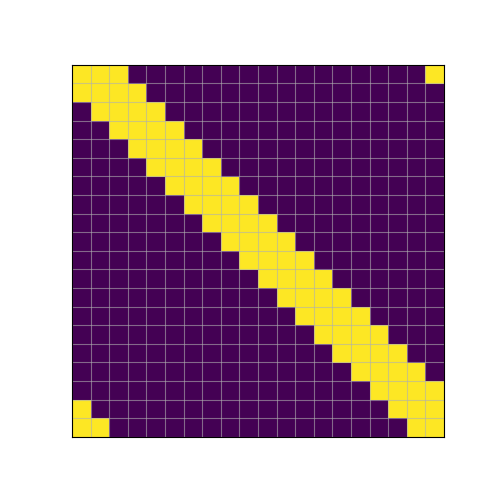}
    \caption{The ground-truth DBN from the Lorenz 96 benchmark}
    \label{}
  \end{subfigure}
  \hspace{1cm}
  \begin{subfigure}{0.20\textwidth}
    \includegraphics[width=\textwidth]{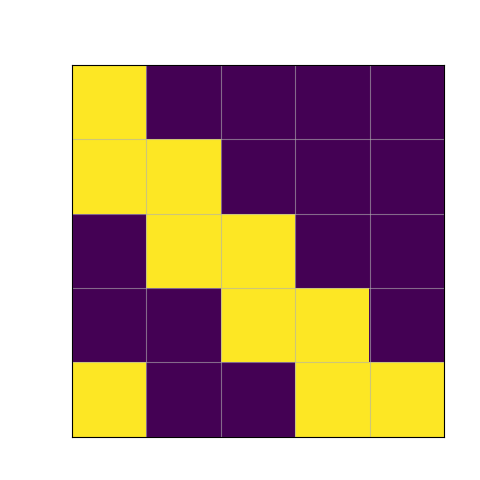}
    \caption{The ground-truth DBN from the fMRI benchmark}
    \label{}
  \end{subfigure}
  \caption{DBNs showing edges from the lag pointing to the instantaneous time step. 
  }
  \label{fig:lorenz96_fmri_color_maps}
\end{figure}

\section{MORE RESULTS WITH LORENZ 96 \& FMRI BENCHMARKS}\label{section_benchmark_more_results}

Besides reporting the F1-score in the main article, we also use SHD, precision and recall to evaluate the methods with benchmark datasets. Please see Table~\ref{tab:lorenz96_fmri_shd},~\ref{tab:lorenz96_fmri_recall},~\ref{tab:lorenz96_fmri_precision}. NTS-NOTEARS achieves the best SHD, recall and precision with both benchmarks. Around 20 combinations of hyperparameter values were used for each method to compute the AUROC in Table~\ref{tab:lorenz96_fmri_auc}.

\begin{table}[h]
\centering
\caption{\textbf{Mean SHDs ($\pm$ SE)} computed with the benchmark datasets.}
    \begin{tabular}{l|c|c}
    \toprule
    Method & Lorenz 96 & fMRI \\
    \midrule
    DYNOTEARS & 20.4 ($\pm$ 1.992) & 10.8 ($\pm$ 0.645) \\
    \midrule
    TCDF & 58.0 ($\pm$ 1.789) & 9.0 ($\pm$ 0.600) \\
    \midrule
    PCMCI+ & 44.6 ($\pm$ 2.492) & 8.1 ($\pm$ 0.624) \\
    \midrule
    NTS-NOTEARS & \textbf{0.6 ($\pm$ 0.358)} & \textbf{6.6 ($\pm$ 0.651)} \\
    \bottomrule
    \end{tabular}
\label{tab:lorenz96_fmri_shd}
\end{table}


\begin{table}[h]
\centering
\caption{\textbf{Mean recalls ($\pm$ SE)} computed with the benchmark datasets.}
    \begin{tabular}{l|c|c}
    \toprule
    Method & Lorenz 96 & fMRI \\
    \midrule
    DYNOTEARS & 0.760 ($\pm$ 0.024) & 0.460 ($\pm$ 0.016) \\
    \midrule
    TCDF & 0.308 ($\pm$ 0.013) & 0.246 ($\pm$ 0.043) \\
    \midrule
    PCMCI+ & 0.495 ($\pm$ 0.032) & 0.406 ($\pm$ 0.043) \\
    \midrule
    NTS-NOTEARS & \textbf{0.993 ($\pm$ 0.004)} & \textbf{0.516 ($\pm$ 0.014)} \\
    \bottomrule
    \end{tabular}
\label{tab:lorenz96_fmri_recall}
\end{table}

\begin{table}[h]
\centering
\caption{\textbf{Mean precisions ($\pm$ SE)} computed with the benchmark datasets.}
    \begin{tabular}{l|c|c}
    \toprule
    Method & Lorenz 96 & fMRI \\
    \midrule
    DYNOTEARS & 0.981 ($\pm$ 0.007) & 0.508 ($\pm$ 0.043) \\
    \midrule
    TCDF & 0.906 ($\pm$ 0.035) & 0.610 ($\pm$ 0.098) \\
    \midrule
    PCMCI+ & 0.903 ($\pm$ 0.006) & 0.754 ($\pm$ 0.070) \\
    \midrule
    NTS-NOTEARS & \textbf{1.000 ($\pm$ 0.000)} & \textbf{0.839 ($\pm$ 0.062)} \\
    \bottomrule
    \end{tabular}
\label{tab:lorenz96_fmri_precision}
\end{table}

\section{REAL-WORLD ICE HOCKEY DATASET}\label{section_ice_hockey_appendix}

\begin{figure}[ht]
  \centering
  \includegraphics[width=0.5\textwidth]{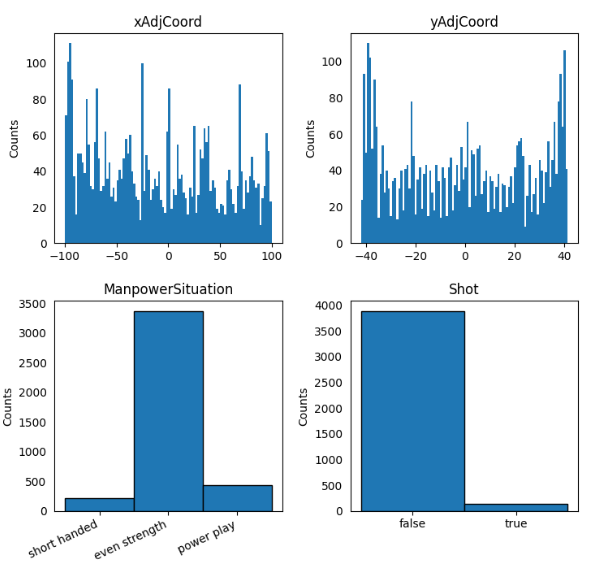}
  \caption{The distributions of two non-Gaussian continuous variables and two discrete variables in the ice hockey dataset.}
  \label{fig:data_distributions}
\end{figure}

Please see Table~\ref{tab:variable_description} for data description and Figure~\ref{fig:data_distributions} for data distribution plots. A nominal variable with $K$ values can be converted to $K$ binary variables using one-hot encoding. The following hyperparameter values are used in Section~\ref{section_ice_hockey}. With these hyperparameter values, NTS-NOTEARS and DYNOTEARS generate graphs with the same number of edges.
\begin{itemize}

  \item NTS-NOTEARS
  \begin{itemize}
    \item $\bm{\lambda_1} = \bm{0.001}$
    \item $\lambda_2 = 0.005$
    \item $K = 1$
    \item $m=d$
    \item the number of hidden layers $=1$
    \item $\bm{W}_{\textit{thres}} = \bm{0.5}$
  \end{itemize}
  
  \item DYNOTEARS
  \begin{itemize}
    \item $ \lambda_{w} = 0.01 $
    \item $ \lambda_{a} = 0.01 $
    \item $p = 1$
    \item $\textit{weight threshold} = 0.03$
  \end{itemize}
\end{itemize}

\begin{table}[h]
\centering
\caption{The variables in the ice hockey dataset.}
  \begin{tabular}{l|l|p{2cm}}
    \toprule
    Variables & Type & Range \\
    \midrule
    time remaining in seconds& continuous & [0, 3600]\\
    \midrule
    adjusted x coordinate of puck & continuous & [-100, 100]\\
    \midrule
    adjusted y coordinate of puck & continuous & [-42.5, 42.5]\\
    \midrule
    score differential & categorical & ($-\infty$, $+\infty$)\\
    \midrule
    manpower situation & categorical & \{short handed, even strength, power play\}\\
    \midrule
    x velocity of puck & continuous & ($-\infty$, $+\infty$)\\
    \midrule
    y velocity of puck & continuous & ($-\infty$, $+\infty$)\\
    \midrule
    event duration & continuous & [0, $+\infty$)\\
    \midrule
    angle between puck and net & continuous & [$-\pi$, $+\pi$]\\
    \midrule
    home team taking possession  & binary & \{true, false\}\\
    \midrule
    shot & binary & \{true, false\}\\
    \midrule
    goal & binary & \{true, false\}\\
  \bottomrule
\end{tabular}
\label{tab:variable_description}
\end{table}

\begin{figure}[ht]
    \centering
    \begin{subfigure}{0.3\textwidth}
        \centering
        \includegraphics[width=\textwidth]{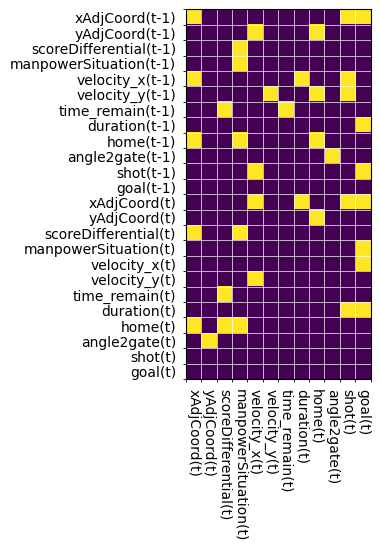}
        \caption{NTS-NOTEARS}
        \label{}
    \end{subfigure}
    \hspace{1cm}
    \begin{subfigure}{0.3\textwidth}
        \centering
        \includegraphics[width=\textwidth]{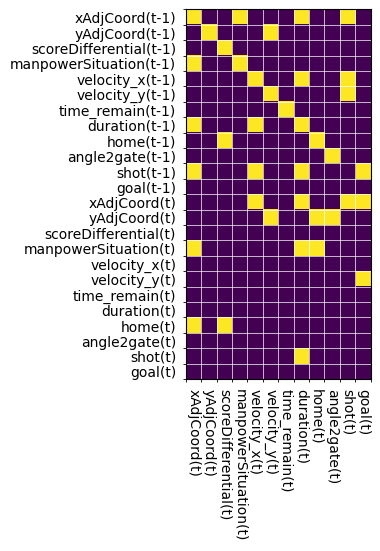}
        \caption{DYNOTEARS}
        \label{}
    \end{subfigure}
  \caption{The DBNs estimated by NTS-NOTEARS and DYNOTEARS with real-world ice hockey data.
  }
  \label{fig:ice_hockey_colourmaps}
\end{figure}

\vfill

\end{document}